\documentclass[10pt,twocolumn,letterpaper]{article}

\usepackage[pagenumbers]{iccv} 

\definecolor{iccvblue}{rgb}{0.21,0.49,0.74}
\usepackage[pagebackref,breaklinks,colorlinks,allcolors=iccvblue]{hyperref}
\usepackage{pifont}
\usepackage{pgfplots}
\usepackage{pgfplotstable}
\pgfplotsset{compat=1.18}

\def\paperID{11386} 
\def\confName{ICCV}
\def\confYear{2025}

\title{Only-Style: Stylistic Consistency in Image Generation \\ without Content Leakage
\vspace{.4cm}} 

\author{
Tilemachos Aravanis\textsuperscript{1} \quad
Panagiotis Filntisis\textsuperscript{2,3} \quad
Petros Maragos\textsuperscript{1, 2, 3} \quad
George Retsinas\textsuperscript{2, 3} \\
\textsuperscript{1}School of Electrical \& Computer Engineering, National Technical University of Athens, Greece \\
\textsuperscript{2}Robotics Institute, Athena Research Center, 15125 Maroussi, Greece \\
\textsuperscript{3}HERON - Center of Excellence in Robotics, Athens, Greece
}

\begin{document}

\maketitle
\begin{abstract}

Generating images in a consistent reference visual style remains a challenging computer vision task. State-of-the-art methods aiming for style-consistent generation struggle to effectively separate semantic content from stylistic elements, leading to content leakage from the image provided as a reference to the targets. To address this challenge, we propose Only-Style: a method designed to mitigate content leakage in a semantically coherent manner while preserving stylistic consistency. Only-Style works by localizing content leakage during inference, allowing the adaptive tuning of a parameter that controls the style alignment process, specifically within the image patches containing the subject in the reference image. This adaptive process best balances stylistic consistency with leakage elimination. Moreover, the localization of content leakage can function as a standalone component, given a reference-target image pair, allowing the adaptive tuning of any method-specific parameter that provides control over the impact of the stylistic reference. In addition, we propose a novel evaluation framework to quantify the success of style-consistent generations in avoiding undesired content leakage. Our approach demonstrates a significant improvement over state-of-the-art methods through extensive evaluation across diverse instances, consistently achieving robust stylistic consistency without undesired content leakage. \href{https://tilemahosaravanis.github.io/Only-Style-PP/}{\textcolor{purple}{\emph{Project-Page}}} 

\vspace{-3pt}

\end{abstract}    
\section{Introduction}
\label{sec:intro}

\begin{figure}[t]
  \centering
    \begin{tabular}{c@{\hspace{.1cm}}c@{\hspace{.1cm}}c@{\hspace{.1cm}}c}
    
        \scriptsize ``A car..." & \scriptsize ``A bear..." & \scriptsize ``A circus tent..."& \scriptsize ``A tiger..." \\
        
        \begin{minipage}{0.11\textwidth}
            \includegraphics[width=\textwidth]{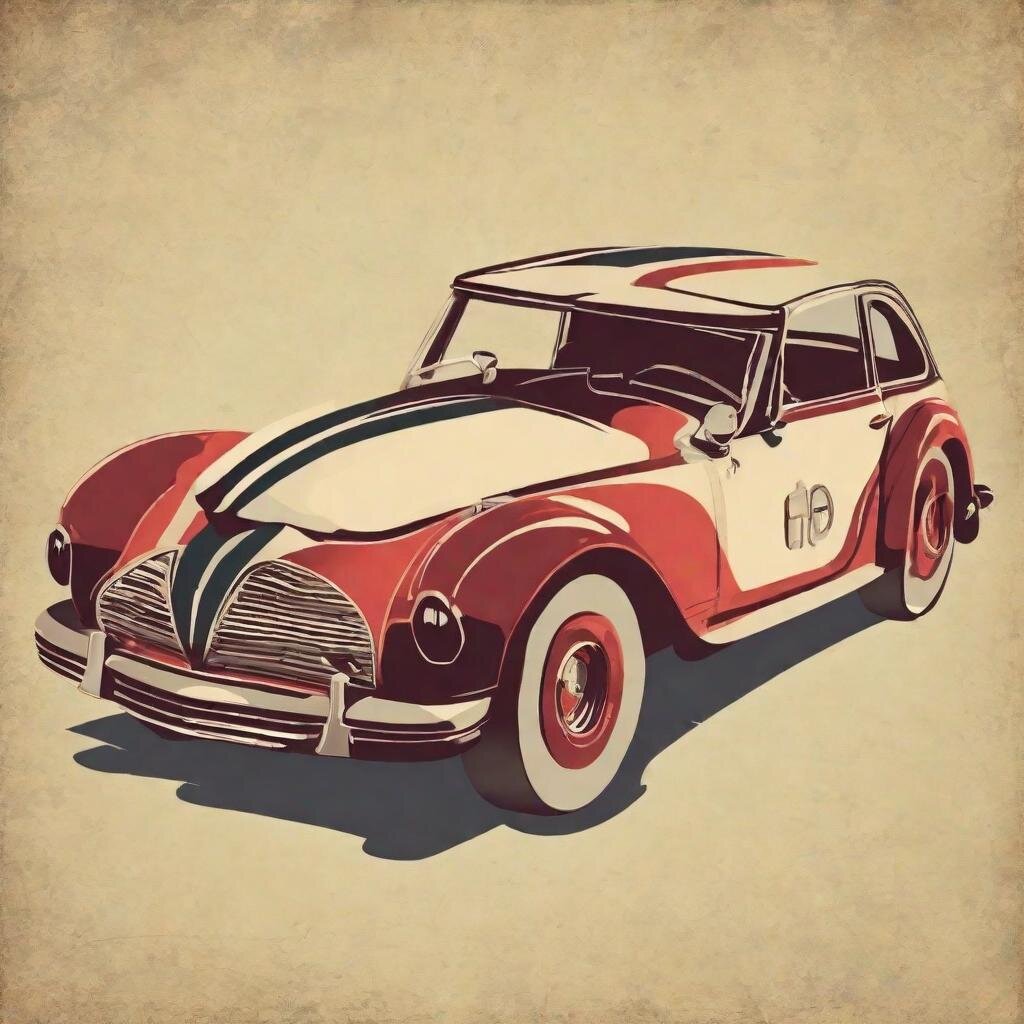}
        \end{minipage} &
        \begin{minipage}{0.11\textwidth}
            \includegraphics[width=\textwidth]{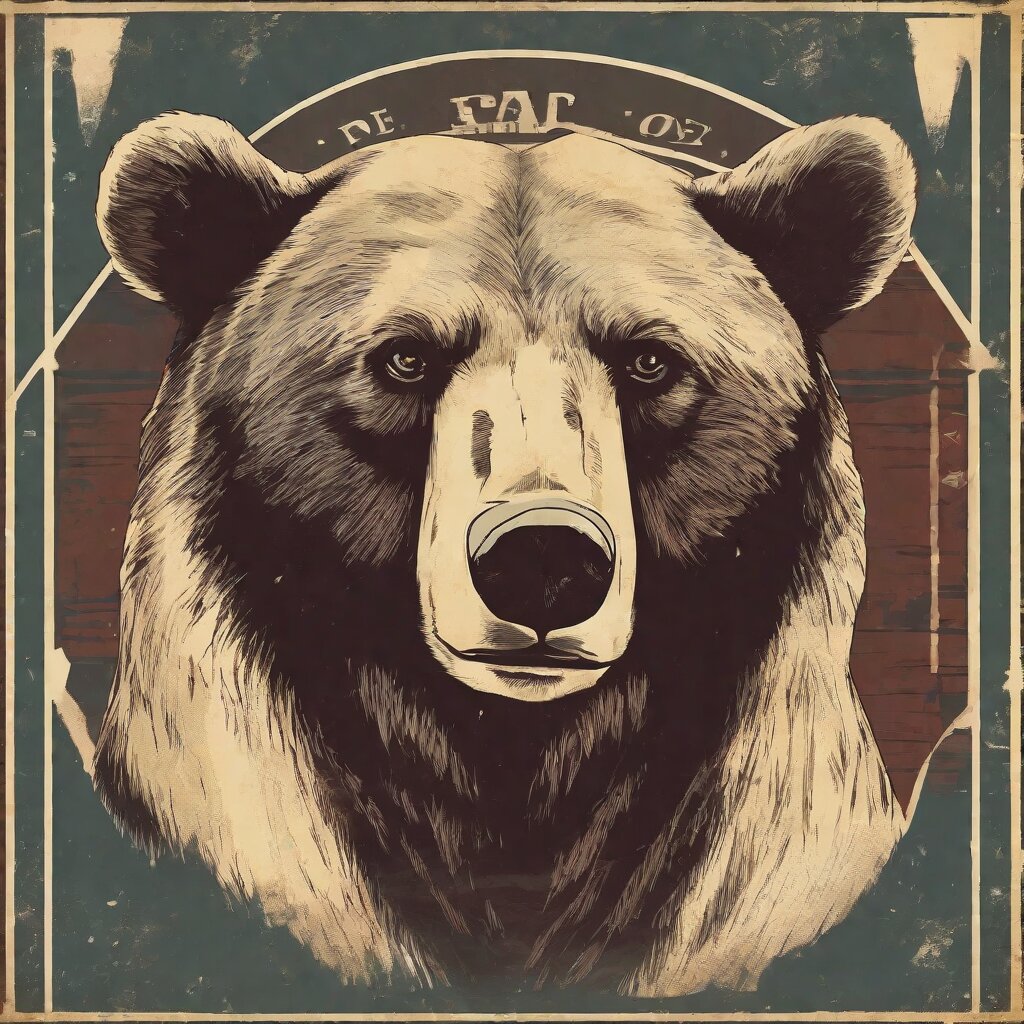}
        \end{minipage} &
        \begin{minipage}{0.11\textwidth}
            \includegraphics[width=\textwidth]{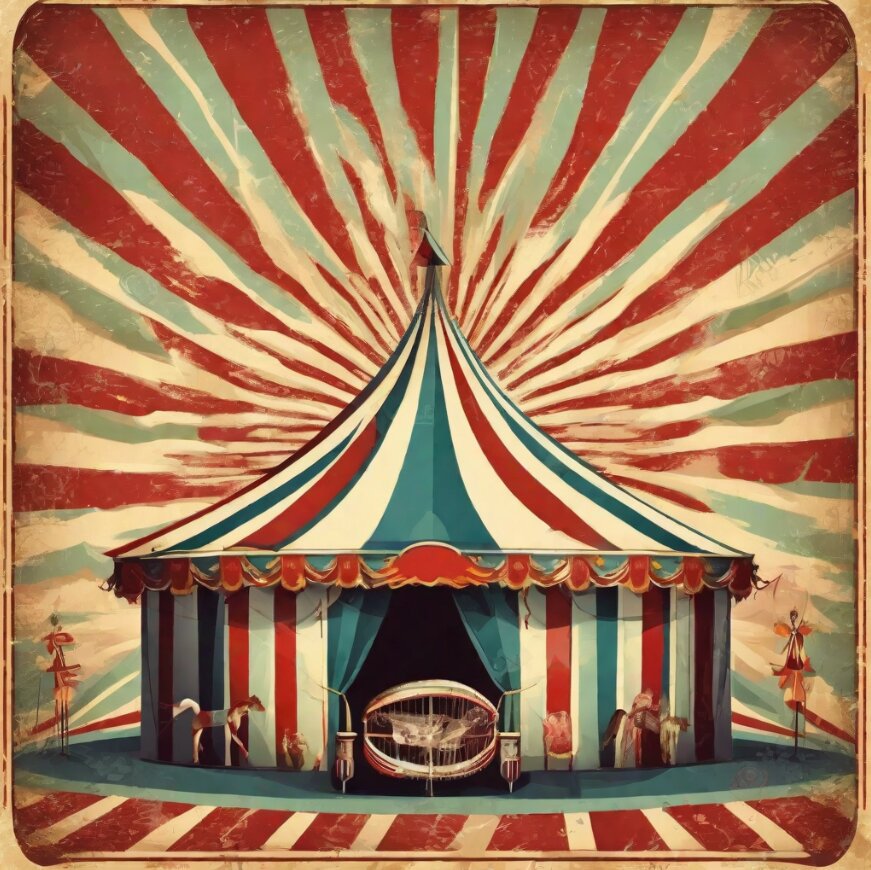}
        \end{minipage} &
        \begin{minipage}{0.11\textwidth}
            \includegraphics[width=\textwidth]{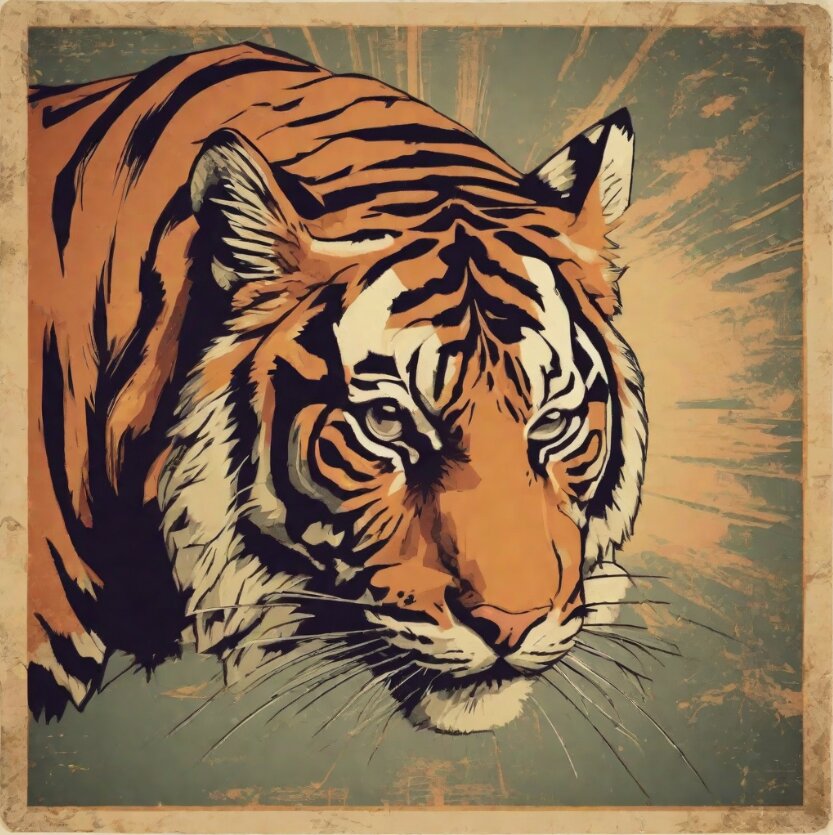}
        \end{minipage} 
    \end{tabular}

    \vspace{.1cm}

    \begin{tabular}{c@{\hspace{.1cm}}c@{\hspace{.1cm}}c@{\hspace{.1cm}}c}
        
        \begin{minipage}{0.11\textwidth}
            \centering
            \scriptsize Initial stylistic \\ 
            \centering
            alignment
        \end{minipage} &
        \begin{minipage}{0.11\textwidth}
            \includegraphics[width=\textwidth]{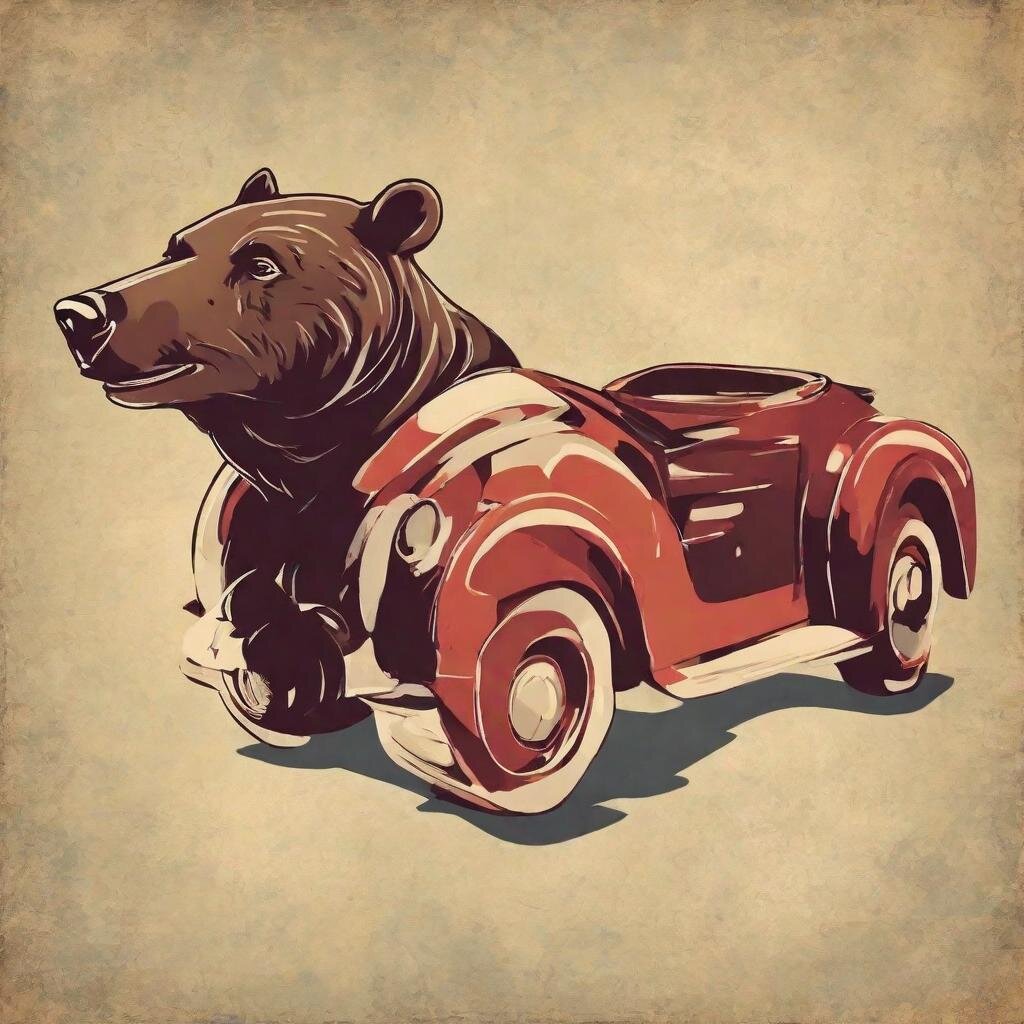}
        \end{minipage} &
        \begin{minipage}{0.11\textwidth}
            \includegraphics[width=\textwidth]{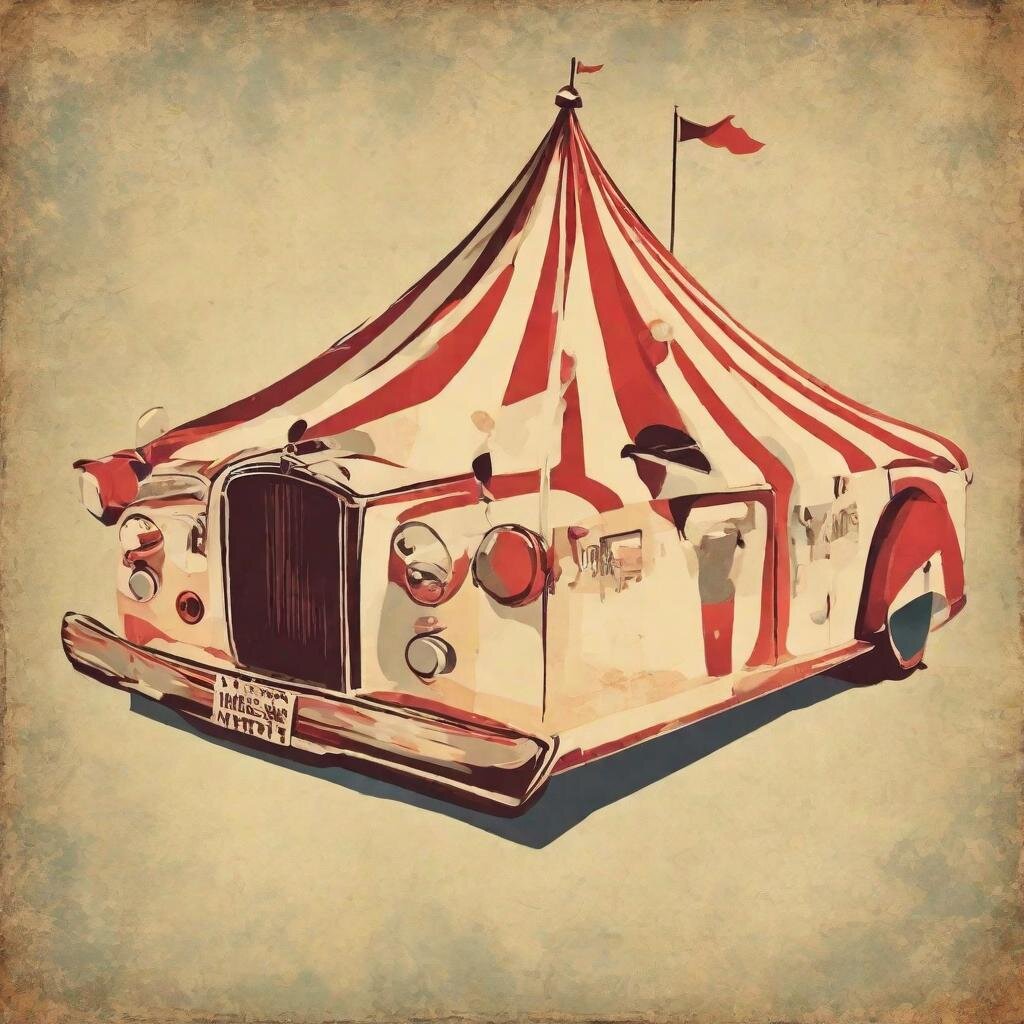}
        \end{minipage} &
        \begin{minipage}{0.11\textwidth}
            \includegraphics[width=\textwidth]{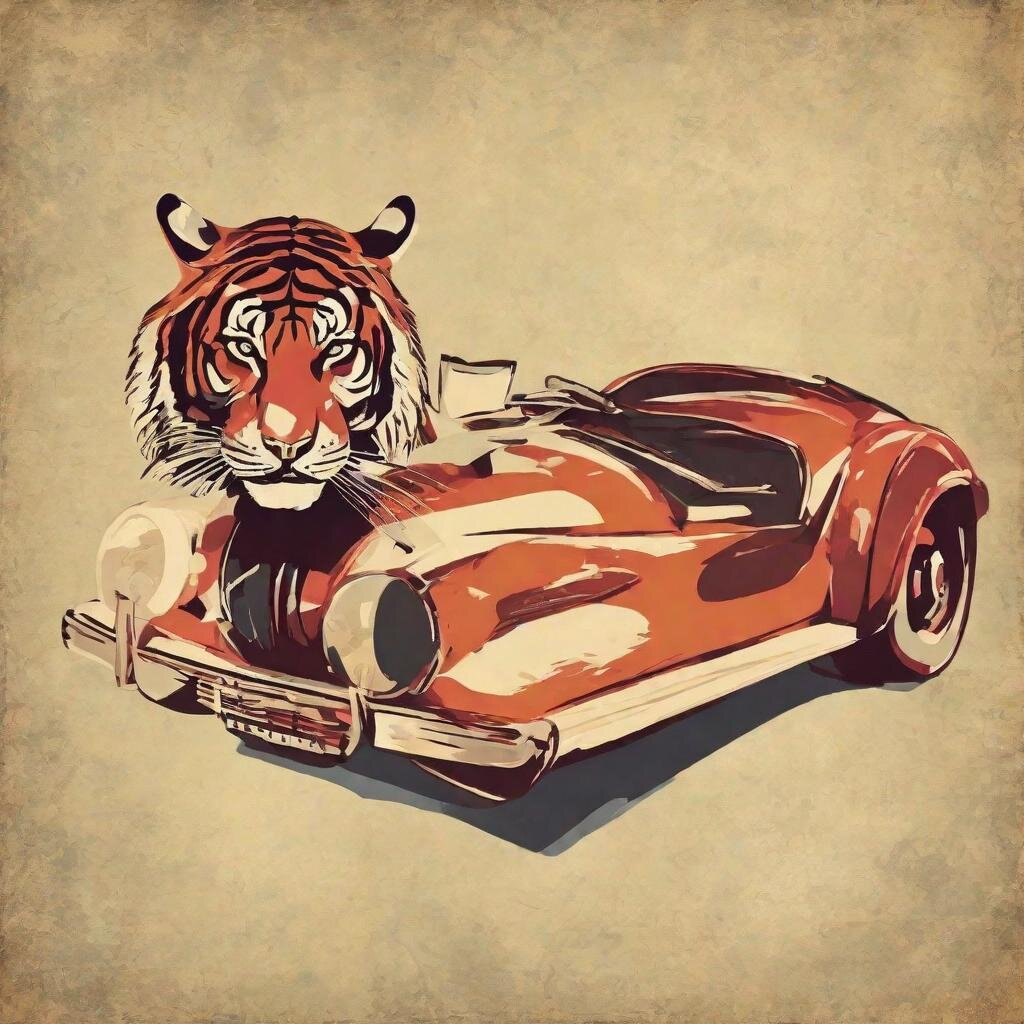}
        \end{minipage} 
    \end{tabular}

    \vspace{.1cm}

    \begin{tabular}{c@{\hspace{.1cm}}c@{\hspace{.1cm}}c@{\hspace{.1cm}}c}  
        \begin{minipage}{0.11\textwidth}
            \centering
            \scriptsize Content leakage
        \end{minipage} &
        \begin{minipage}{0.11\textwidth}
            \includegraphics[width=\textwidth]{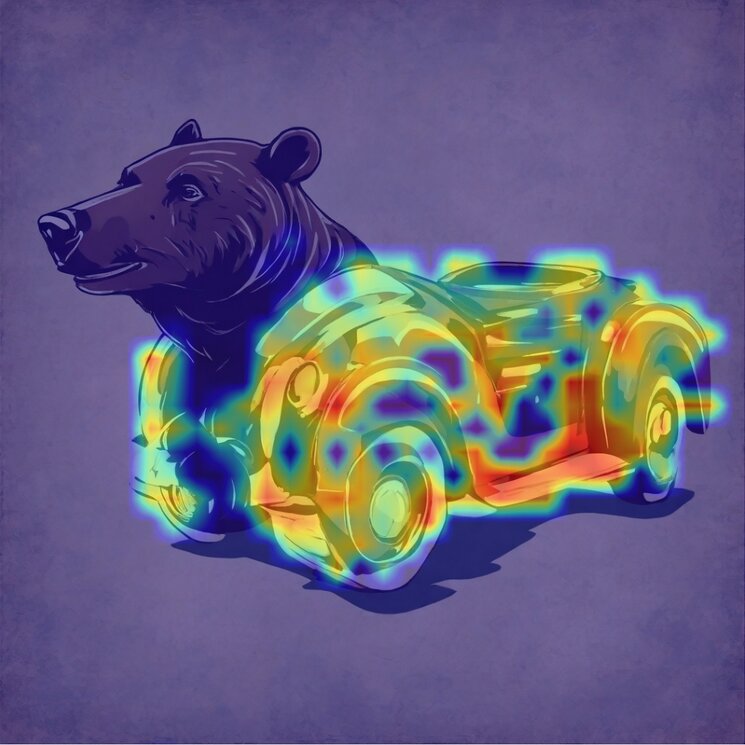}
        \end{minipage} &
        \begin{minipage}{0.11\textwidth}
            \includegraphics[width=\textwidth]{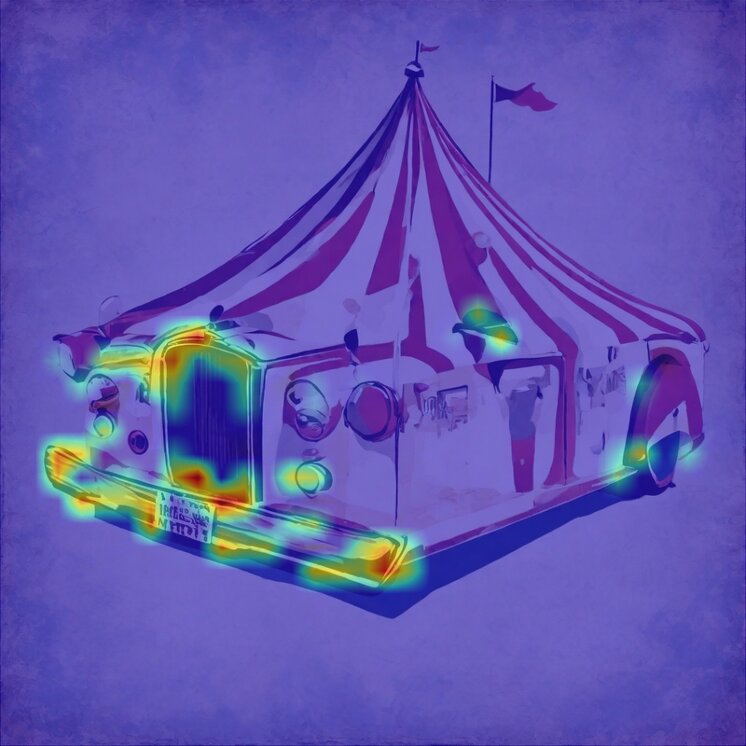}
        \end{minipage} &
        \begin{minipage}{0.11\textwidth}
            \includegraphics[width=\textwidth]{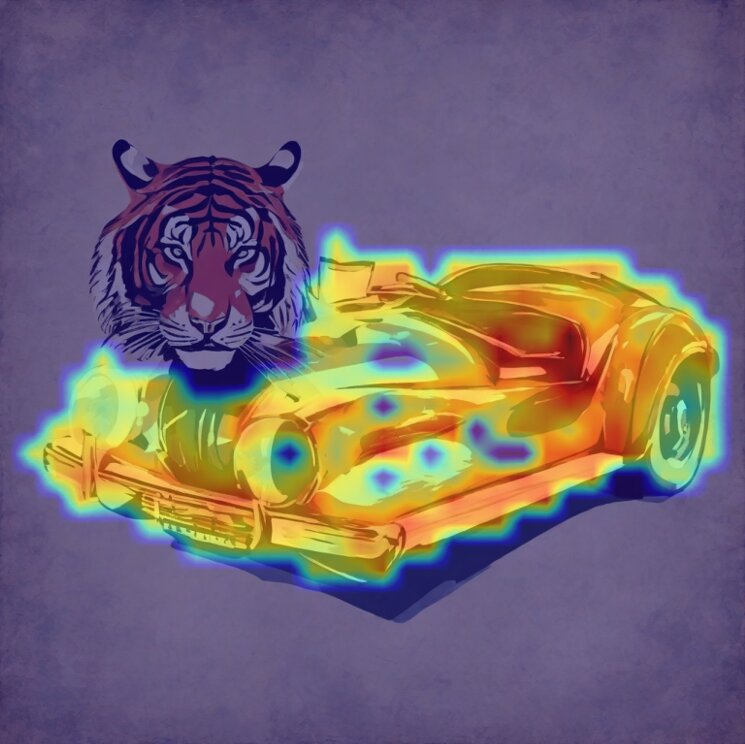}
        \end{minipage} 
    \end{tabular}

    \vspace{.1cm}

    \begin{tabular}{c@{\hspace{.1cm}}c@{\hspace{.1cm}}c@{\hspace{.1cm}}c}
        \begin{minipage}{0.11\textwidth}
            \centering
            \scriptsize \emph{Only-Style}
        \end{minipage} &
        \begin{minipage}{0.11\textwidth}
            \includegraphics[width=\textwidth]{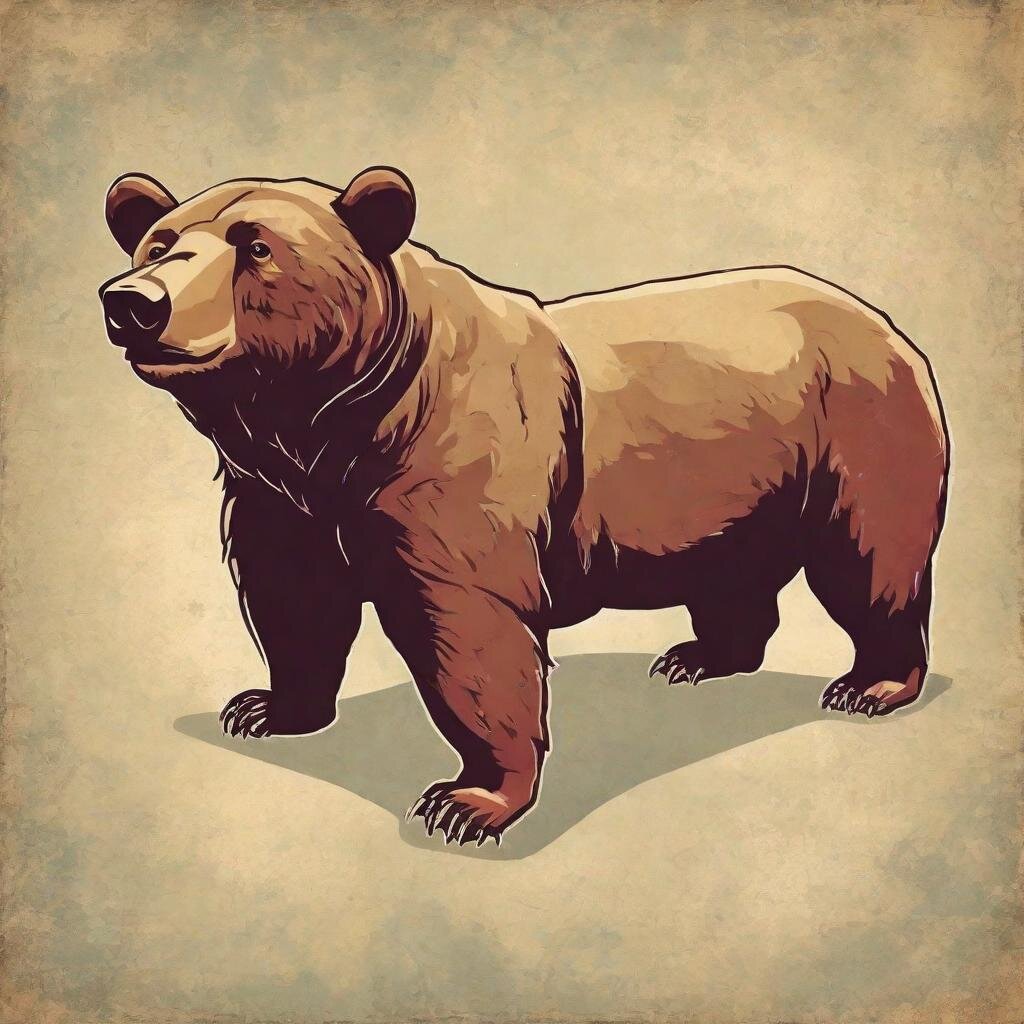}
        \end{minipage} &
        \begin{minipage}{0.11\textwidth}
            \includegraphics[width=\textwidth]{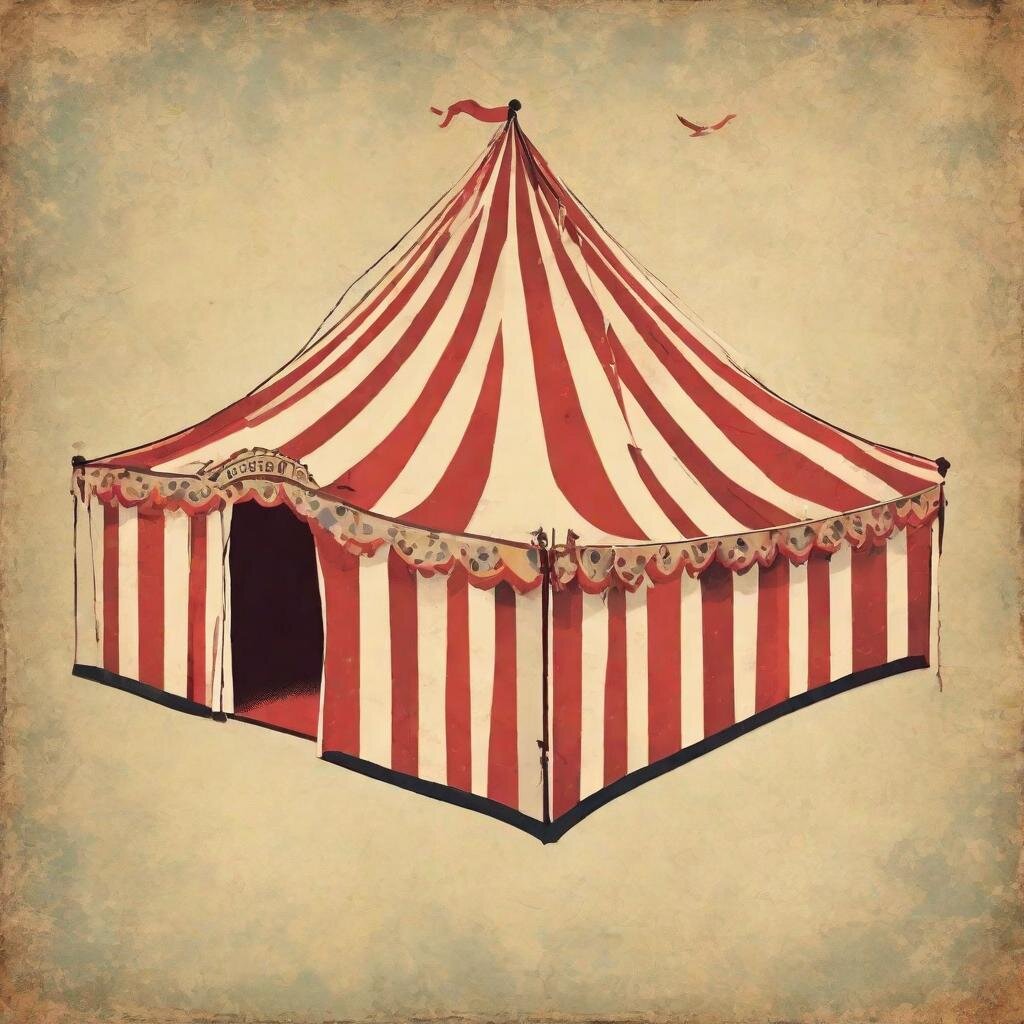}
        \end{minipage} &
        \begin{minipage}{0.11\textwidth}
            \includegraphics[width=\textwidth]{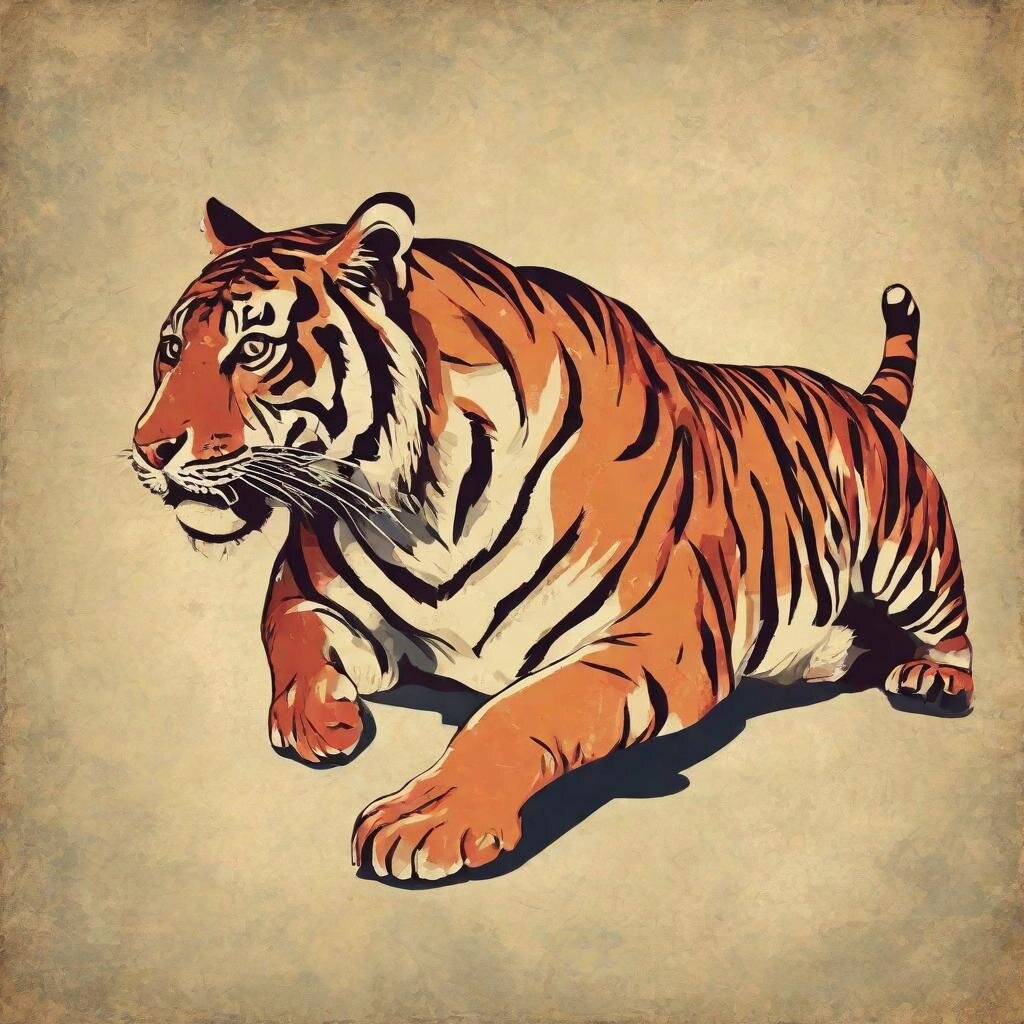}
        \end{minipage}
    \end{tabular}
    \vspace{-.1cm}
    \caption[Only-Style]{\textbf{Only-Style}: The top row shows images independently generated by a text-to-image model using the style descriptor \textit{``in vintage poster style"}. Applying a state-of-the-art method (here StyleAligned \cite{hertz2024style}) to align these images stylistically with the first image (the car) leads to unintended content leakage, causing visual elements of the car to infiltrate the other images. \emph{Only-Style} addresses this by first localizing the semantic content of the reference subject in the target images (third row) and then guiding the alignment process to eliminate this undesired effect (fourth row).}
    \label{fig:Cover}
    \vspace{-0.3cm}
\end{figure}

State-of-the-art text-to-image (T2I) models \cite{ramesh2021zeroshot, chang2023muse, podell2023sdxlimprovinglatentdiffusion, saharia2022photorealistic, pmlr-v235-esser24a} demonstrate impressive results in transforming text into compelling visual outputs. However, such models do not hand the user control over specific visual stylistic results, often producing widely varying interpretations of the same textual descriptor, as shown in the first row of Fig. \ref{fig:Cover}.  

For this reason, several works aim to provide visual thematic consistency across different generated concepts, as they were created or performed in the same manner or technique (e.g., from the same artist). 
Even though these methods achieve the desired stylistic alignment of a reference image with a target one, they frequently exhibit content leakage. In other words, unintended semantic elements from the reference image subject, appear in the target image. Such cases are evident in the second row of Fig.~\ref{fig:Cover}. 
 
The main motivation of this work is to find a semantically meaningful way to remove this content leakage while preserving consistency in style. 
To achieve this, we introduce a method to control the transfer of the reference subject patches in the target image, through a simple scaling of their representations.  
This scaling operation is adaptively tuned via the localization of the reference subject in the target image to determine whether the transfer of the reference subject features should be further restrained (see third row of Fig.~\ref{fig:Cover}). 
The final stylistic alignment is illustrated in the bottom row of Fig.~\ref{fig:Cover}. As it can be observed, the resulting target images exhibit no content leakage - thus we dubbed our method \emph{Only-Style}. 

Additionally, we present two novel evaluation methods to measure the impact of content leakage in style consistent image generation, both coarsely via an encoder-based metric (CL) and fine-grained via Large Vision-Language Models (LVLMs). The former quantifies the semantic correlation between the target image and the reference subject, thereby assessing content leakage, while the latter detects even subtle leakage cases by prompting LVLMs.
We benchmark many state-of-the-art methods, showing that content leakage is a main challenge shared across all of them.
To our knowledge, although content leakage has been recognized as an issue in stylistic consistency, no prior work has proposed a way to measure the appearance of content leakage cases — a very useful tool to understand the efficacy of methods that promote stylistic consistency. 

Our main contributions are:
$\bullet$ 
We introduce fine-grained control over the leakage of reference semantic elements to the target image, tailored for attention sharing approaches~\cite{hertz2024style}.  
$\bullet$ 
We propose a novel method for subtle leakage localization, applicable to any style consistent generation approach.
$\bullet$ We design an end-to-end method that achieves style-consistent image generation by adaptively scaling down the contribution of the reference subject, addressing the problem of content leakage.
$\bullet$ We propose an evaluation framework that quantifies content leakage using two distinct metrics (coarse and fine-grained version).  We release this benchmark to serve as a standardized baseline, addressing a notable gap in the community and facilitating more consistent comparisons of style alignment methods.

\section{Related Work}
\label{sec:related}

\textbf{Text-to-Image diffusion models.} Diffusion models \cite{song2021scorebased, song2019generative, sohldickstein2015deep, ho2020denoisingdiffusionprobabilisticmodels} have transformed the field of image generation, producing highly diverse and visually striking outputs. Further, text-conditioned diffusion models \cite{nichol2021glide, rombach2022highresolutionimagesynthesislatent} enable the generation process to be guided by natural language prompts, leveraging these powerful generative capabilities.

\vspace{2pt}
\noindent\textbf{Controlling the attention in T2I diffusion models.} The attention mechanism is the common underlying ingredient within neural network backbones in T2I diffusion models. Recent works have explored how the self-attention and cross-attention layers can be harnessed to define both the layout and semantic content of text-generated images \cite{hertz2022prompttoprompt, chefer2023attendandexcite, patashnik2023localizingobjectlevelshapevariations, 10.1145/3658157}. Additionally, attention mechanisms have been widely applied for editing text-generated images \cite{cao2023masactrl, mokady2023nulltext, NEURIPS2023_3469b211, tumanyan2023plugandplay, patashnik2024consolidatingattentionfeaturesmultiview}. Building on insights from these methods, we utilize the attention layers to disentangle content and style, addressing the challenge of style-consistent generation.

\vspace{2pt}
\noindent\textbf{Style Transfer.} Style transfer is a long-standing challenge in computer vision \cite{Hertzmann2001ImageAnalogies, Efros2001ImageQuilting} that refers to the process of transforming the visual style of an input image while preserving its content. Neural Style Transfer leverages deep features from pretrained networks to alter the style of a target image based on a reference \cite{Jing2019NeuralStyleTransfer, Gatys2016ImageStyleTransfer}. Moreover, GAN-based techniques have been developed to transfer images across different stylistic domains \cite{Isola2017ImageToImage, Park2020ContrastiveLearning, Katzir2020CrossDomainCascaded, Zhu2017CycleGAN}.

\vspace{2pt}
\noindent\textbf{Consistent style Generation.} With the advent of diffusion models stylization research has focused on generating target text-specified concepts using either real or synthesized stylistic reference images. Different approaches have emerged to tackle this task:
$\bullet$
One family of approaches involves training diffusion models to incorporate conditioning from the output representation of a pretrained image encoder\cite{Ye2023IPAdapter, Wang2024StyleAdapter, Wang2024InstantStyle, Xing2024CSGO}, such as CLIP\cite{radford2021learningtransferablevisualmodels}. However these methods require significant computational resources to train this conditioning and tend to drive the model away from its training distribution. Following this paradigm and conceptually close to our work, InstantStyle~\cite{Wang2024InstantStyle}, in a coarse attempt to reduce content leakage, injects the CLIP image embedding of the stylistic reference, subtracted by the CLIP text embedding of the reference subject, into specific blocks within the diffusion model. 
$\bullet$
Another line of recent works developed optimization techniques over one or more images that let the model capture certain visual features \cite{gal2022image, kumari2023multiconcept, 10377873, ruiz2023dreamboothfinetuningtexttoimage, sohn2023styledrop, frenkel2024blora} such as a style interpretation. For example B-LoRA \cite{frenkel2024blora} trains specific LoRAs within the diffusion backbone to capture separately the content and style of an image. 
$\bullet$
Closer to our work, to circumvent the computationally intensive pretraining or fine-tuning process per instance, recent approaches utilize self-attention layers of the model's backbone to allow communication between images within a batch, and thus the transfer of stylistic features from a single reference to other images \cite{hertz2024style, jeong2024visual}. Building upon these state-of-the-art approaches, our method addresses the persistent challenge of content leakage.
\section{Proposed Method: Only-Style}

\subsection{Overview of Only-Style}

In the forthcoming analysis, we consider the following setup\footnote{Our method can be easily extended to support multi-image and multi-subject generation (see Suppl. Mat. and Fig.~\ref{fig:Qualitative_ours})}: $\bullet$ The goal is the generation of two images $I_{ref}$ and $I_{tgt}$ with visually ``aligned" style, but driven from different prompts $P_{ref}$ and $P_{tgt}$.
$\bullet$ The considered prompts have a specific structure of \textit{\{subject\} + \{style\}}. 
Specifically, we have the textual descriptions of the reference subject $S_{ref}$ (e.g., ``a cat") and the target subject $S_{tgt}$ (e.g., ``a train"), which are combined with the desired stylistic description $P_{stl}$ (e.g., ``in realistic 3D render").

\begin{figure}[t]
    \centering
    \includegraphics[width=1.0\linewidth]{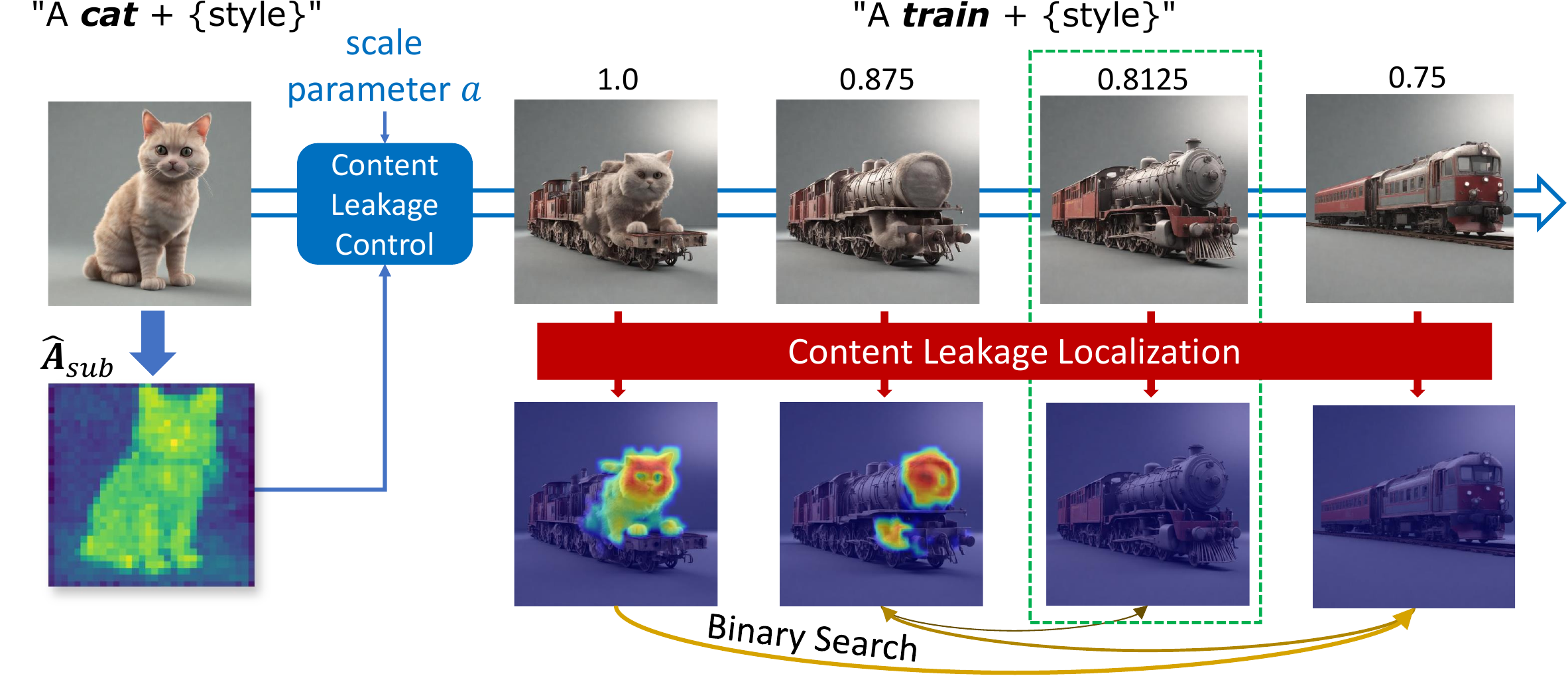}
    \vspace{-0.4cm}
    \caption[Overview]{\textbf{Overview of \emph{Only-Style}:} By localizing semantic content leakage, we adaptively tune a scaling parameter $\alpha$ that controls style sharing in image patches containing the reference subject (subject map $\hat{\mathbf{A}}_{sub}$), resulting in the optimal value that eliminates leakage while preserving stylistic consistency.}
    \label{fig:Overview}
    \vspace{-0.2cm}
\end{figure}

Content leakage occurs when attributes of $S_{ref}$ visually ``leak" into patches of \(I_{tgt}\) leading to unwanted semantic content overlap.
The core idea behind \emph{Only-Style} is to detect these image patches associated with the reference subject and \emph{adaptively} reduce their contribution to the shared style generation, as shown in Fig.~\ref{fig:Overview}. 
The algorithm consists of three main steps that are performed at inference time:

\begin{itemize}
\item 
\emph{Content Leakage Control} (Sec. \ref{sec:style-control}): First, we identify the patches in $I_{ref}$ that are relevant to the reference subject. 
This way, their contribution to the shared style generation can be reduced by simply scaling them down. 
    
\item 
\emph{Content Leakage Localization}  (Sec. \ref{sec:Content leakage loc}):
Next, we detect the patches in $I_{tgt}$ that are more relevant to $S_{ref}$ than to $S_{tgt}$, denoting content leakage. 
     
\item 
\emph{Adaptive Scaling} (Sec. \ref{sec:Optimal_scaling}):
Finally, we combine the steps above to determine the optimal scaling, eliminating content leakage while retaining the stylistic alignment, via a binary search process.
\end{itemize}

\subsection{Preliminaries}

\textbf{Attention in T2I Diffusion Models.} State-of-the-art T2I diffusion models \cite{chang2023muse, podell2023sdxlimprovinglatentdiffusion, saharia2022photorealistic} typically use a U-Net \cite{ronneberger2015unet} architecture as the backbone\footnote{some models, such as \cite{pmlr-v235-esser24a}, use a Transformer \cite{vaswani2023attention} backbone - which is in line with the proposed framework since it relies on attention layers}. 
These image-to-image architectures are augmented with transformer blocks, each one of them consisting of a \emph{self-attention layer} followed by a \emph{cross-attention} layer. The latter contextualizes the deep image features with the text token embeddings. The proposed method is employed on these transformer blocks.

Following the typical attention layer conventions~\cite{vaswani2023attention}, deep features are projected into queries $\mathbf{Q} \in \mathbb{R}^{n \times d_k}$, keys $\mathbf{K} \in \mathbb{R}^{m \times d_k}$, and values $\mathbf{V} \in \mathbb{R}^{m \times d_v}$. The output of the attention layer is computed as:
\begin{equation} \text{Attention}(\mathbf{Q}, \mathbf{K}, \mathbf{V}) = \text{softmax}\left( \frac{\mathbf{Q} \mathbf{K}^\top}{\sqrt{d_k}}\right) \mathbf{V} = \mathbf{A} \mathbf{V}, 
\label{eq:att}
\end{equation}
where $\mathbf{A} \in \mathbb{R}^{n \times m}$ is the output of the softmax operator, referred to as the \emph{attention probabilities} and essentially describing the correlation between $\mathbf{Q}$ and $\mathbf{K}$. In self-attention, $\mathbf{Q}$, $\mathbf{K}$, and $\mathbf{V}$ are all derived from the same image features $\mathbf{F}$, while in cross-attention, $\mathbf{Q}$ comes from the image features and $\mathbf{K}$, $\mathbf{V}$ come from the text token embeddings.


\subsection{Content Leakage Control}
\label{sec:style-control}

The first step in mitigating content leakage is to regulate the influence of the reference image. Most existing approaches incorporate hyperparameters that serve this purpose (see Fig.\ref{fig:Generalization} and the relevant discussion).
Aiming for a finer control, \emph{Only-Style}  builds upon an \emph{attention-based} method that achieves consistent style generation, StyleAligned~\cite{hertz2024style}. 
Our objective is to leverage the attention mechanism to selectively scale down reference subject patches while maintaining style consistency, even under scaling adjustments.

Specifically, 
this control module consists of two steps: \textbf{1)} detection of image patches in $I_{ref}$ that visually correspond to the subject $S_{ref}$, and \textbf{2)} scaling down the contribution of these subject patches only, according to a given scale parameter $a$. The first step requires the inference of the reference image $I_{ref}$ according to the reference prompt $P_{ref} = S_{ref} + P_{stl}$, and is performed within the \emph{cross-attention} layers of the transformer blocks. The second step is then performed on the \emph{self-attention layers}, by scaling down the reference keys $\mathbf{K}_{ref}$ of the shared attention mechanism employed in~\cite{hertz2024style} (see suppl. material) during the generation of $I_{tgt}$.

\textbf{Detecting Subject Patches.}
As evident in Fig.~\ref{fig:Cover}, 
content leakage in $I_{tgt}$ originates from the transfer of patches in $I_{ref}$ that are semantically close to the reference subject $S_{ref}$.
Our key observation is that cross-attention layers, which serve as a semantic explanation in T2I models\cite{hertz2022prompttoprompt}, can be leveraged to annotate these patches in $I_{ref}$.

Thus, to access and control the ``leakage" of a subject, an aggregated attention visual map $\mathbf{a}_{sub} \in \mathbb{R}^{n}
\equiv \hat{\mathbf{A}}_{sub} \in \mathbb{R}^{H \times W}$ is required, referred to as \emph{subject map}, where $\mathbf{a}_{sub}$ is the flattened version across all the patches, while $\hat{\mathbf{A}}_{sub}$ is reshaped to match the image’s spatial structure.
Specifically, we use the cross attention probabilities ($\mathbf{A}^{l, t}$, see Eq.~\ref{eq:att}) at iteration $t$ and in the layer $l$. 
The considered layers are the bottleneck layers of the U-Net backbone, 
known to contain rich semantic information~\cite{kwon2023diffusionmodelssemanticlatent, patashnik2023localizingobjectlevelshapevariations}. 

Given a set of bottleneck attention layers $B$ and iterations $t\in[1, T]$, we compute the averaged cross attention probabilities 
with respect to the subject token $S_{ref}$ as:
\begin{equation}
\mathbf{a}_{sub} = \frac{1}{T|B|} \Bigl( \sum_{t} \sum_{l \in B} \mathbf{A}^{l, t}\Bigr)  \mathbf{e}_s 
\end{equation}
where $\mathbf{e}_s$ isolates the column corresponding to the subject.

The output of this step is a binary mask $\mathbf{R} \in \mathbb{R}^{H \times W}$, identifying patches relevant to the subject.
Thus, given the subject map $\hat{\mathbf{A}}_{sub}$, we aim to separate the patches content-related patches (\textit{source of leakage}) from unrelated ones.
Since it is impossible to a priori specify a good thresholding value across all cases (slightly different text prompts lead to different attention values, even for the same subject token), we perform the separation via a K-means clustering method with two centroids. See Suppl. Mat. for details.

\textbf{Controlling Content Leakage.} 
Following ~\cite{hertz2024style}, content leakage can be mitigated by reducing the contribution of reference key features $\mathbf{K}_{ref}$ in shared self-attention layers.

However, scaling all the patches is not optimal, since style contribution can be affected too (see ablation on Suppl. Mat.).
Instead, we selectively scale only the key features of ``content patches," as determined by the subject mask $\mathbf{R}$.
Using a single scalar parameter $\alpha \in [0, 1]$, we scale the key features at each iteration $t$ and layer $l \in B$ as:

\begin{equation}
    \hat{\mathbf{K}}_{ref} = (1 - \mathbf{R}) \odot \mathbf{K}_{ref} + \alpha \mathbf{R} \odot \mathbf{K}_{ref}
    \label{eq:scaling}
\end{equation}

Intuitively, by reducing $\alpha$, this weighting makes the attention distribution on the reference subject patches more uniform, resulting in a global stylistic alignment rather than a polarised local ``semantic" transfer. 
As shown in Fig.~\ref{fig:Overview}, decreasing $\alpha$ progressively reduces content leakage in a semantically explainable manner.

\subsection{Content Leakage Localization}
\label{sec:Content leakage loc}

Reducing the influence of reference subject patches may lead to stylistic misalignment, as shown in Fig.~\ref{fig:Ablation_2}. This necessitates finding a scaling parameter high enough for accurate style transfer yet low enough to minimize content leakage. To achieve this, we must measure content leakage in target images to establish a lower bound for scaling.

To this end, we introduce a patch-level content leakage localization method at inference, applied in two consecutive diffusion iterations. Our method relies on a simple premise: 
\emph{determining whether a target image patch contains more information about the reference than the target subject.}
To implement this, we need to define the following: \textbf{1)} how to extract robust and faithful representations $\mathbf{v}$ for both subjects, \textbf{2)} how to use them to detect leakages.

\textbf{Extracting Subject Representations.} 
The CLIP token embeddings that guide the generation via the cross-attention, are not expressive enough to localize subtle reference subject that overlap with the target. 
To improve this, we use cross-attention maps $\hat{\mathbf{A}}_{sub}$ (Sec.~\ref{sec:style-control}), averaged on a single iteration, to pool one representation per subject before each self-attention layer. 
Since directly using $\hat{\mathbf{A}}_{sub}$ does not reliably localize the most relevant features to the subject, we refine this via clustering and percentile thresholding, forming a binary mask $\mathbf{M}_{sub}$ of subject-relevant patches\footnote{Check Suppl. Mat. for more details on this step.}. 
The refined subject-relevant attention map $\Tilde{\mathbf{A}}_{sub}$ is then extracted as:

\vspace{-1pt}
\begin{align}
    \Tilde{\mathbf{A}}_{sub} &= (\mathbf{M}_{sub} \odot \hat{\mathbf{A}}_{sub}) / \sum (\mathbf{M}_{sub} \odot \hat{\mathbf{A}}_{sub}) 
\end{align}
\vspace{-1pt}

In the iteration following the extraction of $\Tilde{\mathbf{A}}_{sub}$, each layer $l$ uses the feature map $\mathbf{F}^{l}$ before the self-attention layer to extract a per-layer visual representation $\mathbf{v}^{l} \in R^d$ that best describes the subject in layer $l$:
\begin{align}
    \mathbf{v}_{sub}^{l} = \sum_i \sum_j [\mathbf{F}^{l} \odot \Tilde{\mathbf{A}}_{sub}]_{ij}
\end{align}
\vspace{-1pt}
\noindent This gives us two sets of representation vectors: $\{\mathbf{v}_{ref}^{l}\}$ corresponding to $S_{ref}$ in $I_{ref}$ and $\{\mathbf{v}_{tgt}^{l}\}$ for $S_{tgt}$ in $I_{tgt}$.

\begin{figure}[t]
    \centering
    \begin{subfigure}{0.11\textwidth}
        \centering
        {\tiny
        $I_{tgt}$
        }\\ 
        \includegraphics[width=\linewidth]{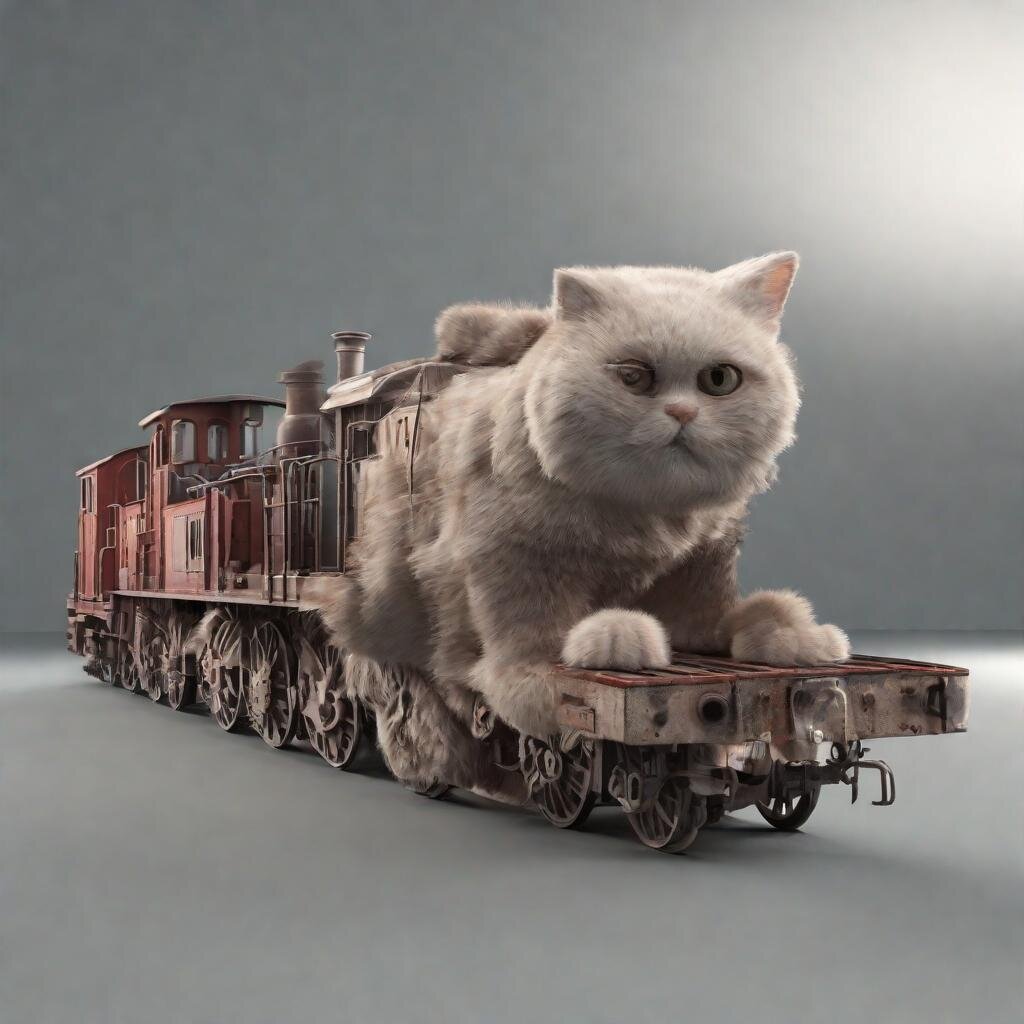}
    \end{subfigure}
    \hfill
    \begin{subfigure}{0.11\textwidth}
        \centering
        {\tiny
        $\mathbf{C}_{ref}$
        }\\ 
        \includegraphics[width=\linewidth]{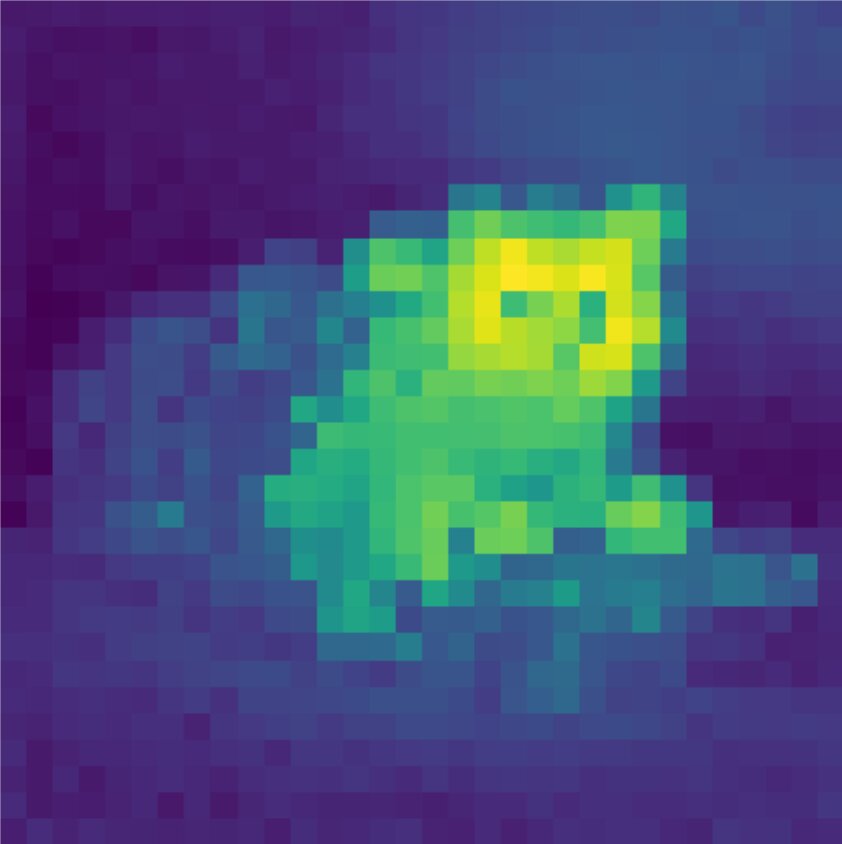}
    \end{subfigure}
    \hfill
    \begin{subfigure}{0.11\textwidth}
        \centering
        {\tiny
        $\mathbf{C}_{tgt}$ 
        }\\ 
        \includegraphics[width=\linewidth]{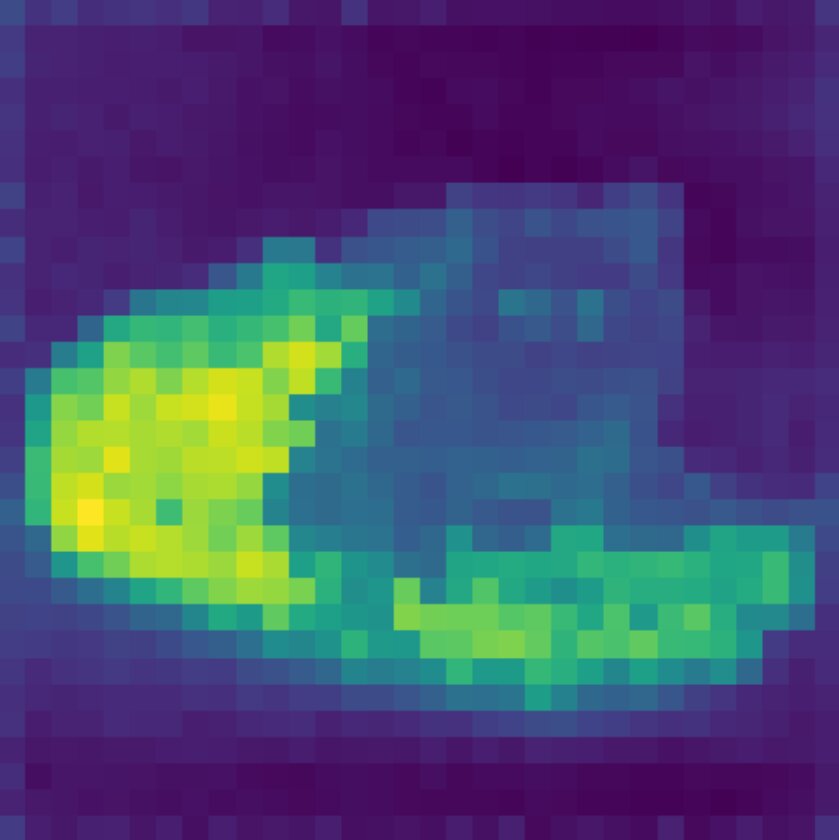}
    \end{subfigure}
    \hfill
    \begin{subfigure}{0.11\textwidth}
        \centering
        {\tiny
        $\mathbf{L} \odot (\mathbf{C}_{ref} - \mathbf{C}_{tgt})$
        }\\ 
        \includegraphics[width=\linewidth]{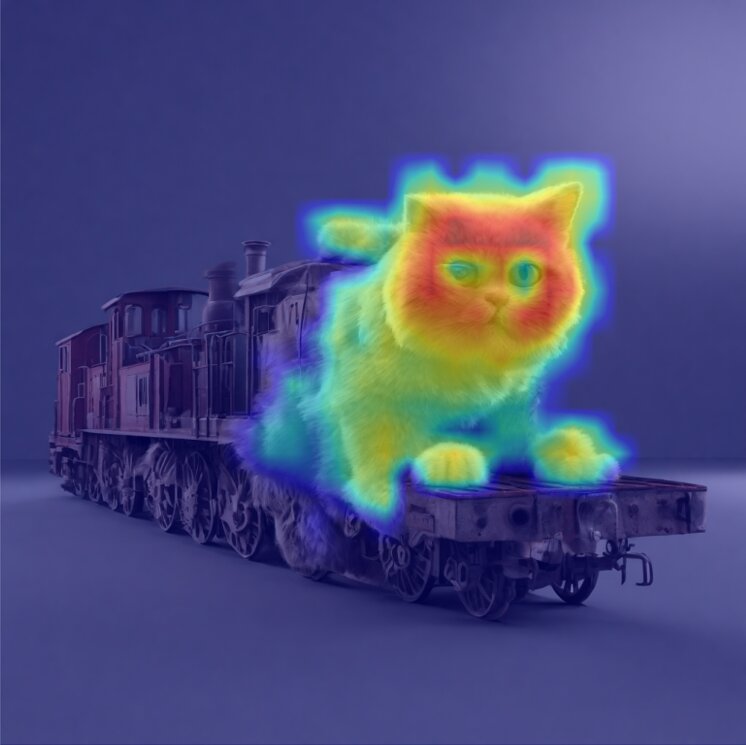}
    \end{subfigure}
    \caption{\textbf{Similarity maps} of the original reference subject (cat) and the target subject (train). By combining these maps we can effectively localize content leakage in the target image.}
    \label{fig:similarity maps}
    \vspace{-0.3cm}
\end{figure}

\textbf{Detecting Leakages.}
The semantic relevance of each patch in $I_{tgt}$ to the subject can then be computed using the cosine similarity scores between the target image features $\mathbf{F}_{tgt}^{l}$ and the subject representations $\mathbf{v}_{sub}^l$.
For each subject, we aggregate this similarity across bottleneck layers $\{B\}$, resulting in the similarity map $\mathbf{C}_{sub} \in \mathbb{R}^{H \times W}$. 
Formally:
\vspace{-1pt}
\begin{align}
    [\mathbf{C}_{sub}]_{ij} = \frac{1}{|B|} \sum_{l \in B} cos([\mathbf{F}_{tgt}^{l}]_{ij}, \mathbf{v}_{sub}^{l}),
\end{align}
\vspace{-1pt}
where $i \in [1, H]$ and $j \in [1, W$] denote spatial patch positions, and $cos$ is cosine similarity. Thus, we obtain the similarity map $\mathbf{C}_{ref}$ of $S_{ref}$ and $\mathbf{C}_{tgt}$ of $S_{tgt}$ (see  Fig.~\ref{fig:similarity maps}).

A patch $p_{ij}$ is marked as content leakage (binary value $L_{ij}$) if it contains more of $S_{ref}$ than $S_{tgt}$:

{
\begin{align}
    \begin{split}
      \mathbf{L}_{ij} = \bigl( [\mathbf{C}_{ref}]_{ij} \geq [\mathbf{C}_{tgt}]_{ij} + t_{leak} \bigr) \, \land \, \\ 
      \bigl( \left( [\mathbf{C}_{tgt}]_{ij} \geq t_{rel} \right) \lor \left( [\mathbf{C}_{ref}]_{ij} \geq t_{rel} \right) \bigr)
        \label{eq:patch_leakage}
    \end{split}
\end{align}
}

where $t_{leak}$ determines the minimum difference for leakage detection, and $t_{rel}$ filters out irrelevant/background patches.

Finally, we can obtain the overall leakage value as the logical addition of $\mathbf{L}_{ij}$: $L_{o} = \bigvee_{ij} \mathbf{L}_{ij}$. Both thresholds remain fixed across all experiments ($t_{leak}$=$0.1$ \& $t_{rel}$=$0.4$).

\begin{figure}[t]
    \centering
    \scriptsize
    \resizebox{.45\textwidth}{!}{
    \begin{tabular}{c@{\hspace{.1cm}}c@{\hspace{.1cm}}c@{\hspace{.1cm}}c}  
        Reference & Adaptive $\alpha$ & $\alpha = .9$ & $\alpha = .5$ \\

        \begin{minipage}{0.11\textwidth}
            \includegraphics[width=\textwidth]{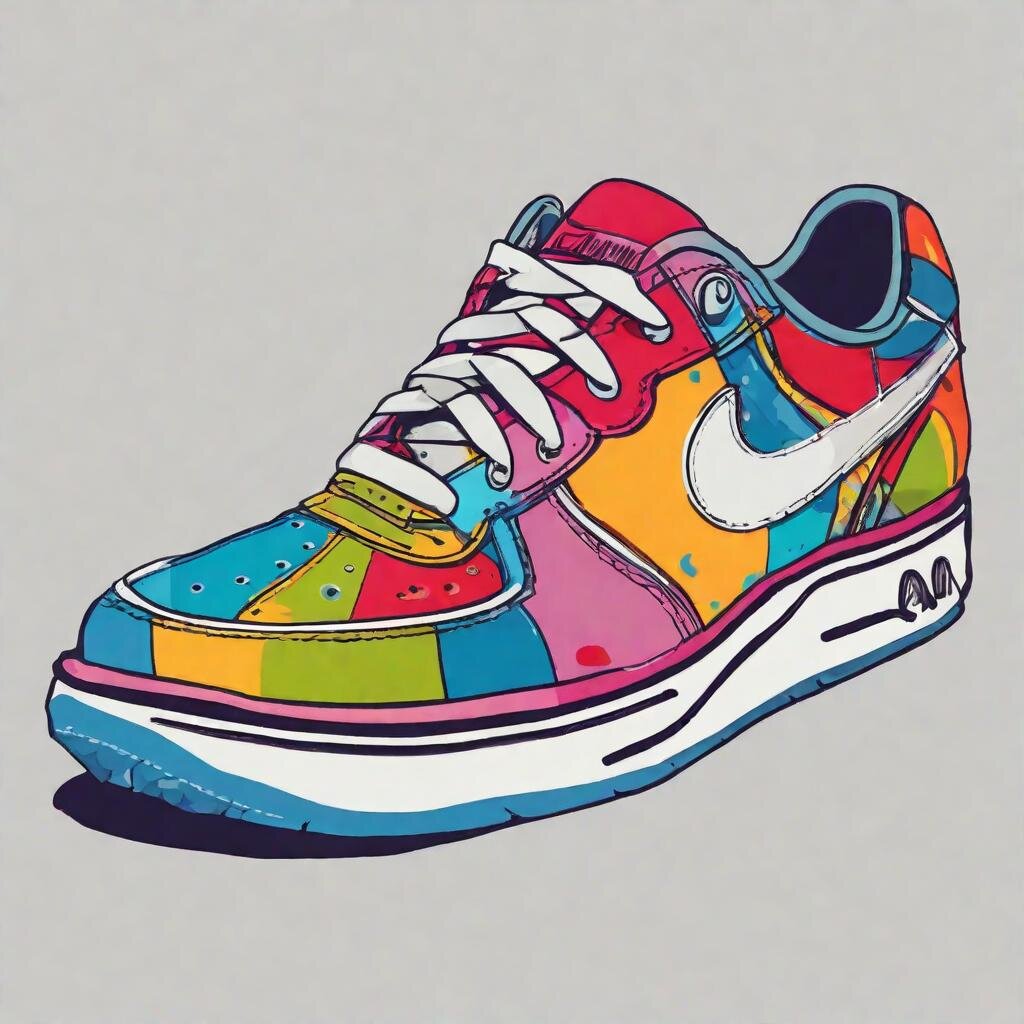}
        \end{minipage} &
        \begin{minipage}{0.11\textwidth}
            \includegraphics[width=\textwidth]{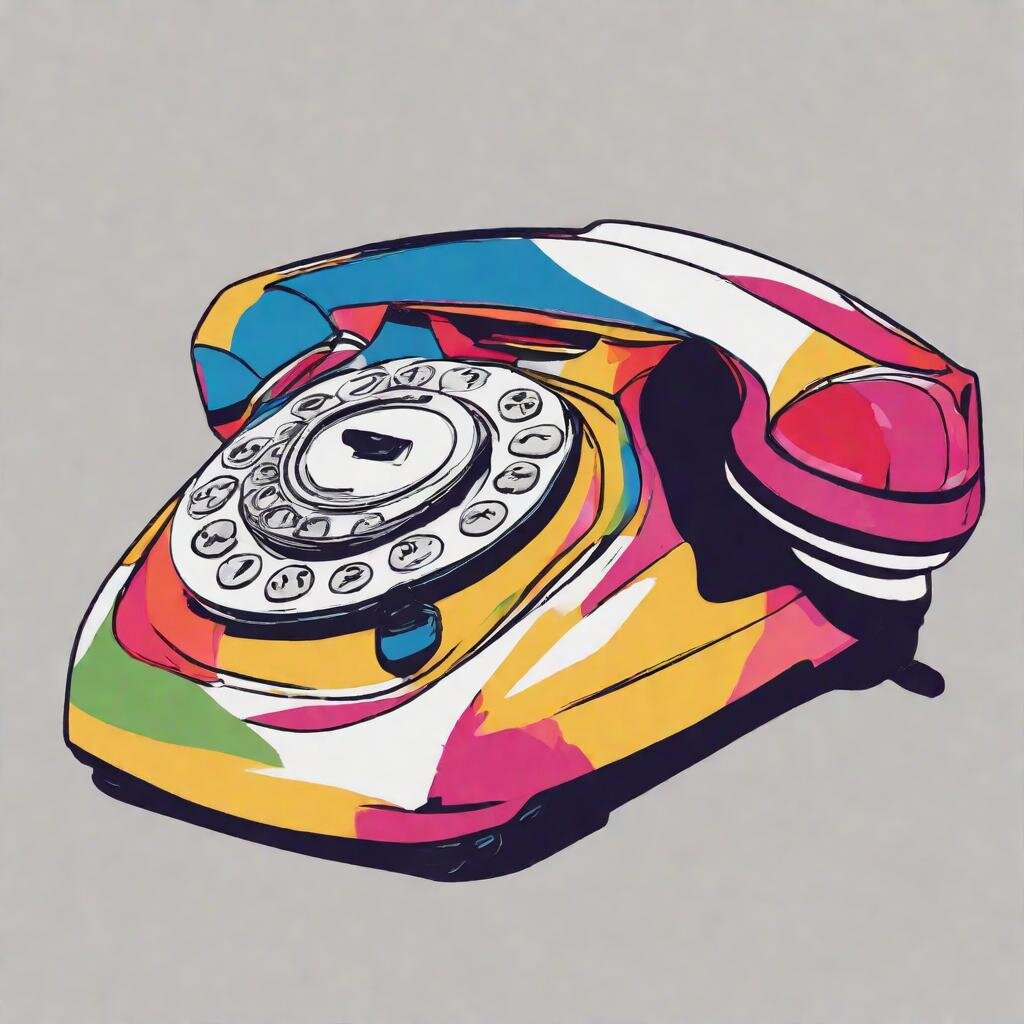}
        \end{minipage} &
        \begin{minipage}{0.11\textwidth}
            \includegraphics[width=\textwidth]{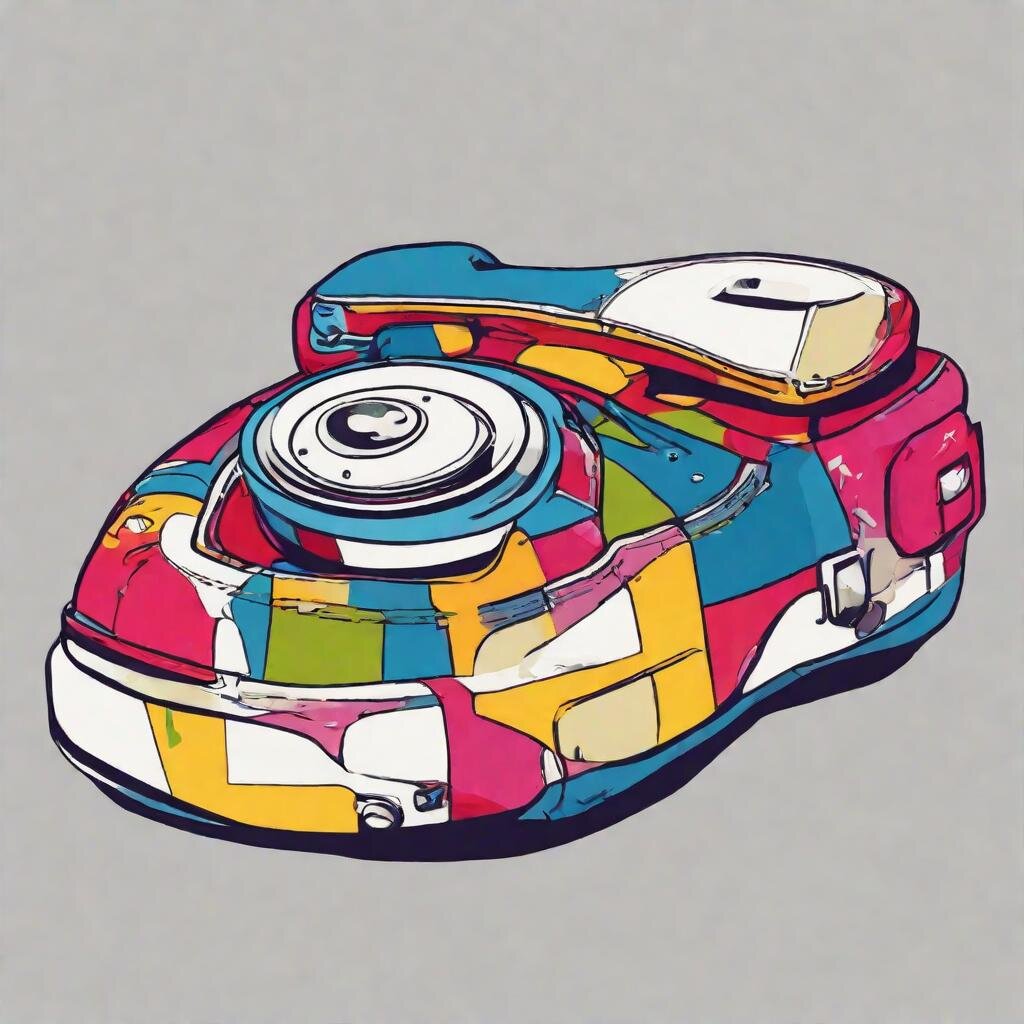}
        \end{minipage} & 
        \begin{minipage}{0.11\textwidth}
            \includegraphics[width=\textwidth]{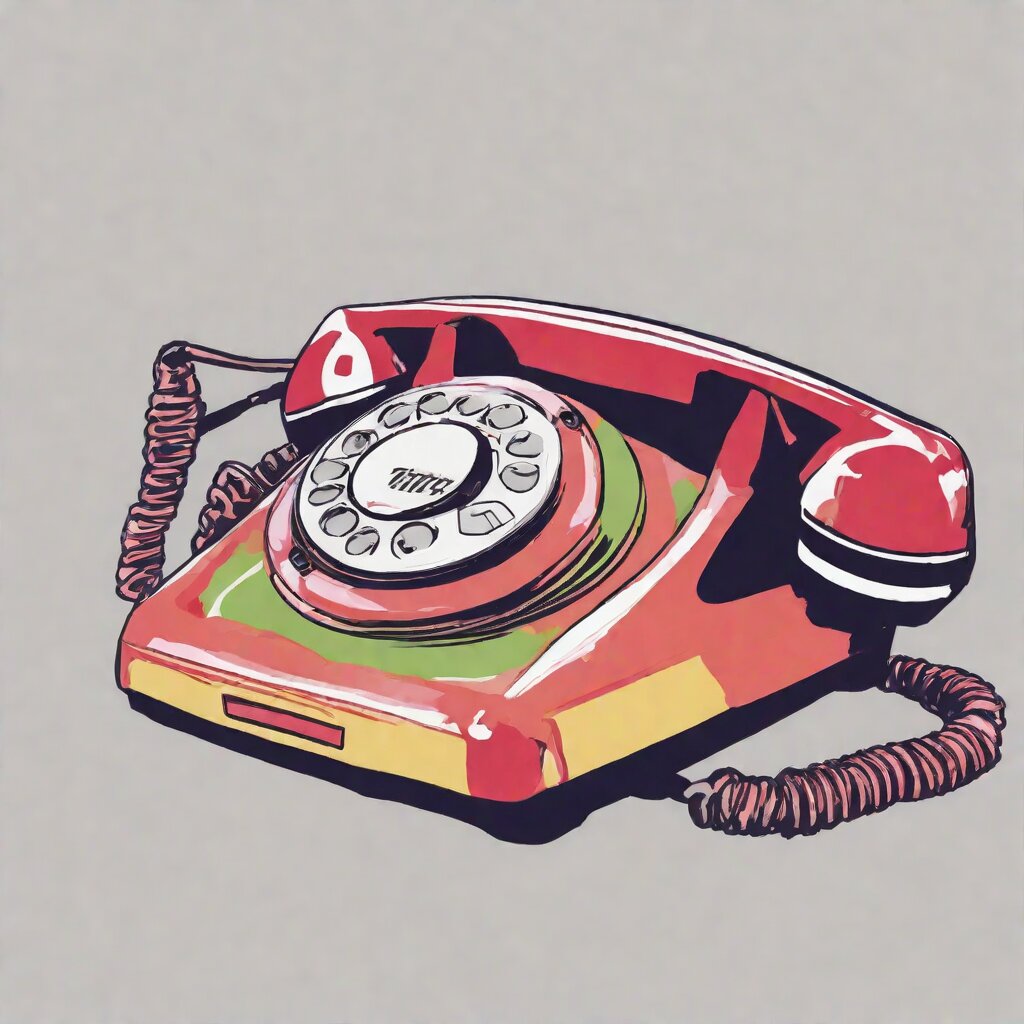}
        \end{minipage} 
        \vspace{1px}
        \\
        \vspace{1px}
        \footnotesize ``A shoe" & \footnotesize ``A phone" &
        \multicolumn{2}{c}{\footnotesize ``... in pop art style."} \\
        \begin{minipage}{0.11\textwidth}
            \includegraphics[width=\textwidth]{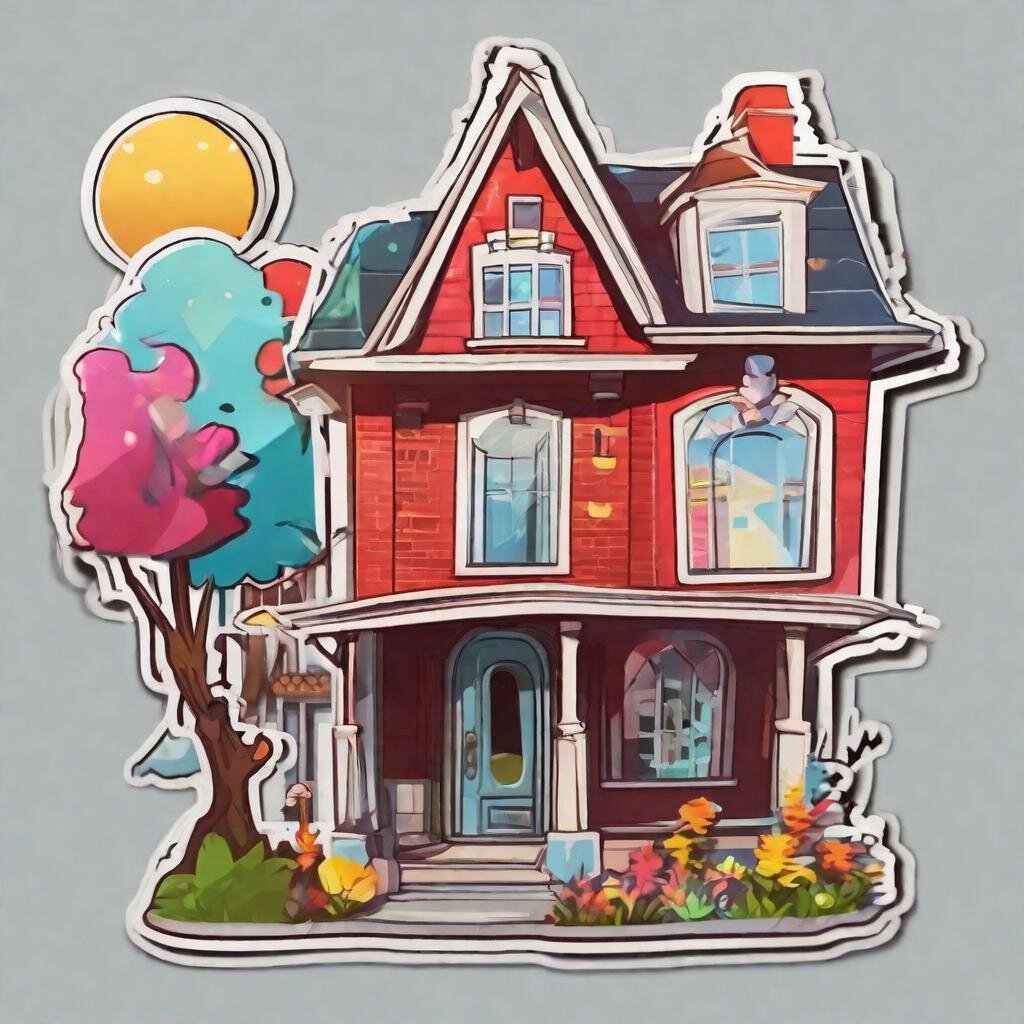}
        \end{minipage} &
        \begin{minipage}{0.11\textwidth}
            \includegraphics[width=\textwidth]{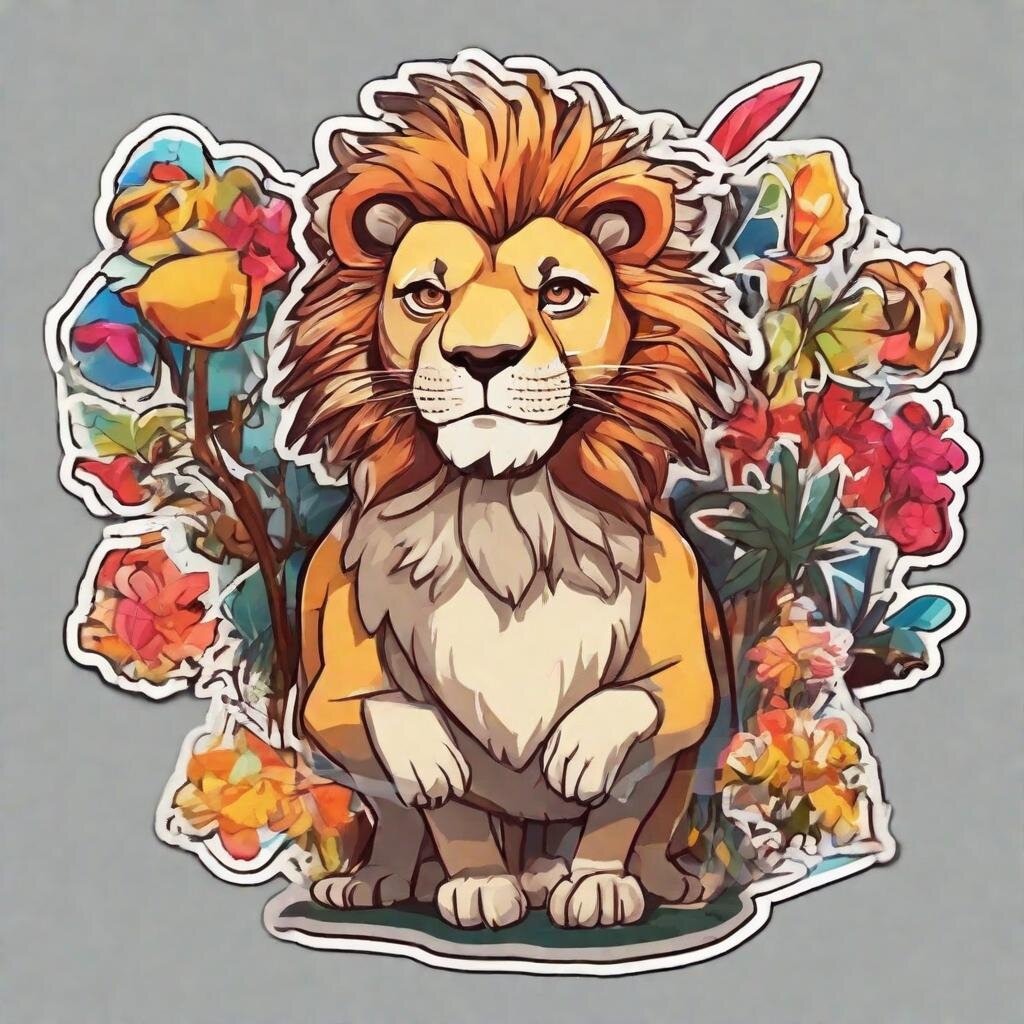}
        \end{minipage} &
        \begin{minipage}{0.11\textwidth}
            \includegraphics[width=\textwidth]{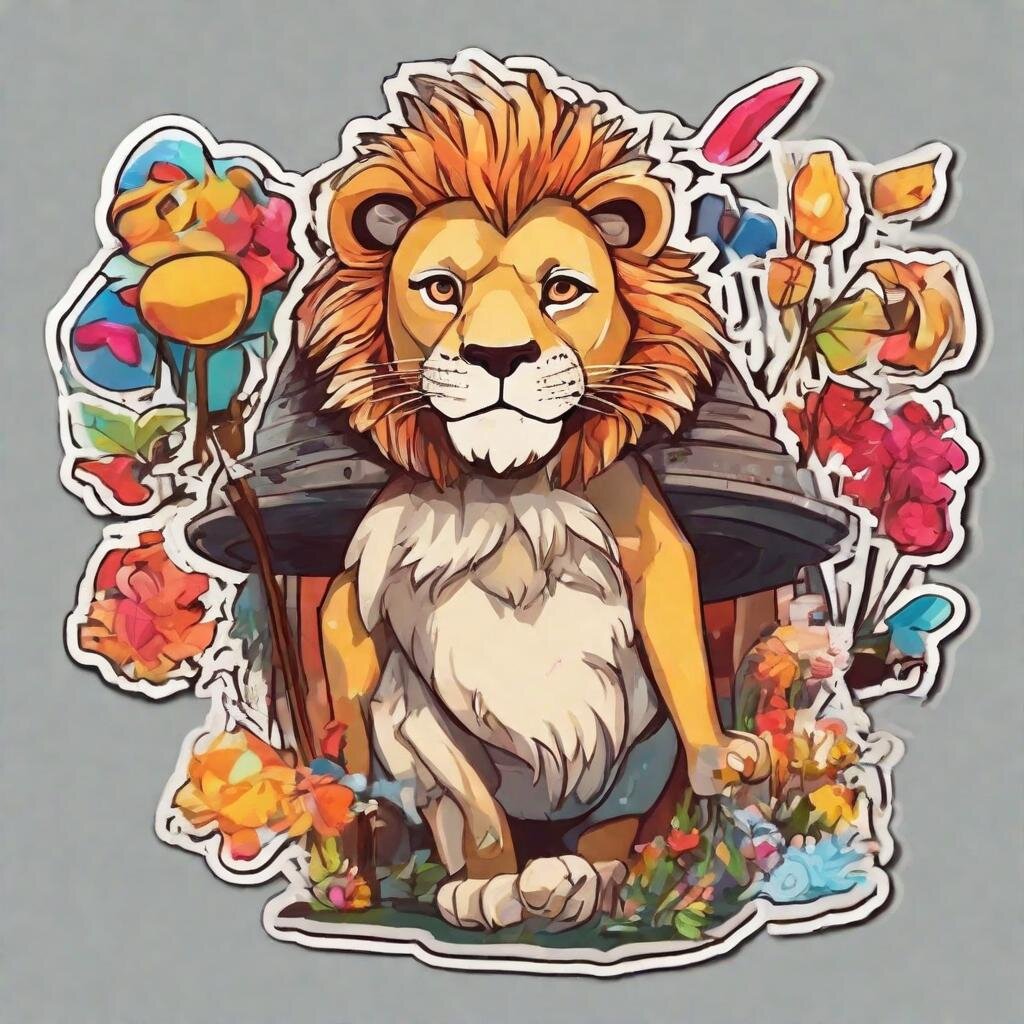}
        \end{minipage} & 
        \begin{minipage}{0.11\textwidth}
            \includegraphics[width=\textwidth]{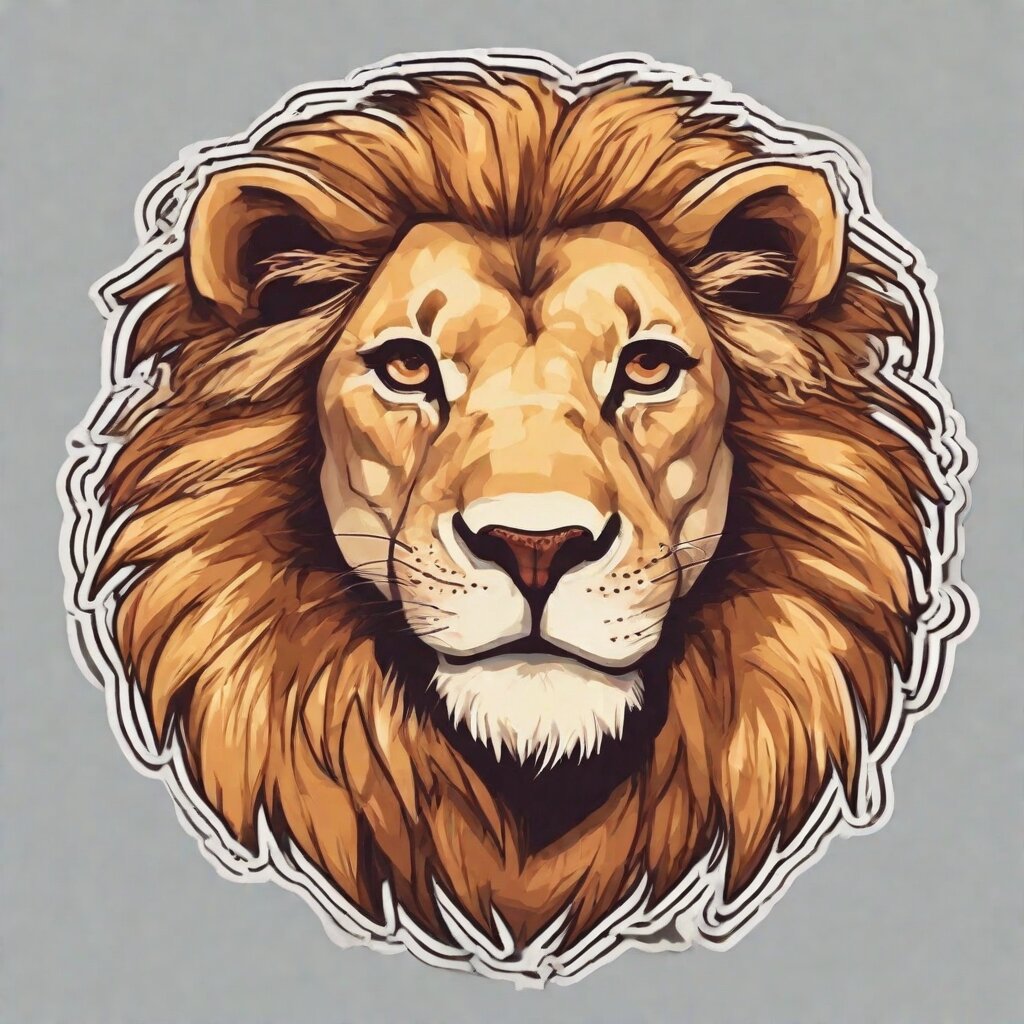}
        \end{minipage} 
        \vspace{1px}
        \\
        \vspace{1px}
        \footnotesize ``A house" & \footnotesize ``A lion" &
         \multicolumn{2}{c}{\footnotesize ``.. in stickers style."} \\
        \begin{minipage}{0.11\textwidth}
            \includegraphics[width=\textwidth]{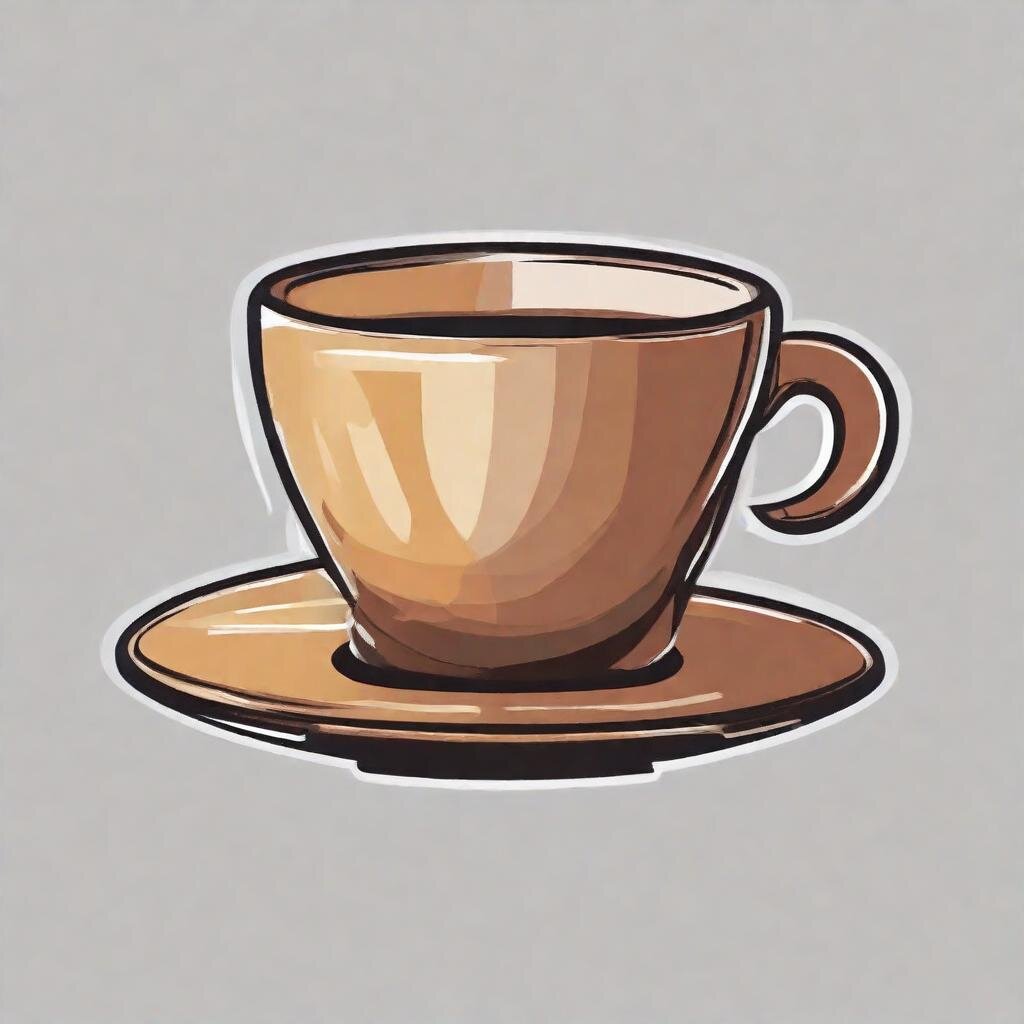}
        \end{minipage} &
        \begin{minipage}{0.11\textwidth}
            \includegraphics[width=\textwidth]{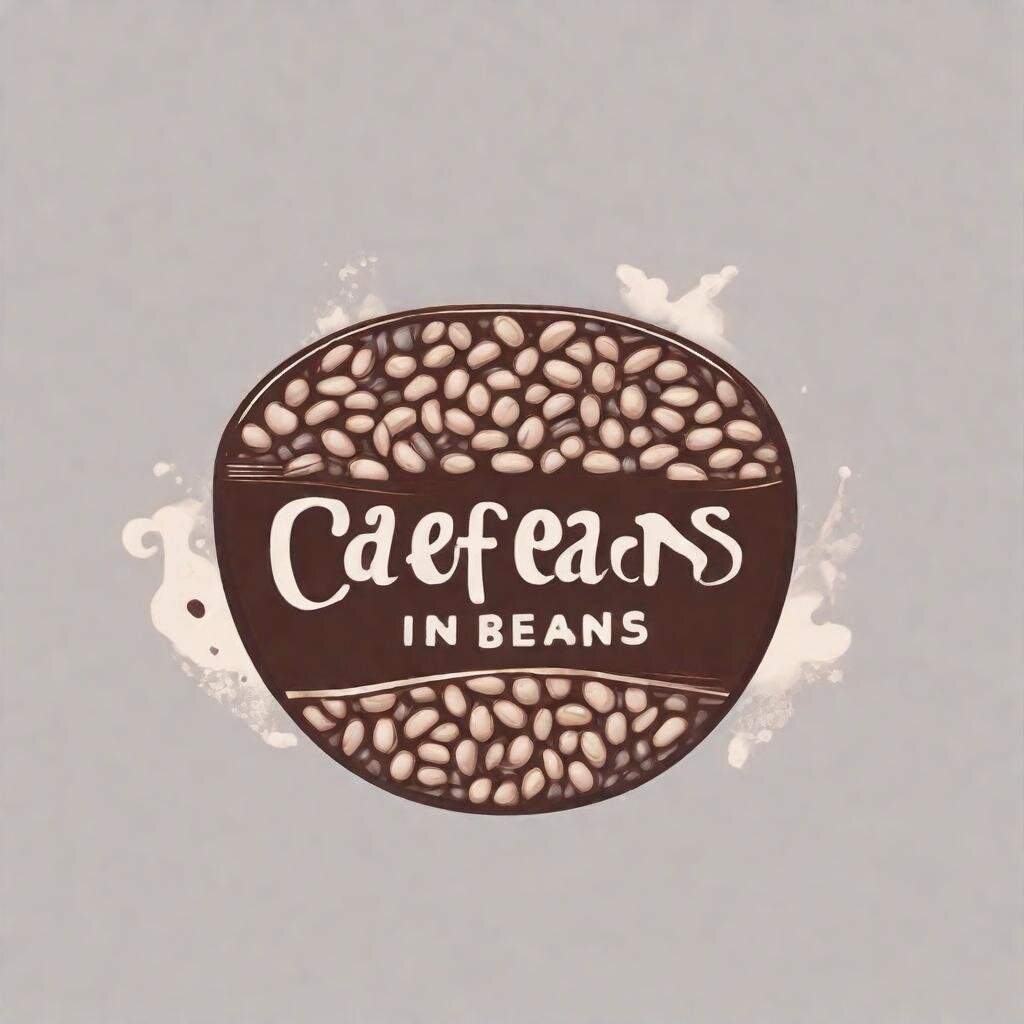}
        \end{minipage} &
        \begin{minipage}{0.11\textwidth}
            \includegraphics[width=\textwidth]{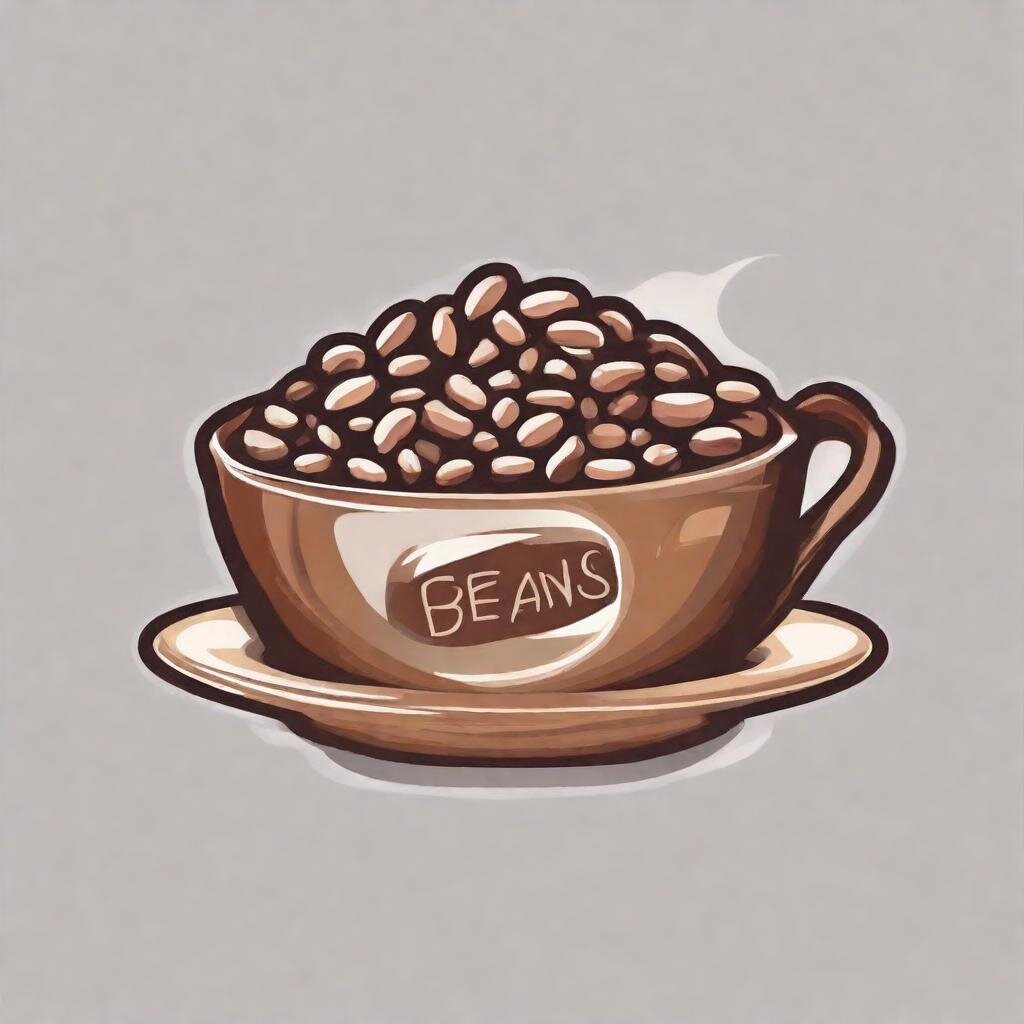}
        \end{minipage} &
        \begin{minipage}{0.11\textwidth}
            \includegraphics[width=\textwidth]{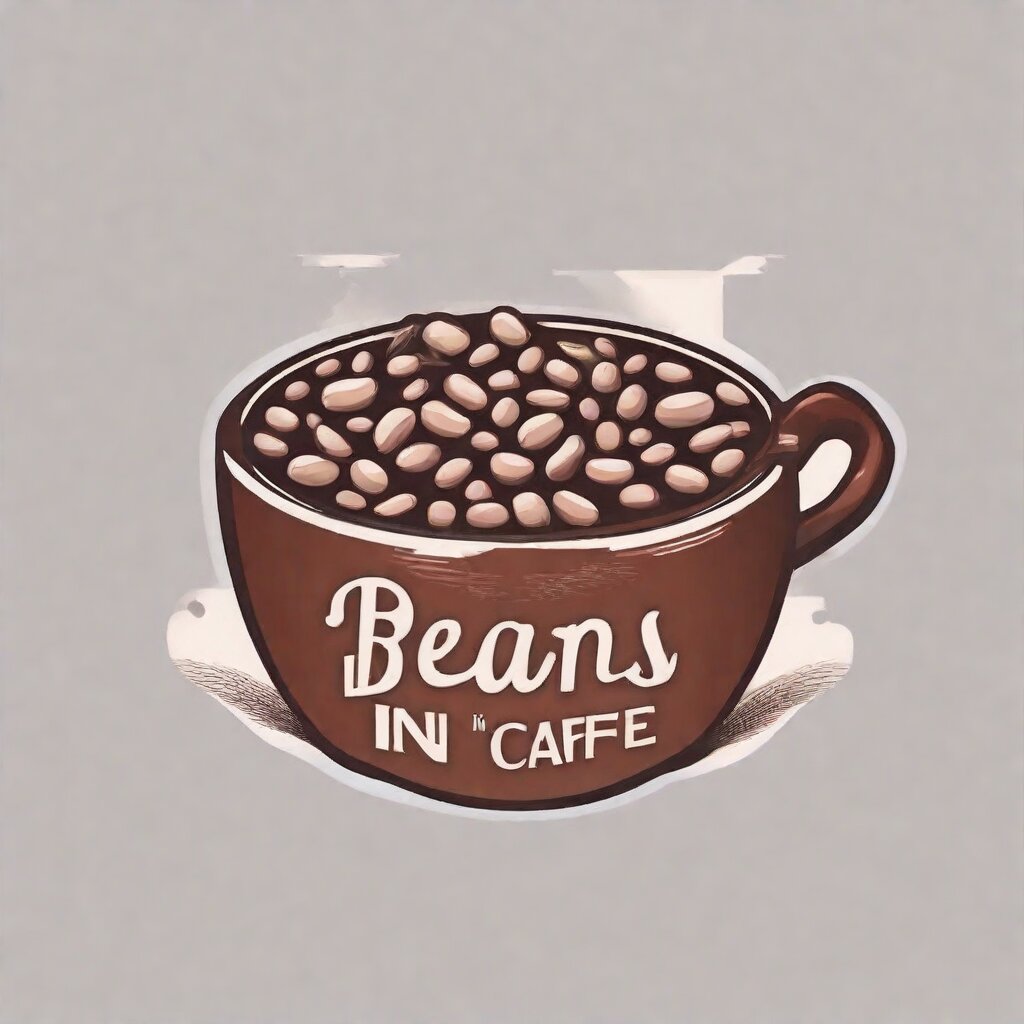}
        \end{minipage} 
        \vspace{1px}
        \\
        \vspace{1px}
        \footnotesize ``A cup" & \footnotesize ``Beans" &
        \multicolumn{2}{c}{\footnotesize``... in cafe logo style."} \\
        \begin{minipage}{0.11\textwidth}
            \includegraphics[width=\textwidth]{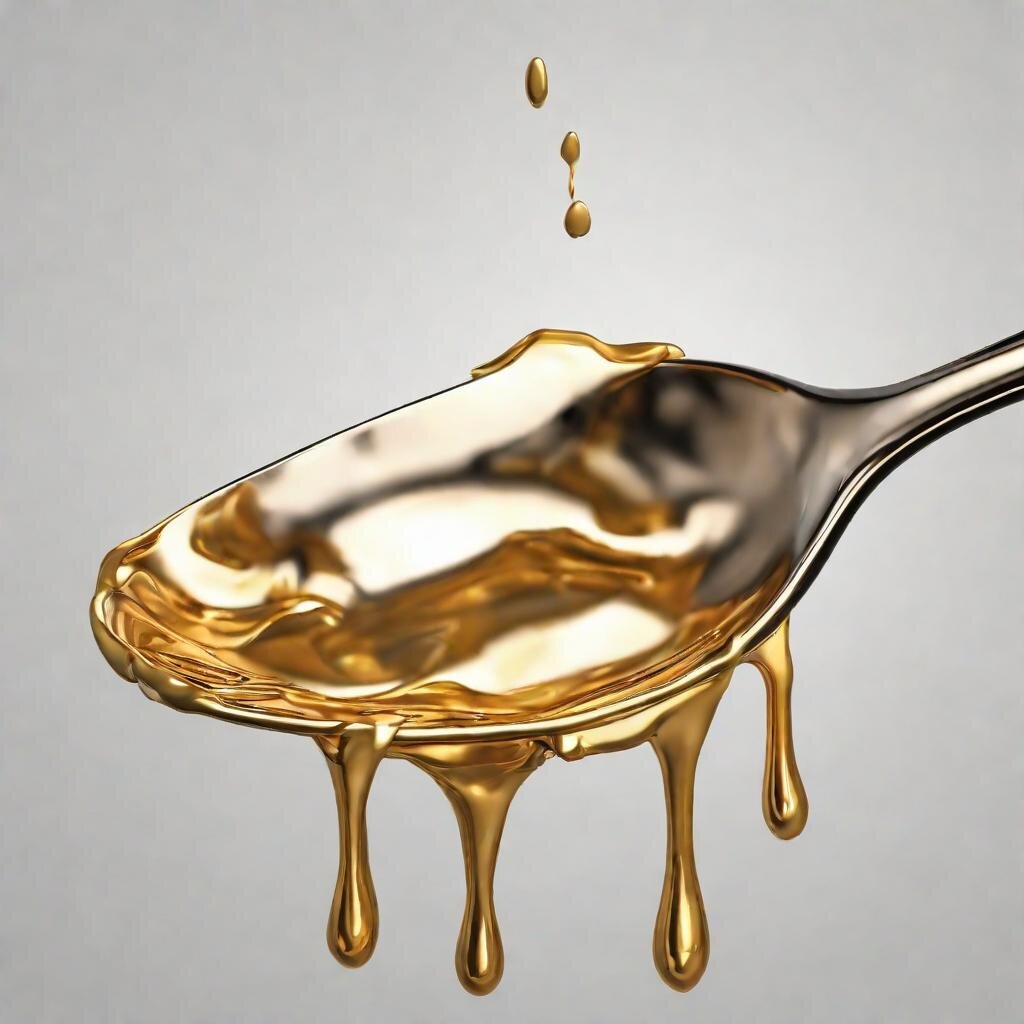}
        \end{minipage} &
        \begin{minipage}{0.11\textwidth}
            \includegraphics[width=\textwidth]{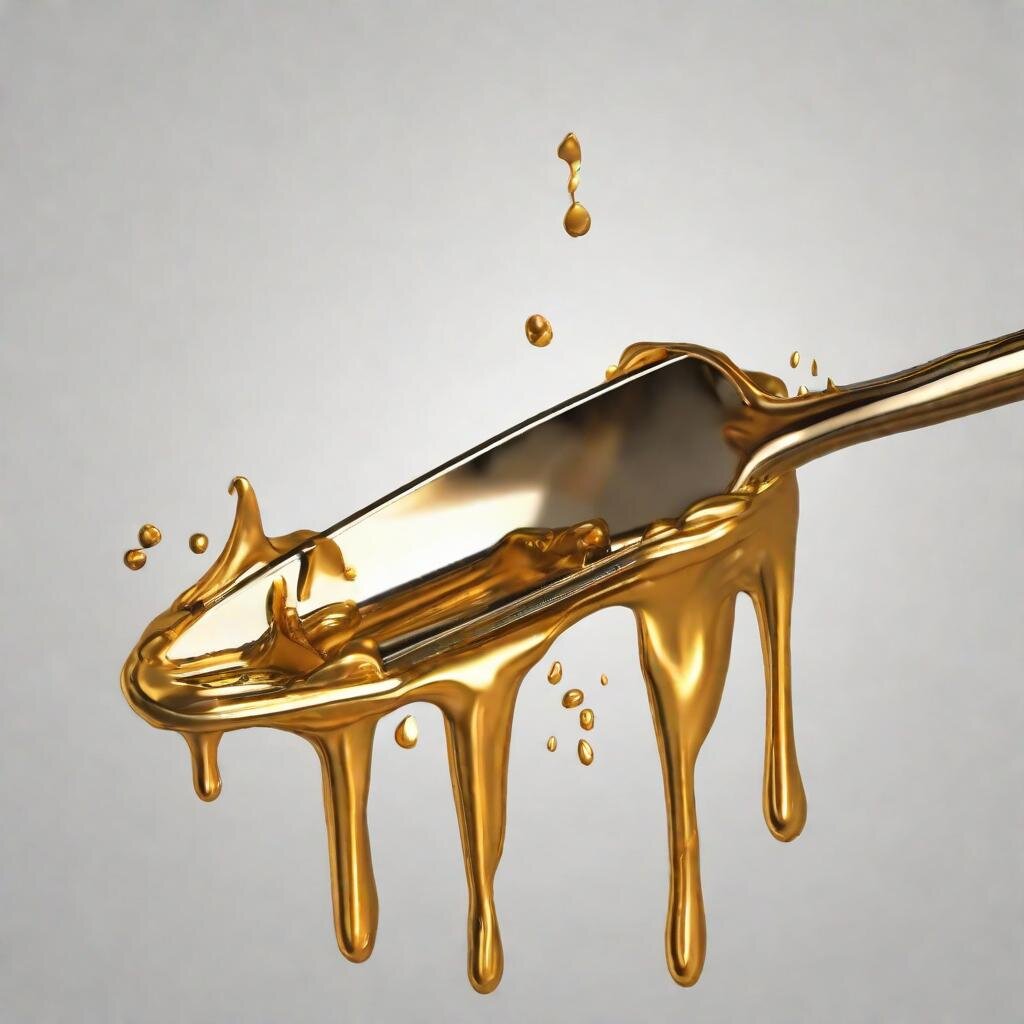}
        \end{minipage} &
        \begin{minipage}{0.11\textwidth}
            \includegraphics[width=\textwidth]{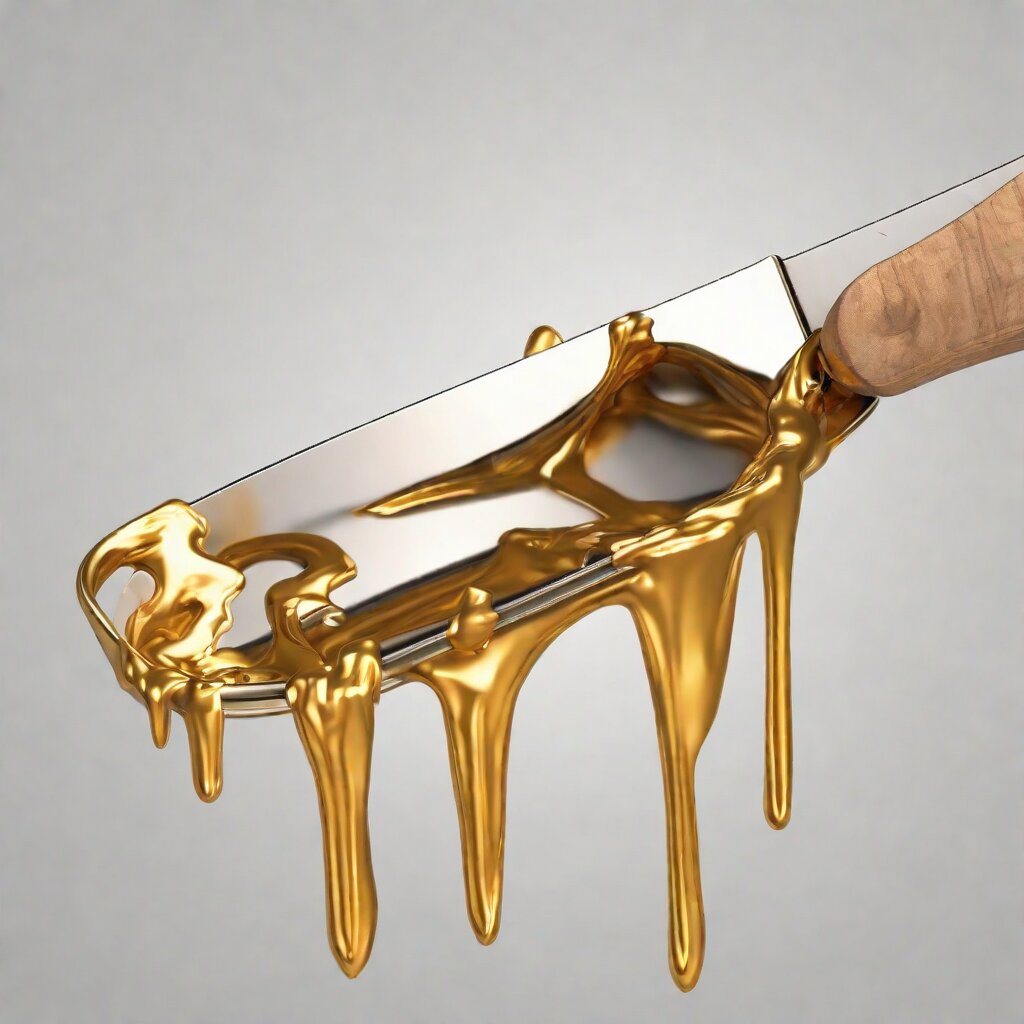}
        \end{minipage} & 
        \begin{minipage}{0.11\textwidth}
            \includegraphics[width=\textwidth]{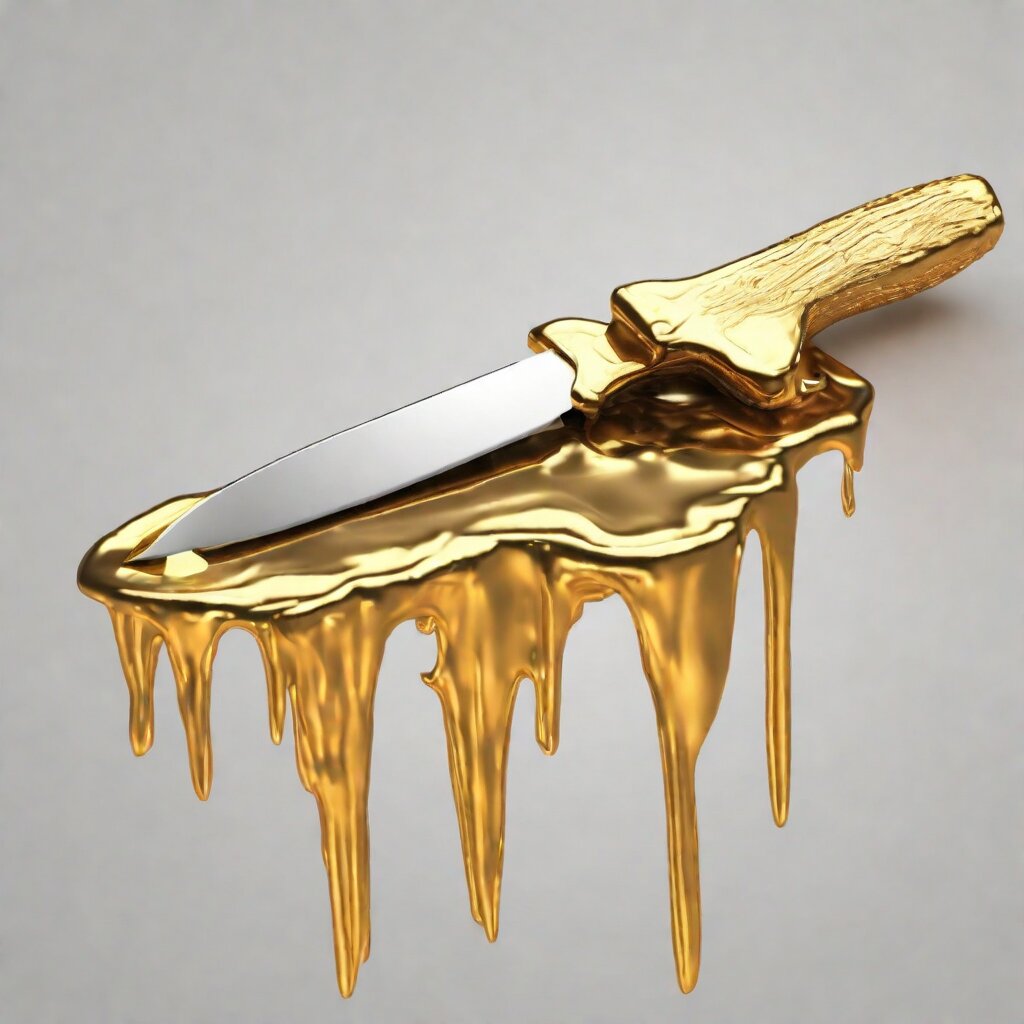}
        \end{minipage}
        \vspace{1px}
        \\
        \vspace{1px}
        \footnotesize ``A spoon" & \footnotesize ``A knife" &
        \multicolumn{2}{c}{\footnotesize ``... in melting golden 3D style."} \\
    
    \end{tabular}
    }
    \vspace{-5pt}
    \caption[Ablation study on the essence of adaptivity in style alignment]{
    \textbf{Adaptive vs Fixed Scaling}: Fixing the scaling parameter that controls shared attention does not yield consistent results across all instances, often failing to prevent content leakage or accurately align the desired style with the target subject.}
    \vspace{-15pt}
    \label{fig:Ablation_2}
\end{figure}

A key property of generative diffusion models is that structure and semantics emerge early in the process~\cite{patashnik2023localizingobjectlevelshapevariations}. Since content leakage is inherently semantic, we can apply the proposed process early in the denoising stage, bypassing full generation.
Thus, to \emph{increase efficiency}, we perform the proposed localization approach at $t = T/2$, as our experiments indicate that leakage is observable by this stage.

\textbf{Generalization.} The proposed localization approach can be applied in the output of any style consistency diffusion pipeline, used as a standalone component. 
Specifically, given the images $I_{ref}$ and $I_{tgt}$ along with their subjects $S_{ref}$ and $S_{tgt}$ we can use DDIM inversion to obtain latents $z_{T}, z_{T-1}$ for each image.
Then, we simulate the final two diffusion steps to localize content leakage as described in this section.
This has minimal computational overhead and can serve as a post-processing step for any style consistency method.

\subsection{Adaptive Scaling}
\label{sec:Optimal_scaling}

As mentioned before, we would like to choose the maximum value of scale $\alpha$ that resolves content leakage, as faithfully aligning the style of the reference ``subject" with the target one is also necessary in many cases, and it is intuitively natural that lower values of scale, lower that alignment. To avoid a linear search, we exploit the monotonicity of $\alpha$ with respect to the binary leakage indicator $L_o$, which answers the decision problem: \textit{``Is there any content leakage on $I_{tgt}$?"}. Given this property, we apply binary search on $\alpha$ to efficiently minimize leakage. This process has a multiplicative computational overhead of $\Theta(|\log(p)|)$, requiring $|\log(p)|$ style aligning generations, where $p$ is the enforced precision of $\alpha$.
\section{Experiments and Evaluation}
\label{sec:Experiments_and_Evaluation}

\begin{figure}[t]
    \centering
    \scriptsize
    \begin{tabular}{c@{\hspace{-.1cm}}c@{\hspace{.1cm}}c@{\hspace{.1cm}}c@{\hspace{.1cm}}c}  
         & Reference & $\beta = 0.875$ & Content Leakage & Adaptive\\

        \begin{minipage}{0.02\textwidth}
            \rotatebox[origin=c]{90}{\scriptsize B-LoRA}
        \end{minipage} &
        \begin{minipage}{0.11\textwidth}
            \includegraphics[width=\textwidth]{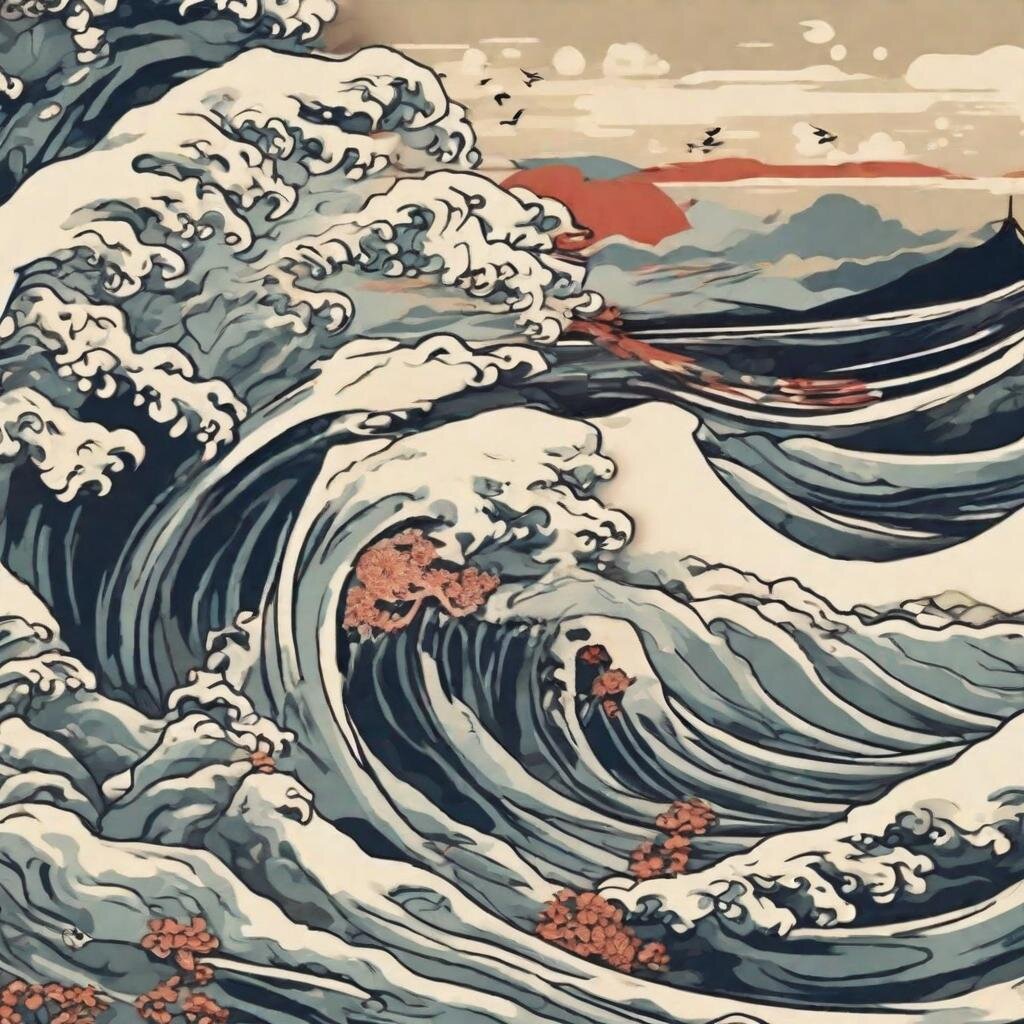}
        \end{minipage} &
        \begin{minipage}{0.11\textwidth}
            \includegraphics[width=\textwidth]{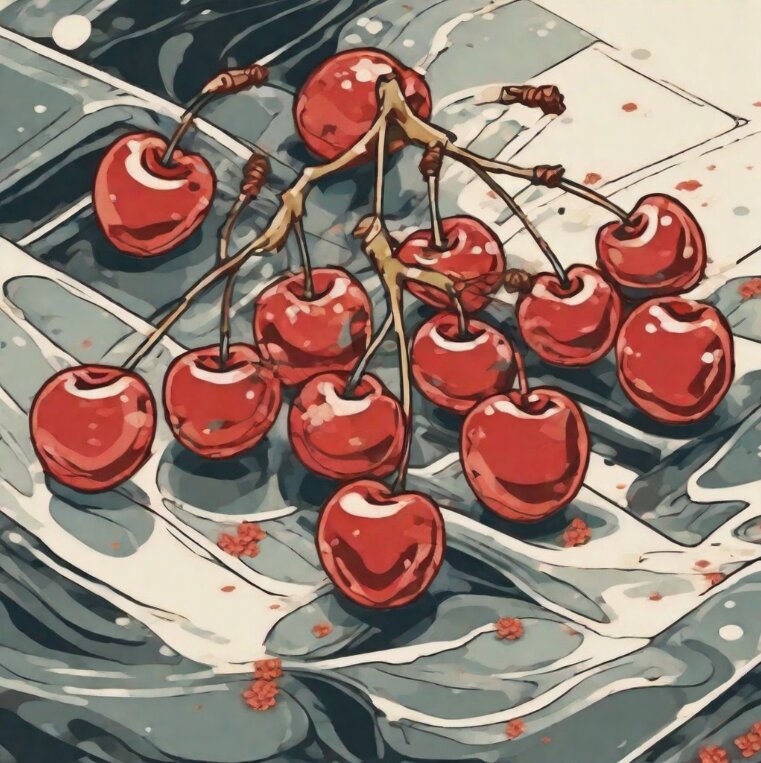}
        \end{minipage} &
        \begin{minipage}{0.11\textwidth}
            \includegraphics[width=\textwidth]{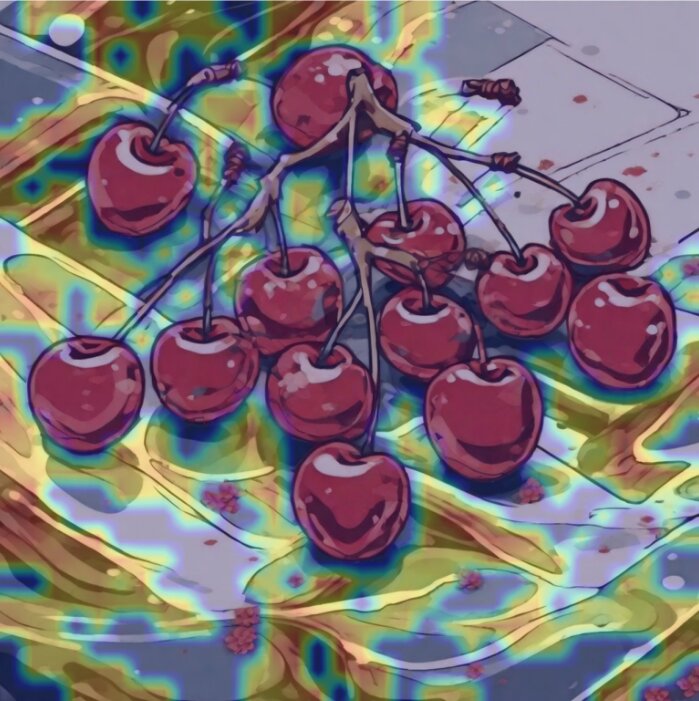}
        \end{minipage} & 
        \begin{minipage}{0.11\textwidth}
            \includegraphics[width=\textwidth]{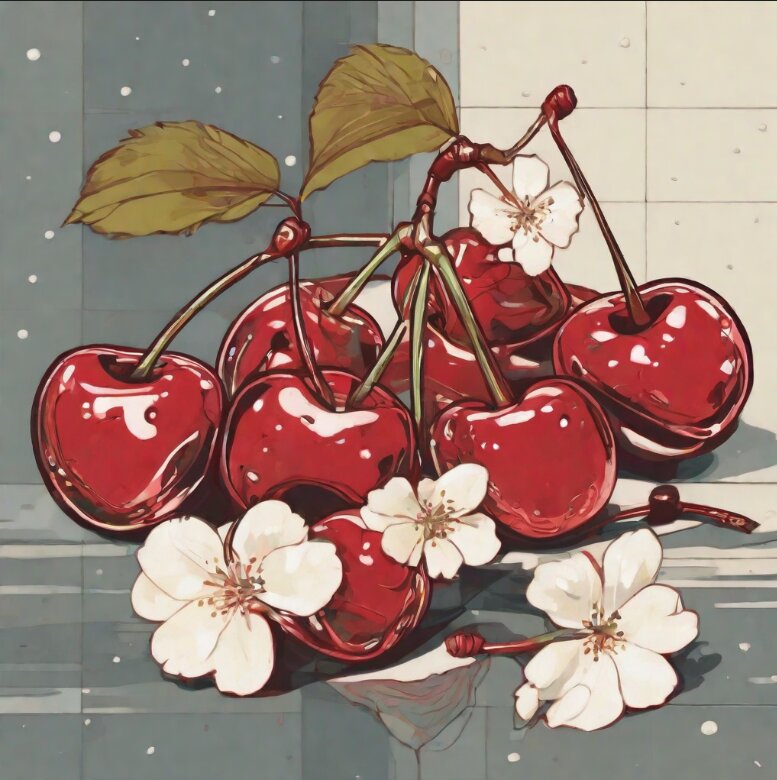}
        \end{minipage} 
        \vspace{1.5px}
        \\
        \vspace{0.5px}
         & & $\sigma = 0.25$ &  &  \\
        
        \begin{minipage}{0.02\textwidth}
            \rotatebox[origin=c]{90}{\scriptsize InstantStyle}
        \end{minipage} &
        \begin{minipage}{0.11\textwidth}
            \includegraphics[width=\textwidth]{graphics/results/generalization/ref.jpg}
        \end{minipage} &
        \begin{minipage}{0.11\textwidth}
            \includegraphics[width=\textwidth]{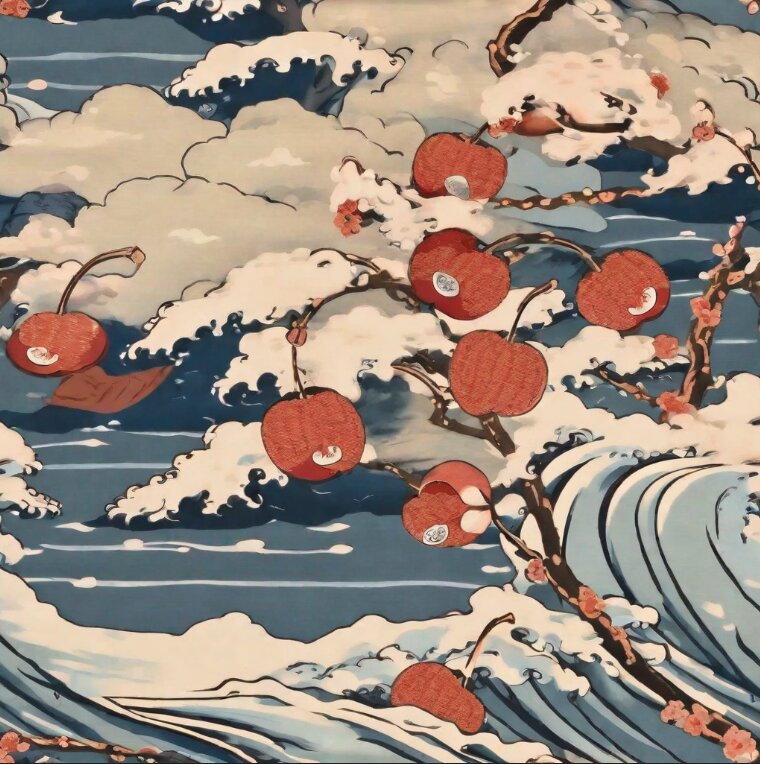}
        \end{minipage} &
        \begin{minipage}{0.11\textwidth}
            \includegraphics[width=\textwidth]{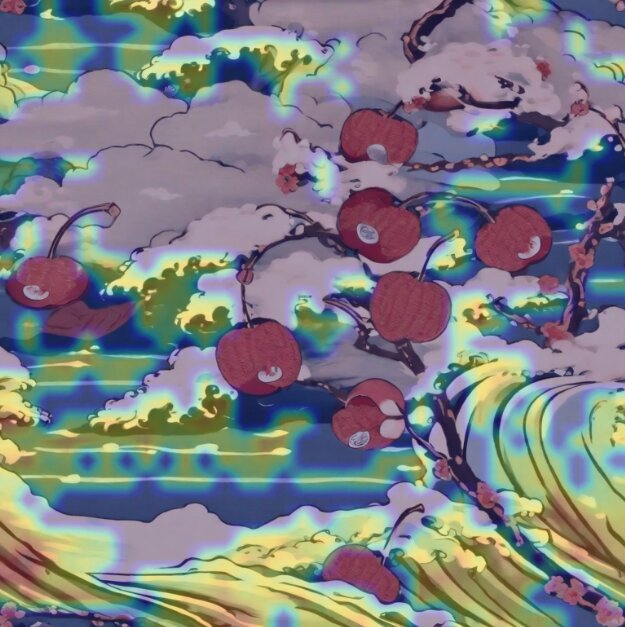}
        \end{minipage} & 
        \begin{minipage}{0.11\textwidth}
            \includegraphics[width=\textwidth]{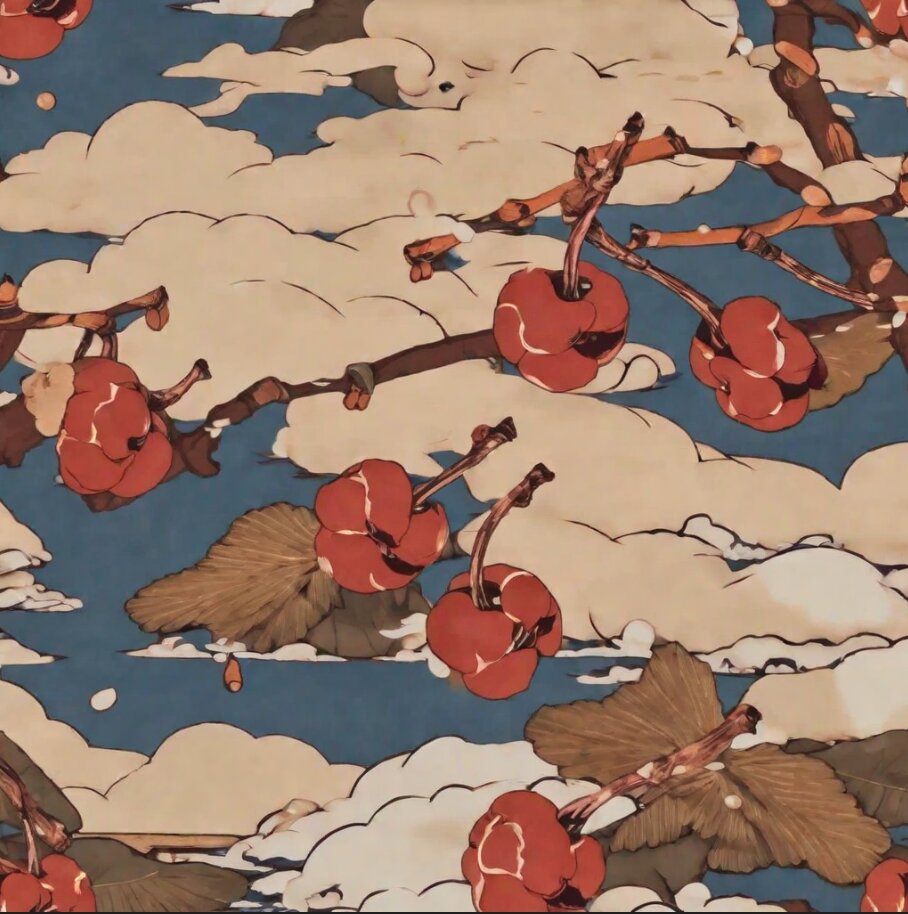}
        \end{minipage} 
        \vspace{1.5px}
        \\
        \vspace{0.5px}
        & & $\alpha = 0.875$ &  & \\

        \begin{minipage}{0.02\textwidth}
            \rotatebox[origin=c]{90}{\scriptsize \emph{Only-Style}}
        \end{minipage} &
        \begin{minipage}{0.11\textwidth}
            \includegraphics[width=\textwidth]{graphics/results/generalization/ref.jpg}
        \end{minipage} &
        \begin{minipage}{0.11\textwidth}
            \includegraphics[width=\textwidth]{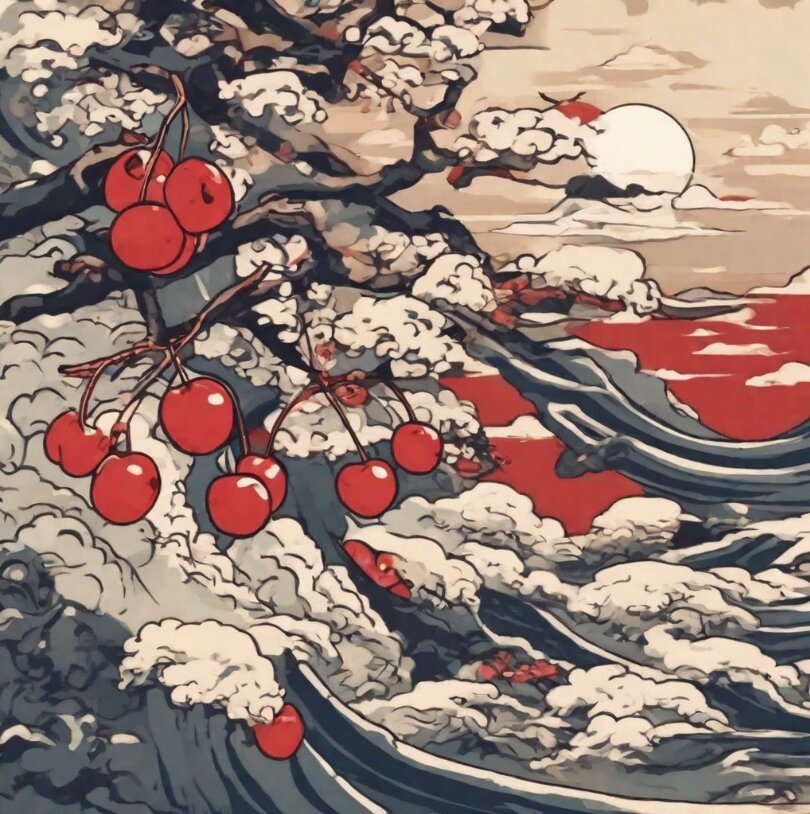}
        \end{minipage} &
        \begin{minipage}{0.11\textwidth}
            \includegraphics[width=\textwidth]{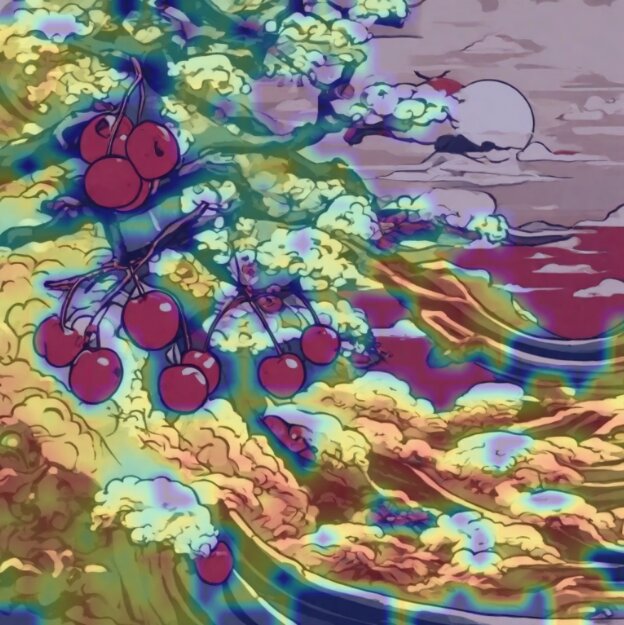}
        \end{minipage} &
        \begin{minipage}{0.11\textwidth}
            \includegraphics[width=\textwidth]{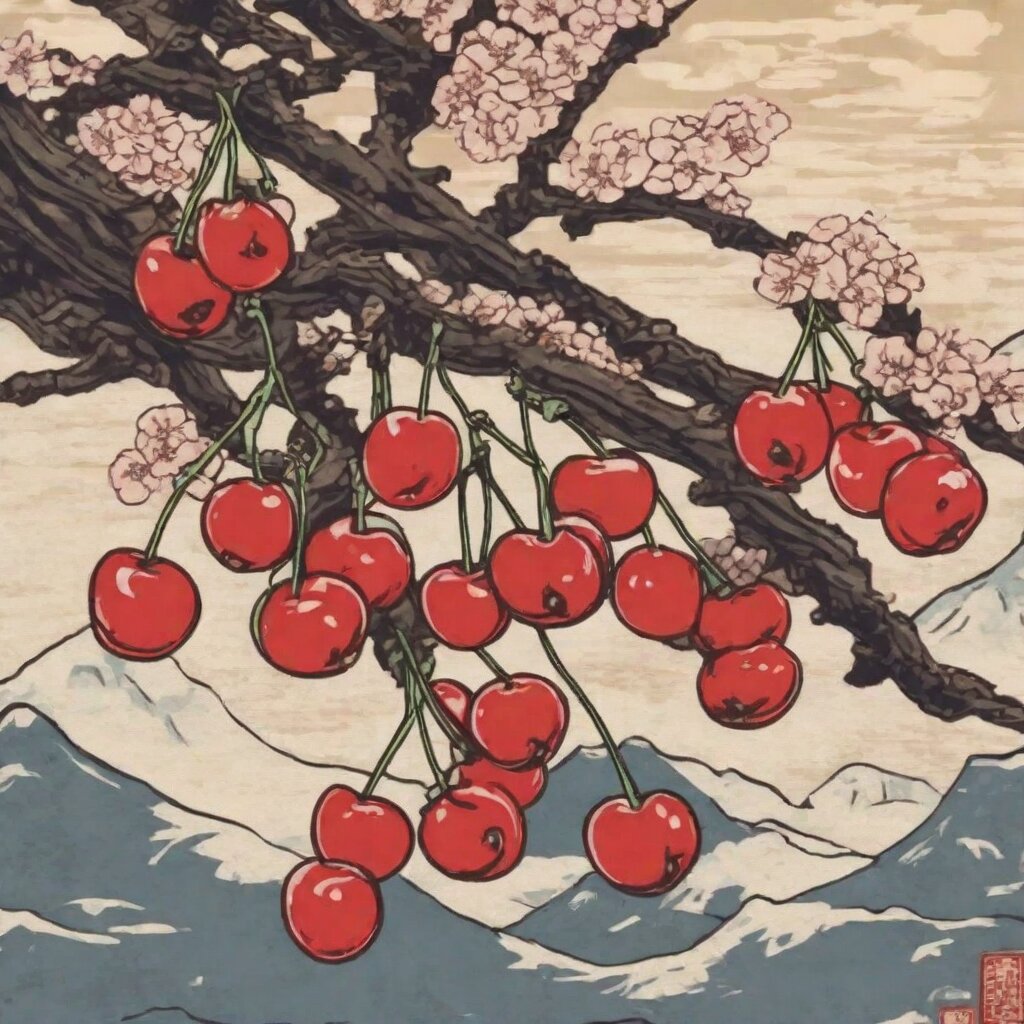}
        \end{minipage} 
        \vspace{2pt} 
        \\
        \vspace{-1pt}
        & \scriptsize ``A wave" & \scriptsize ``Cherries" &  
        \multicolumn{2}{c}{\scriptsize ``...in Japanese Ukiyo-e style."} \\
        
    \end{tabular}
    \vspace{-5pt}
    \caption[hyperparameter tuning.]{
    \textbf{Hyperparameter tuning to mitigate content leakage.} We apply the leakage detection of Sec.\ref{sec:Content leakage loc} to adaptively tune hyperparameters presented in B-LoRA\cite{frenkel2024blora} and InstantStyle\cite{Wang2024InstantStyle}. While these methods reduce leakage, they distort the reference style. \emph{Only-Style} preserves both style and content integrity.} 
    \vspace{-15pt}
    \label{fig:Generalization}
\end{figure}

\subsection{Experimental setup}

\textbf{Implementation.} We implement our method on top of StyleAligned \cite{hertz2024style} which uses Stable Diffusion XL (SDXL) \cite{podell2023sdxlimprovinglatentdiffusion} at its core. The proposed leakage control process (for a single scale $\alpha$) takes 31 seconds in a NVIDIA GeForce RTX 3090, only 2 seconds more than the base StyleAligned generation. We set a fixed precision value of \(p = 0.03125\) for the binary search of Sec.~\ref{sec:Optimal_scaling}, which means that the whole process involves 4 half generations (see Sec.~\ref{sec:Content leakage loc}), each one of them determining the binary indication of leakage, and one whole, to generate the final style-consistent pair. Thus, our method runs for approximately 1.5 minutes per style alignment instance. Details on time requirements of sota in the Suppl. Mat.

\textbf{Evaluation prompt set.} We create an evaluation set of 100 prompts, by extracting the 100 creative style descriptors used in the evaluation set of StyleAligned \cite{hertz2024style}, and then employing ChatGPT to generate highly diverse objects (4 per style) that could appear in the specific style context. The unified style prompt looks like the following: \textit{\{`A clock', $\dots$, `A cupcake'\} in abstract rainbow colored flowing smoke wave design}.
The textual descriptions of the objects are kept minimal without modifiers such as adjectives and prepositional phrases, as the linguistic identification of the subject text tokens within a prompt, that is necessary for the localization of content leakage, remains out of the scope of our work. The set is presented in the Suppl. Mat.

\textbf{Comparison with state-of-the-art methods.} We compare our work with the following state-of-the-art style-consistent generation methods, implemented on top of generative diffusion models. Apart from StyleAligned (SA)~\cite{hertz2024style}, we considered InstantStyle (IS)~\cite{Wang2024InstantStyle} as an adapter-based method that focuses on avoiding content leakage, B-LoRA~\cite{frenkel2024blora} and StyleDrop (SDRP), as two optimization-based baselines that yield state-of-the-art results. For B-LoRA~\cite{frenkel2024blora} we utilize only the style adaptation and employ it for text-based style consistent generation.
For the first three we utilize their official implementations, while we implement SDRP on top of SDXL. All methods use SDXL as their base model. Comparisons with additional baselines can be found in the Suppl. Mat.

\subsection{Ablation study}

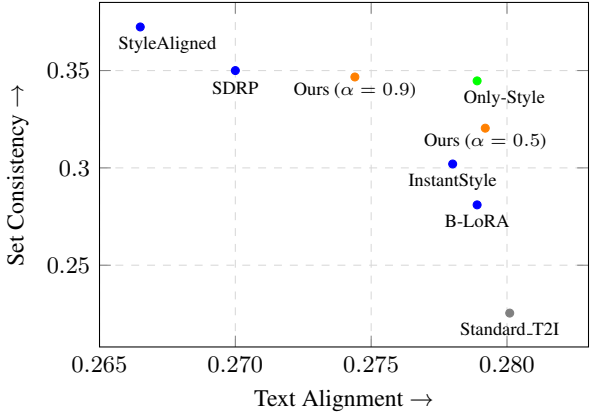
\begin{figure}[t]
    \centering
    \begin{tikzpicture}
        \begin{axis}[
            scale only axis,
            width=0.78\linewidth,
            height=0.55\linewidth,
            xlabel={Text Alignment →},
            ylabel={Set Consistency →},
            xmin=0.265, xmax=0.283,
            ymin=0.208, ymax=0.385,
            grid=both,
            grid style={dashed,gray!30},
            tick label style={font=\small},
            label style={font=\small},
            xticklabel style={
                /pgf/number format/.cd,
                fixed zerofill,
                precision=3
             },
            scaled x ticks=false, 
        ]
            \addplot[only marks, mark=*, mark size=1.5pt, color=blue] coordinates {
                (0.2665, 0.3724)  
                (0.2700, 0.3500)  
                (0.2780, 0.3020)  
                (0.2789, 0.2810)  
            };

            \node[font=\scriptsize] at (axis cs:0.2665+0.001, 0.3724-0.009    ) {StyleAligned};
            \node[font=\scriptsize] at (axis cs:0.2700+0.000, 0.3500-0.009     ) {SDRP};
            \node[font=\scriptsize] at (axis cs:0.2780+0.000, 0.3020-0.009     ) {InstantStyle};
            \node[font=\scriptsize] at (axis cs:0.2789+0.000, 0.2810-0.008     ) {B-LoRA};

            \addplot[only marks, mark=*, mark size=1.5pt, color=orange] coordinates {
                (0.2792, 0.3204)  
                (0.2744, 0.3467)  
            };

            \node[font=\scriptsize] at (axis cs:0.2792+0.000, 0.3204-0.007     ) {Ours ($\alpha=0.5$)};

            \node[font=\scriptsize] at (axis cs:0.2744+0.000, 0.3467-0.007 ) {Ours ($\alpha=0.9$)};

            \addplot[only marks, mark=*, mark size=1.5pt, color=green] coordinates {
                (0.2789, 0.3447)  
            };

            \node[font=\scriptsize] at (axis cs:0.2789+0.001, 0.3447-0.009     ) {Only-Style};

            \addplot[only marks, mark=*, mark size=1.5pt, color=gray] coordinates {
                (0.2801, 0.2254)  
            };
           
            \node[font=\scriptsize] at (axis cs:0.2801+0.000, 0.2254-0.008     ) {Standard\_T2I};

        \end{axis}
    \end{tikzpicture}
    \caption[Quantitative results of our method]{\textbf{Text Alignment vs Stylistic Set Consistency}: We compare three state-of-the-art methods (blue marks), a baseline without stylistic alignment (grey mark), our two ablation variants (orange marks) and \emph{Only-Style} (green mark) in terms of text alignment (CLIP similarity) and set consistency (DINO similarity).
    }
    \vspace{-0.3cm}
    \label{fig:Quantitative_ours}
\end{figure}

\begin{figure*}[t]
\vspace{-2pt}
    \centering
    \resizebox{.95\linewidth}{!}{
    \begin{tabular}{c@{\hspace{.1cm}}c@{\hspace{.1cm}}c@{\hspace{.1cm}}c@{\hspace{.1cm}}c@{\hspace{.1cm}}c@{\hspace{.1cm}}c}
        \scriptsize Reference & \scriptsize B-LoRA~\cite{frenkel2024blora} & \scriptsize InstantStyle~\cite{Wang2024InstantStyle} & \scriptsize StyleDrop~\cite{sohn2023styledrop} & \scriptsize StyleAligned~\cite{hertz2024style} & \scriptsize Content Leakage & \scriptsize \emph{Only-Style} (Ours)\\
        \begin{minipage}{0.12\textwidth}
            \includegraphics[width=\textwidth]{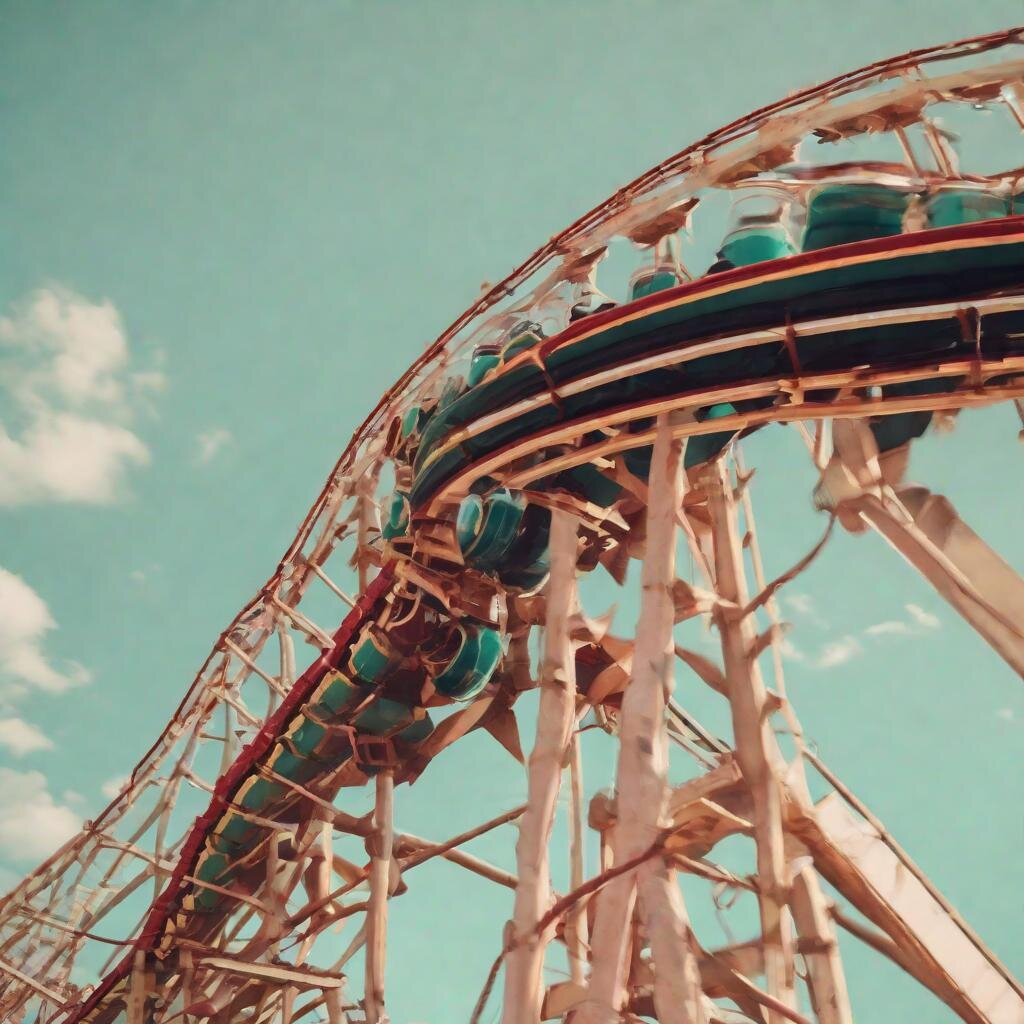}
        \end{minipage} &
        \begin{minipage}{0.12\textwidth}
            \includegraphics[width=\textwidth]{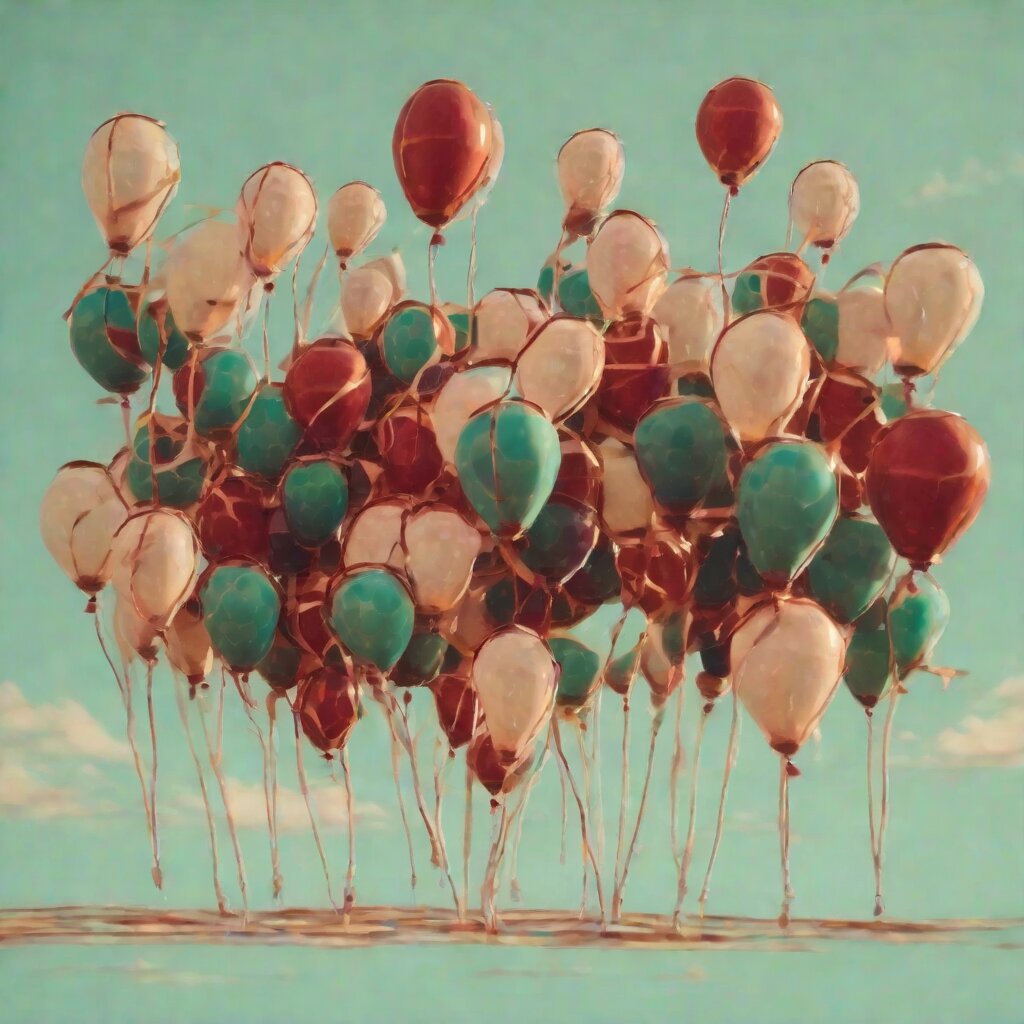}
        \end{minipage} &
        \begin{minipage}{0.12\textwidth}
            \includegraphics[width=\textwidth]{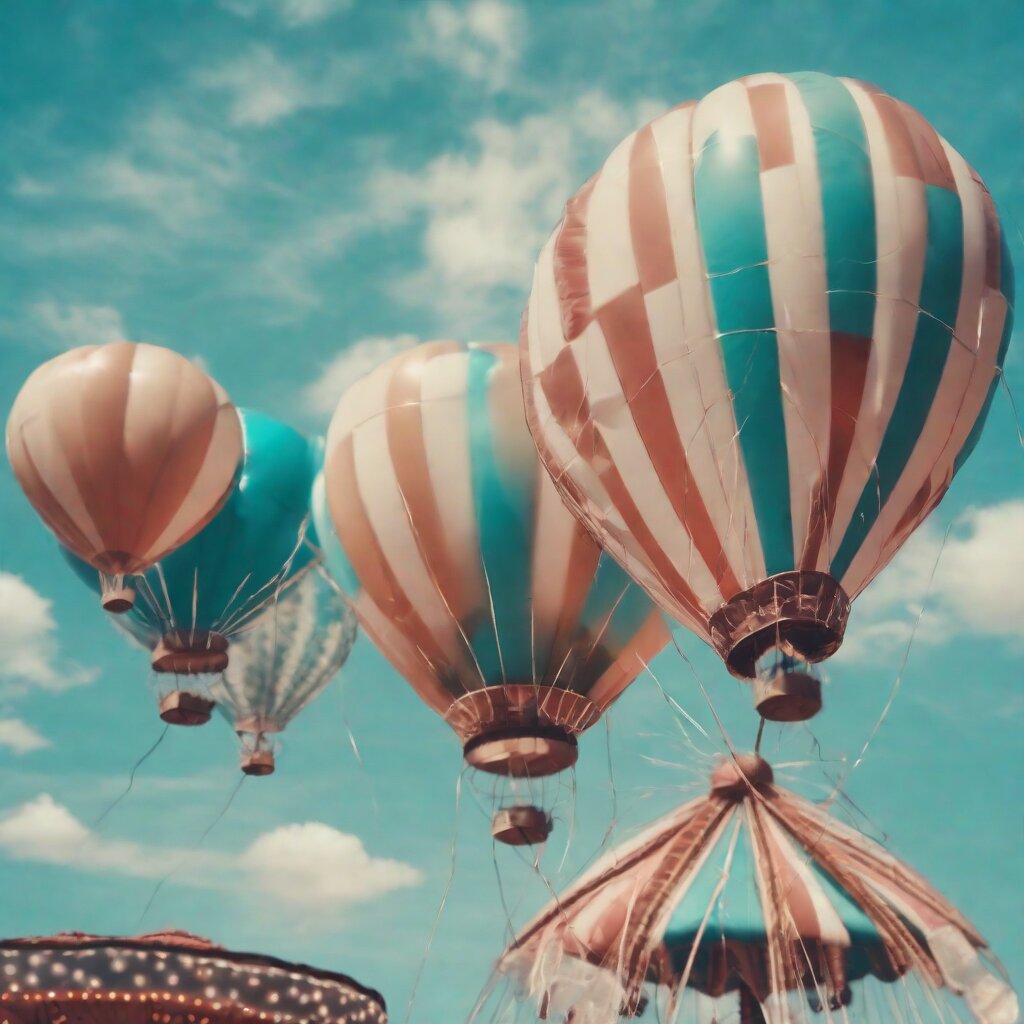}
        \end{minipage} &
        \begin{minipage}{0.12\textwidth}
            \includegraphics[width=\textwidth]{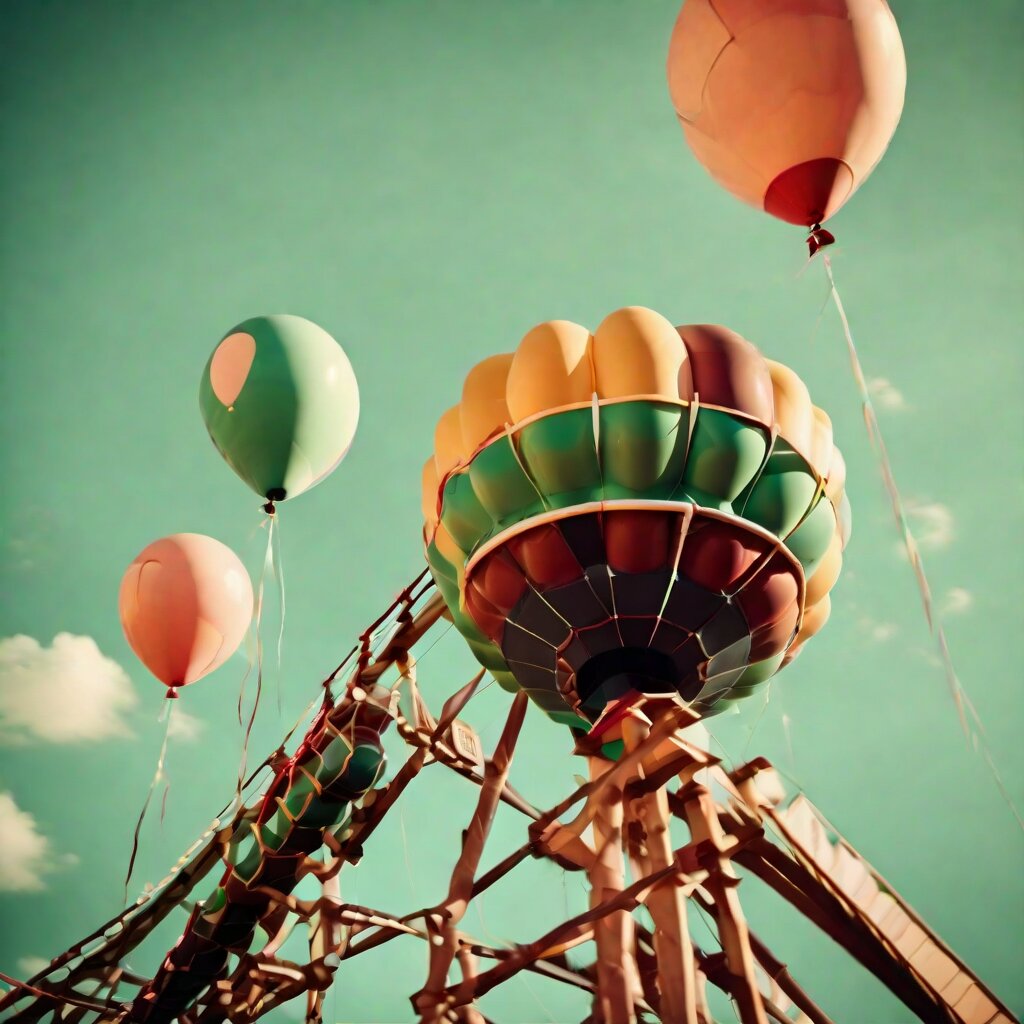}
        \end{minipage} &
        \begin{minipage}{0.12\textwidth}
            \includegraphics[width=\textwidth]{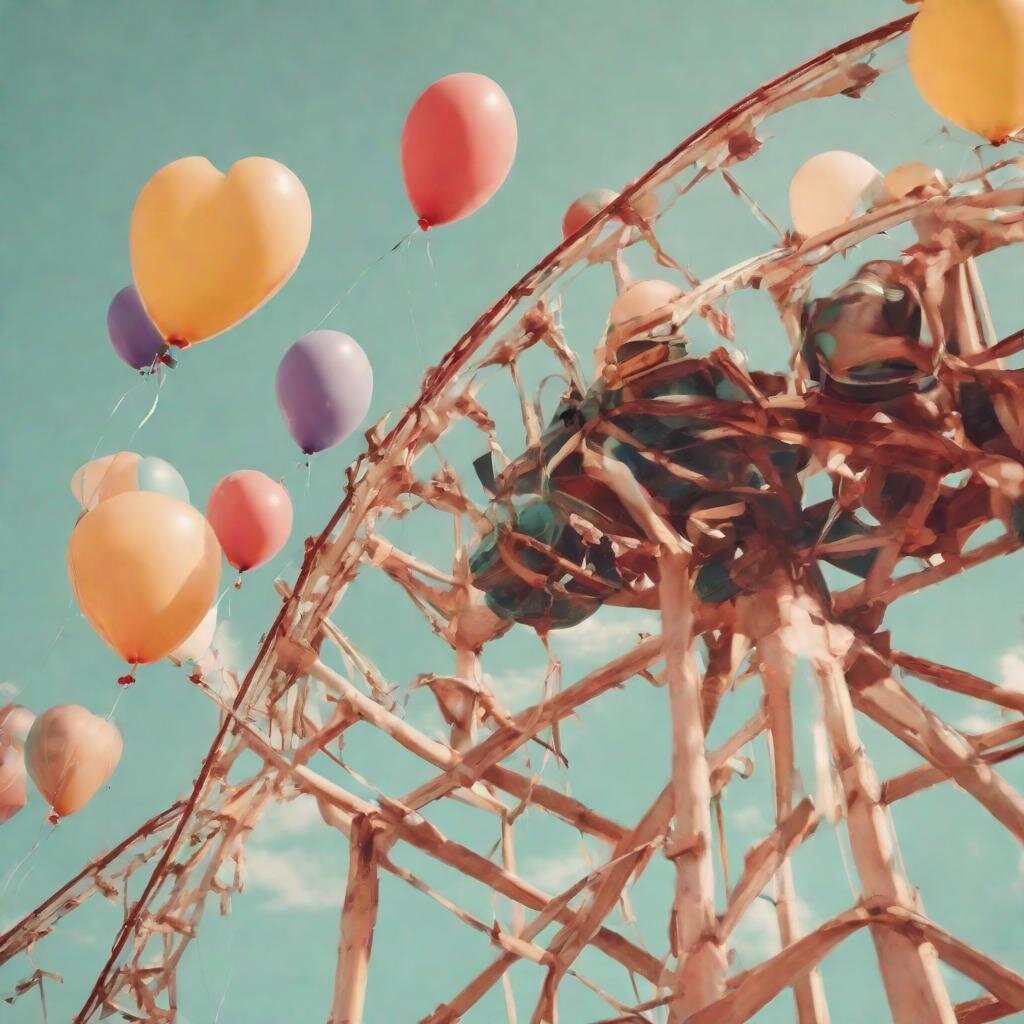}
        \end{minipage} &
        \begin{minipage}{0.12\textwidth}
            \includegraphics[width=\textwidth]{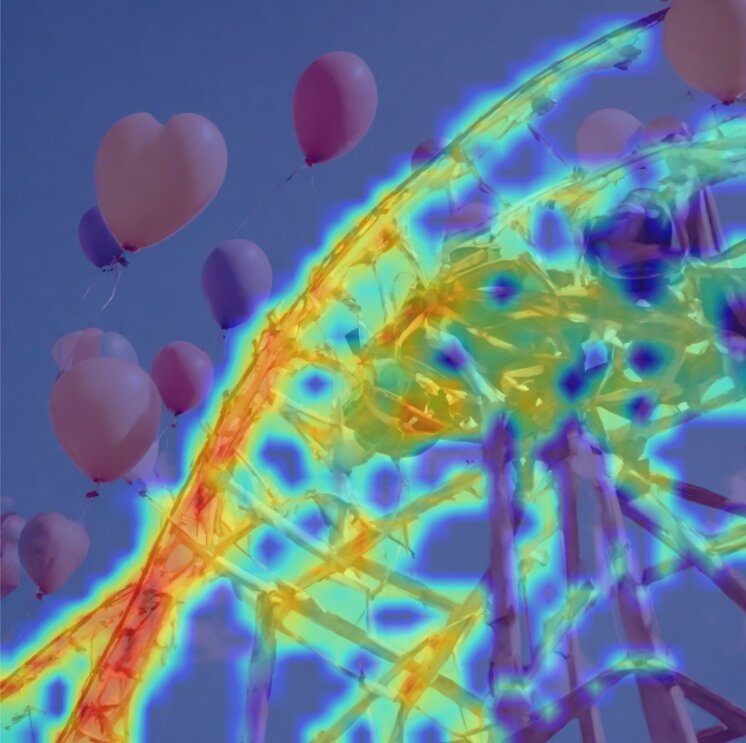}
        \end{minipage} &
        \begin{minipage}{0.12\textwidth}
            \includegraphics[width=\textwidth]{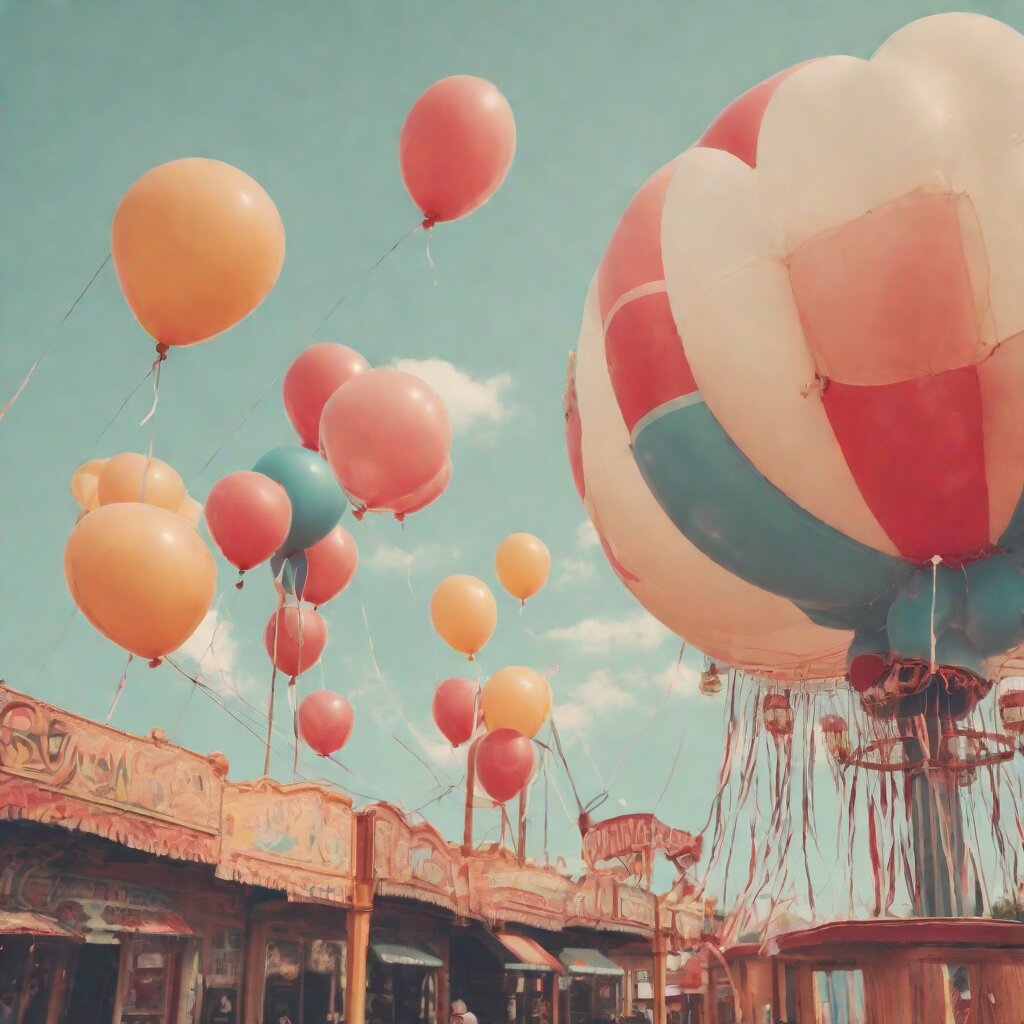}
        \end{minipage} 
        \vspace{-2pt} 
        \\
        \vspace{-1pt} 
        \scriptsize ``A rollercoaster" & \scriptsize ``Balloons" &
        \multicolumn{5}{c}{\scriptsize ``...in retro amusement park style."} \\
        \begin{minipage}{0.12\textwidth}
            \includegraphics[width=\textwidth]{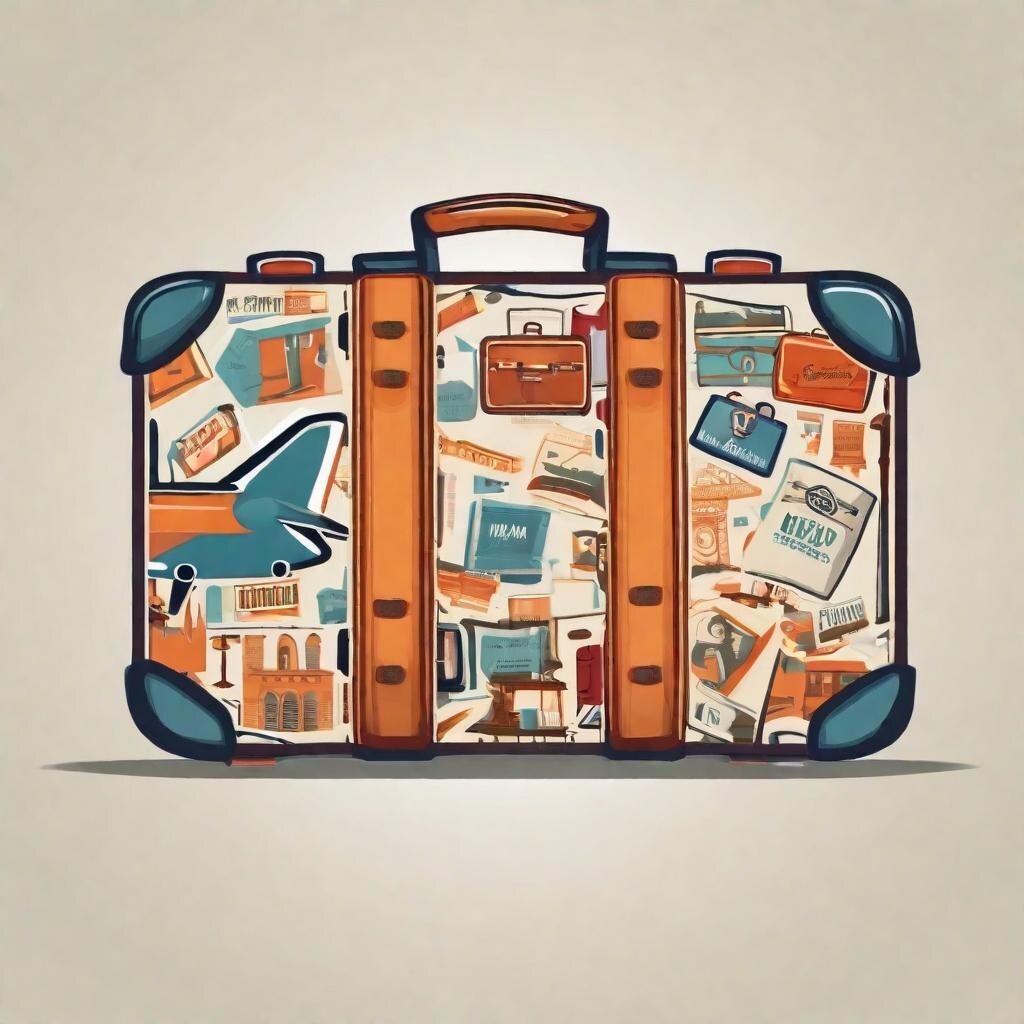}
        \end{minipage} &
        \begin{minipage}{0.12\textwidth}
            \includegraphics[width=\textwidth]{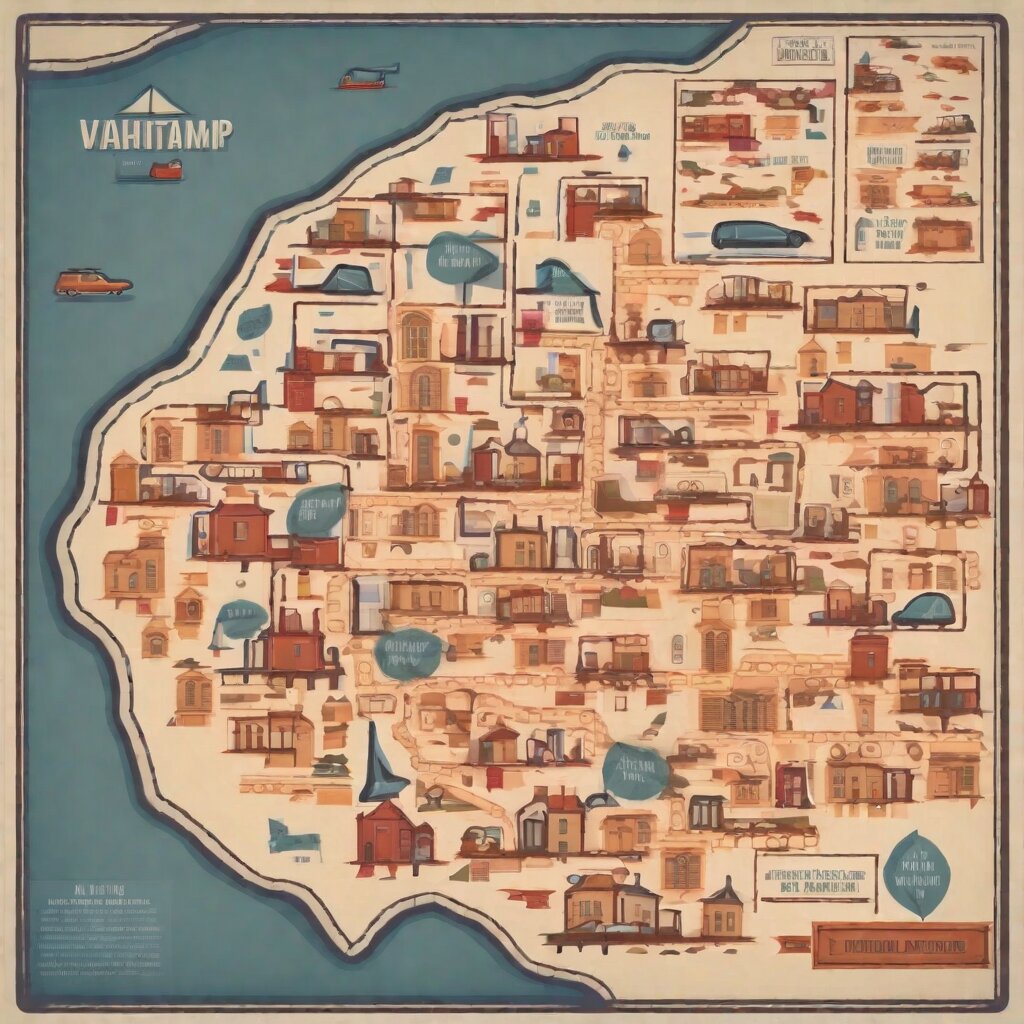}
        \end{minipage} &
        \begin{minipage}{0.12\textwidth}
            \includegraphics[width=\textwidth]{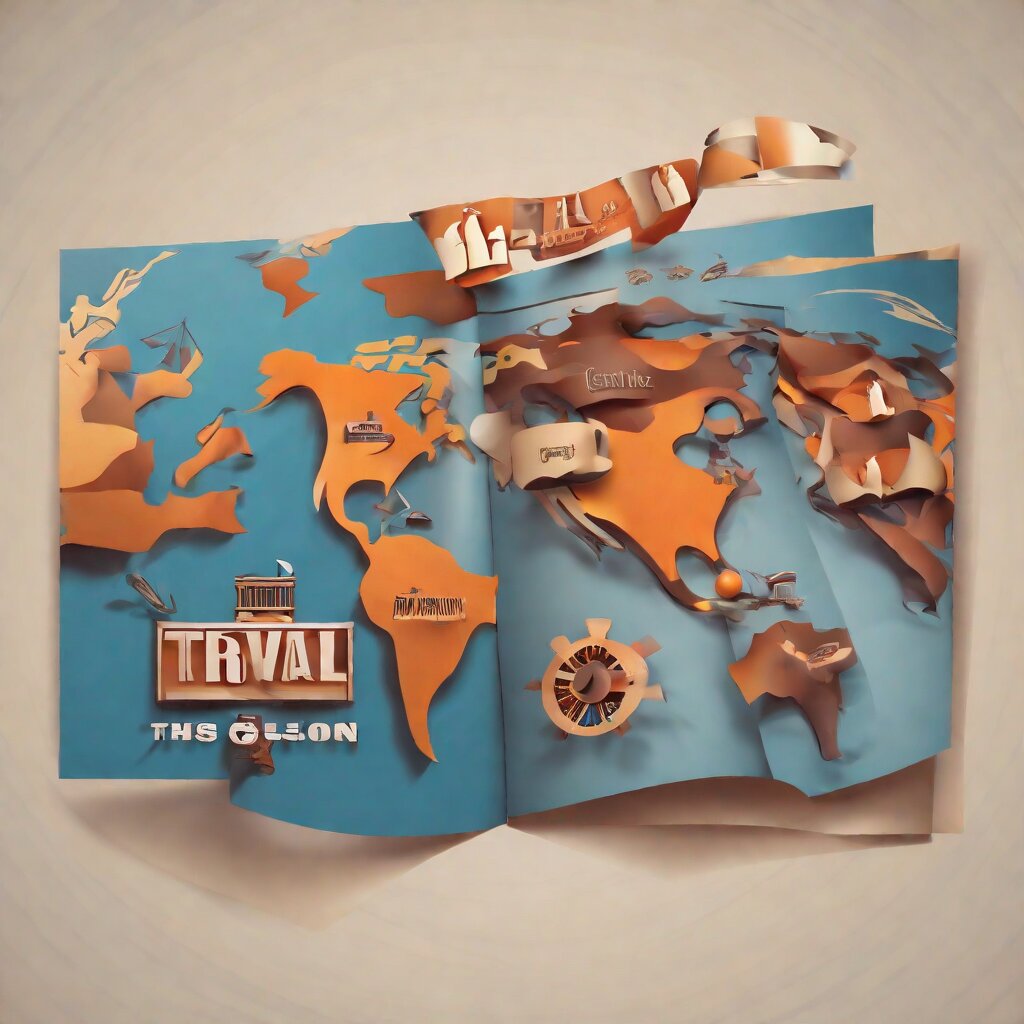}
        \end{minipage} &
        \begin{minipage}{0.12\textwidth}
            \includegraphics[width=\textwidth]{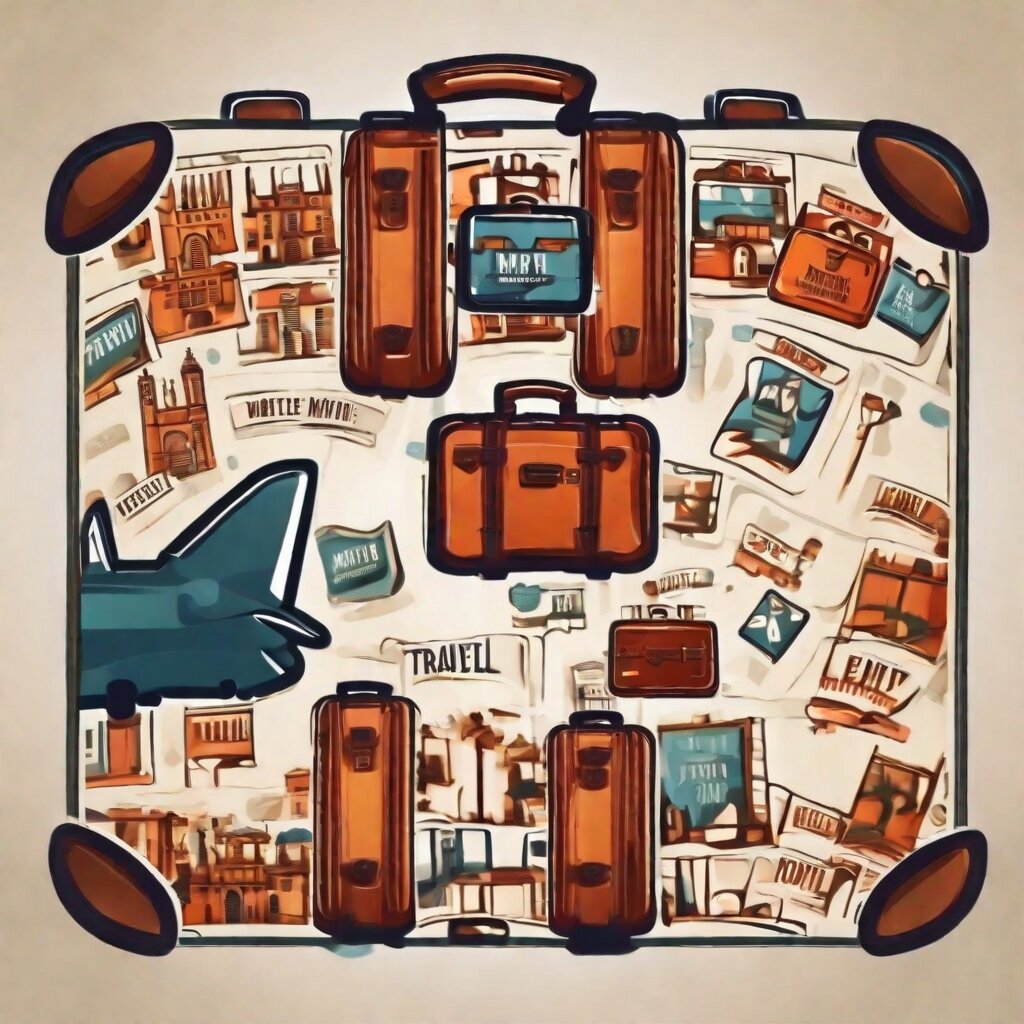}
        \end{minipage} &
        \begin{minipage}{0.12\textwidth}
            \includegraphics[width=\textwidth]{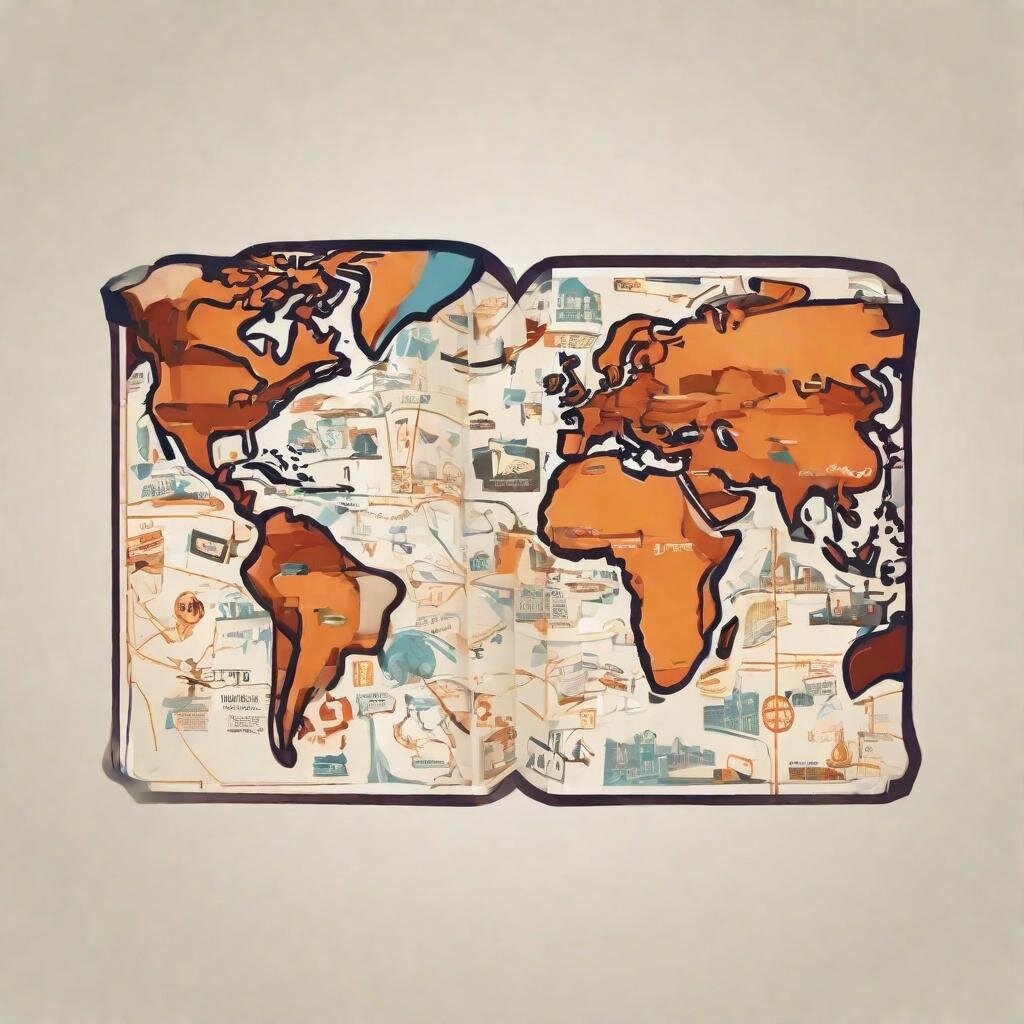}
        \end{minipage} &
        \begin{minipage}{0.12\textwidth}
            \includegraphics[width=\textwidth]{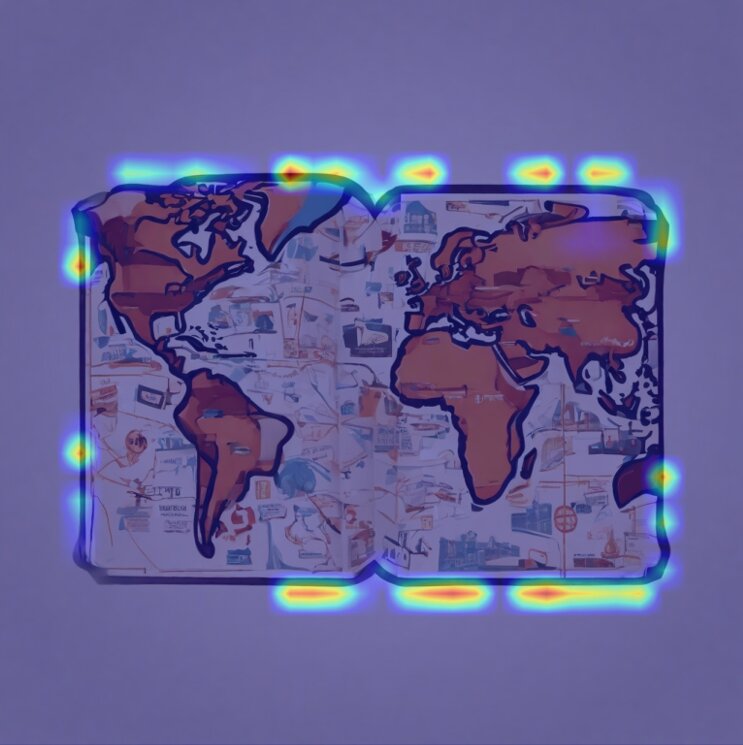}
        \end{minipage} &
        \begin{minipage}{0.12\textwidth}
            \includegraphics[width=\textwidth]{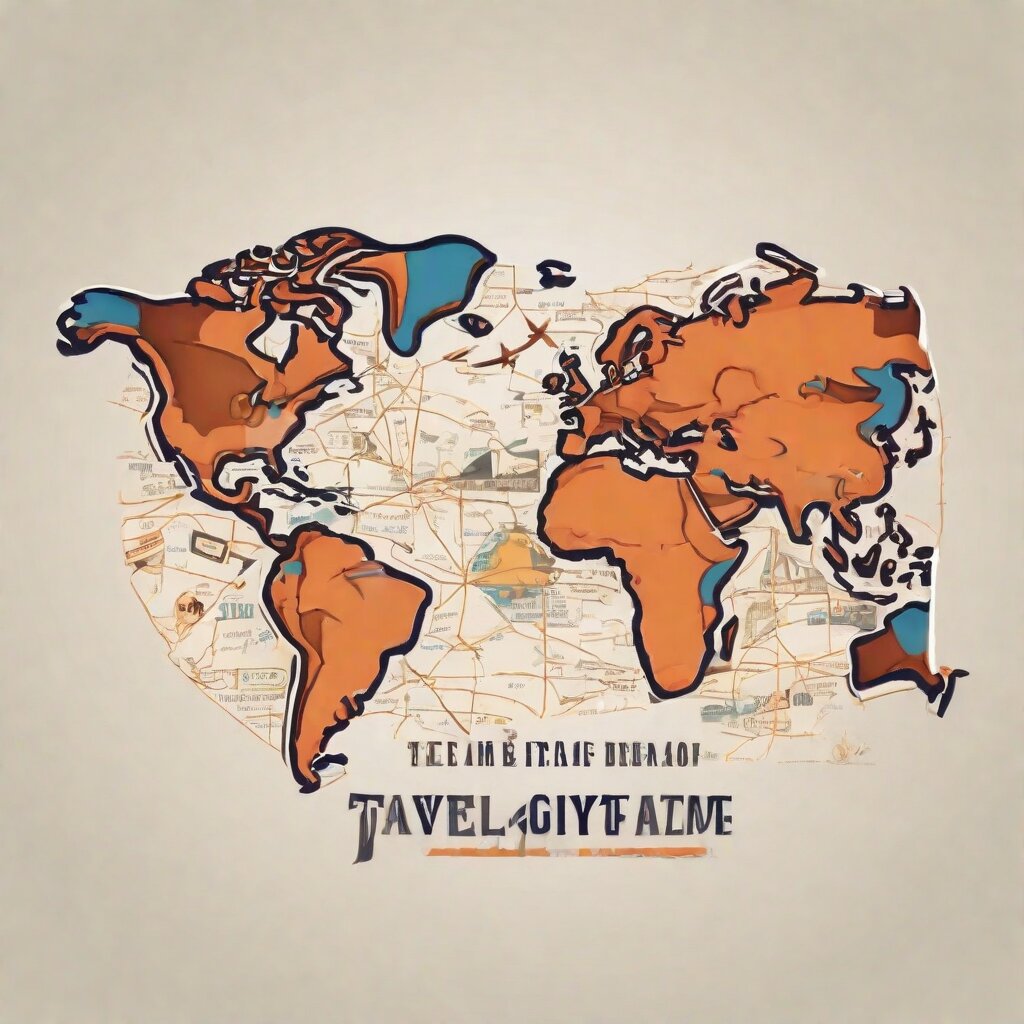}
        \end{minipage} 
        \vspace{-2pt} 
        \\
        \vspace{-1pt} 
        \scriptsize ``A suitcase" & \scriptsize ``A map" &
        \multicolumn{5}{c}{\scriptsize ``...in travel agency logo style."} \\

        \begin{minipage}{0.12\textwidth}
            \includegraphics[width=\textwidth]{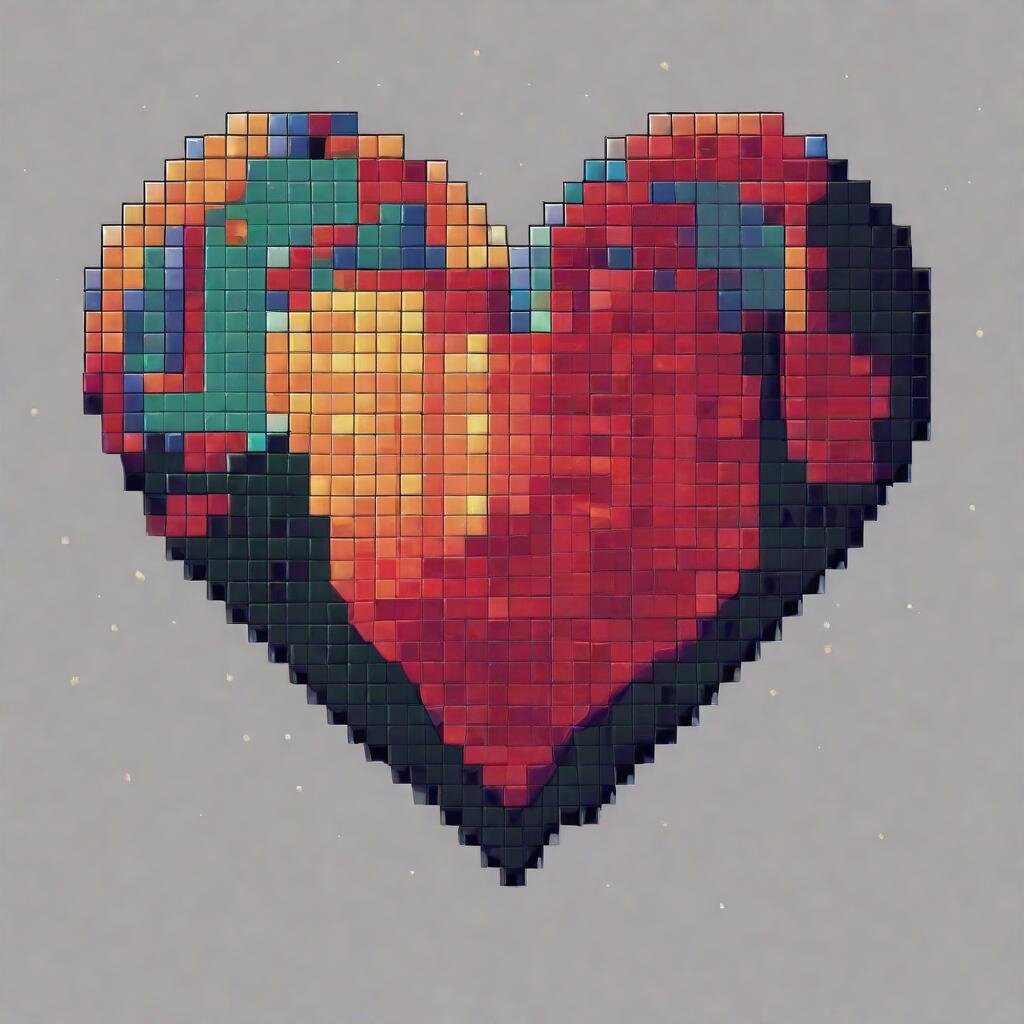}
        \end{minipage} &
        \begin{minipage}{0.12\textwidth}
            \includegraphics[width=\textwidth]{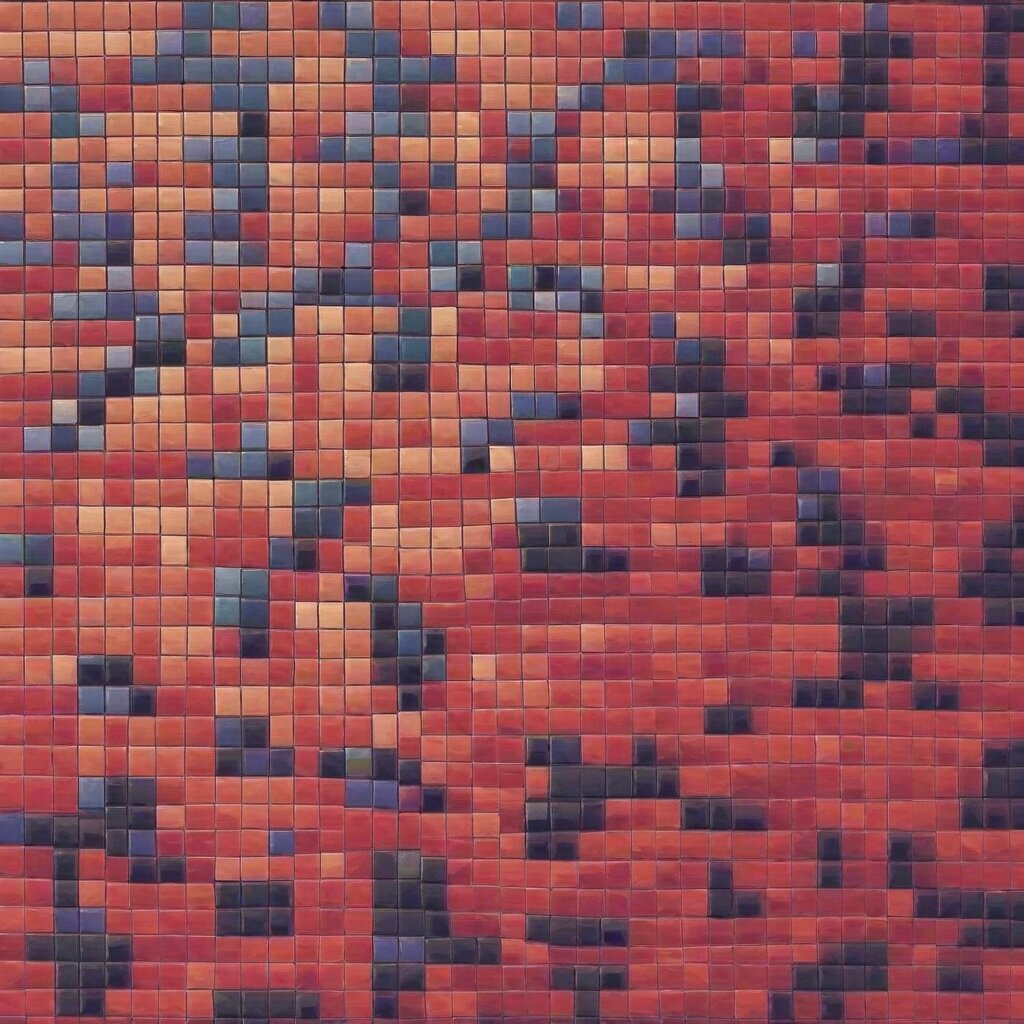}
        \end{minipage} &
        \begin{minipage}{0.12\textwidth}
            \includegraphics[width=\textwidth]{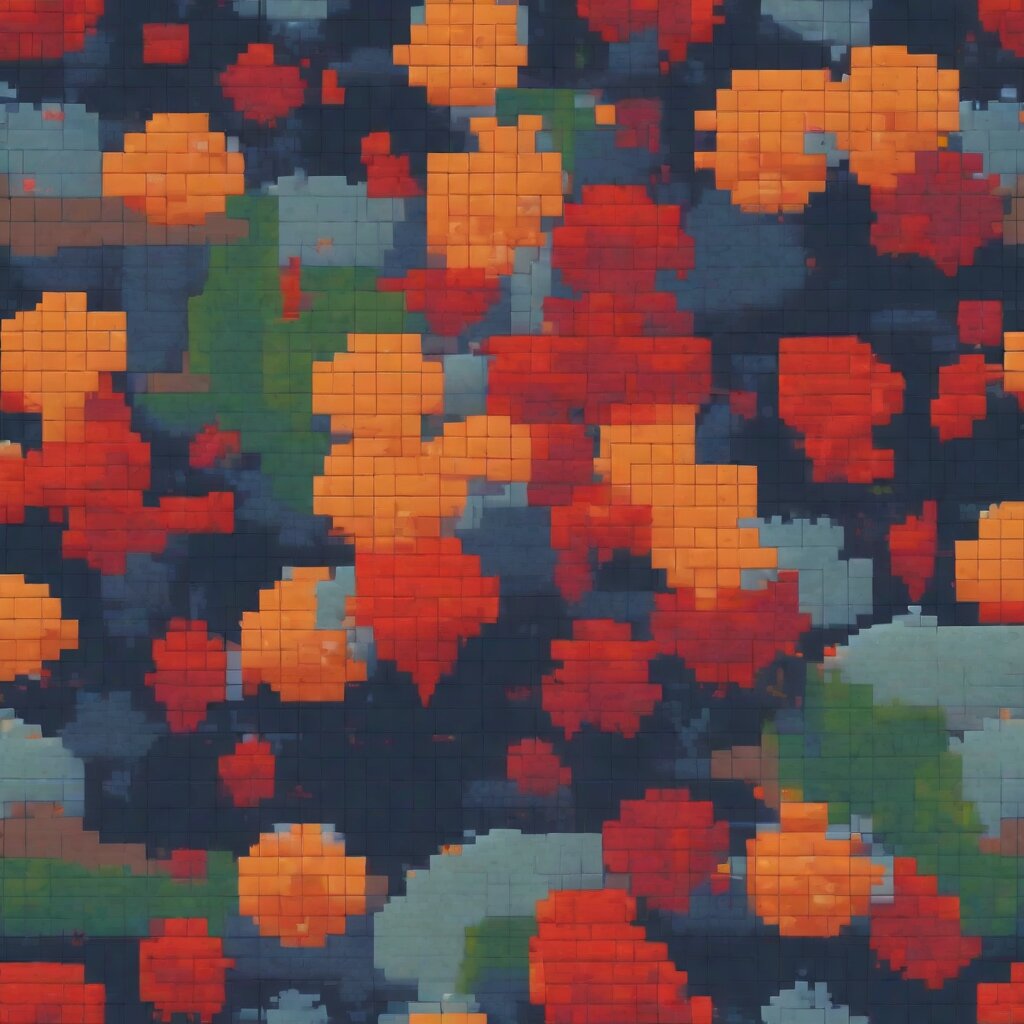}
        \end{minipage} &
        \begin{minipage}{0.12\textwidth}
            \includegraphics[width=\textwidth]{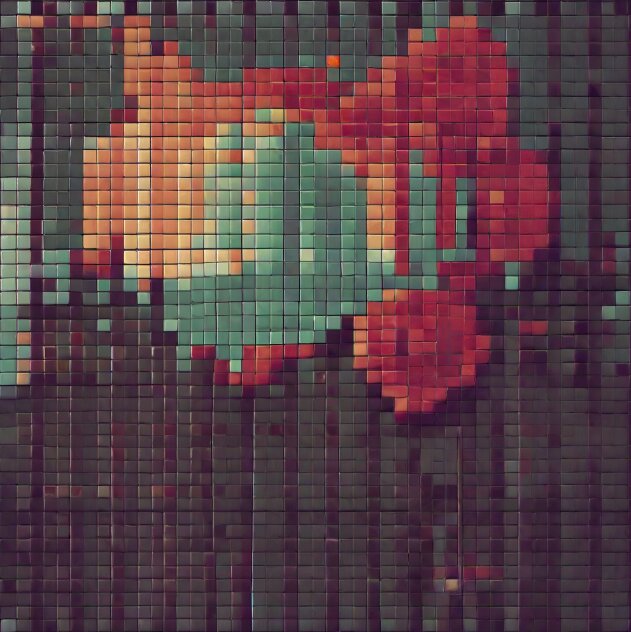}
        \end{minipage} &
        \begin{minipage}{0.12\textwidth}
            \includegraphics[width=\textwidth]{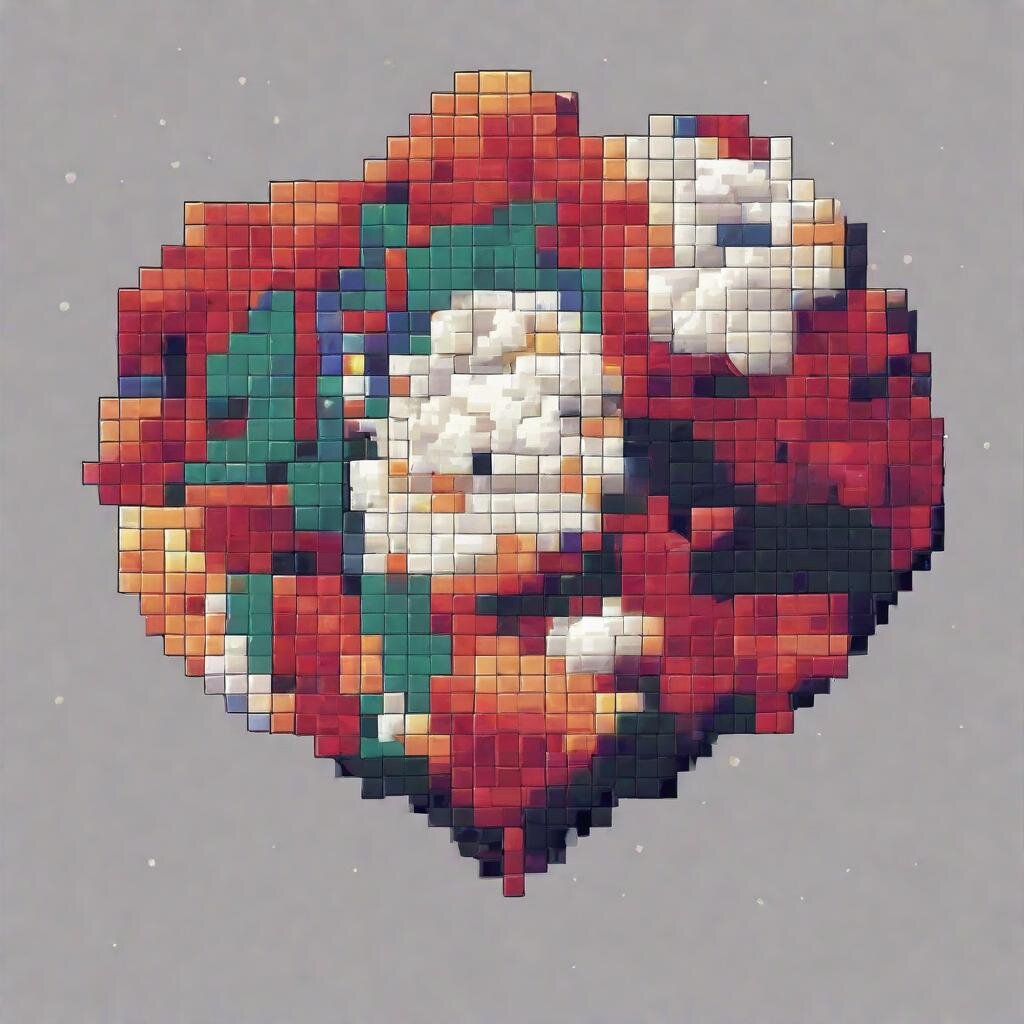}
        \end{minipage} &
        \begin{minipage}{0.12\textwidth}
            \includegraphics[width=\textwidth]{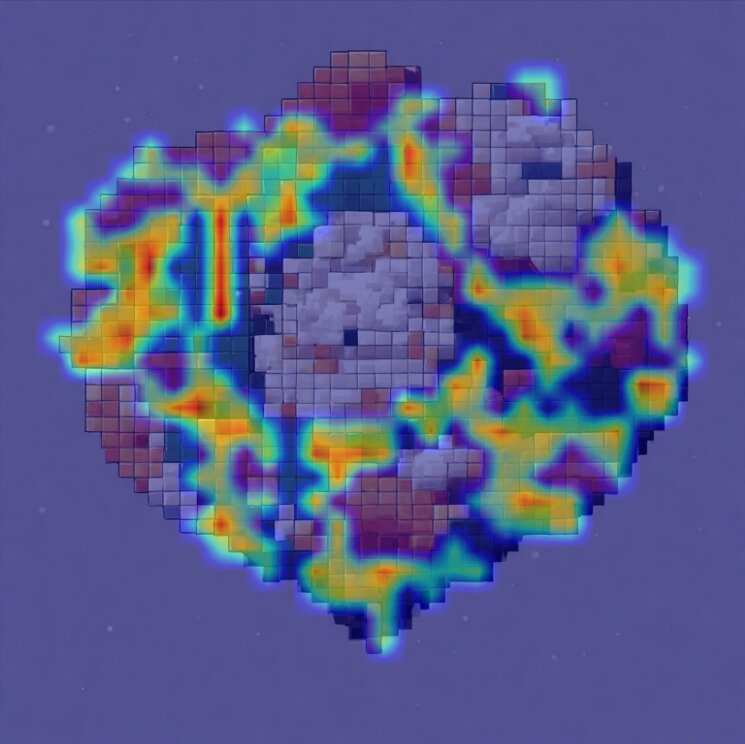}
        \end{minipage} &
        \begin{minipage}{0.12\textwidth}
            \includegraphics[width=\textwidth]{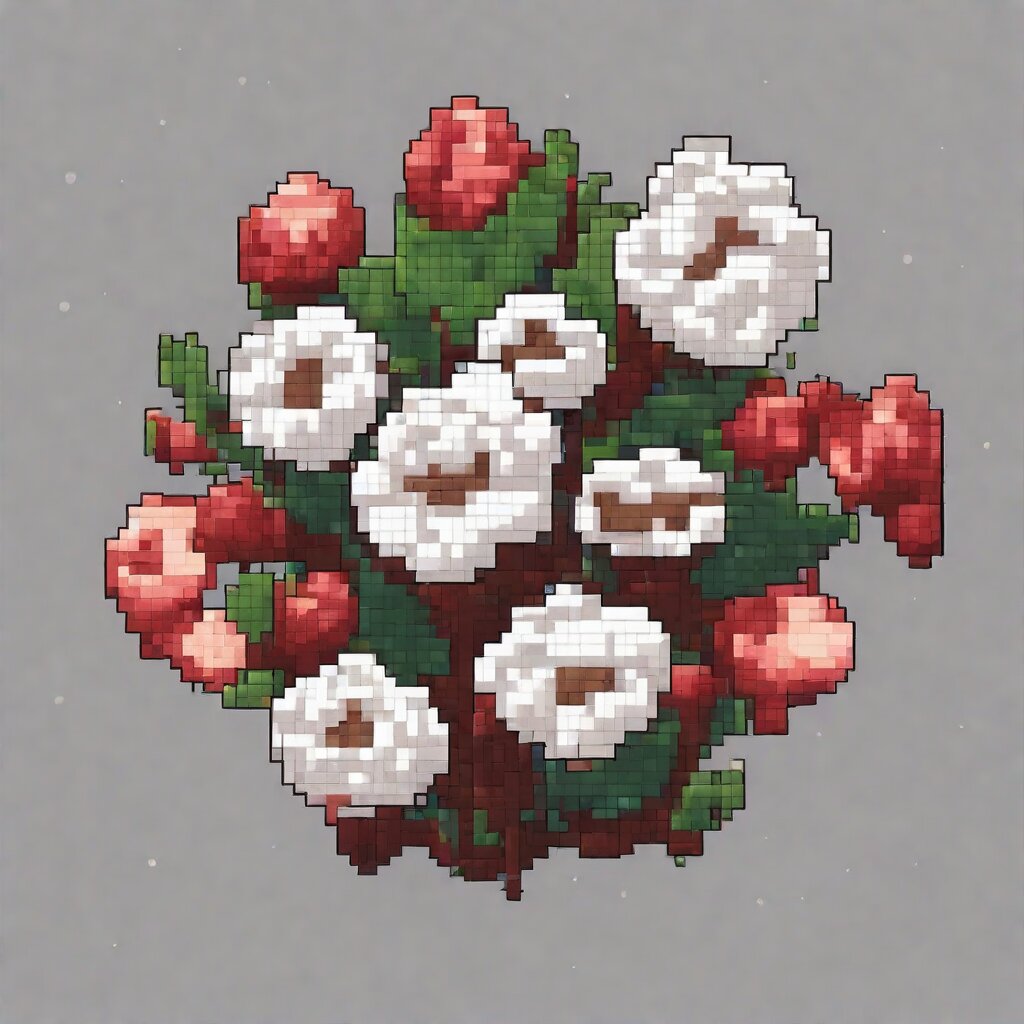}
        \end{minipage} 
        \vspace{-2pt} 
        \\
        \vspace{-1pt} 
        \scriptsize ``A heart" & \scriptsize ``Cotton" &
        \multicolumn{5}{c}{\scriptsize ``...in pixel art style."} \\

        \begin{minipage}{0.12\textwidth}
            \includegraphics[width=\textwidth]{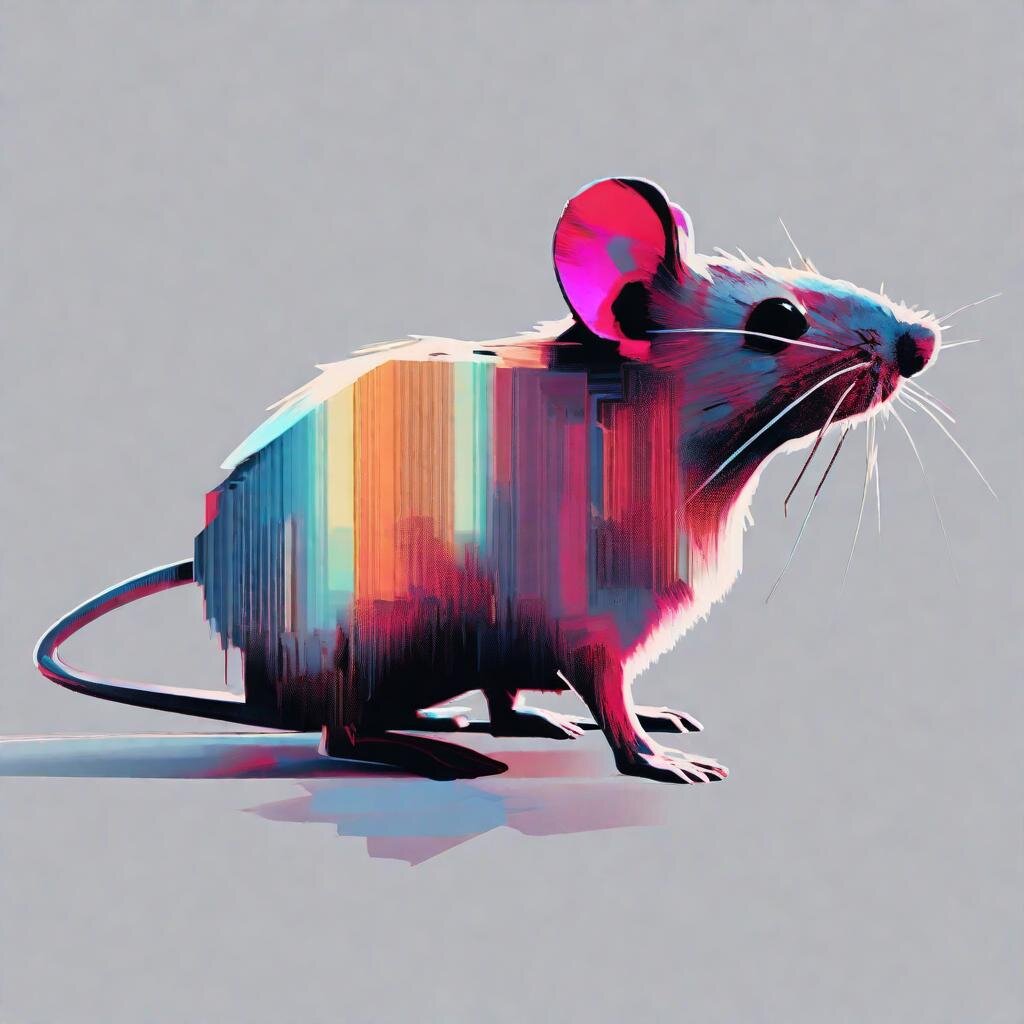}
        \end{minipage} &
        \begin{minipage}{0.12\textwidth}
            \includegraphics[width=\textwidth]{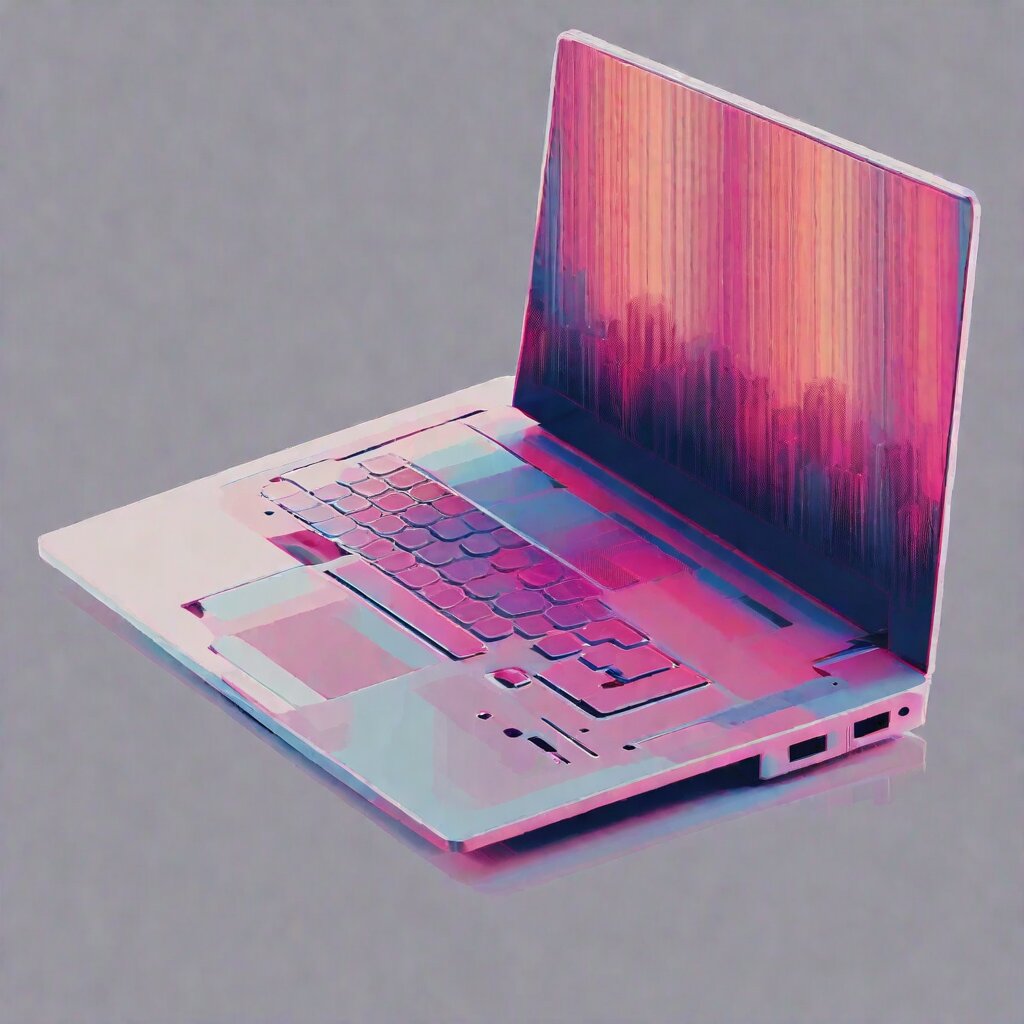}
        \end{minipage} &
        \begin{minipage}{0.12\textwidth}
            \includegraphics[width=\textwidth]{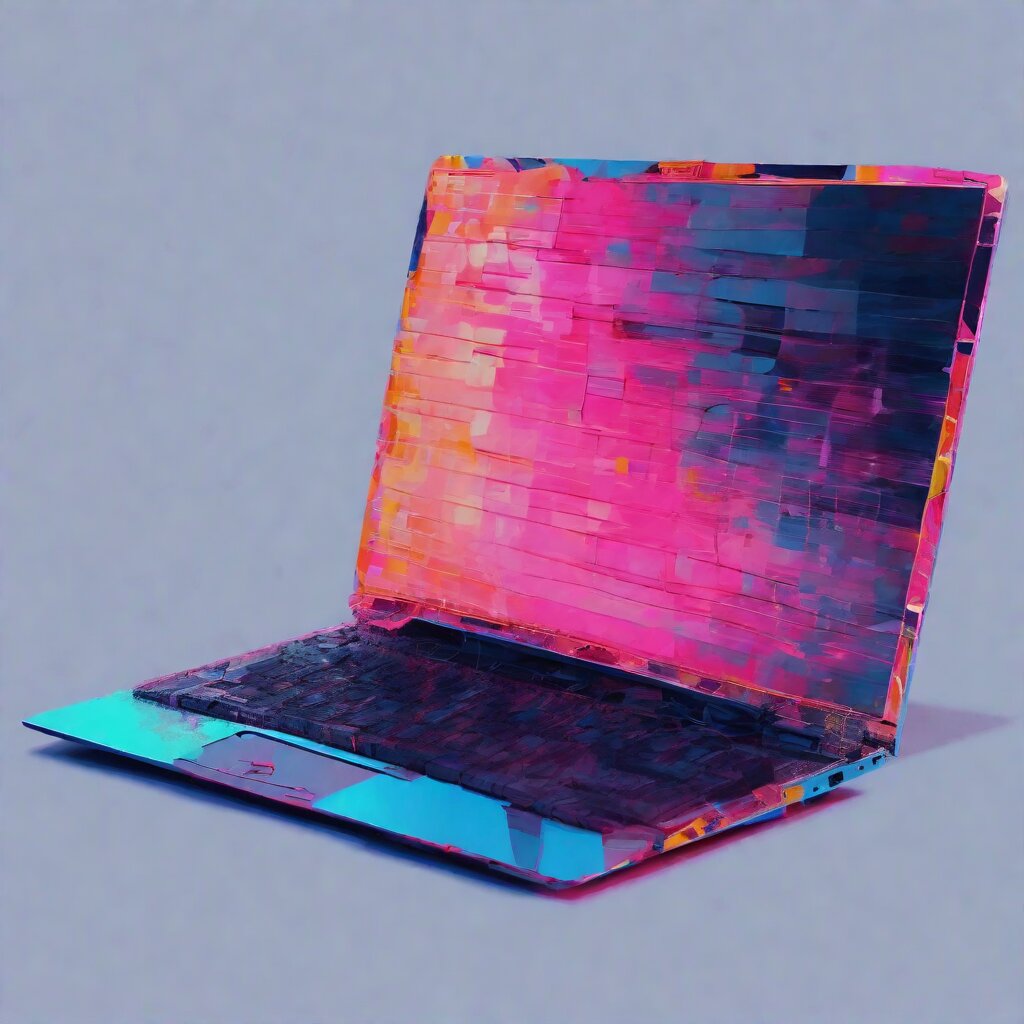}
        \end{minipage} &
        \begin{minipage}{0.12\textwidth}
            \includegraphics[width=\textwidth]{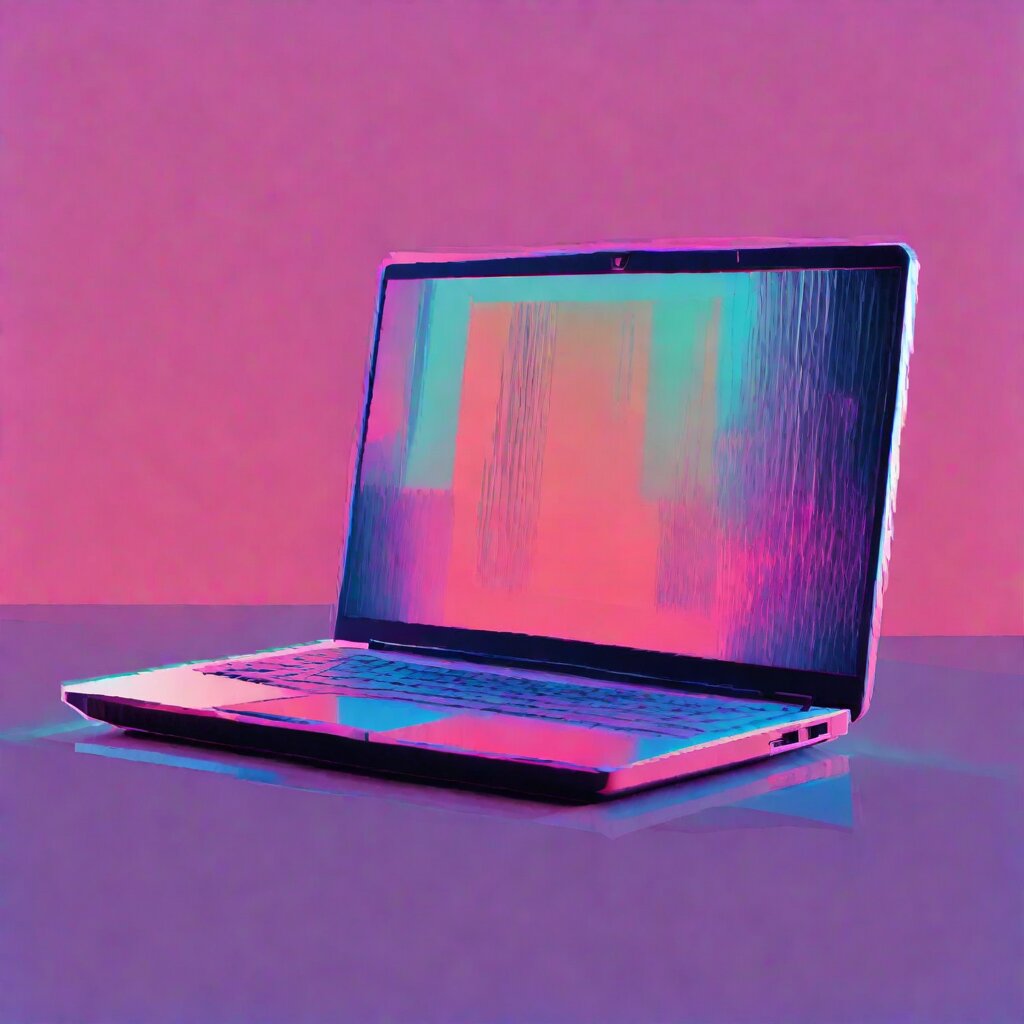}
        \end{minipage} &
        \begin{minipage}{0.12\textwidth}
            \includegraphics[width=\textwidth]{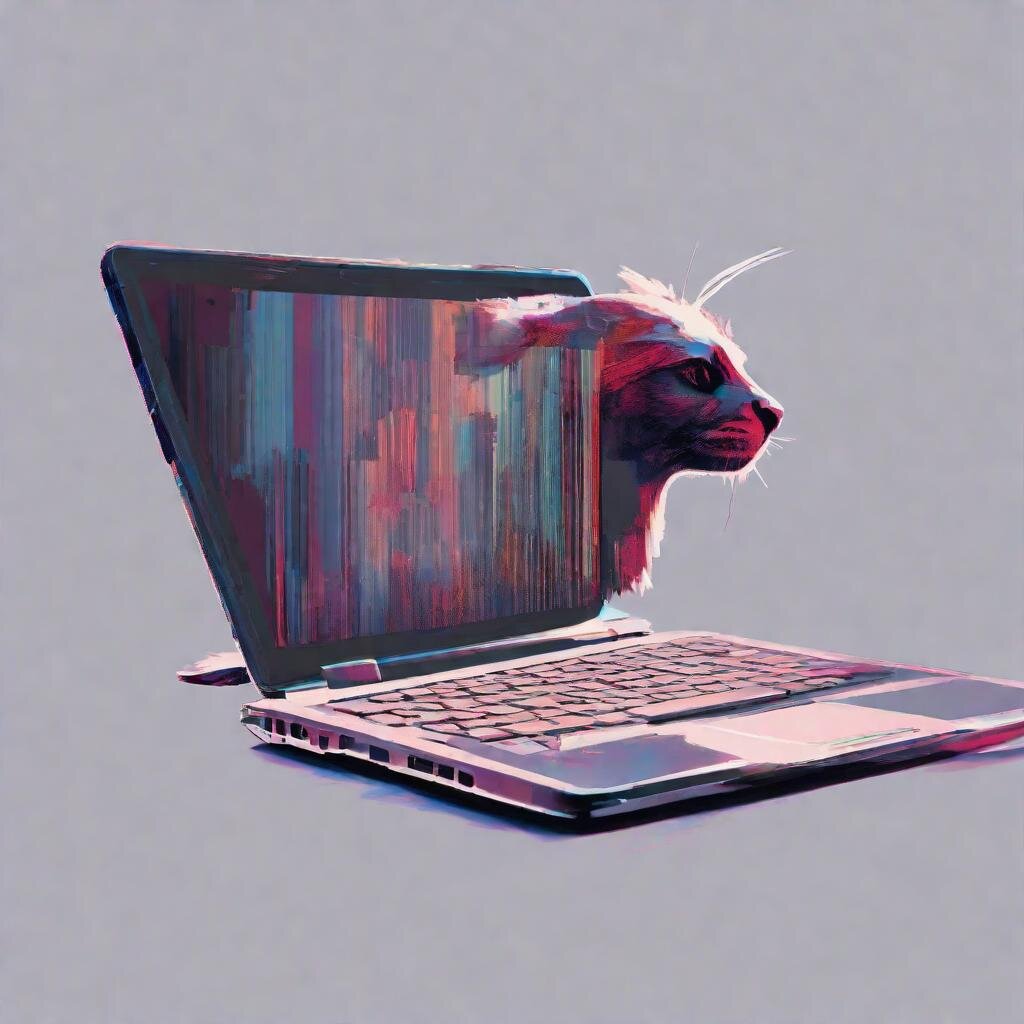}
        \end{minipage} &
        \begin{minipage}{0.12\textwidth}
            \includegraphics[width=\textwidth]{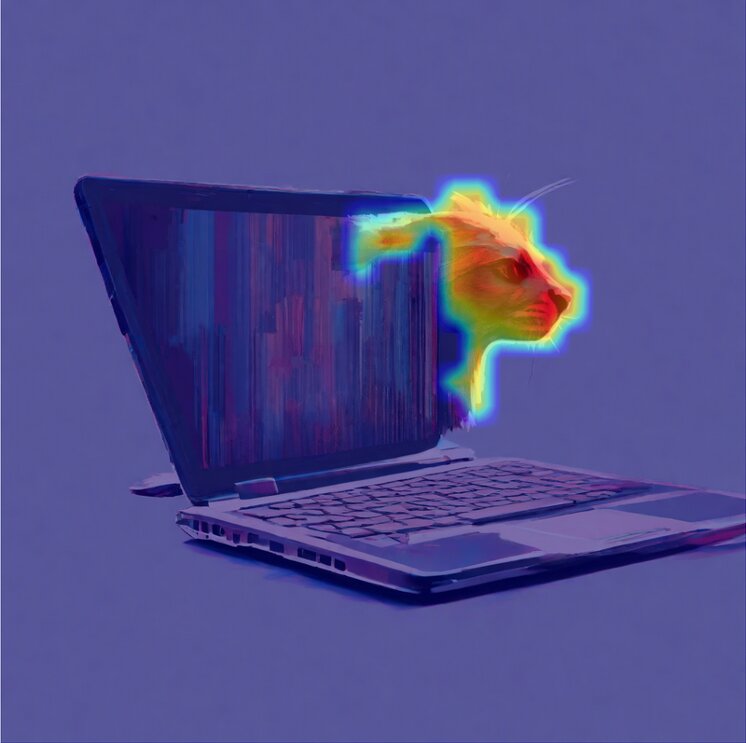}
        \end{minipage} &
        \begin{minipage}{0.12\textwidth}
            \includegraphics[width=\textwidth]{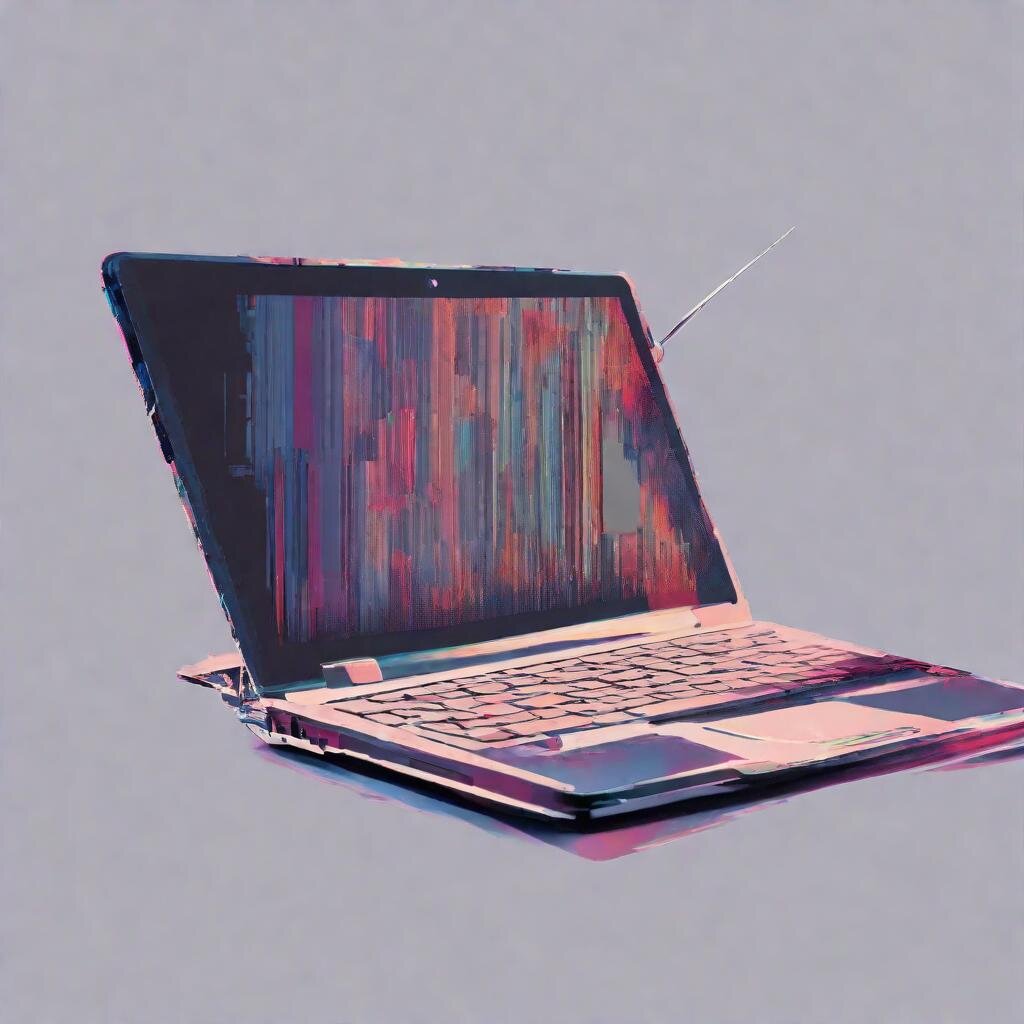}
        \end{minipage} 
        \vspace{-2pt} 
        \\
        \vspace{-1pt} 
        \scriptsize ``A mouse" & \scriptsize ``A laptop" &
        \multicolumn{5}{c}{\scriptsize ``...in digital glitch style."} \\
        \begin{minipage}{0.12\textwidth}
            \includegraphics[width=\textwidth]{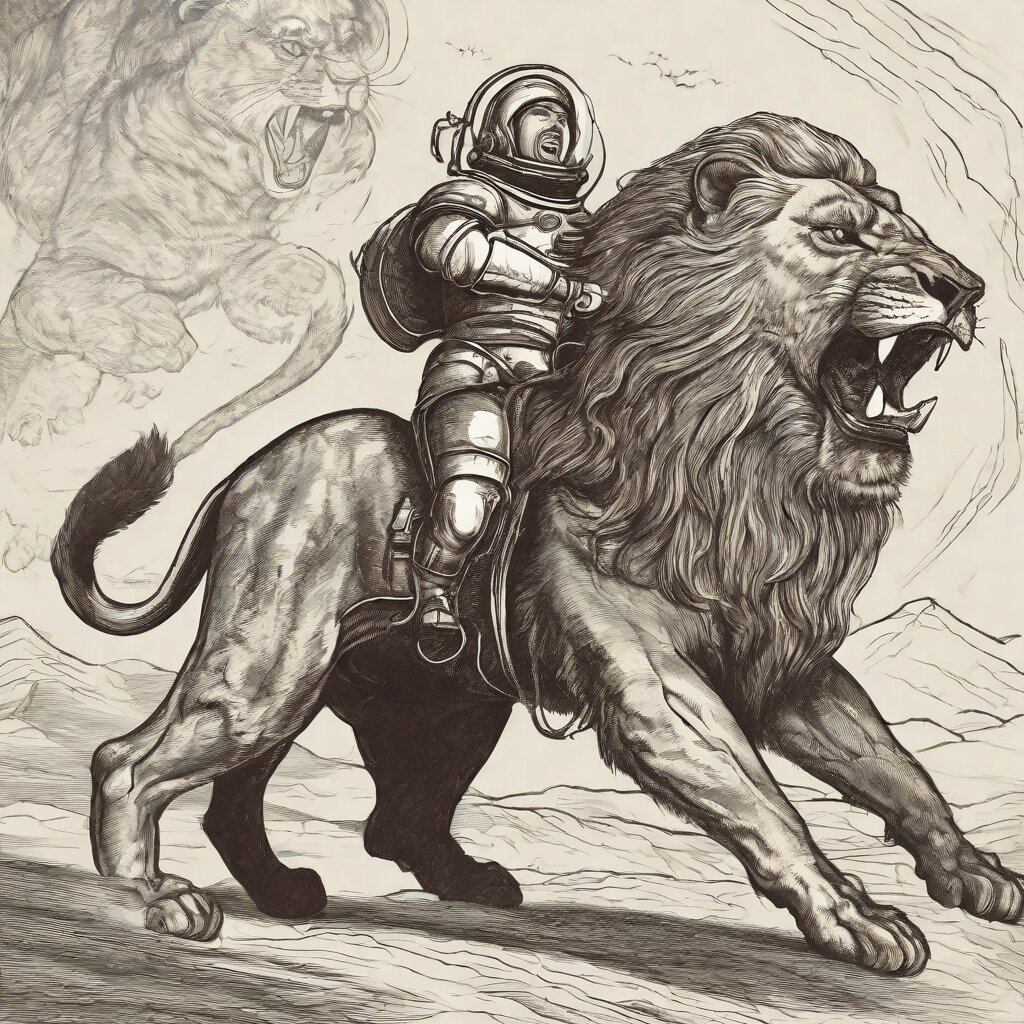}
        \end{minipage} &
        \begin{minipage}{0.12\textwidth}
            \includegraphics[width=\textwidth]{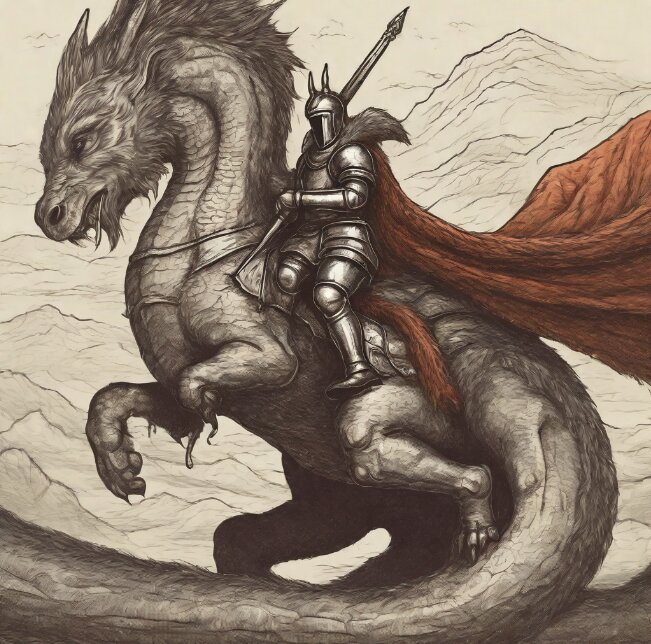}
        \end{minipage} &
        \begin{minipage}{0.12\textwidth}
            \includegraphics[width=\textwidth]{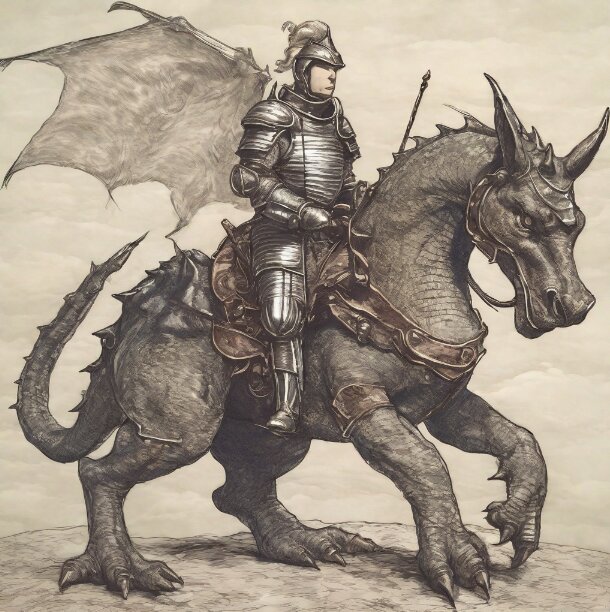}
        \end{minipage} &
        \begin{minipage}{0.12\textwidth}
            \includegraphics[width=\textwidth]{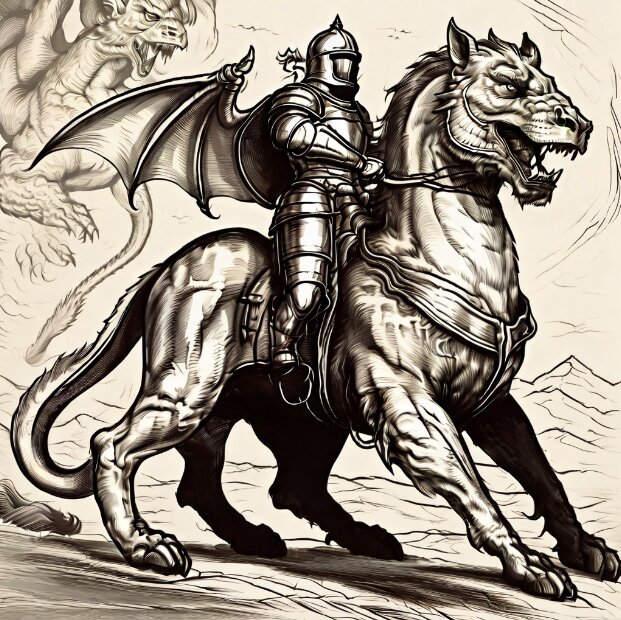}
        \end{minipage} &
        \begin{minipage}{0.12\textwidth}
            \includegraphics[width=\textwidth]{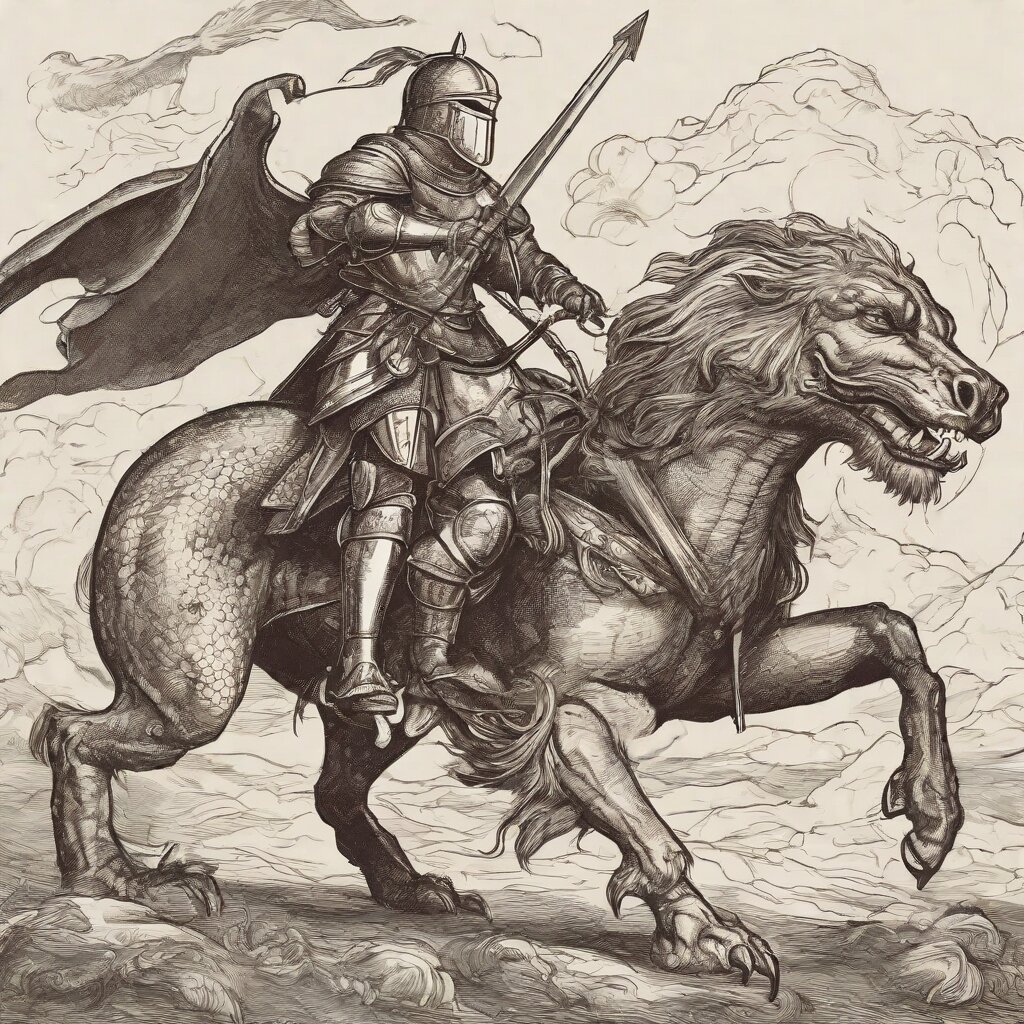}
        \end{minipage} &
        \begin{minipage}{0.12\textwidth}
            \includegraphics[width=\textwidth]{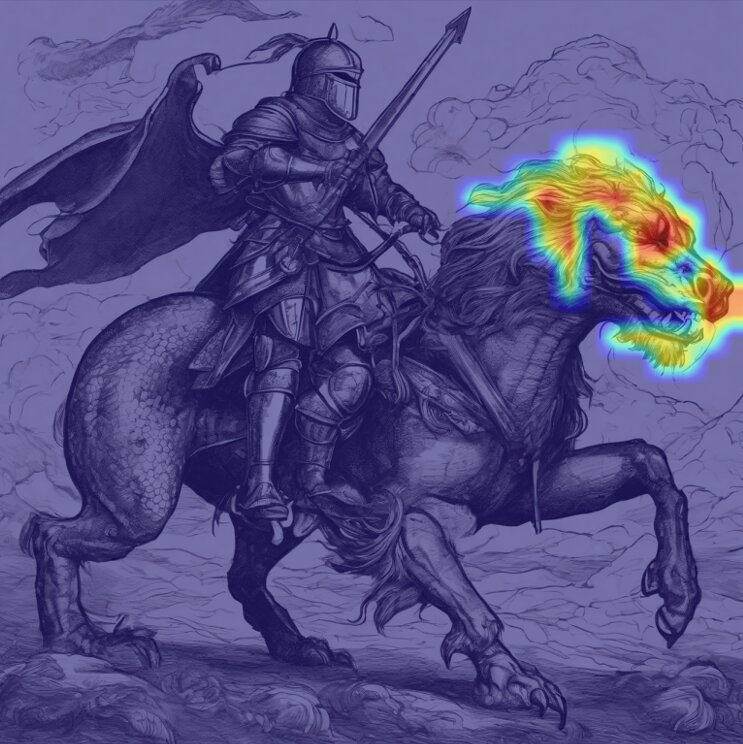}
        \end{minipage} &
        \begin{minipage}{0.12\textwidth}
            \includegraphics[width=\textwidth]{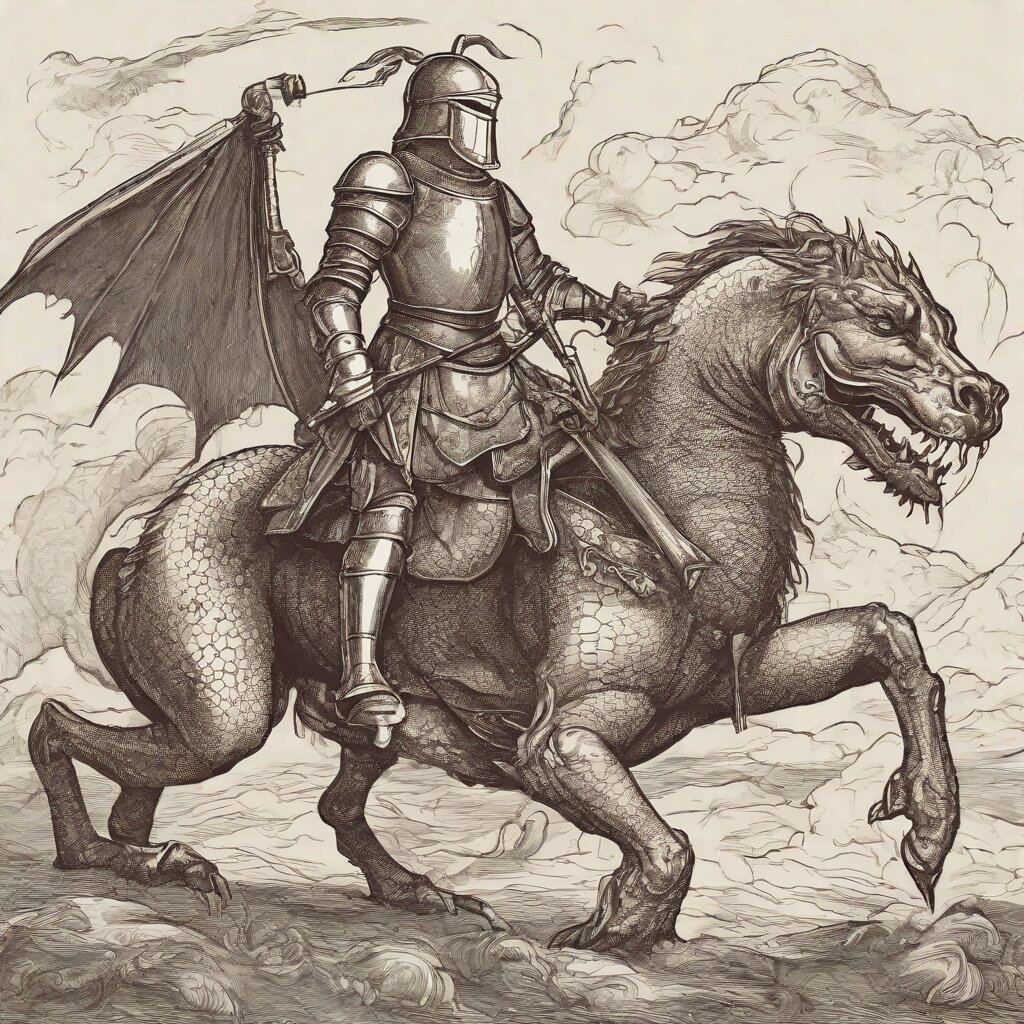}
        \end{minipage} 
        
        \\[-2pt]
        
        \multicolumn{2}{c}{\scriptsize ``An austronaut on a roaring lion"} & \multicolumn{2}{c}{\scriptsize ``A knight on a dragon"} & \multicolumn{3}{c}{\scriptsize ``...in medieval fantasy illustration style."} \\

        \cmidrule(lr){1-7}
        
        \begin{minipage}{0.12\textwidth}
            \includegraphics[width=\textwidth]{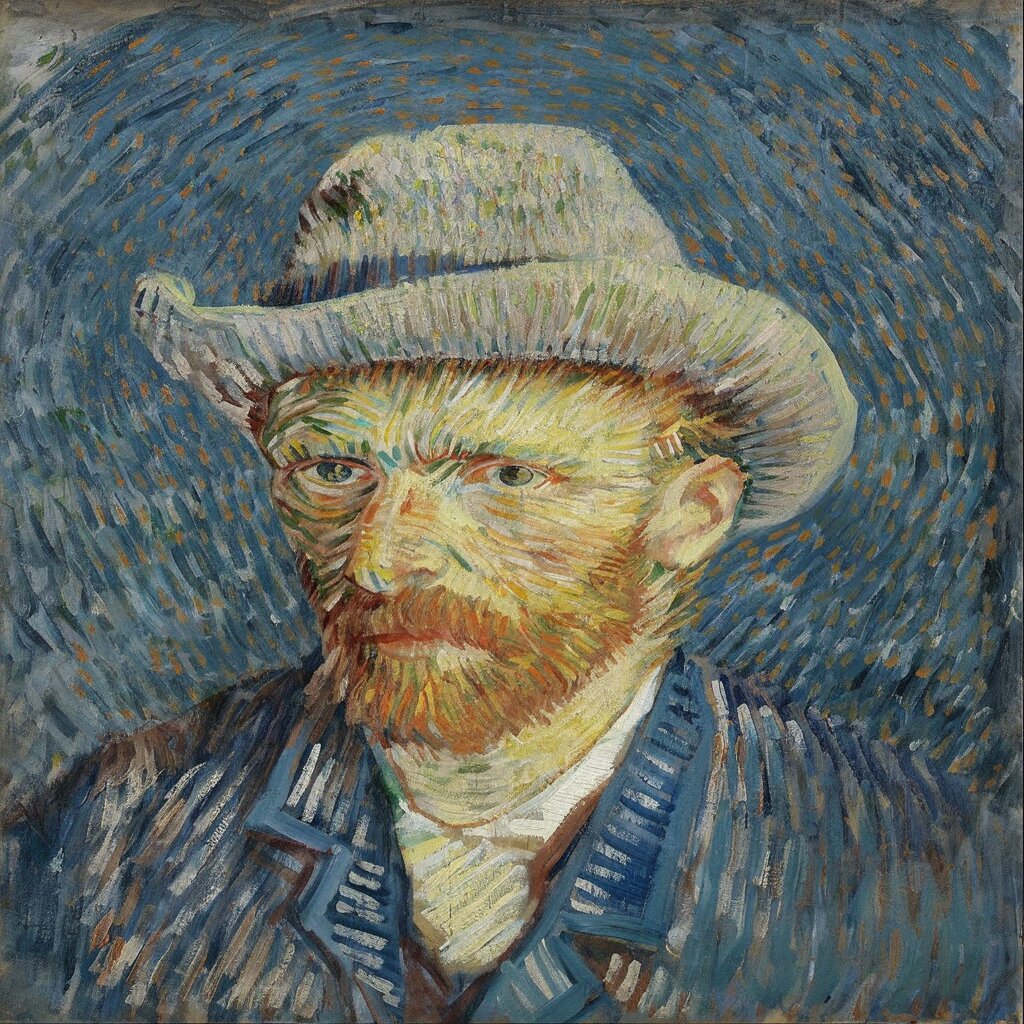}
        \end{minipage} &
        \begin{minipage}{0.12\textwidth}
            \includegraphics[width=\textwidth]{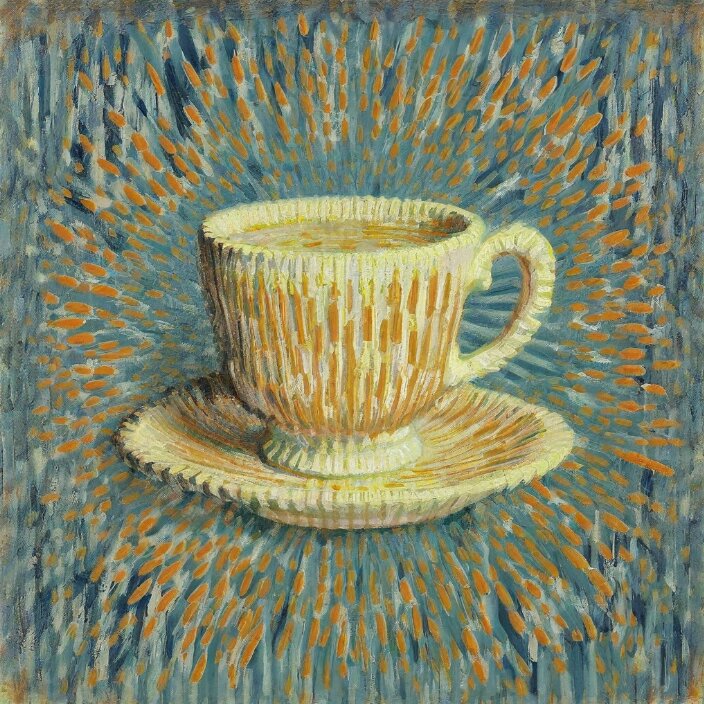}
        \end{minipage} &
        \begin{minipage}{0.12\textwidth}
            \includegraphics[width=\textwidth]{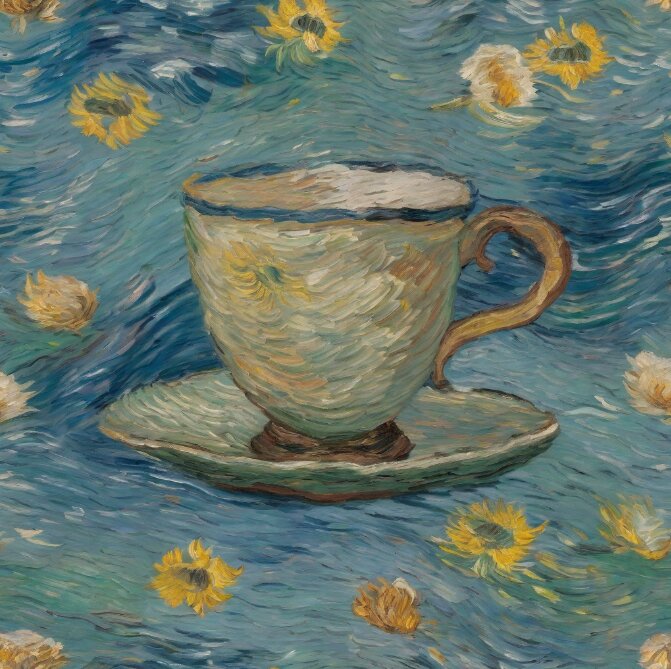}
        \end{minipage} &
        \begin{minipage}{0.12\textwidth}
            \includegraphics[width=\textwidth]{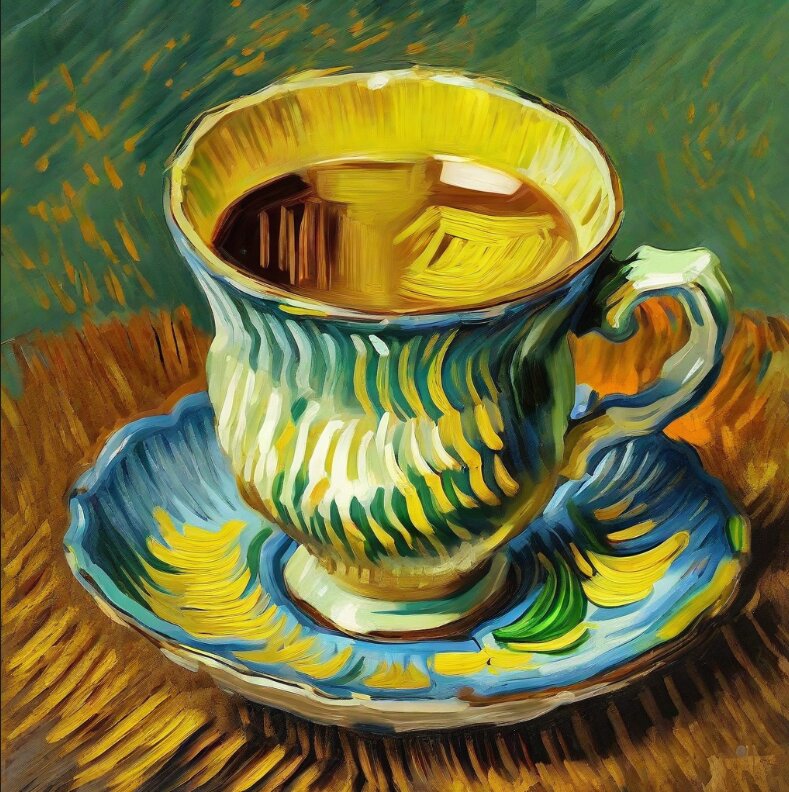}
        \end{minipage} &
        \begin{minipage}{0.12\textwidth}
            \includegraphics[width=\textwidth]{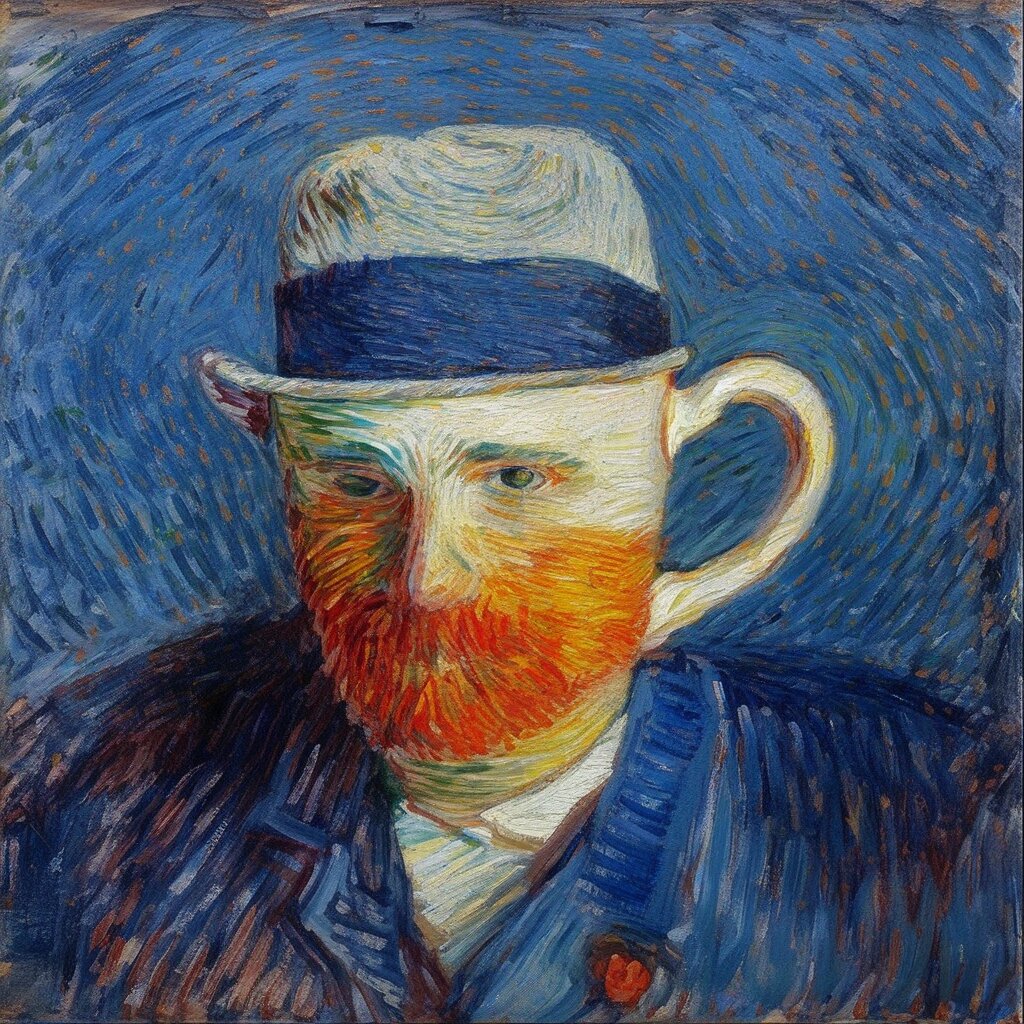}
        \end{minipage} &
        \begin{minipage}{0.12\textwidth}
            \includegraphics[width=\textwidth]{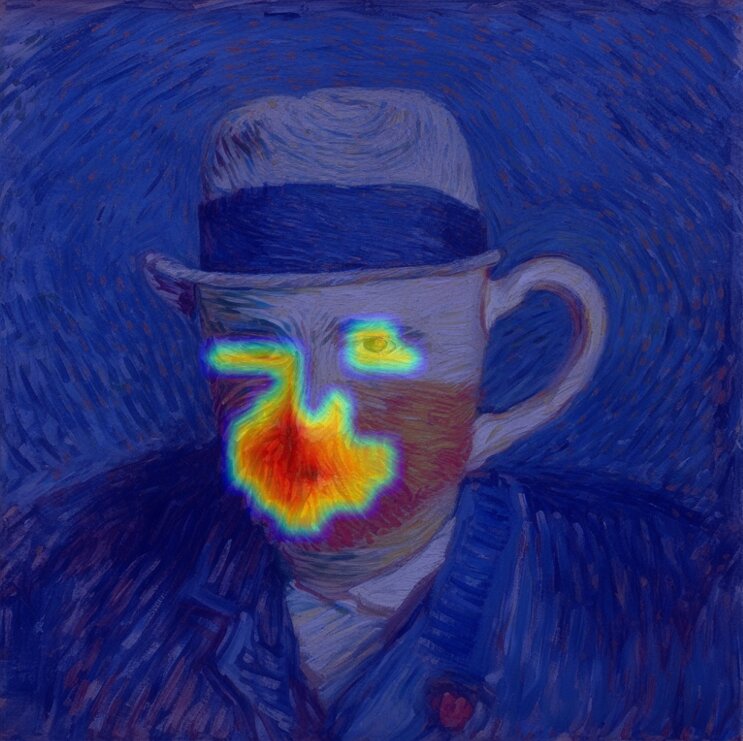}
        \end{minipage} &
        \begin{minipage}{0.12\textwidth}
            \includegraphics[width=\textwidth]{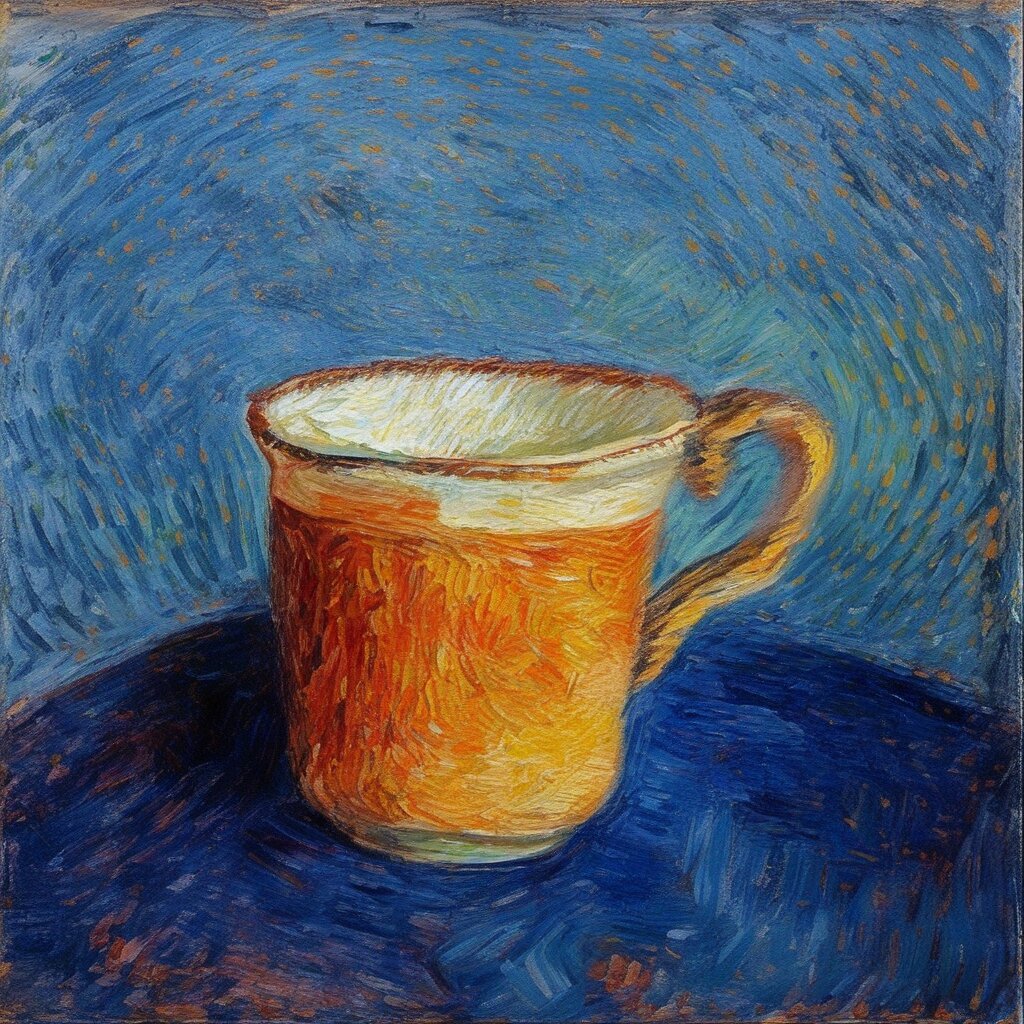}
        \end{minipage}  
        \vspace{-2pt} 
        \\
        \vspace{-1pt} 
        \scriptsize ``A man" & \scriptsize ``A cup" & \multicolumn{5}{c}{\scriptsize ``...in oil painting style."} \\
        
        \begin{minipage}{0.12\textwidth}
            \includegraphics[width=\textwidth]{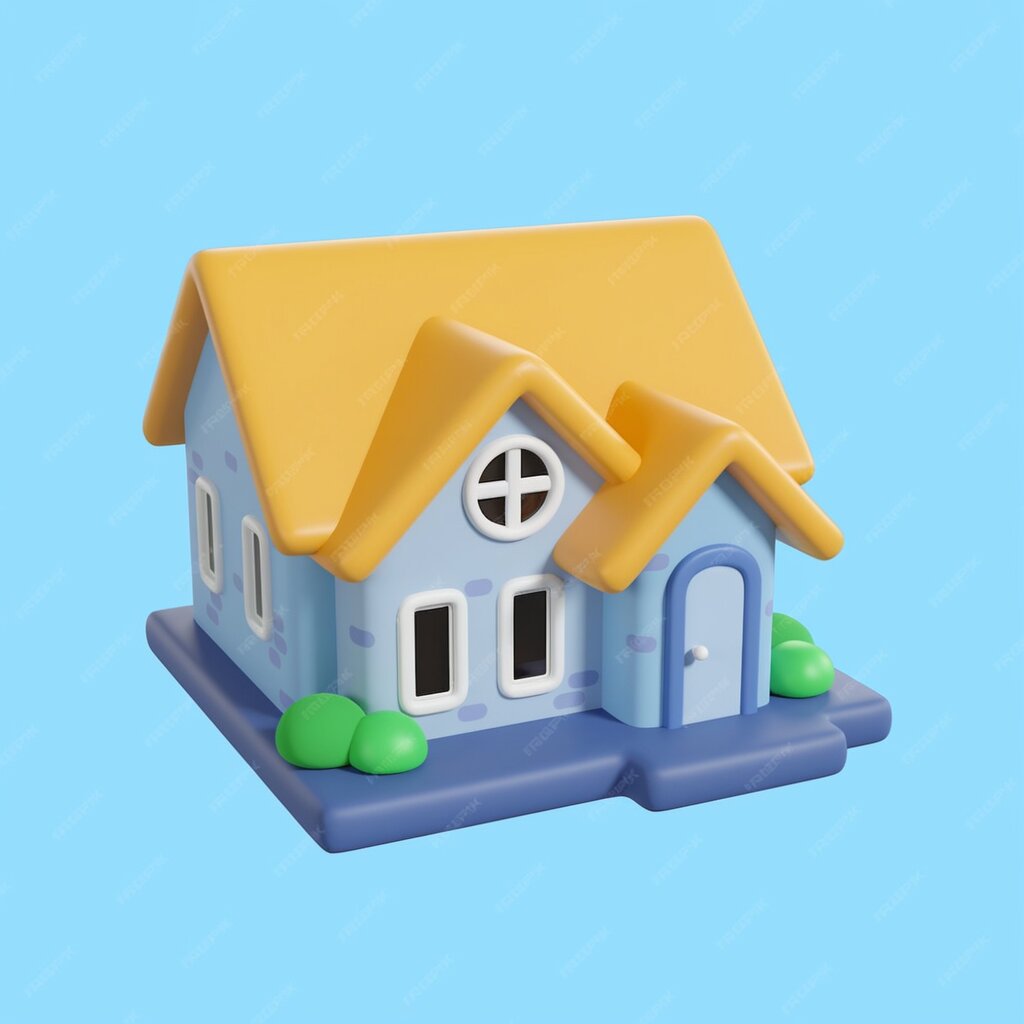}
        \end{minipage} &
        \begin{minipage}{0.12\textwidth}
            \includegraphics[width=\textwidth]{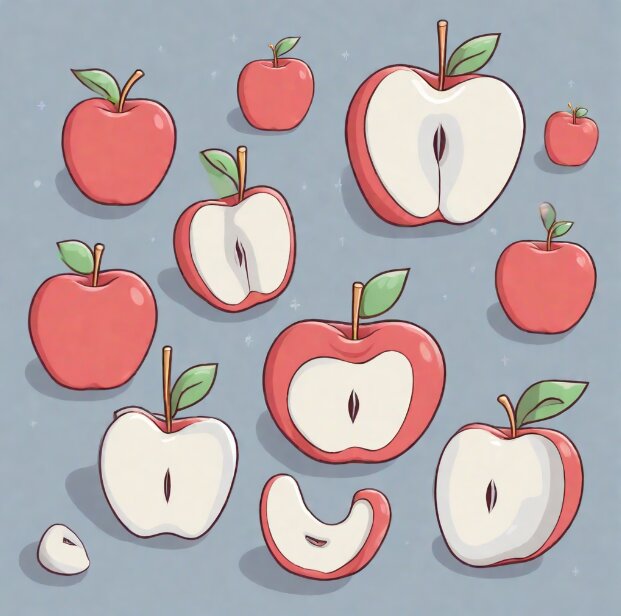}
        \end{minipage} &
        \begin{minipage}{0.12\textwidth}
            \includegraphics[width=\textwidth]{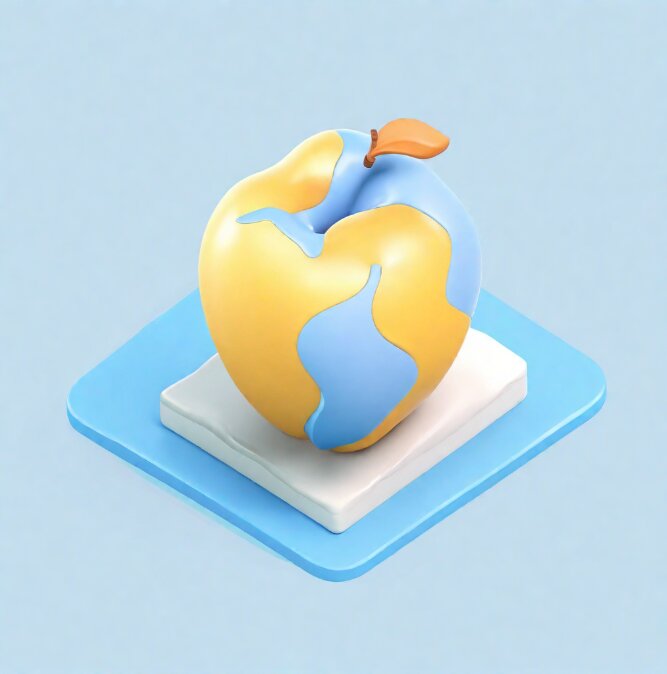}
        \end{minipage} &
        \begin{minipage}{0.12\textwidth}
            \includegraphics[width=\textwidth]{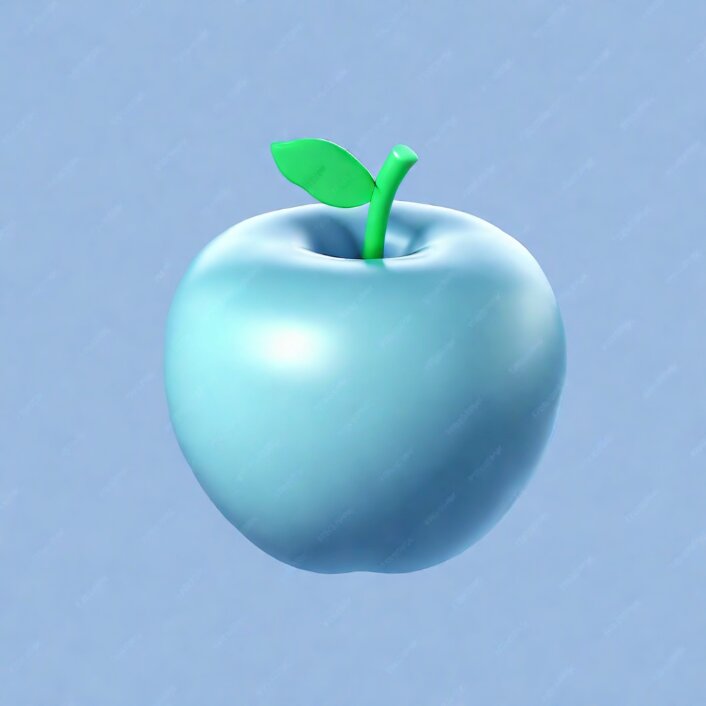}
        \end{minipage} &
        \begin{minipage}{0.12\textwidth}
            \includegraphics[width=\textwidth]{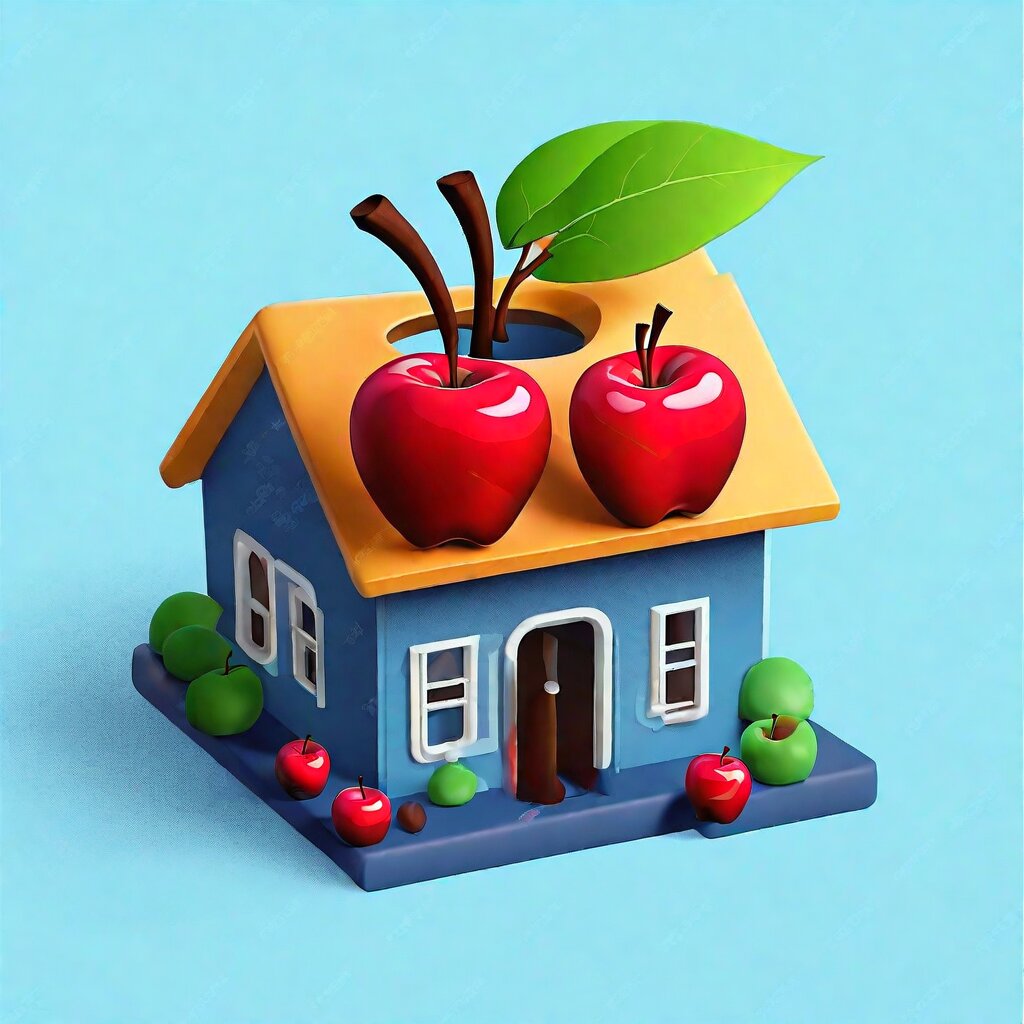}
        \end{minipage} &
        \begin{minipage}{0.12\textwidth}
            \includegraphics[width=\textwidth]{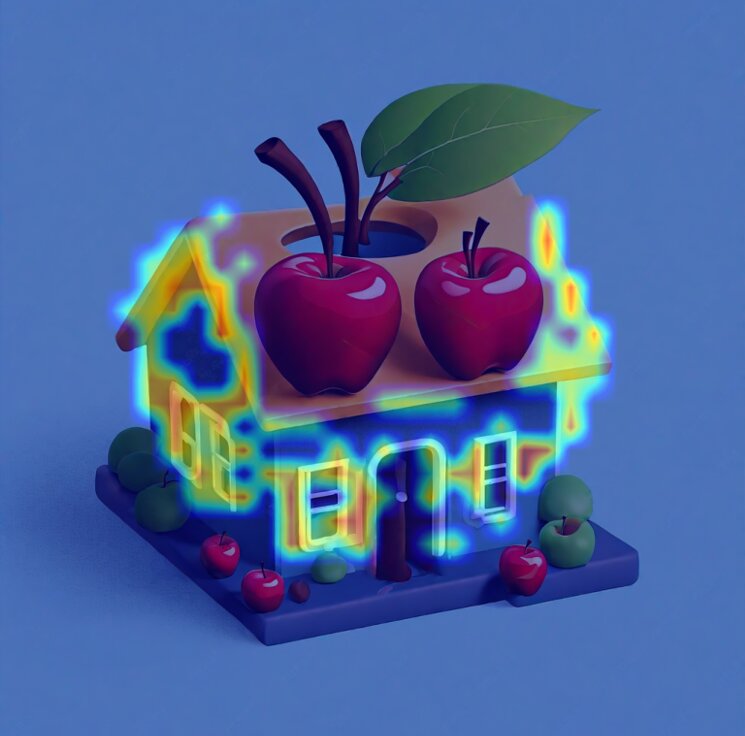}
        \end{minipage} &
        \begin{minipage}{0.12\textwidth}
            \includegraphics[width=\textwidth]{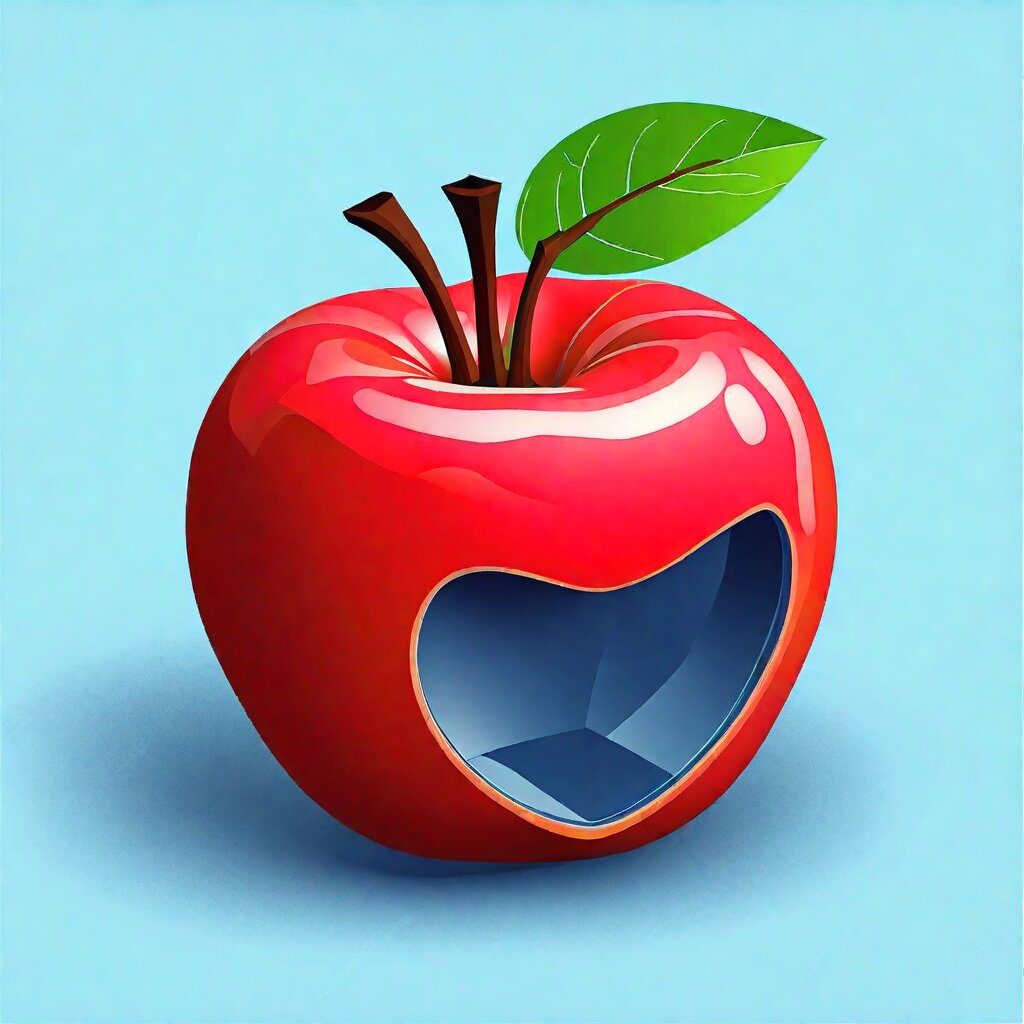}
        \end{minipage} 
        \vspace{-2pt} 
        \\
        \vspace{-1pt} 
        \scriptsize ``A house" & \scriptsize ``An apple" & \multicolumn{5}{c}{\scriptsize ``...in isometric illustration style."} \\
        
        \begin{minipage}{0.12\textwidth}
            \includegraphics[width=\textwidth]{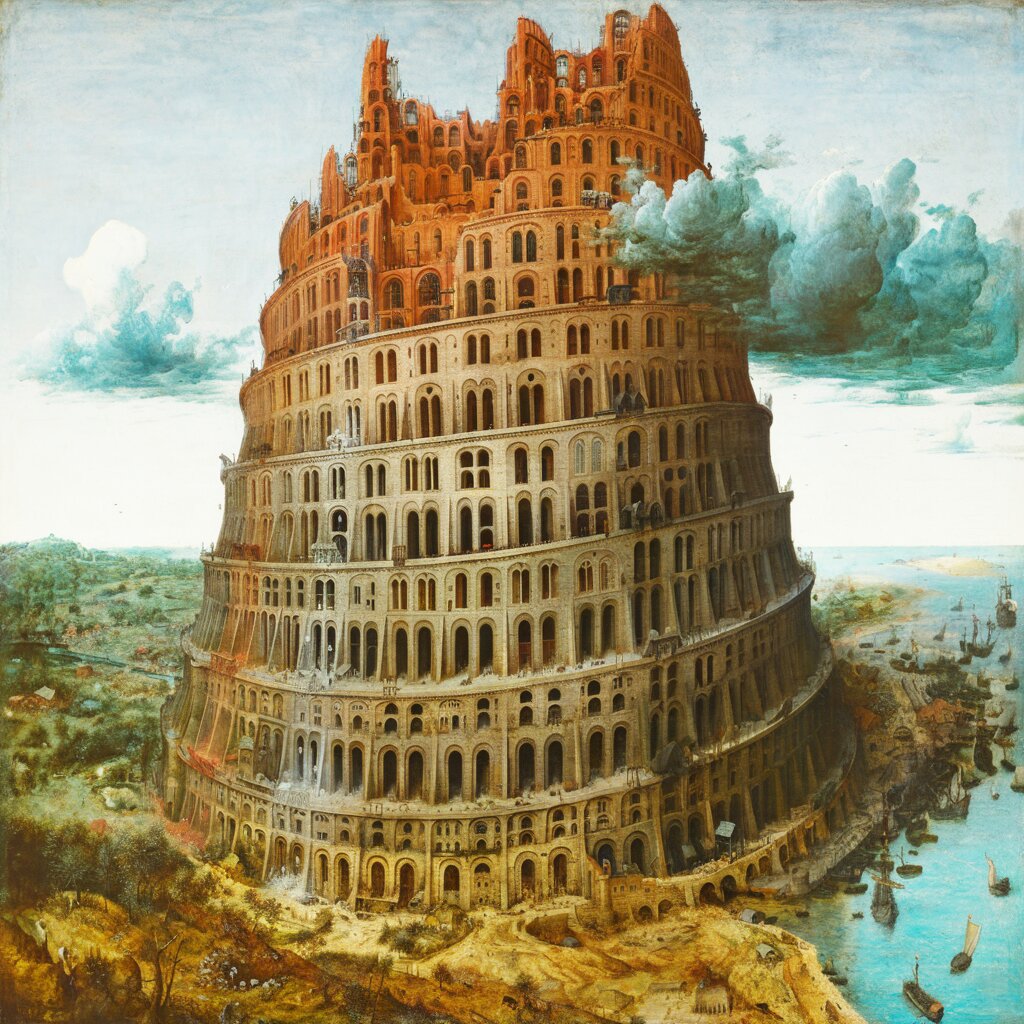}
        \end{minipage} &
        \begin{minipage}{0.12\textwidth}
            \includegraphics[width=\textwidth]{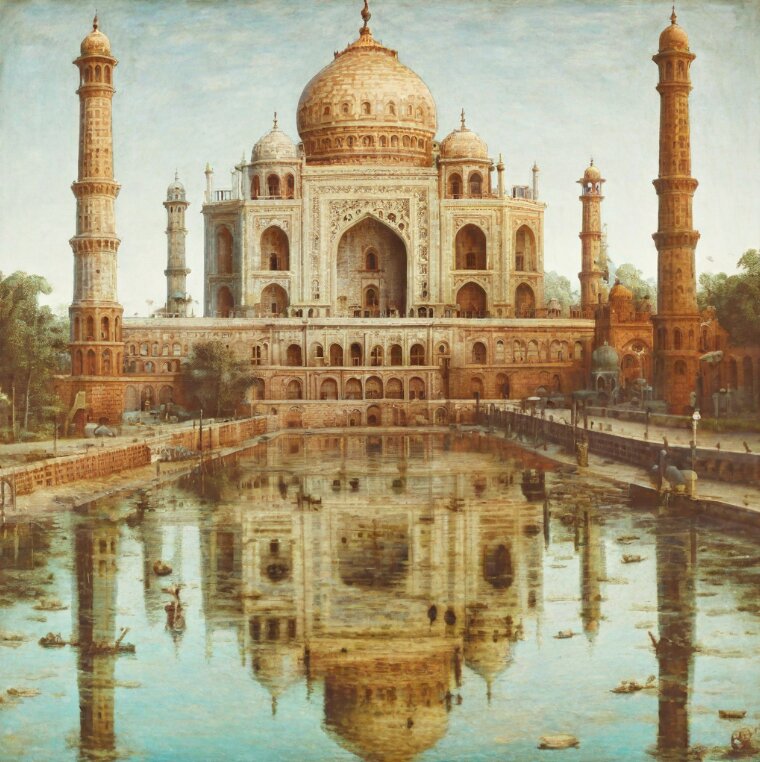}
        \end{minipage} &
        \begin{minipage}{0.12\textwidth}
            \includegraphics[width=\textwidth]{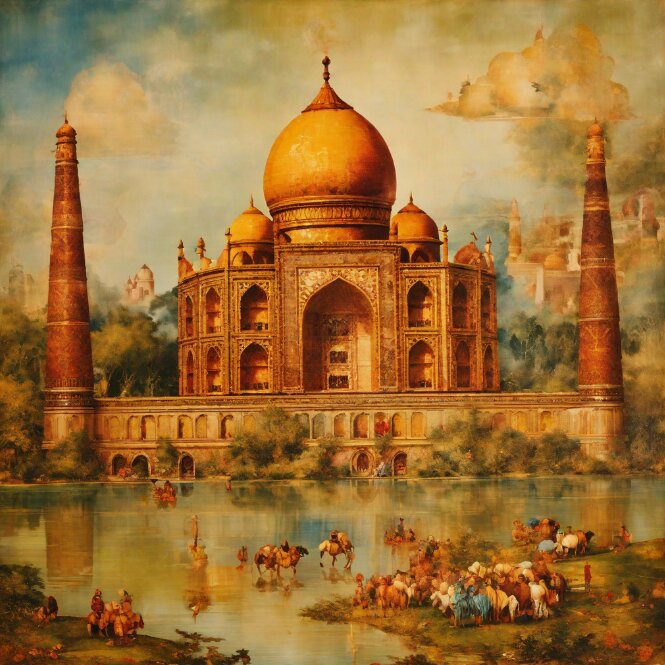}
        \end{minipage} &
        \begin{minipage}{0.12\textwidth}
            \includegraphics[width=\textwidth]{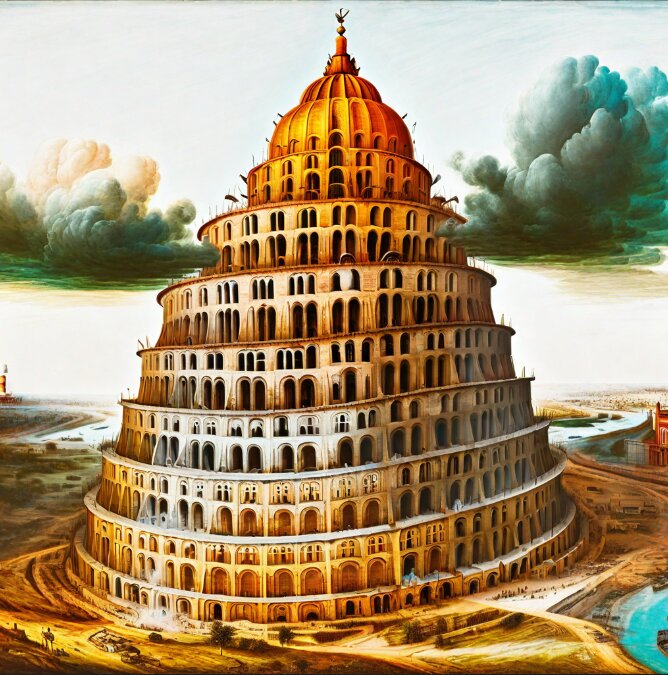}
        \end{minipage} &
        \begin{minipage}{0.12\textwidth}
            \includegraphics[width=\textwidth]{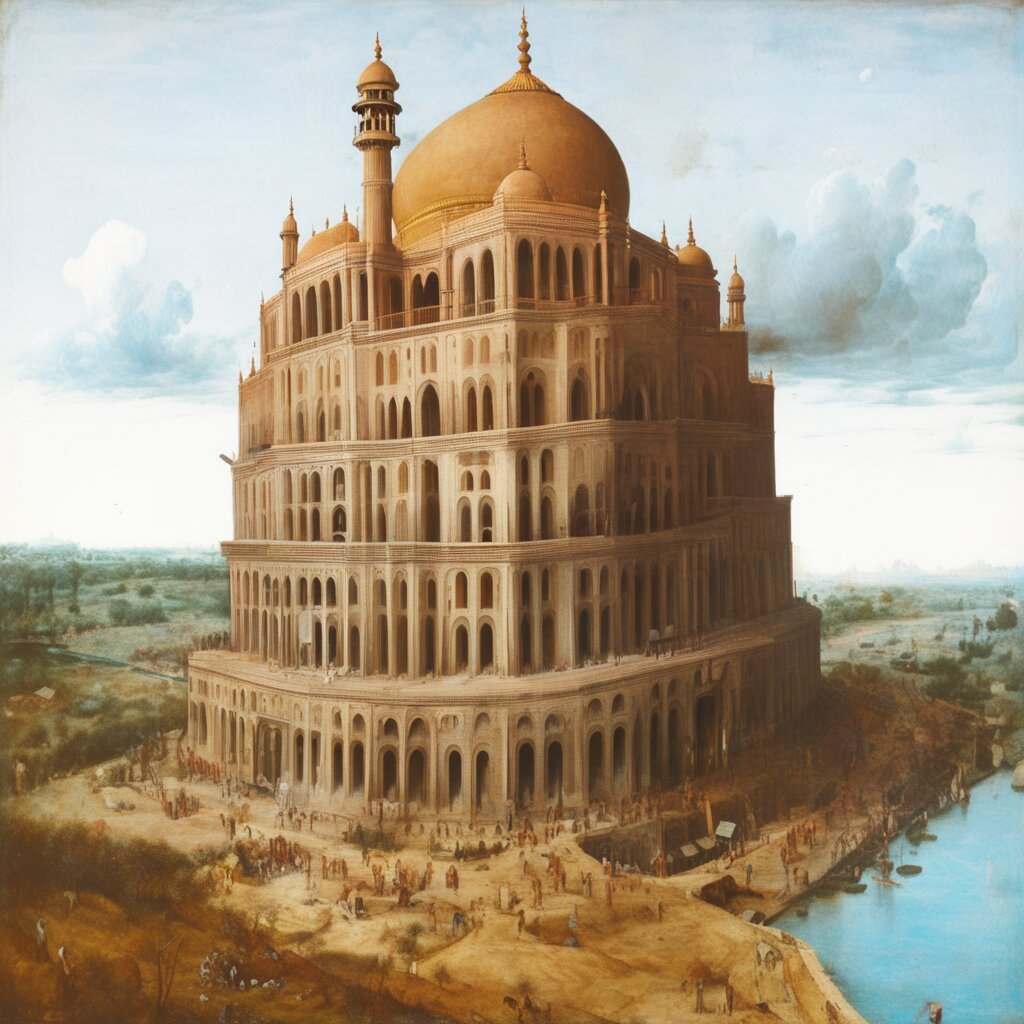}
        \end{minipage} &
        \begin{minipage}{0.12\textwidth}
            \includegraphics[width=\textwidth]{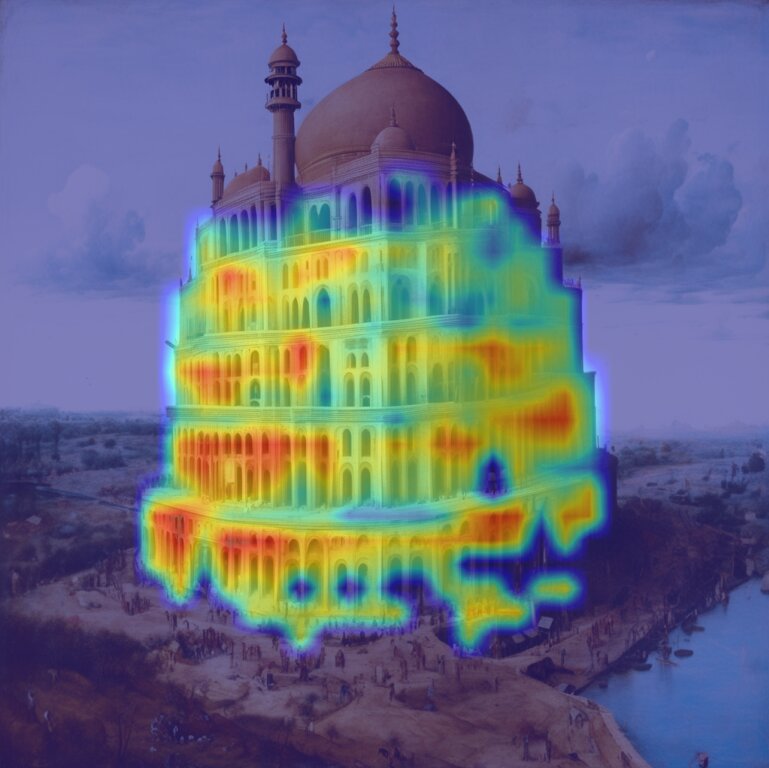}
        \end{minipage} &
        \begin{minipage}{0.12\textwidth}
            \includegraphics[width=\textwidth]{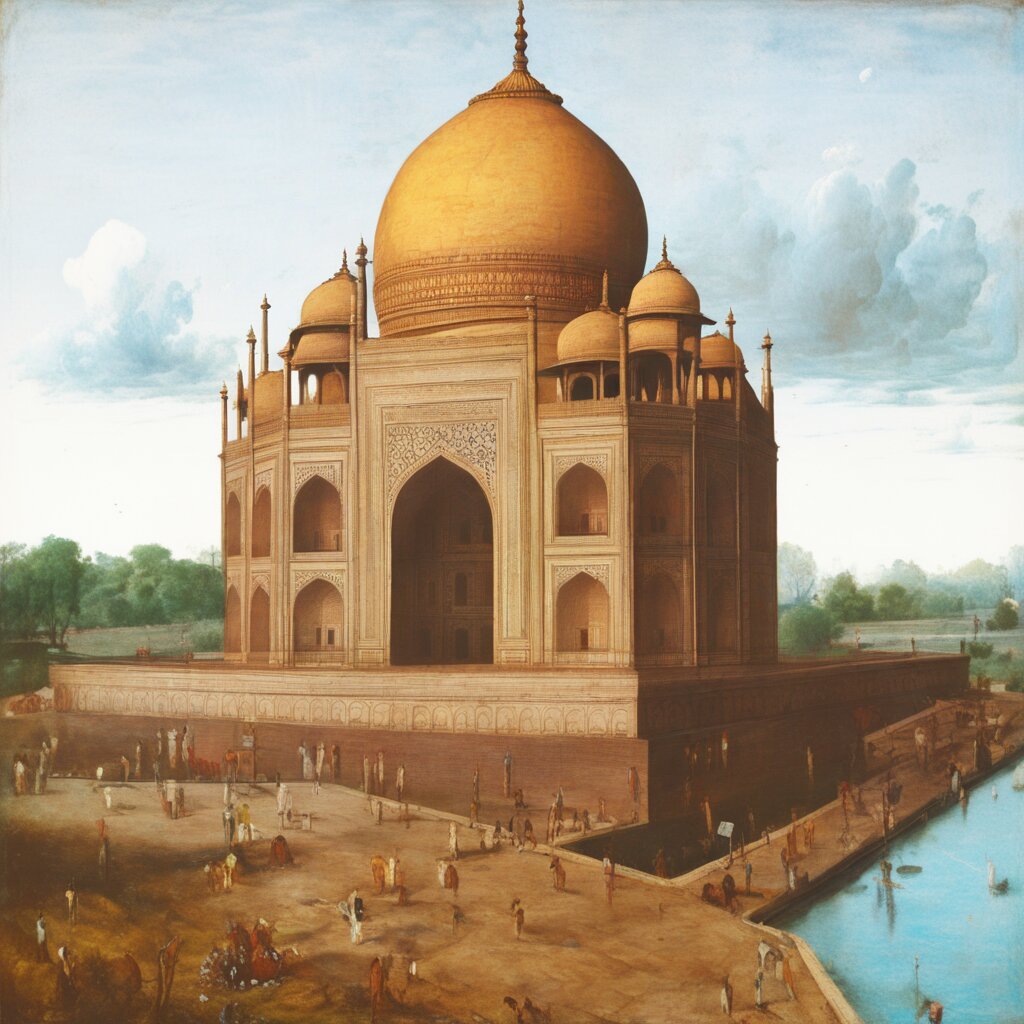}
        \end{minipage}  
        \vspace{-2pt} 
        \\
        \vspace{-1pt} 
        \scriptsize ``The Tower of Babel" & \scriptsize ``The Taj Mahal" & \multicolumn{5}{c}{\scriptsize ``...in Bruegel's painting style."} 
    \end{tabular}
    }
    \vspace{-5pt}
    \caption[Qualitative results of Only-Style]{\textbf{Qualitative results}. We compare Only-Style against StyleAligned~\cite{hertz2024style}, InstantStyle~\cite{Wang2024InstantStyle}, B-LoRA~\cite{frenkel2024blora}. and StyleDrop~\cite{sohn2023styledrop}. In the next-to-last column we also highlight the content leakage observed in StyleAligned, which is localized and effectively mitigated by our method.
    First four rows are examples from our evaluation set, fifth row showcases an intricate multi-subject case, while the last three rows correspond to real reference images.
    }
    \vspace{-10pt}
    \label{fig:Qualitative_ours}
\end{figure*}

\begin{table*}[t]
    \centering
    \caption{\small{\textbf{Content Leakage (CL) Metric Results.} We calculate CL scores across various methods, quantifying content leakage as the cosine similarity between CLIP embeddings of the target image and the reference subject's text description. Lower is better (less leakage).}}
    \vspace{-5pt}
    \label{tab:CL}
    \resizebox{.9\textwidth}{!}{%
    \begin{tabular}{c@{\hspace{1.2em}}|c@{\hspace{1.2em}}c@{\hspace{1.2em}}c@{\hspace{1.2em}}c@{\hspace{1.2em}}|c@{\hspace{1.2em}}c@{\hspace{1.2em}}c@{\hspace{1.2em}}|c}
        \toprule
        full leakage
        & SA~\cite{hertz2024style} & SDRP~\cite{sohn2023styledrop} & IS~\cite{Wang2024InstantStyle} & B-LoRA~\cite{frenkel2024blora} & Ours ($\alpha=0.5$) & Ours ($\alpha=0.9$) & \emph{Only-Style} & no leakage
        \\
        \midrule
        0.28 & 0.231 & 0.227 & 0.220 & 0.223 & 0.214 & 0.219 & \textbf{0.215} & 0.21 \\
        \bottomrule
    \end{tabular}%
    }
    \vspace{-6pt}
\end{table*}

\begin{table*}[t]
    \centering
    \caption{\small{\textbf{Content Leakage Measurements using LVLM-Based Prompting.} Our method shows very low content leakage, closely matching the performance of standard T2I. For all questions, the numbers denote success rate.}}
    \vspace{-5pt}
    \label{tab:Quant_LVLM}
      \resizebox{\textwidth}{!}{%
        \begin{tabular}{lcccc|ccc|c}
            \toprule
            Question & SA~\cite{hertz2024style} & SDRP~\cite{sohn2023styledrop} & IS~\cite{Wang2024InstantStyle} & B-LoRA~\cite{frenkel2024blora} & Ours ($\alpha = 0.5$) & Ours ($\alpha = 0.9$) & \emph{Only-Style} & Standard T2I~\cite{podell2023sdxlimprovinglatentdiffusion} \\
            \midrule
            \textit{``Are there any \{$S_{ref}$\} visual features in this \{$S_{tgt}$\} image?"} (Q1) & 0.470 & 0.553 & 0.647 & 0.643 & 0.663 & 0.603 & \textbf{0.683} & \textbf{0.703} \\
            \textit{``Is there any \{$S_{ref}$\} in this image?"} (Q2) & 0.583 & 0.687 & 0.793 & 0.786 & 0.823 & 0.770 & \textbf{0.830} & \textbf{0.833} \\
            \textit{``Is there any \{$S_{tgt}$\} in this image?"} (Q3) & 0.850 & 0.903 & 0.94 & 0.947 & 0.943 & 0.930 & \textbf{0.957} & \textbf{0.963} \\
            \bottomrule
        \end{tabular}%
        }
        \vspace{-7pt}
\end{table*}

\textbf{Adaptive vs Fixed scaling.} 
To highlight the critical role of adaptivity in style alignment, we fix the scaling parameter $\alpha$ into distinct values, essentially performing only the approach presented in Sec.~\ref{sec:style-control}.
Specifically, we select two different fixed values ($\alpha = 0.9$ and $\alpha = 0.5$) and qualitatively compare the results with those of \emph{Only-Style}, as shown in Fig. \ref{fig:Ablation_2} - quantitative results with these fixed value alternatives will be presented in upcoming sections. 
It is evident that in both cases, the fixed scaling parameter is either unnecessarily low, ruining stylistic alignment, or not low enough to erase the effect of content leakage. 
On the contrary, \emph{Only-Style} faithfully removes content leakage while maintaining the desired stylistic alignment. More ablation studies can be found in the Suppl. Mat.

\subsection{Comparisons}

\textbf{Qualitative Comparisons.}Figure~\ref{fig:Qualitative_ours} presents qualitative comparisons between \emph{Only-Style} and the considered baselines. 
The first four rows showcase results from our evaluation prompt set, the fifth row depicts an example of an intricate multi-subject reference and target scene, 
while the bottom three rows include real stylistic reference images, a common application scenario in the literature.
Concerning the case of real images,  we employ a DDIM-based inversion technique to transfer the style of real images to the generated images of different target prompts, following the paradigm of StyleAligned~\cite{hertz2024style}.
Please refer to the Suppl. Mat. for details on the multi-subject extensions, as well as more qualitative results.

\textbf{Tuning method-specific hyperparameters.} Many methods that tackle style-consistent image generation have introduced hyperparameters to control the impact of the stylistic reference to the target. For example B-LoRA \cite{frenkel2024blora} introduces a scalar $\beta \in [0,1]$ to reduce the influence of the style-LoRA adapter $\delta W$: \( W = W_0 + \beta \cdot \Delta W \). On the other hand, InstantStyle~\cite{Wang2024InstantStyle} addresses leakage by modulating the subtraction of the subject CLIP text embedding from the reference image CLIP embedding using a scalar $\sigma \in [0,1]$: $\text{CLIP}_{\text{img}}(I_{\text{ref}}) - \sigma \cdot \text{CLIP}_{\text{txt}}(S_{\text{ref}})$. We compare the effects of these parameters with our approach in Fig.~\ref{fig:Generalization}. Additionally, we apply the proposed adaptive tuning algorithm—leveraging the generalized leakage localization method presented in Sec. 3.4—to determine the maximum values of these parameters that prevent content leakage. While this effectively eliminates content leakage, we observe that such parameters degrade the stylistic alignment with the reference. In contrast, \emph{Only-Style} preserves the intended stylistic alignment while mitigating content leakage.

\textbf{Text Alignment and Stylistic Set Consistency.}
Following \cite{hertz2024style, frenkel2024blora, jeong2024visual}, we use CLIP~\cite{radford2021learningtransferablevisualmodels} cosine similarity to measure the \emph{text alignment} between each target image $I_{tgt}$ and the text description of the target subject $S_{tgt}$, while as \emph{stylistic set consistency} we measure the cosine similarity between DINO~\cite{caron2021emerging} embeddings of the generated target images $I_{tgt}$ with their reference $I_{ref}$. 
The results prompt set can be seen in Figure~\ref{fig:Quantitative_ours}, where we compare our method to the four state-of-the-art baselines we mentioned, a baseline generation method without any stylistic alignment (Standard T2I~\cite{podell2023sdxlimprovinglatentdiffusion}) and the two fixed scale variants of Sec.~\ref{sec:style-control}. \emph{Only-Style} exhibits a notable balance between retaining the style of the reference image (set consistency) and faithfully depicting the target subject (text alignment). 
Despite an expected drop in set consistency compared to StyleAlign, which generates aligned images at the cost of exhibiting leakages, our method achieves almost identical text alignment with standard T2I, supporting our claim on keeping the semantics of the target subject intact. On the other hand InstantStyle~\cite{Wang2024InstantStyle} and B-LoRA~\cite{frenkel2024blora} preserve alignment with the target subject but compromise stylistic consistency.

Additionally, we observe that fixed scaling results in lower stylistic consistency, as it indiscriminately scales instances that might not exhibit any leakage, while the text alignment can be negatively affected by a high scale value (0.9) that allows leakage cases. 
Even though this experiment offers a strong indication of the effectiveness of our method, the capability to eliminate content leakage is not captured properly by the above metrics. This is because the cosine similarity of the image DINO embeddings is favored by semantic content leakage (details on Suppl. Mat.).

\textbf{Content Leakage.}
The aforementioned metrics fail to quantify content leakage in a straightforward way. 
To address this, we introduce a novel metric, called \emph{CL}, which is defined as the cosine similarity between the CLIP embeddings of the target image $I_{tgt}$ and the text description of the reference image subject $S_{ref}$, namely quantifying the correlation of the reference subject with the target image. 
Results of the \emph{CL} metric can be found in Table \ref{tab:CL}. Standard T2I is used as a lower bound for the metric, which denotes the ``no leakage" case, since no style sharing takes place.
Likewise, we use standard T2I to generate images of $S_{ref}$ and compute \emph{CL} again in order to define the upper bound, denoting the ``full leakage" case, when we expect to detect the subject in the image. 
As we can see, Only-Style obtains a score very close to the ``no leakage" case, showing minor bias towards generating the reference subject. 
On the contrary, other methods exhibit considerably higher CL scores, indicating non-trivial leakage of the reference subject.
As expected, InstantStyle (IS)~\cite{Wang2024InstantStyle}, that tries to address the leakage issue, has the best performing score out of the compared methods, but still is out-performed by \emph{Only-Style}. 

\textbf{LVLM-based Quantitative Evaluation Protocol.} 
The aforementioned metrics are not adept in capturing fine-grained details. 
For example, subtle content leakage may not be penalized by the cosine similarity between the CLIP representations. 
For this reason, we introduce a novel evaluation protocol based on Large Vision-Language Models. 
Specifically, we employ LLaVA~\cite{li2024llava,liu2024llavanext, liu2023improvedllava,liu2023llava} in order to unveil content leakage by prompting the model to identify the desired target subject and possibly the undesired reference subject in the generated target image $I_{tgt}$. 
Specifically we provide the LVLM the generated images $I_{tgt}$ along with following questions.
\textbf{Q1:} \textit{``Are there any \{$S_{ref}$\} visual features in this \{$S_{tgt}$\} image?"} - if the answer is positive, it suggests the existence of undesired semantic features of the reference subject.
\textbf{Q2:} \textit{``Is there any \{$S_{ref}$\} in this image?"} - a more robust question that naturally exposes only severe content leakage cases.
\textbf{Q3:} \textit{``Is there any \{$S_{tgt}$\} in this image?"} - check if the desired target subject is not rendered at all in the target image. 

We present the results of the aforementioned evaluation in Table~\ref{tab:Quant_LVLM}. We explicitly ask the LVLM to \textit{``Choose one: Yes or No"} along with the question and the image subjected in stylistic alignment in order to measure the success rate of the respective method in the question we pose. We observe, again, significant content leakage across state-of-the-art methods and almost identical to standard T2I performance from our method, suggesting almost no leakage. Please refer to the Suppl. Mat. for qualitative examples indicating the performance of the proposed framework as well as benchmarking of additional methods that highlights the presence of content leakage in style-consistent generation.

\textbf{User Study.} We conducted a user study where participants were shown randomized triplets consisting of a reference image and two target images generated by Only-Style and one competitor method. Participants were asked to select their preferred image based on the following criteria: stylistic alignment, text alignment, and overall image quality. An option ``Cannot Decide" was also provided. We collected 800 pairwise method comparisons across 100 users and show the results in Table~\ref{tab:study1}. Only-Style was significantly preferred over all other baselines, indicating the effectiveness of resolving content leakage in terms of human preference. More information provided in the Suppl. Mat.

\begin{table}[!ht]
\centering
\caption{
\textbf{User study:} `a/b' indicates that Ours (left) was preferred \textit{a} times, while the competing method was chosen \textit{b} times. Only-Style was the preferred method by the participants.
}
\vspace{-3pt}
\label{tab:study1}
\resizebox{.98\linewidth}{!}{
\begin{tabular}{l|c|c|c|c}
     & StyleAligned~\cite{hertz2024style}   & IS~\cite{Wang2024InstantStyle}  & B-LoRA~\cite{frenkel2024blora} & SDRP~\cite{sohn2023styledrop}   \\\toprule
\emph{Only-Style} & \textbf{357}/137& \textbf{319}/210 &  \textbf{419}/155 &\textbf{321}/202  \\ 
\end{tabular}
}
\vspace{-10pt}
\end{table}


\section{Conclusion}

We introduced \emph{Only-Style}, a novel approach designed to pinpoint and mitigate unwanted semantic content leakage in style-consistent generation. Extensive experiments demonstrate that \emph{Only-Style} prevents content leakage, ensuring stylistic consistency in target images with the reference style. Finally, we proposed a framework to quantitatively assess this issue within style alignment methods, providing a structured approach to evaluate their effectiveness.

{
    \small
    \bibliographystyle{ieeenat_fullname}
    \bibliography{main}
}

\clearpage
\setcounter{page}{1}
\maketitlesupplementary

\section{Methodology Details}

\subsection{Preliminaries: StyleAligned}

As we discussed in the main manuscript, recent state-of-the-art style alignment methods in image generation \cite{hertz2024style, jeong2024visual} leverage the self-attention layers of T2I models during inference to facilitate communication between images within a batch, thereby aligning their styles. We will provide further details on the operations involved in these methods and the underlying intuition, focusing on StyleAligned \cite{hertz2024style}, which our method builds upon.

StyleAligned employs an attention sharing operation between a stylistic reference image (typically the first one within a batch) and the target images (other images within the same batch). 
This operation is only applied to the self-attention layers of the attention-augmented UNet backbone.
On such an attention layer of the model's backbone, the queries $\mathbf{Q}_{tgt}$ and keys $\mathbf{K}_{tgt}$ of the target image are normalized using the queries $\mathbf{Q}_{ref}$ and keys $\mathbf{K}_{ref}$ of the reference, with the adaptive instance normalization operation (AdaIN) \cite{huang2017arbitrarystyletransferrealtime}, which essentially aligns the target features with respect to the first and second moments of the reference features. Formally, we have:

\begin{equation}
    \text{AdaIN}(\mathbf{X}, \mathbf{Y}) = \sigma(\mathbf{Y}) \left(\frac{\mathbf{X} - \mu(\mathbf{X})}{\sigma(\mathbf{X})}\right) + \mu(\mathbf{Y}) \nonumber
    \label{eq5}
\end{equation}
\begin{equation}
    \hat{\mathbf{Q}}_{tgt} = \text{AdaIN}(\mathbf{Q}_{tgt}, \mathbf{Q}_{ref}),\,\, \hat{\mathbf{K}}_{tgt} = \text{AdaIN}(\mathbf{K}_{tgt}, \mathbf{K}_{ref}) \nonumber
\end{equation} 

Then, to further promote sharing, the attention operation is applied to concatenated versions of the keys and the values that include both reference and target features. 
This way, the sharing is performed in a ``natural" way, where features from both reference and target images are mingled together, essentially providing style context from the reference images to the target one. 
More specifically, the target queries are replaced by the normalized ones $\hat{\mathbf{Q}}_{tgt}$, the target keys are replaced by the concatenation of the reference keys $\mathbf{K}_{ref}$ with the normalized target ones $\hat{\mathbf{K}}_{tgt}$ and finally the target values are replaced by the concatenation of the reference values $\mathbf{V}_{ref}$ with the target ones $\mathbf{V}_{tgt}$. 
The concatenation is performed at a token level, duplicating the context length in the attention layer. 
Following the notation of ~\cite{hertz2024style}, the substituted shared self-attention layer is denoted as $\text{Attention}(\hat{\mathbf{Q}}_{tgt}, \mathbf{K}_{rt}, \mathbf{V}_{rt})$, where:
\begin{equation}
    \mathbf{K}_{rt} = \begin{bmatrix} \mathbf{K}_{ref} \\ \hat{\mathbf{K}}_{tgt} \end{bmatrix} , \, \mathbf{V}_{rt} = \begin{bmatrix} \mathbf{V}_{ref} \\ \mathbf{V}_{tgt} \end{bmatrix} \nonumber
\end{equation}
Note that this concatenation does not affect the size of the output, since the patch length of the queries is not affected.

The concatenation of the target features with the reference ones at a token level allows a minimal contextualization of the target image features with the reference, effectively aligning the two images. 
Meanwhile, applying AdaIN to the target keys using the reference boosts the attention similarity scores between the target features and the reference, facilitating a smoother attention flow from the reference to the target.

\begin{figure}[t]
    \centering
    \includegraphics[width=1.0\linewidth]{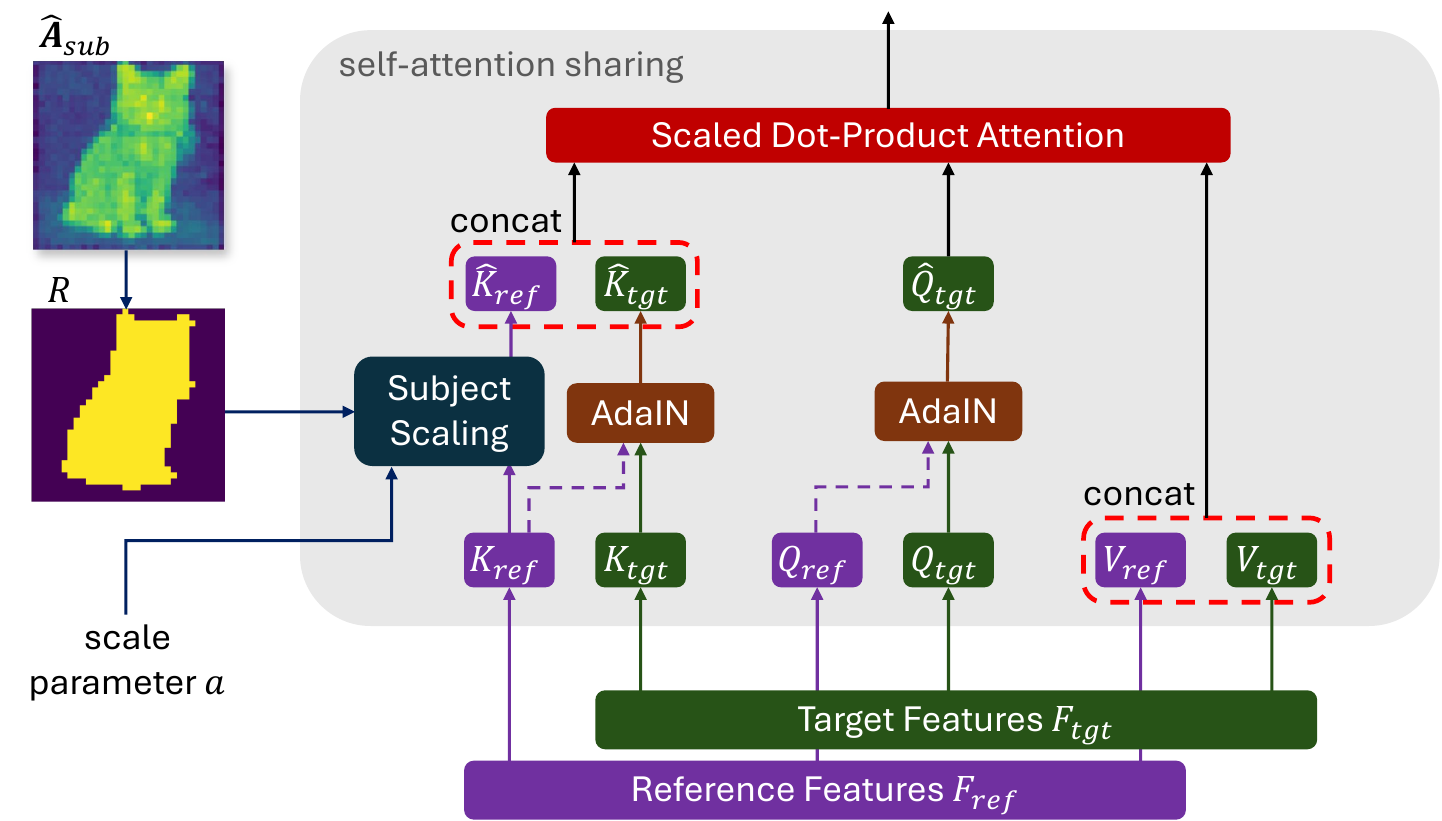} 
    \vspace{-0.4cm}
    \caption{\textbf{Content Leakage Control}: Content leakage is mitigated by applying a weighting of the localized reference subject Key representations $\mathbf{K}_{ref}$, in every self-attention module that is used to align the style of a reference image with a target.}
    \label{fig:control}
    \vspace{-0.2cm}
\end{figure}

\subsection{Extracting the Subject Mask R} 

\begin{figure}[t]
    \centering
    \begin{subfigure}{0.15\textwidth}
        \centering
        {\scriptsize
        $\hat{\mathbf{A}}_{sub}$
        }\\ 
        \includegraphics[width=\linewidth]{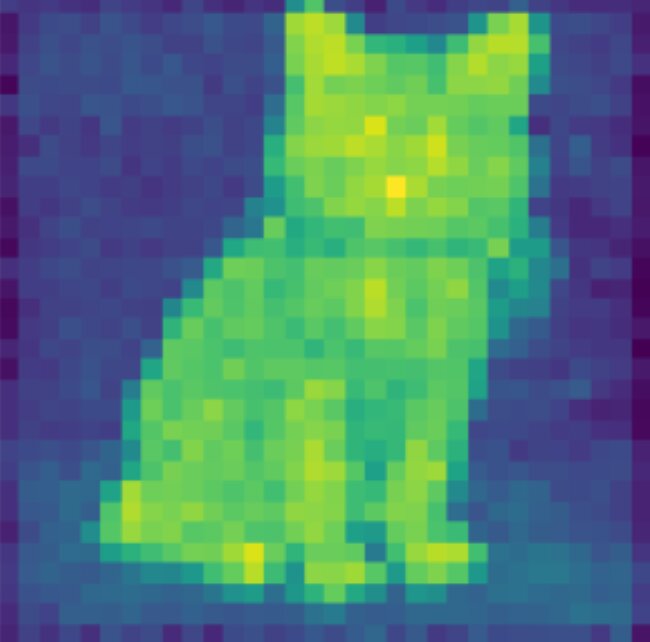}
    \end{subfigure}
    \hfill
    \begin{subfigure}{0.15\textwidth}
        \centering
        {\scriptsize
        2-means clustering
        }\\ 
        \includegraphics[width=\linewidth]{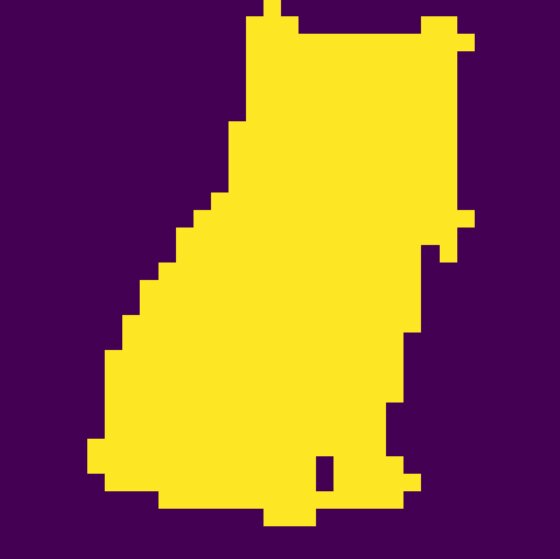}
    \end{subfigure}
    \hfill
    \begin{subfigure}{0.15\textwidth}
        \centering
        {\scriptsize
        Closed mask ($\mathbf{R}$)
        }\\ 
        \includegraphics[width=\linewidth]{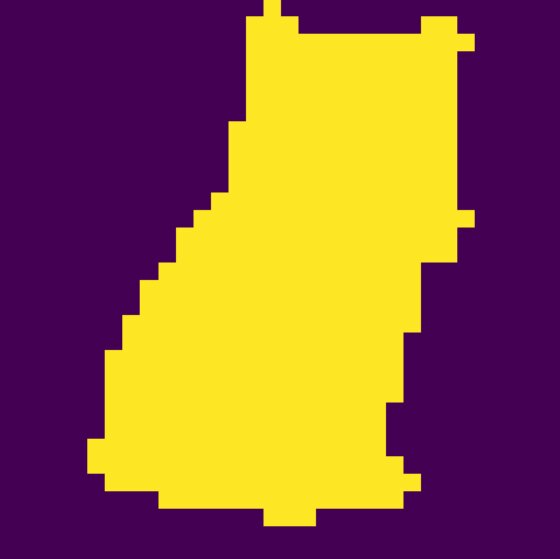}
    \end{subfigure}
    
    \caption{\textbf{Visualization of the intermediate results in the extraction of mask} $\mathbf{R}$. We cluster the aggregated cross-attention probabilities $\hat{\mathbf{A}}_{sub}$ using K-means with two centroids and then apply morphological closing to fill small gaps in the foreground.}
    \label{fig:R}
\end{figure}

As we discussed in Sec. 3.3 of the main manuscript, given the subject map $\hat{\mathbf{A}}_{sub} \in \mathbb{R}^{H \times W}$ (illustrated in Fig.~\ref{fig:R}), we aim to separate the patches into two distinct groups: one that is semantically related to the reference subject (and is the source of content leakage) and one unrelated.

Specifically, we consider the one-dimensional semantic representations of the image patches in $\hat{\mathbf{A}}_{sub}$ and use a K-means clustering method with two centroids to separate them, fixed across all of our experiments. 
Retrieving the patches grouped in the cluster with the maximum value centroid gives us the annotated subject of the image.
This is equivalent to a binarization approach with a threshold depending on the image and its subject map $\hat{\mathbf{A}}_{sub}$~\cite{10.1145/3658157}\footnote{A similar mask extraction was employed in \cite{10.1145/3658157} to extract a subject mask and then preserve the identity of this subject across multiple images, following a ``dual" direction of aligning subjects and not style.},  as opposed to a fixed threshold approach across all images~\cite{patashnik2023localizingobjectlevelshapevariations} which typically under performs (see Suppl. Sec.~2.1).
To ensure that all the subject patches are obtained, we apply a denoising morphological closing in the binary subject mask, filling small holes and gaps in the foreground. The resulting binary mask $\mathbf{R} \in \mathbb{R}^{H \times W}$, takes true values if the corresponding patch is deemed relevant to the subject. The intermediate results of this process are illustrated in Fig.~\ref{fig:R} 

Then, we use this binary subject mask to scale down the influence of the reference key features $\mathbf{K}_{ref}$ on the shared self-attention layers. 
As outlined in Eq. 3 of the main manuscript, we apply a uniform scalar value across all subject patches for scaling, following a "hard" decision rationale instead of using a ``soft" scaling via the cross-attention probabilities $\hat{\mathbf{A}}_{sub}$ for those patches. 
Such ``hard" choice allows the scaling parameter to be set to $\alpha = 1$ when no leakage is detected, effectively replicating the base StyleAligned~\cite{hertz2024style} process in our implementation.
In other words, we wanted to keep the functionality of StyleAligned as it is if no leakage is detected, rather than modifying the subject contribution every time regardless the leakage.

\subsection{Leakage Control over StyleAligned}

The scaling of the content patches is performed using the following equation, as it was derived in Sec.~3.3 of the main manuscript. 

\begin{equation}
    \hat{\mathbf{K}}_{ref} = (1 - \mathbf{R}) \odot \mathbf{K}_{ref} + \alpha \mathbf{R} \odot \mathbf{K}_{ref}
    \label{eq:scaling_supp}
\end{equation}

This way, following the notation of \cite{hertz2024style}, we effectively control the self-attention distribution $\mathbf{A}$, between $\hat{\mathbf{Q}}_{tgt}$ and the updated $\hat{\mathbf{K}}_{rt} = [\hat{\mathbf{K}}_{ref} \, \hat{\mathbf{K}}_{tgt}]^\top$, thus controlling the transfer of the value representations $\mathbf{V}_{rt}$, and more precisely their subset that corresponds to the reference subject patches, in the target image. 
Note that when $\alpha\,$=1, we have the exact same behavior with StyleAligned~\cite{hertz2024style}.

The proposed functionality of the scaling operation over the shared attention mechanism of \cite{hertz2024style} is depicted in Fig.~\ref{fig:control}.

\begin{figure}[t]
    \centering
    \begin{subfigure}{0.15\textwidth}
        \centering
        {\scriptsize
        $\hat{\mathbf{A}}_{sub}$
        }\\ 
        \includegraphics[width=\linewidth]{graphics/method/A_sub.jpg}
    \end{subfigure}
    \hfill
    \begin{subfigure}{0.15\textwidth}
        \centering
        {\scriptsize
        3-means clustering
        }\\ 
        \includegraphics[width=\linewidth]{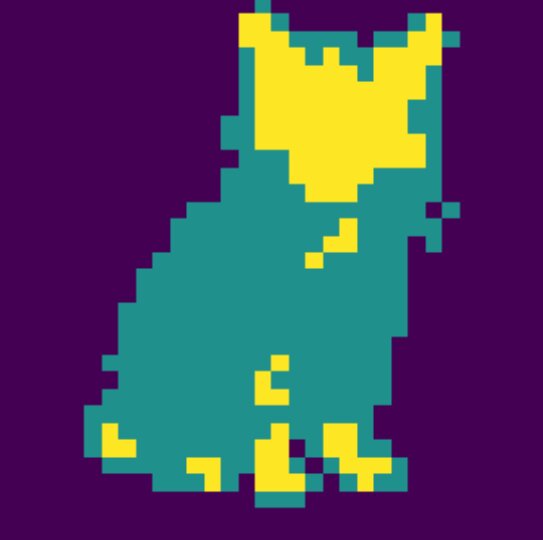}
    \end{subfigure}
    \hfill
    \begin{subfigure}{0.15\textwidth}
        \centering
        {\scriptsize
        $\mathbf{M}_{sub} \odot \hat{\mathbf{A}}_{sub}$
        }\\ 
        \includegraphics[width=\linewidth]{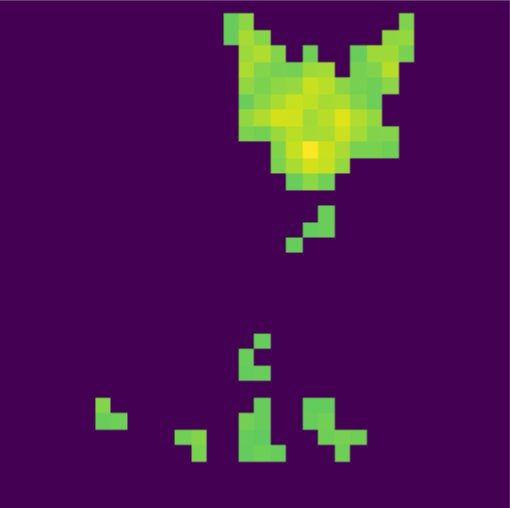}
    \end{subfigure}

    \caption{\textbf{Visualization of intermediate steps in extracting the mask} $\mathbf{M}_{sub}$. Using K-means clustering with 3 centroids, we segment the subject map $\hat{\mathbf{A}}{sub}$ to identify semantically rich subject patches (yellow-labeled $\mathbf{M}_{sub}$). Cross-attention values from these patches (third image) are then used to compute a weighted average of image representations during inference, yielding the subject representation.}
    \label{fig:M_sub}
\end{figure}

\subsection{Extracting the Subject Description Mask M} 

As outlined in Sec. 3.4 of the main manuscript, we focus on isolating a subset of the subject map $\hat{\mathbf{A}}_{sub}$ to pool the representations of image patches, thereby extracting a representation of the image's subject. 
This is achieved again using a binary mask $\mathbf{M}_{sub}$, which contains true values for patches whose representations should be included in the pooling operation.
This subject description mask $\mathbf{M}_{sub}$, differs from the previously defined subject mask $\mathbf{R}$ in its granularity. 
Here, we are interested in more fine-grained localization of patches that are relevant to the subject and can help build robust pooled representations.

To extract this mask we perform again a K-means clustering of the subject map $\hat{\mathbf{A}}_{sub}$, using three clusters this time, one grouping the background patches, one grouping the poor semantic patches, and one grouping the patches with rich semantic information. 
We only use the latter to represent the respective subject, making sure that, the patches do not exceed 10\% of the total image patches in order to retrieve a compact representation and not average-out important semantic features. 
This is performed via percentile thresholding if the resulting cluster with the maximum value centroid exceeds the \(10^{th}\) percentile. 
Note that if the number of subject patches exceeds the 10\%, the clustering operation is redundant, since one can apply percentile thresholding directly on the values of $\hat{\mathbf{A}}_{sub}$. 
Nonetheless, the clustering step is crucial in cases of small objects, as the percentile thresholding would also annotate background patches. 
Again, this is equivalent to a binarization with a threshold dependent on $\hat{\mathbf{A}}_{sub}$, but following a ``stricter" criterion compared to $\mathbf{R}$.
The intermediate results of this process are illustrated in Fig.~\ref{fig:M_sub}.
The proposed mask extraction was deemed helpful in practice, providing robust subject descriptions and thus helping the localization of content leakage, and no further exploration was performed on alternative ways to extract $\mathbf{M}_{sub}$.

\begin{figure}[t]
    \centering
    \scriptsize
    \begin{tabular}{c@{\hspace{.1cm}}c@{\hspace{.1cm}}c@{\hspace{.1cm}}c}  
        Reference & StyleAligned~\cite{hertz2024style} & Leakage & \emph{Only-Style} \\
        \includegraphics[width=0.11\textwidth]{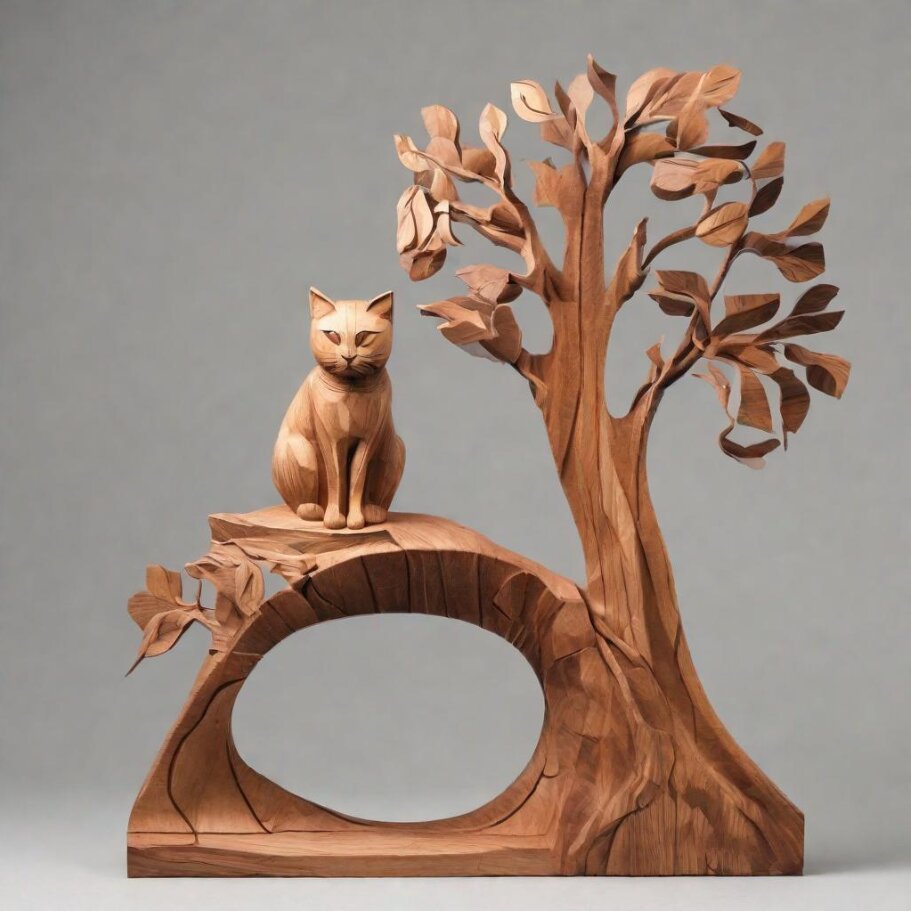} &
        \includegraphics[width=0.11\textwidth]{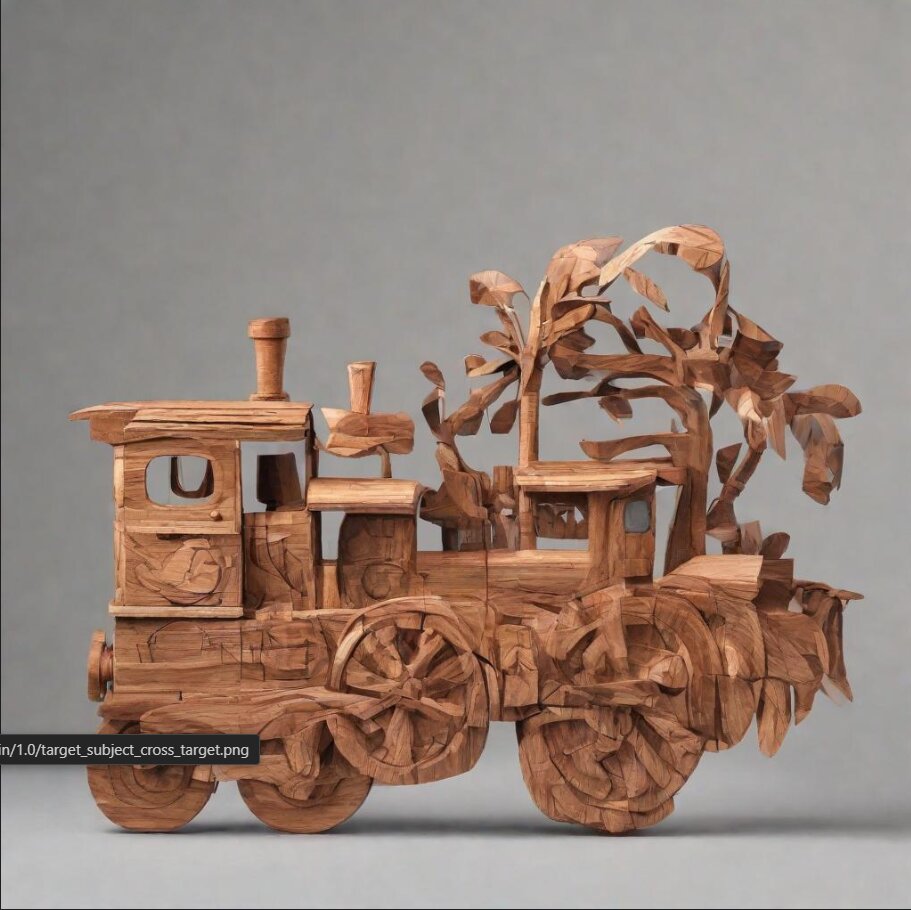} &
        \includegraphics[width=0.11\textwidth]{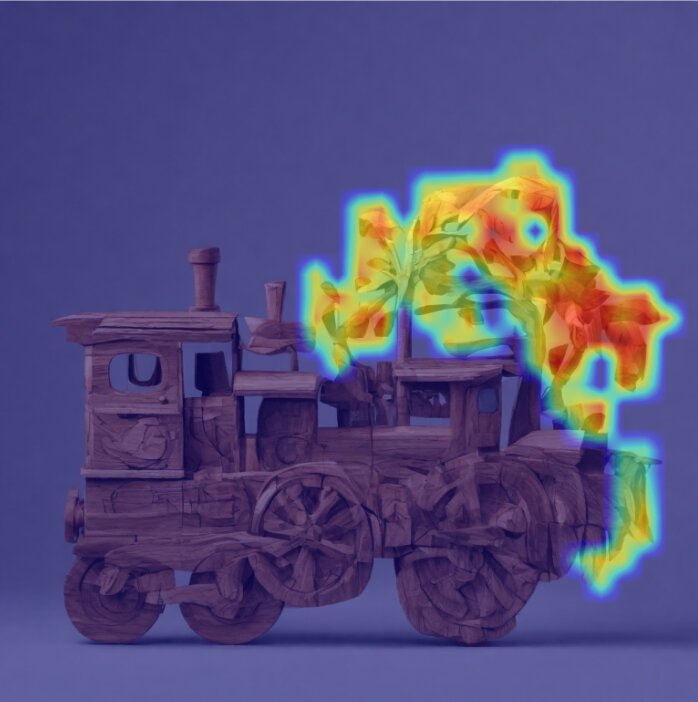} &
        \includegraphics[width=0.11\textwidth]{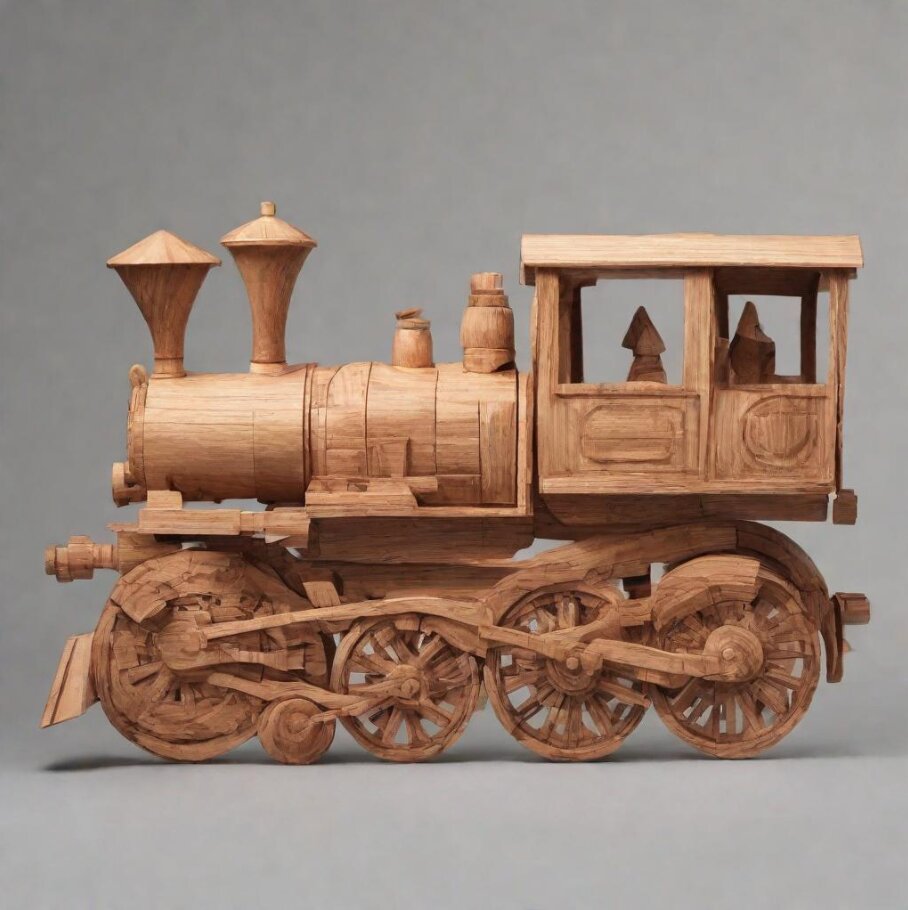} \\
        \footnotesize ``A cat and a tree" & \footnotesize ``A train" & 
        \multicolumn{2}{c}{\footnotesize ``... in wooden sculpture."} \\
        
        \includegraphics[width=0.11\textwidth]{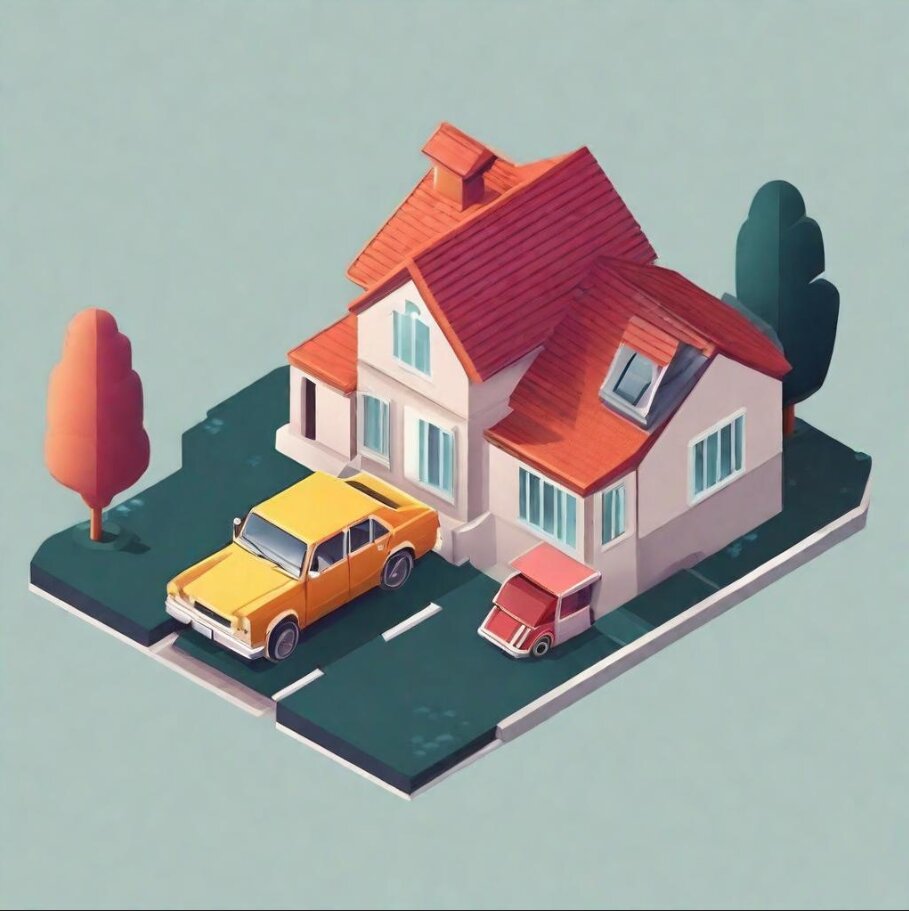} &
        \includegraphics[width=0.11\textwidth]{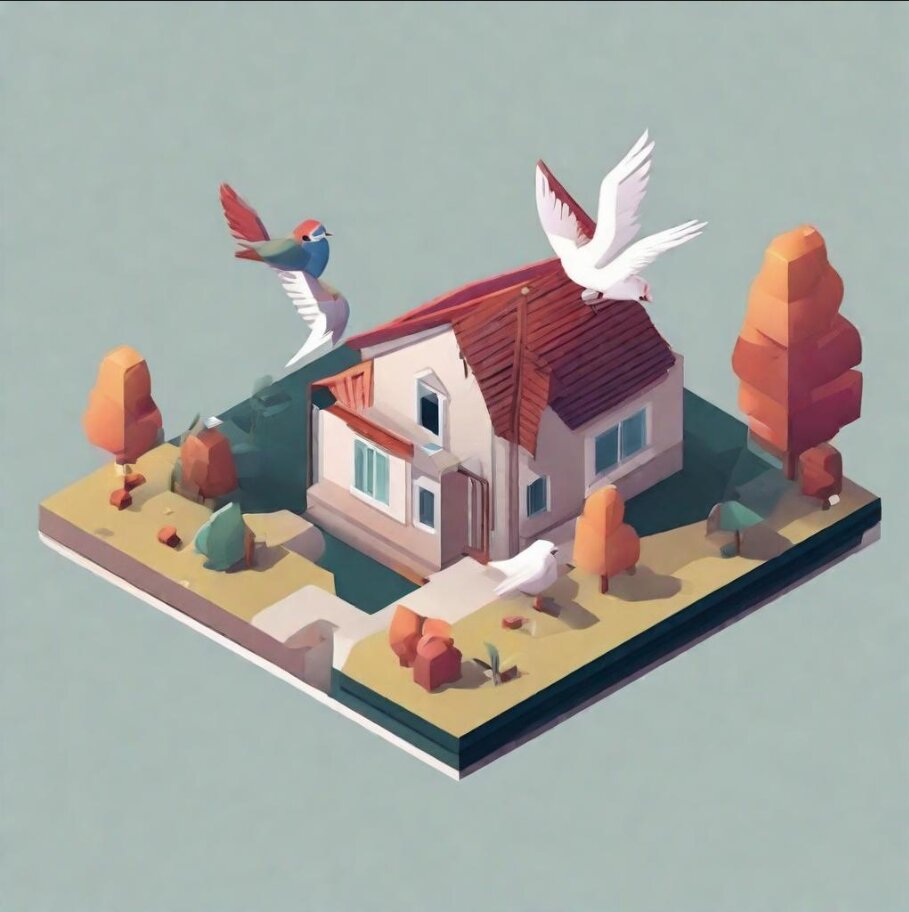} &
        \includegraphics[width=0.11\textwidth]{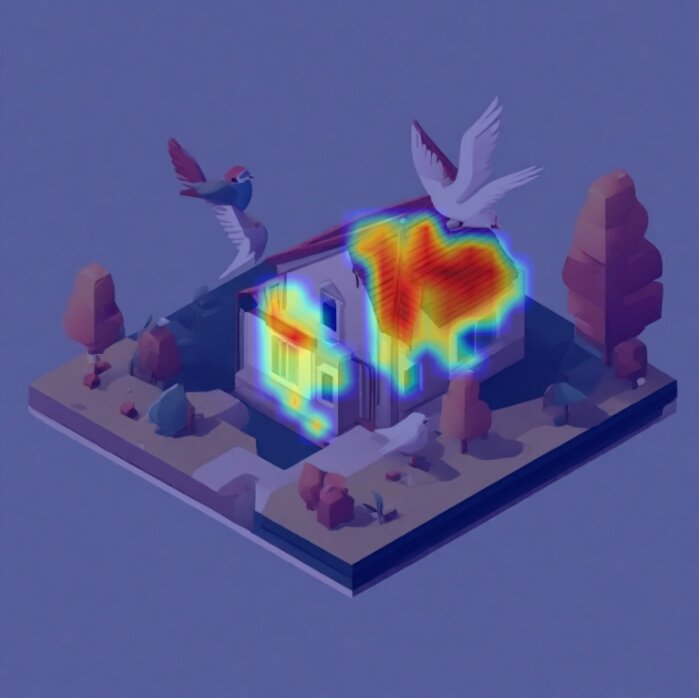} &
        \includegraphics[width=0.11\textwidth]{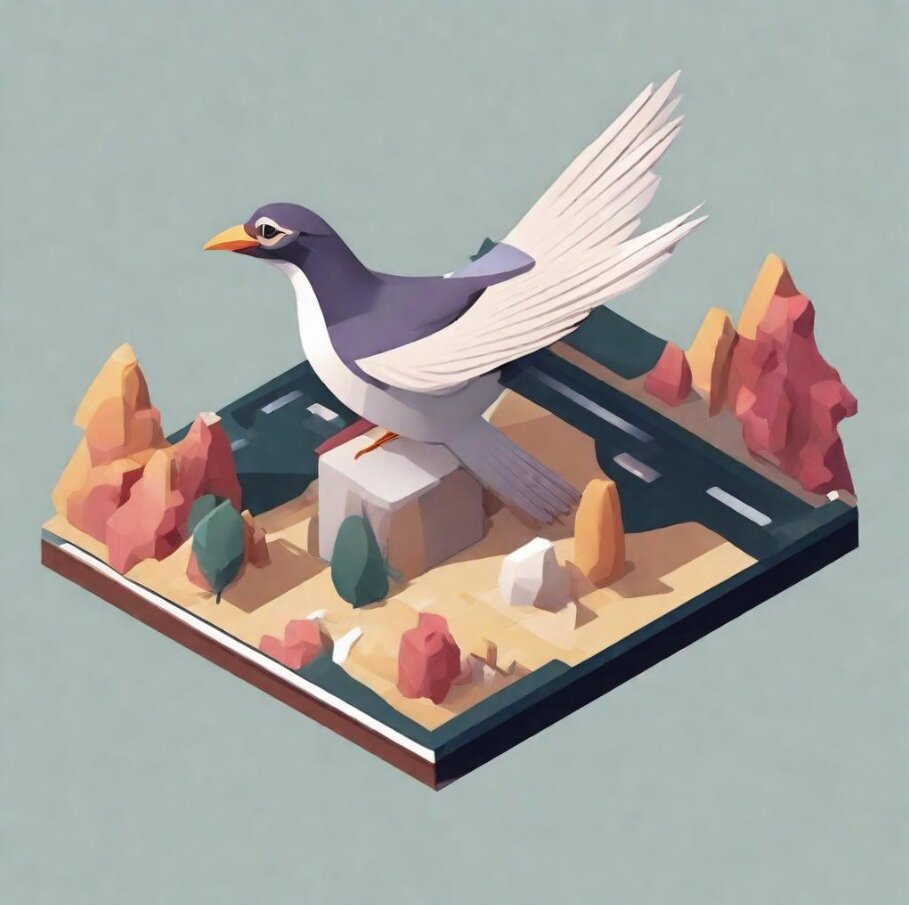} \\
        \footnotesize ``A car and a house" & \footnotesize ``A bird" &
        \multicolumn{2}{c}{\footnotesize ``... in isometric illustration."} \\

        \includegraphics[width=0.11\textwidth]{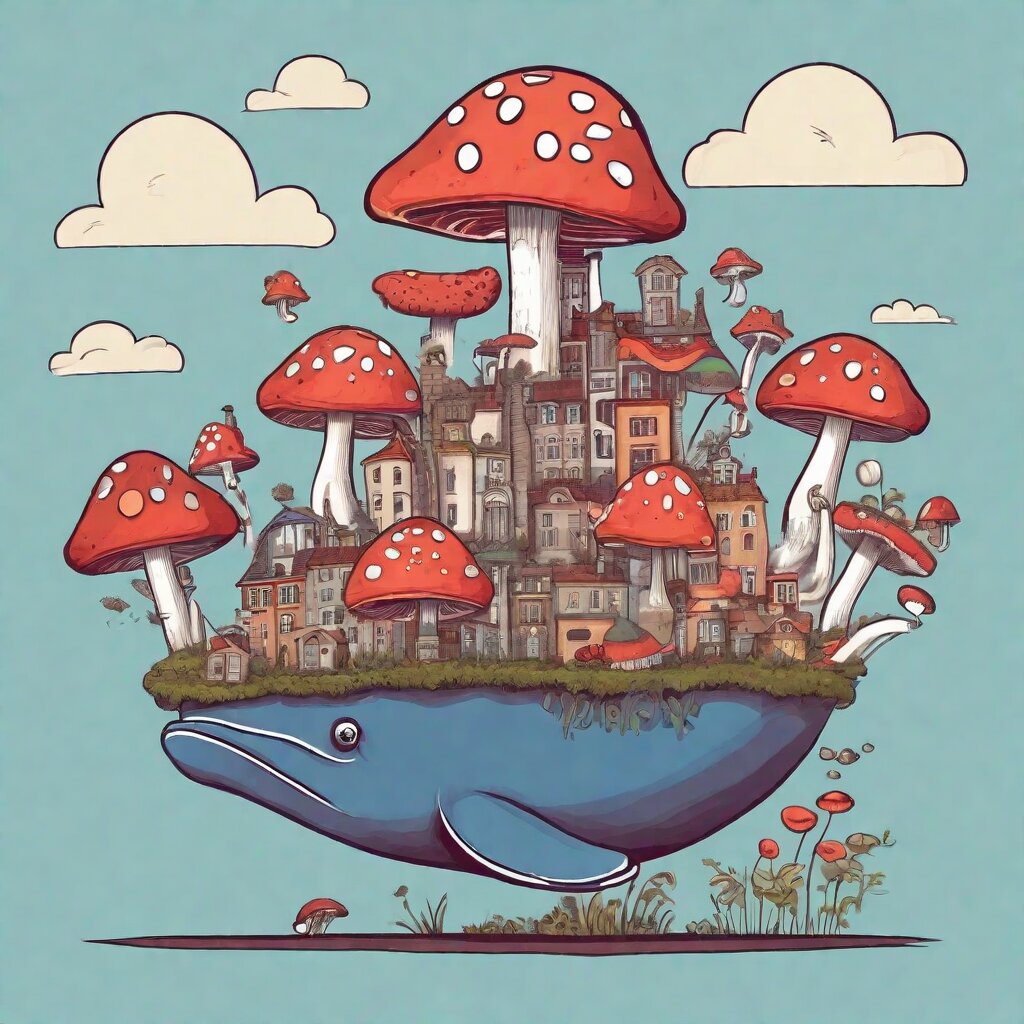} &
        \includegraphics[width=0.11\textwidth]{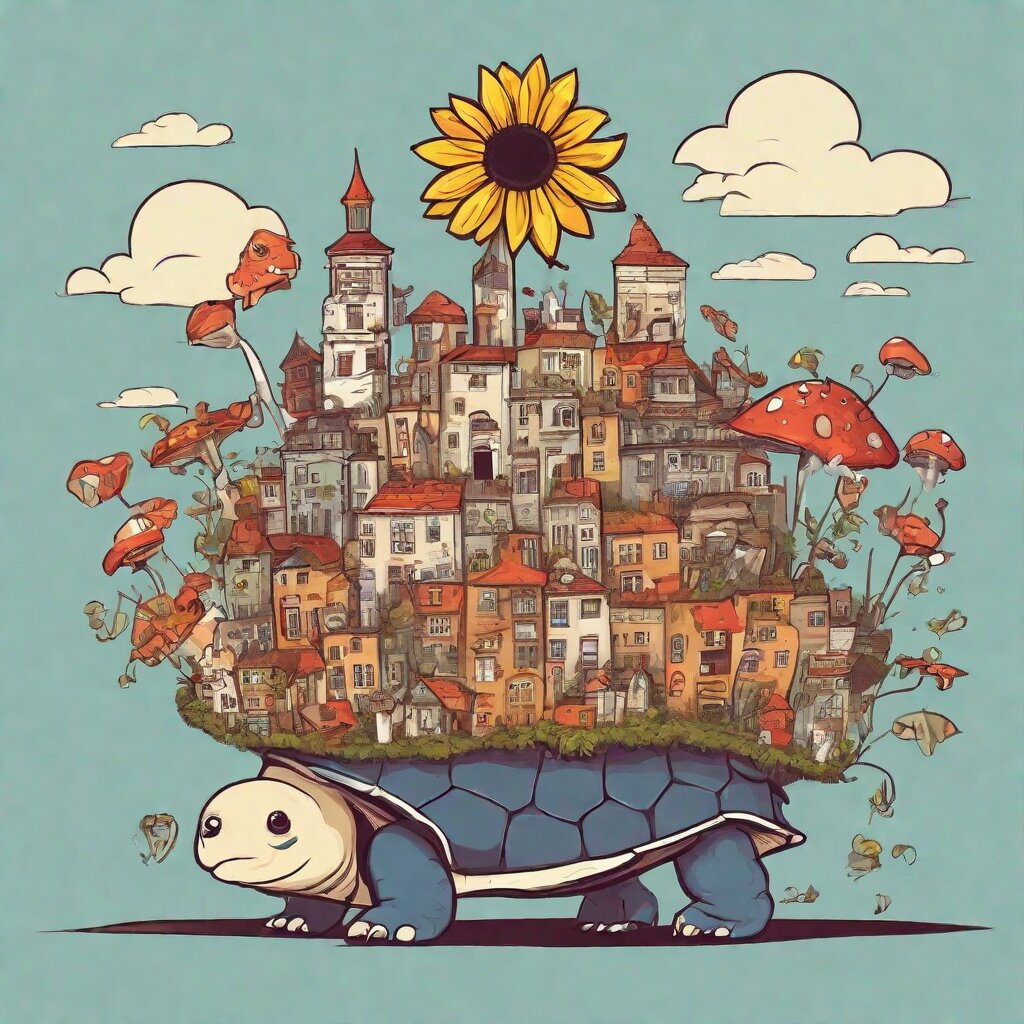} &
        \includegraphics[width=0.11\textwidth]{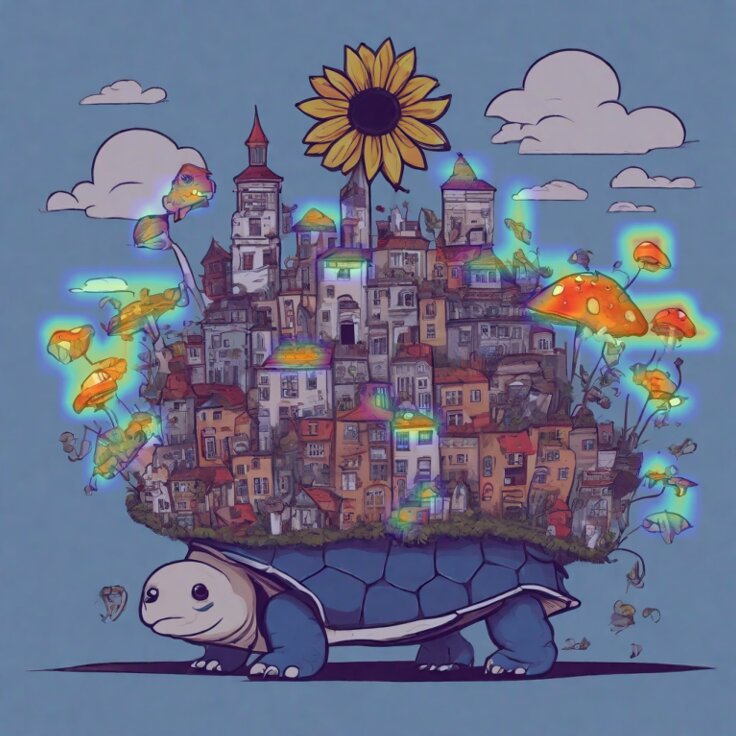} &
        \includegraphics[width=0.11\textwidth]{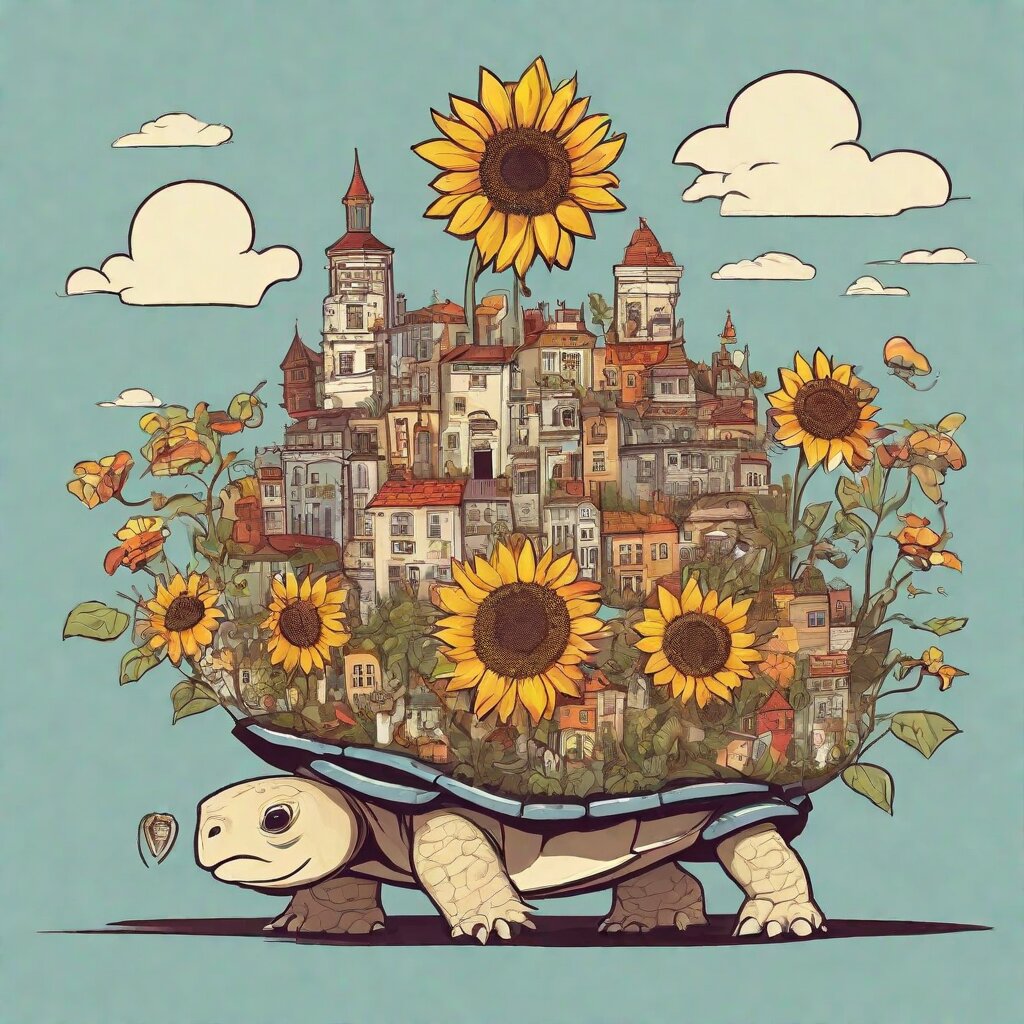} \\
        \multicolumn{2}{c}{\footnotesize Mushroom city on a whale’s back} & \multicolumn{2}{c}{\footnotesize Sunflower city on a turtle’s shell} \\

        \includegraphics[width=0.11\textwidth]{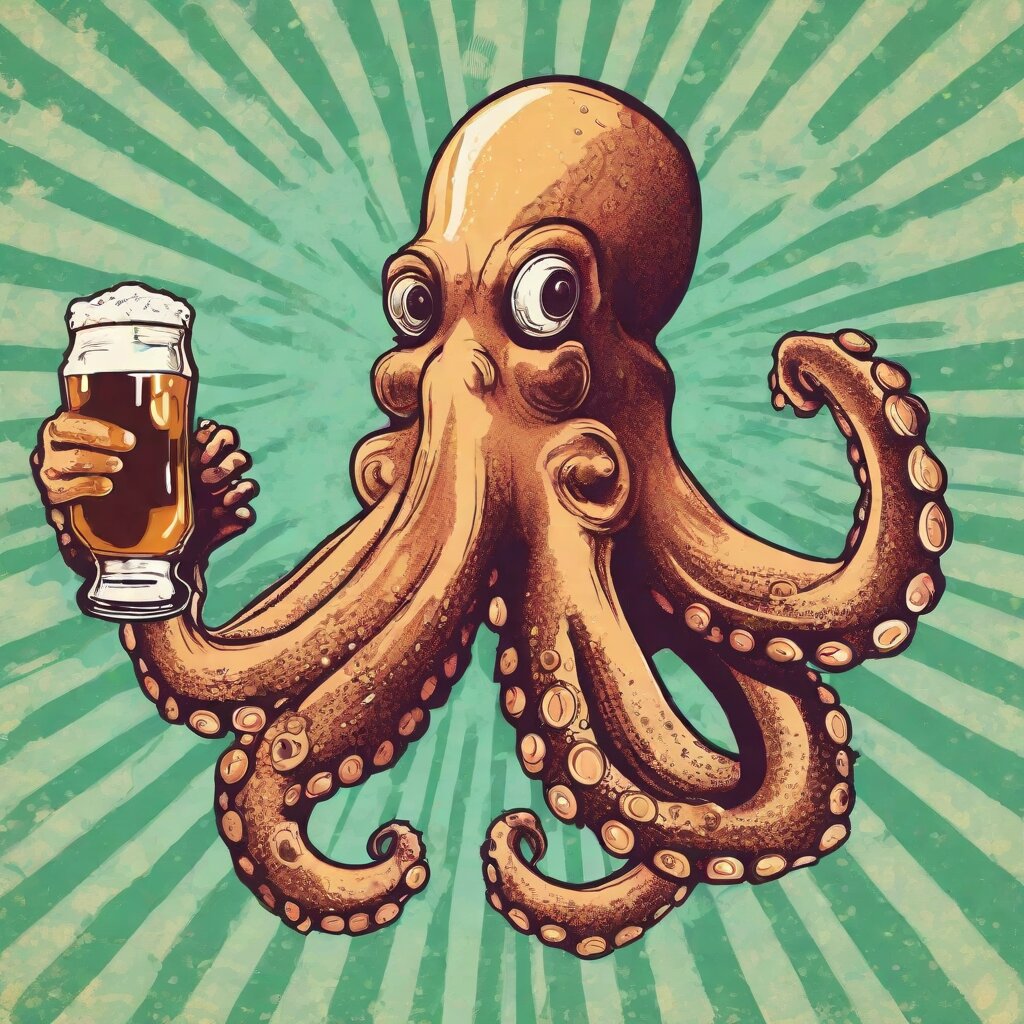} &
        \includegraphics[width=0.11\textwidth]{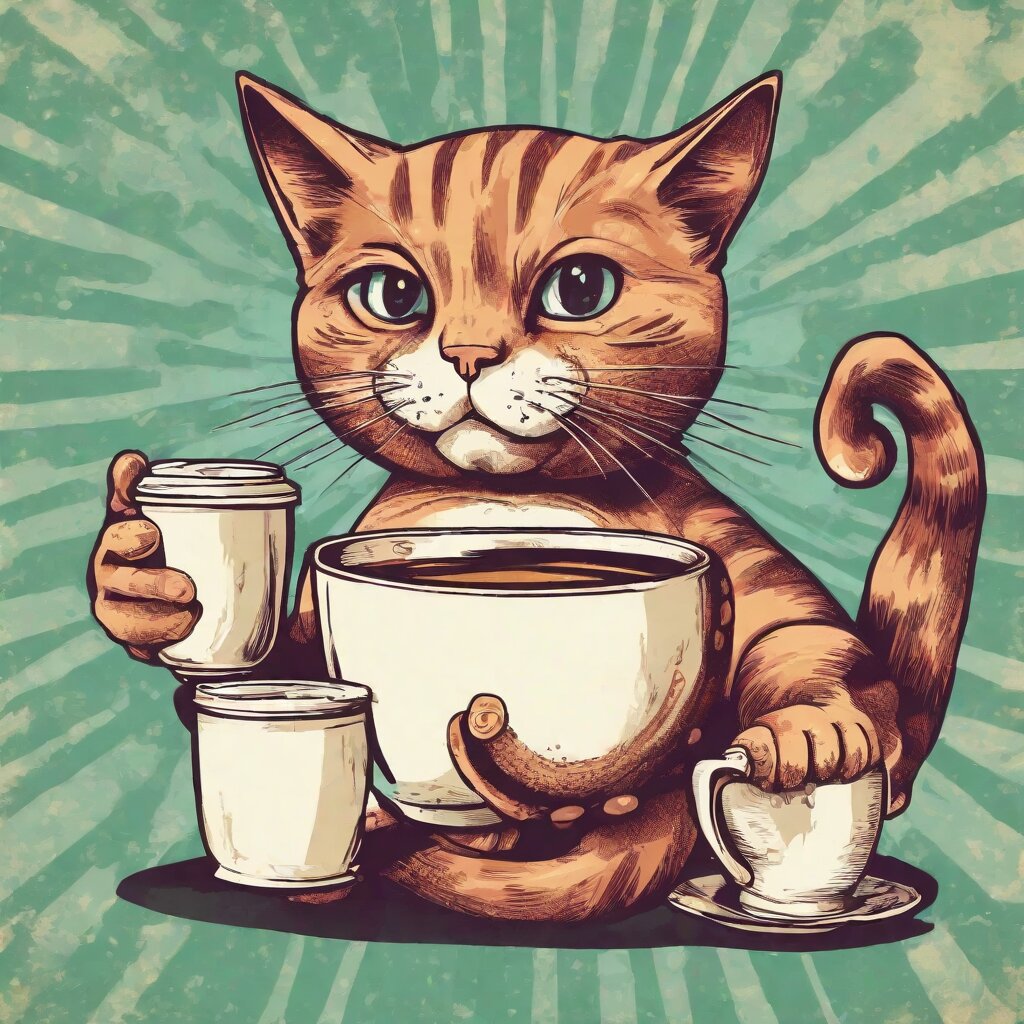} &
        \includegraphics[width=0.11\textwidth]{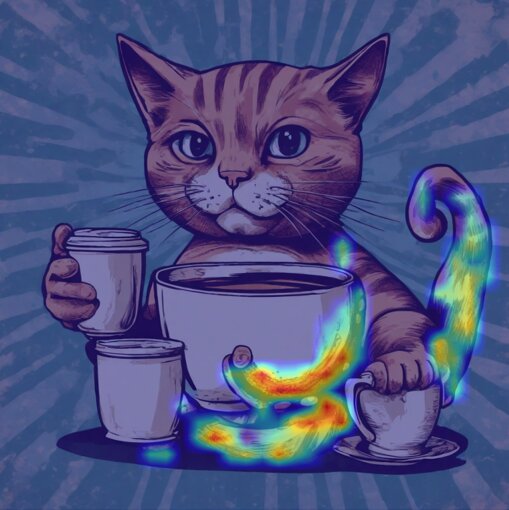} &
        \includegraphics[width=0.11\textwidth]{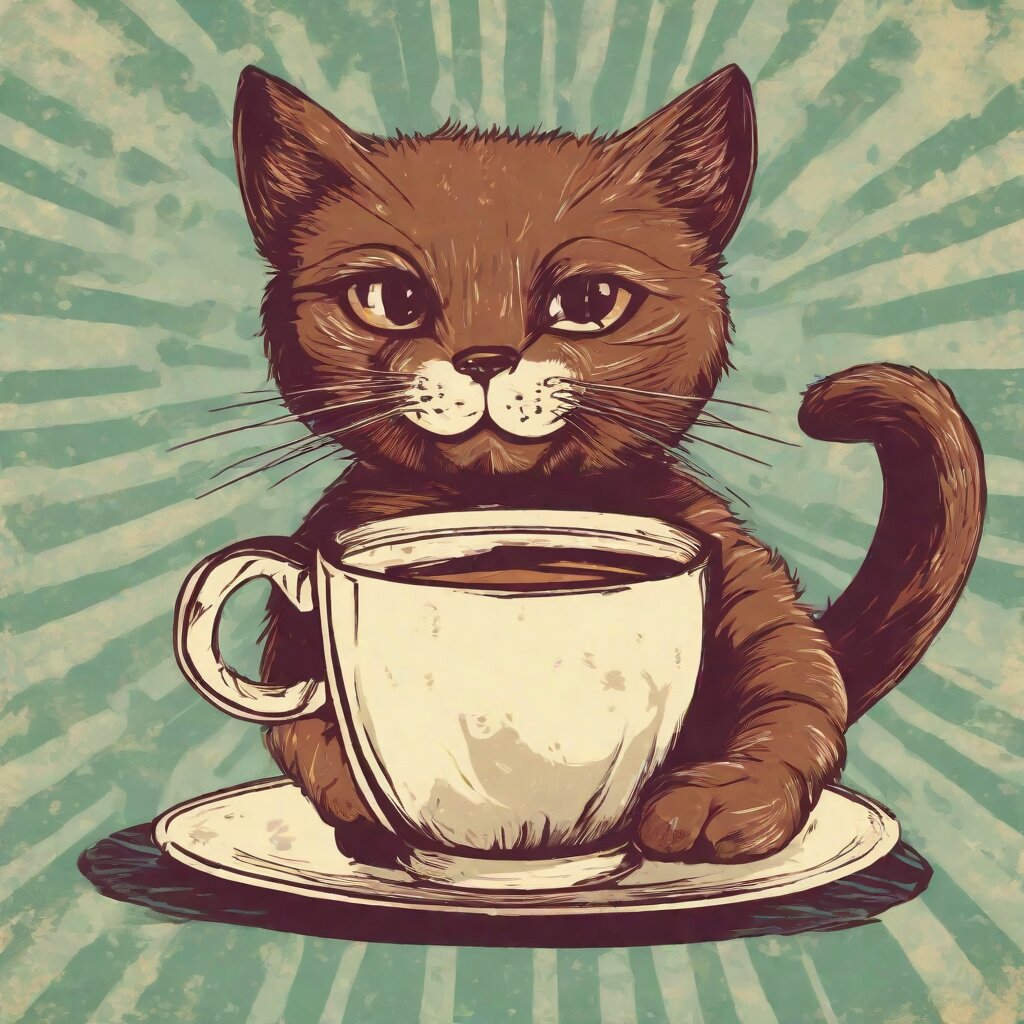} \\
        \multicolumn{2}{c}{\footnotesize ``An octopus holding a beer"} & \multicolumn{2}{c}{\footnotesize ``A cat holding a cup of coffee"} \\

    \end{tabular}
    
    \caption{\textbf{Examples including multi-reference and multi-target subjects.} Only-Style can be directly extended to remove content leakage in multi-reference and multi-target subject scenes.}
    \label{fig:multi}
\end{figure}

\subsection{Extensions}

\textbf{Multi-Image Extension.} 
To create a set of style-consistent images using the same stylistic reference image, we follow the StyleAligned~\cite{hertz2024style} approach by extending the batch with multiple target images. Specifically, the target images attend to the first image in the batch, which serves as the reference. 
Our end-to-end method can be applied independently to each target image by replicating the process described in the main manuscript. 
This involves defining a unique scaling parameter $\alpha$ for each target image and using a binary search algorithm to optimize the scaling by localizing content leakage for each such image. 
Importantly, this approach preserves batch parallelism, as both content leakage control and localization rely on tensor operations that can be executed in parallel. 
An example of a stylistically aligned image set is illustrated in Fig. 1 of the main manuscript.

\textbf{Multi-subject Extension.} 
In the main manuscript, we analyzed the single-subject scenario, where both the reference and the target prompt contained one subject. Nonetheless, our method can easily be extended in multi-subject scenarios. 

For multiple reference subjects, our approach can be generalized by replicating the process outlined in the main manuscript. 
Here, we assume that each subject can have an optimal scaling value independent of the values selected for the other reference subjects - such an assumption stems that in theory the subject masks should be disjoint and scale a different part of the common reference image.
Thus, exactly as in the multi-image scenario, we duplicate the batch and independently apply the end-to-end method to each subject, disregarding the others.
Finally, we combine the optimal scaling parameters $\alpha$ for each reference subject to generate the resulting image, ensuring that no subject experiences leakage. 
It is important to note that this process requires extending the batch to include as many images as there are reference subjects, as well as performing an extra final generation of the optimal scaling set. 
These operations increase computational overhead, both time-wise (the extra generation step leads to a $\times 6$ overhead, including the binary search, to the standard StyleAligned for this batched multi-reference case) and memory-wise (memory requirements are multiplied by the number of reference subjects). 
We illustrate some indicative examples on Fig.~\ref{fig:multi}. Our experimentation with multiple subjects shows that usually only the visually dominant reference subject leaks in the target image, as text-to-image models frequently focus on one subject in multi-subject scenarios \cite{chefer2023attendandexcite}.

For multiple target subjects, we just need to perform the content leakage localization (Sec. 3.4 of the main manuscript) of the reference subject with each target one, distinctly. 
Essentially we check if any patch of the generated target image contains more information about the reference subject than each of the target ones. 
It is worth noting that this requires extracting a subject representation for each target subject, which adds minimal computational overhead, since it is only performed at the last iteration of the generation process (see Sec. 3.4 of the main manuscript).

\begin{figure}[t]
    \centering
    \begin{tikzpicture}
        \begin{axis}[
            scale only axis,
            width=0.78\linewidth,
            height=0.55\linewidth,
            xlabel={Text Alignment →},
            ylabel={Set Consistency →},
            xmin=0.264, xmax=0.283,
            ymin=0.208, ymax=0.395,
            grid=both,
            grid style={dashed,gray!30},
            tick label style={font=\small},
            label style={font=\small},
            xticklabel style={
                /pgf/number format/.cd,
                fixed zerofill,
                precision=3
            },
            scaled x ticks=false,
        ]
          
            \addplot[only marks, mark=*, mark size=1.5pt, color=blue] coordinates {
                (0.2720, 0.3450)  
                (0.2690, 0.3130)  
                (0.2655, 0.3780)  
            };
           
            \node[font=\scriptsize] at (axis cs:0.2720+0.00, 0.3450-0.009) {CSGO};
            \node[font=\scriptsize] at (axis cs:0.2690+0.0006, 0.3130-0.009) {IP-Adapter};
            \node[font=\scriptsize] at (axis cs:0.2655+0.0006, 0.3780-0.009) {DB-LoRA};

            \addplot[only marks, mark=*, mark size=1.5pt, color=green] coordinates {
                (0.2789, 0.3447)  
            };
           
            \node[font=\scriptsize] at (axis cs:0.2789+0.0006, 0.3447-0.009) {Only-Style};

            \addplot[only marks, mark=*, mark size=1.5pt, color=gray] coordinates {
                (0.2801, 0.2254)  
            };
           
            \node[font=\scriptsize] at (axis cs:0.2801+0.000, 0.2254-0.008) {Standard\_T2I};

        \end{axis}
    \end{tikzpicture}
    \caption[Quantitative results]{\textbf{Text Alignment vs Stylistic Set Consistency}: We compare three additional state-of-the-art methods (blue marks), a baseline without stylistic alignment (grey mark) and \emph{Only-Style} (green mark) in terms of text alignment (CLIP similarity) and set consistency (DINO similarity).
    \vspace{-2pt}
    }
    \label{fig:Quantitative_add}
\end{figure}
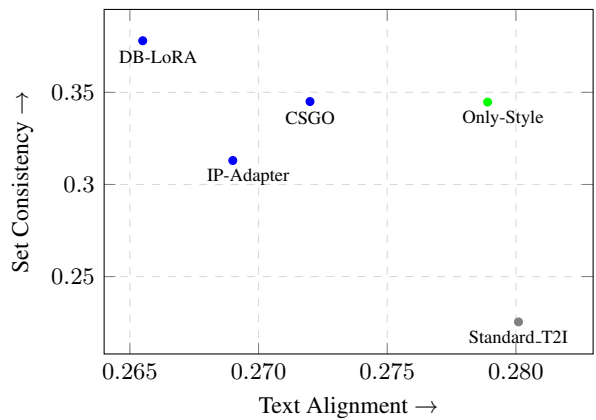

\begin{figure*}[t]
\vspace{-2pt}
    \centering
    \resizebox{.98\linewidth}{!}{
    \begin{tabular}{c@{\hspace{.1cm}}c@{\hspace{.1cm}}c@{\hspace{.1cm}}c@{\hspace{.1cm}}c@{\hspace{.1cm}}c@{\hspace{.1cm}}c}
        \scriptsize Reference & \scriptsize IP-Adapter~\cite{Ye2023IPAdapter} & \scriptsize CSGO~\cite{Xing2024CSGO} & \scriptsize DB-LoRA~\cite{ruiz2023dreamboothfinetuningtexttoimage} & \scriptsize StyleAligned~\cite{hertz2024style} & \scriptsize Content Leakage & \scriptsize Only-Style (Ours)\\
        \begin{minipage}{0.12\textwidth}
            \includegraphics[width=\textwidth]{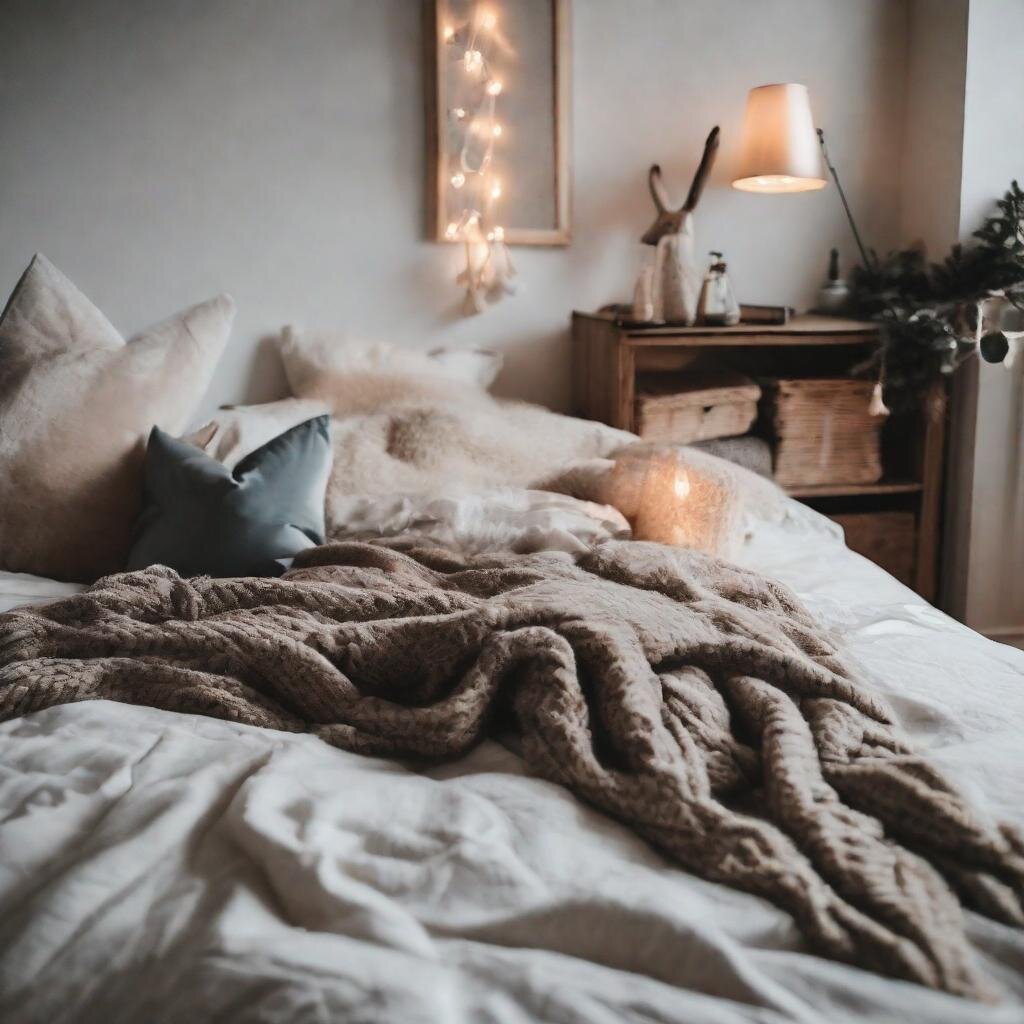}
        \end{minipage} &
        \begin{minipage}{0.12\textwidth}
            \includegraphics[width=\textwidth]{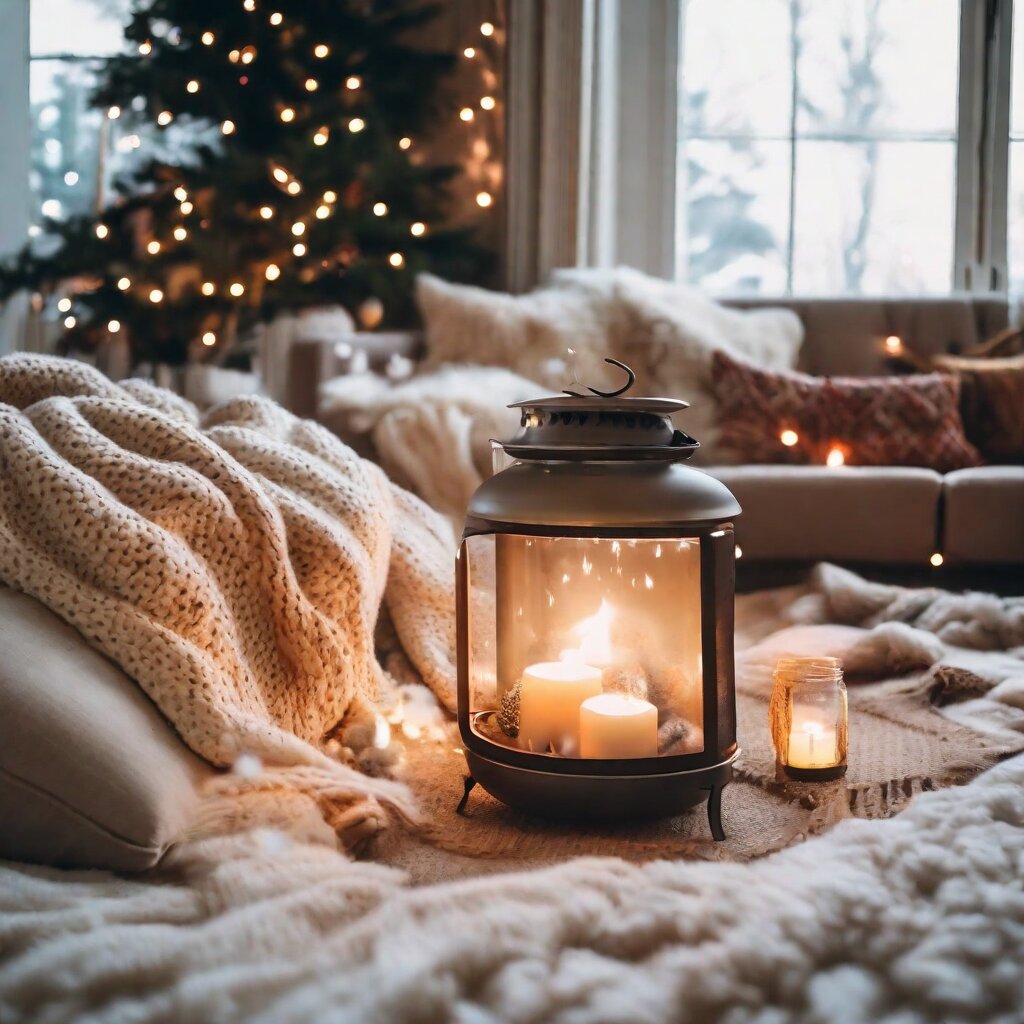}
        \end{minipage} &
        \begin{minipage}{0.12\textwidth}
            \includegraphics[width=\textwidth]{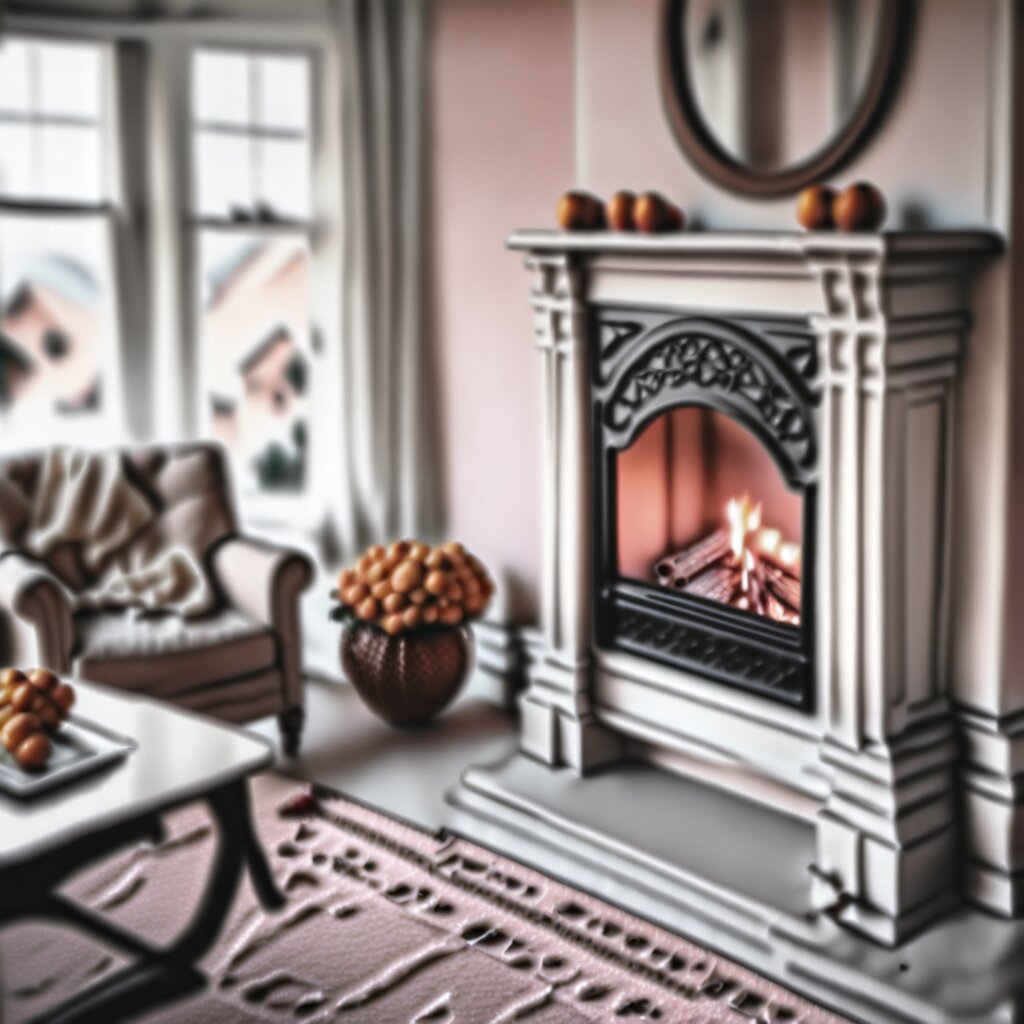}
        \end{minipage} &
        \begin{minipage}{0.12\textwidth}
            \includegraphics[width=\textwidth]{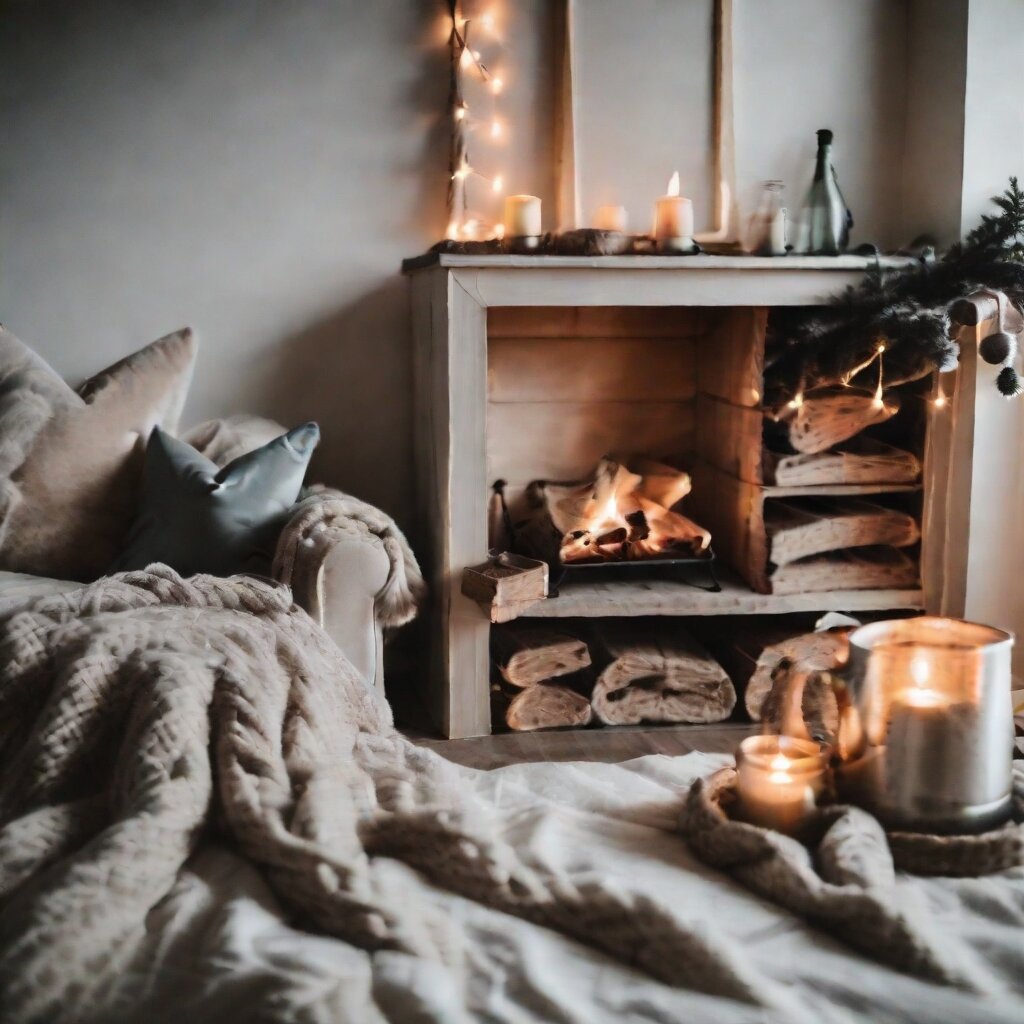}
        \end{minipage} &
        \begin{minipage}{0.12\textwidth}
            \includegraphics[width=\textwidth]{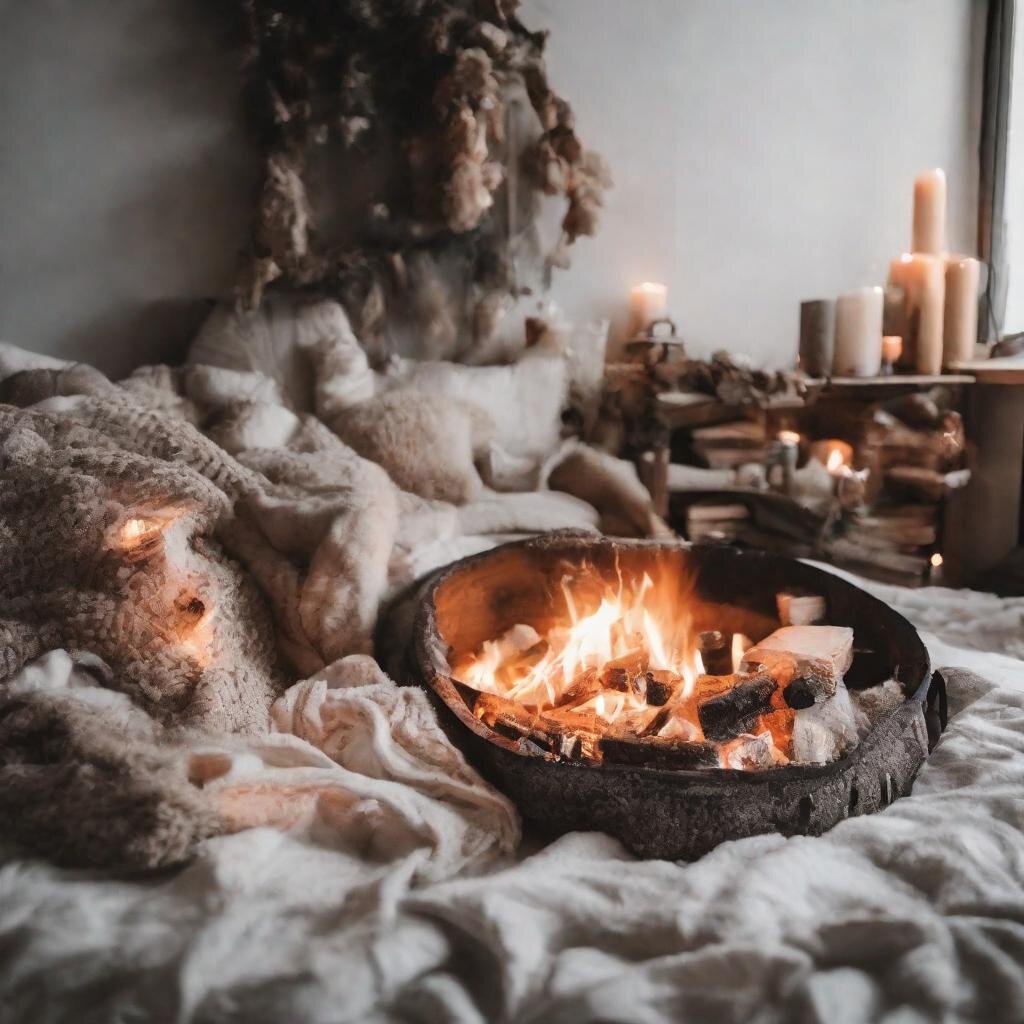}
        \end{minipage} &
        \begin{minipage}{0.12\textwidth}
            \includegraphics[width=\textwidth]{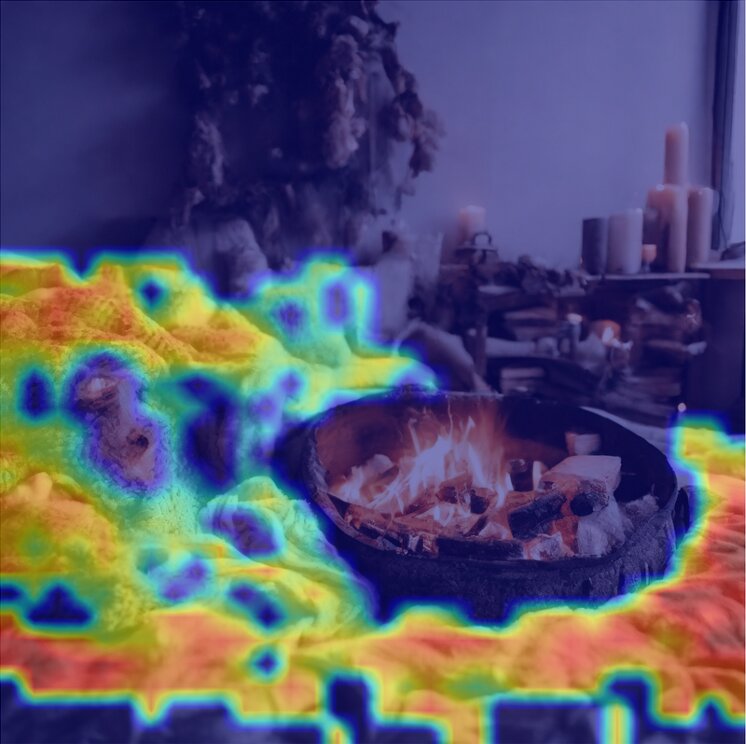}
        \end{minipage} &
        \begin{minipage}{0.12\textwidth}
            \includegraphics[width=\textwidth]{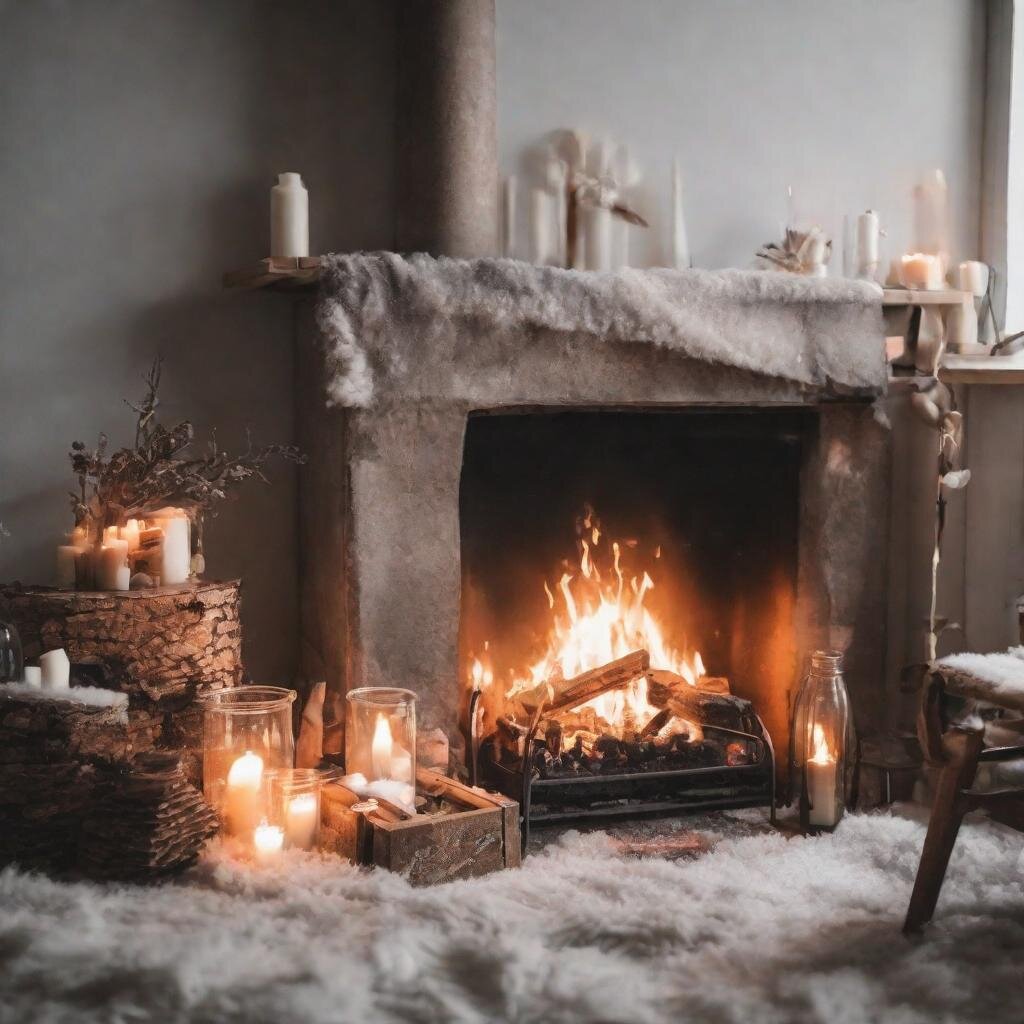}
        \end{minipage} 
        \vspace{-2pt} 
        \\
        \vspace{-1pt} 
        \scriptsize ``A bed" & \scriptsize ``A fireplace" &
        \multicolumn{5}{c}{\scriptsize ``...in cozy winter lifestyle photography style."} \\
        
        \begin{minipage}{0.12\textwidth}
            \includegraphics[width=\textwidth]{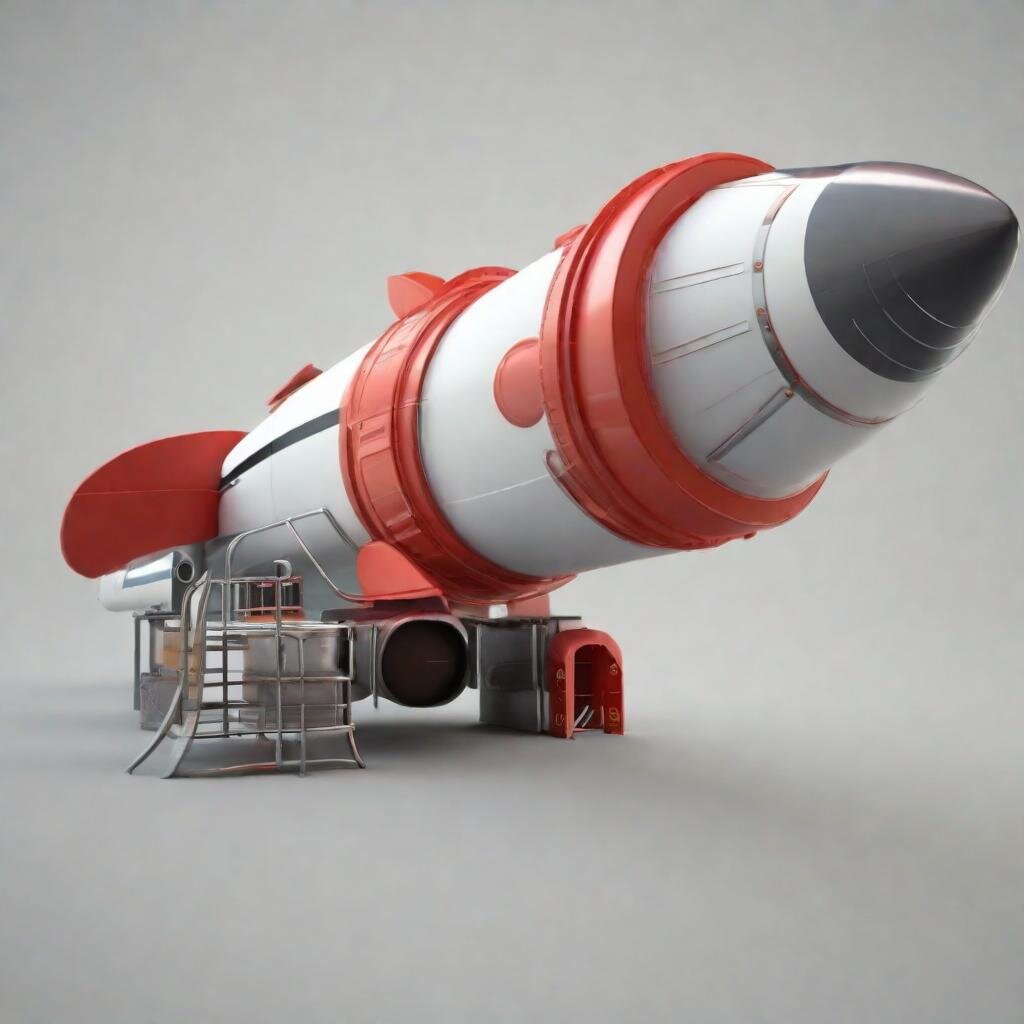}
        \end{minipage} &
        \begin{minipage}{0.12\textwidth}
            \includegraphics[width=\textwidth]{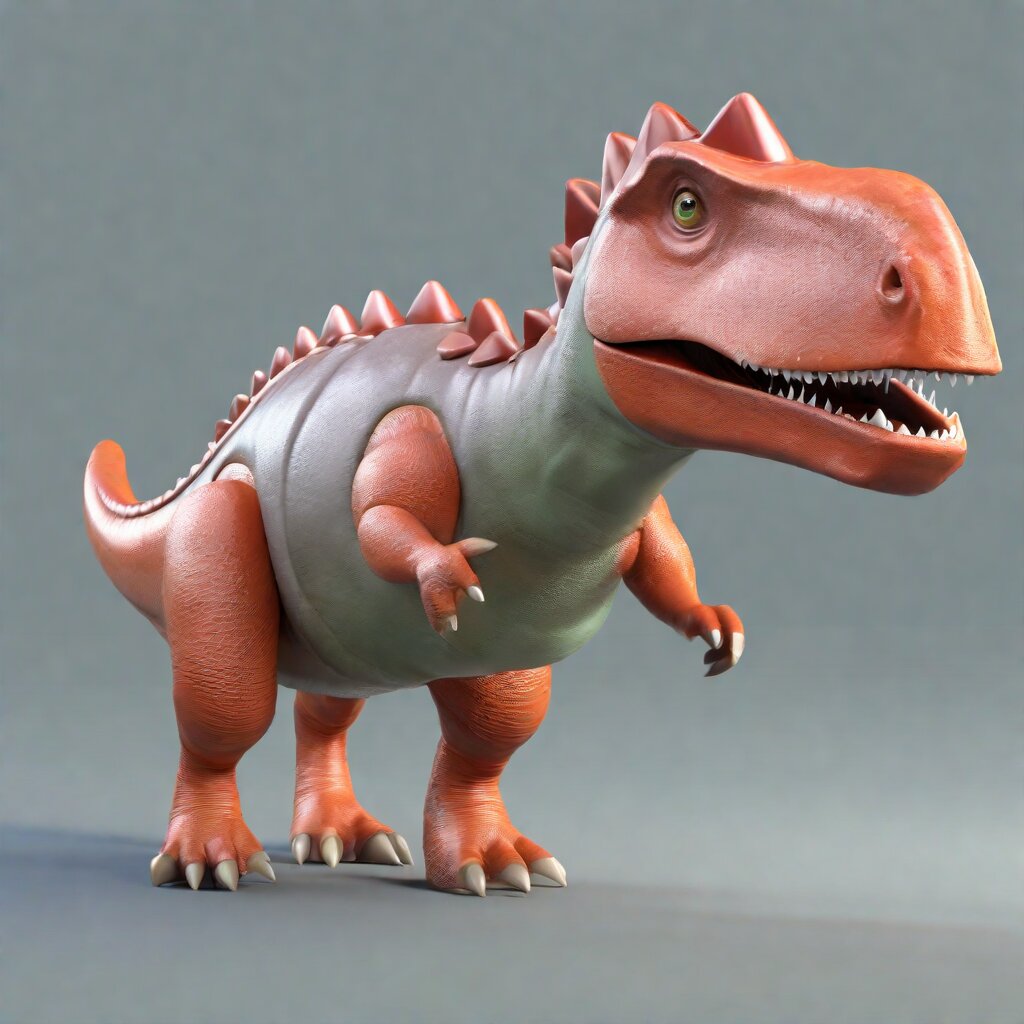}
        \end{minipage} &
        \begin{minipage}{0.12\textwidth}
            \includegraphics[width=\textwidth]{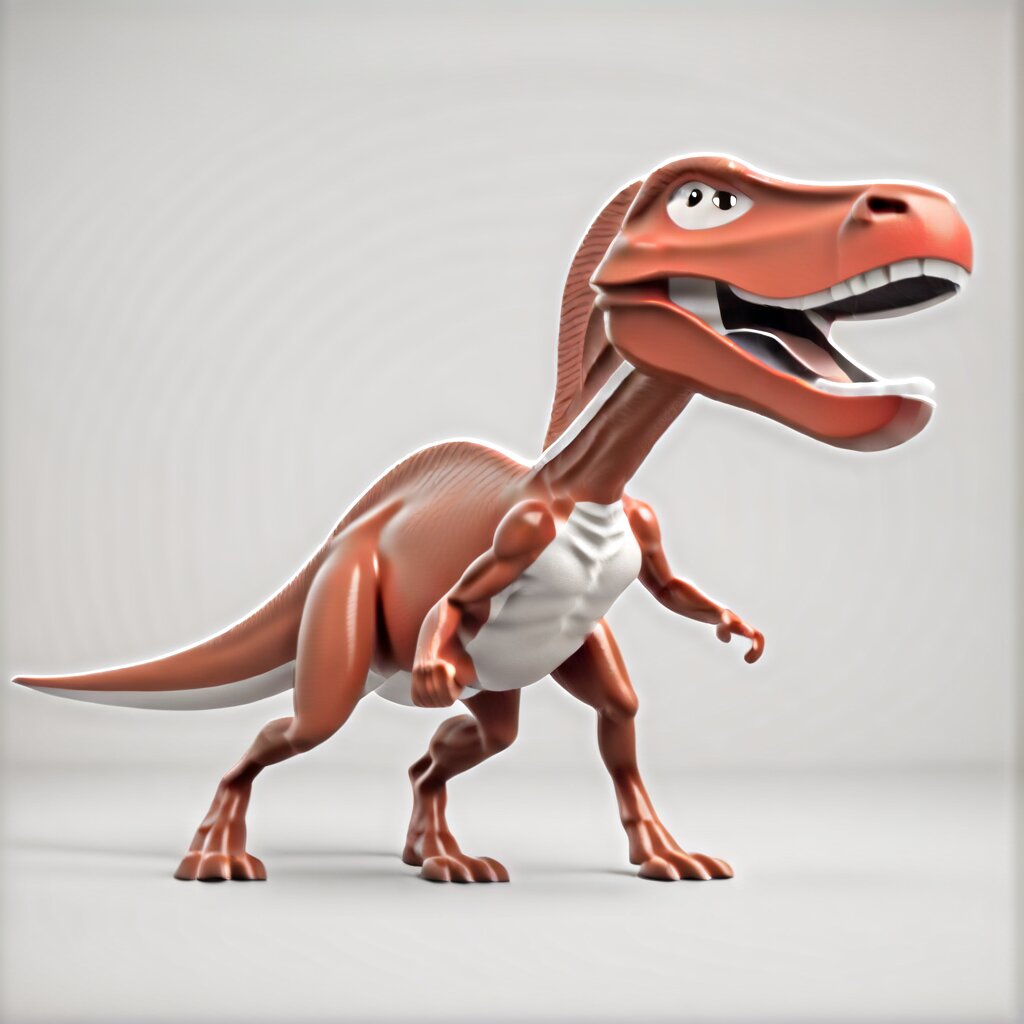}
        \end{minipage} &
        \begin{minipage}{0.12\textwidth}
            \includegraphics[width=\textwidth]{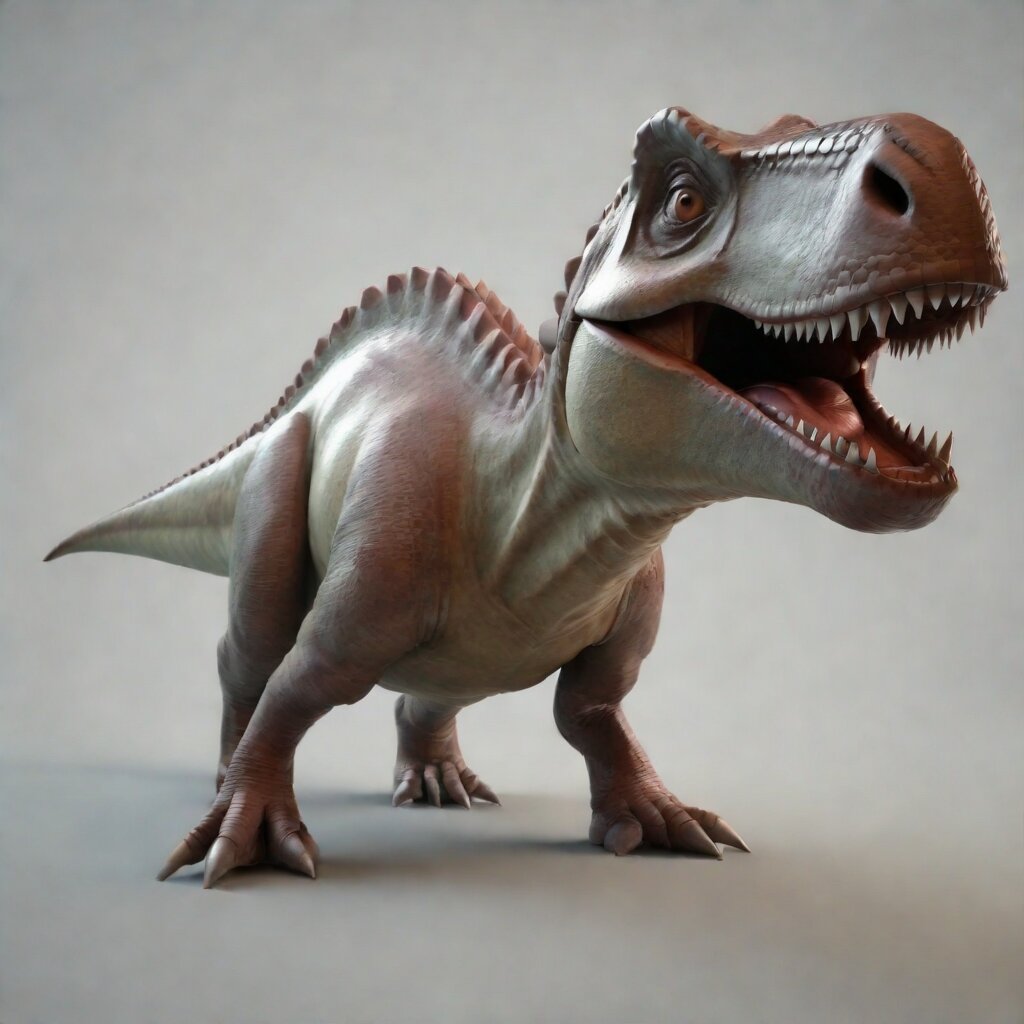}
        \end{minipage} &
        \begin{minipage}{0.12\textwidth}
            \includegraphics[width=\textwidth]{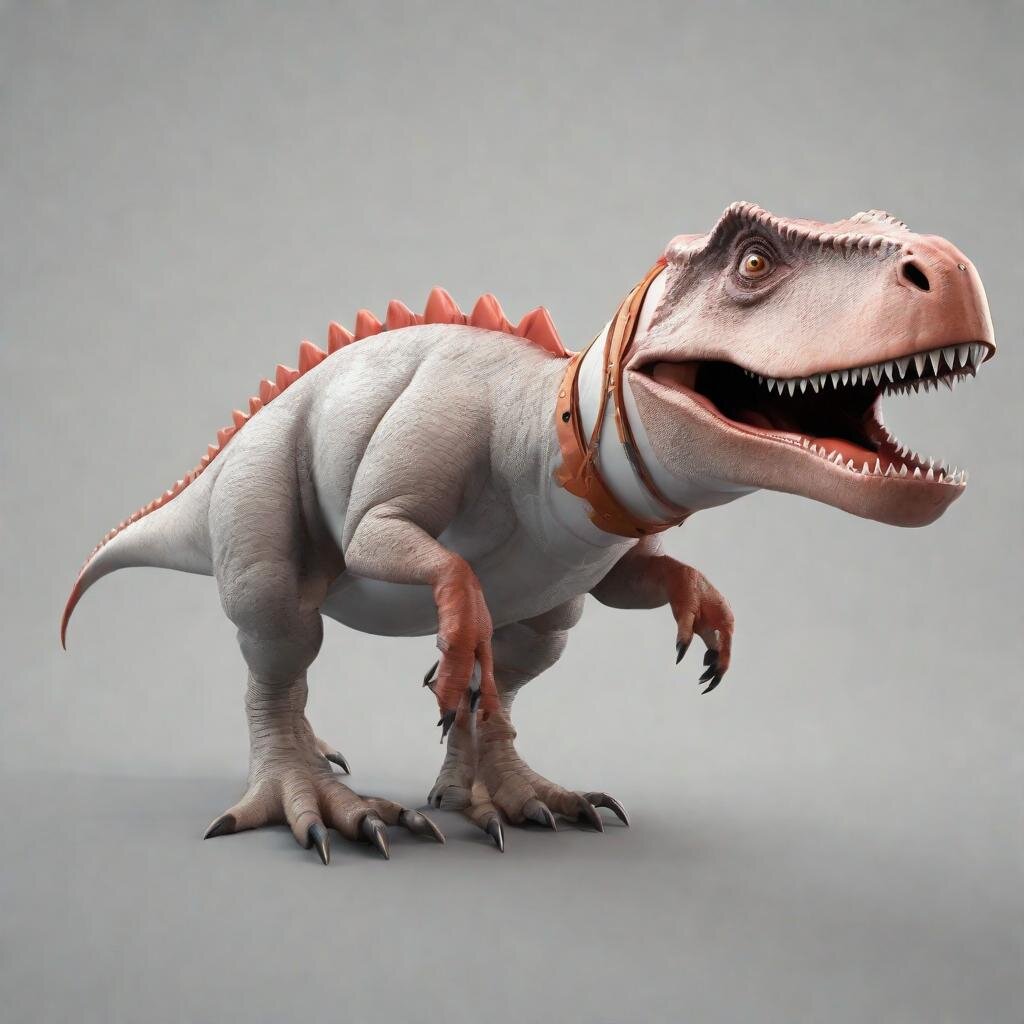}
        \end{minipage} &
        \begin{minipage}{0.12\textwidth}
            \includegraphics[width=\textwidth]{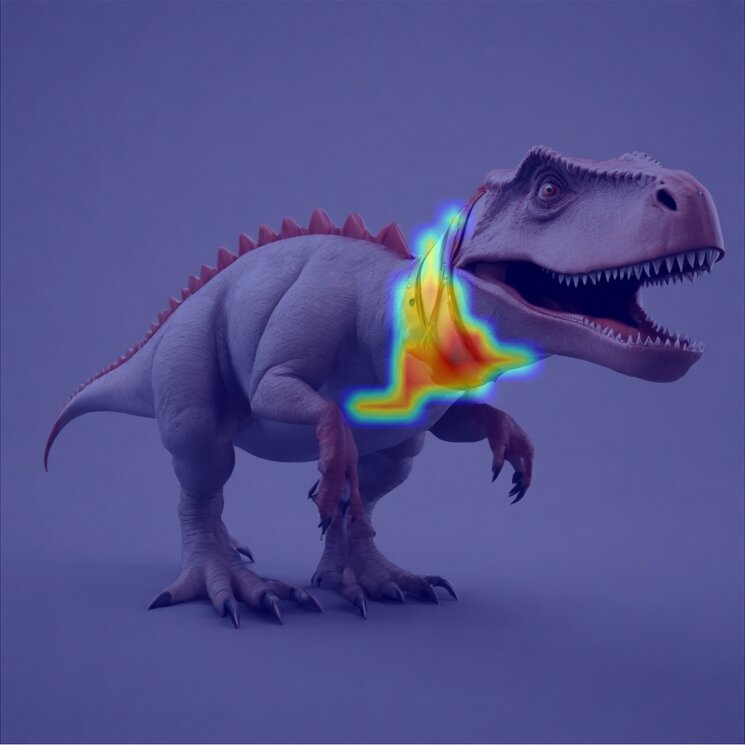}
        \end{minipage} &
        \begin{minipage}{0.12\textwidth}
            \includegraphics[width=\textwidth]{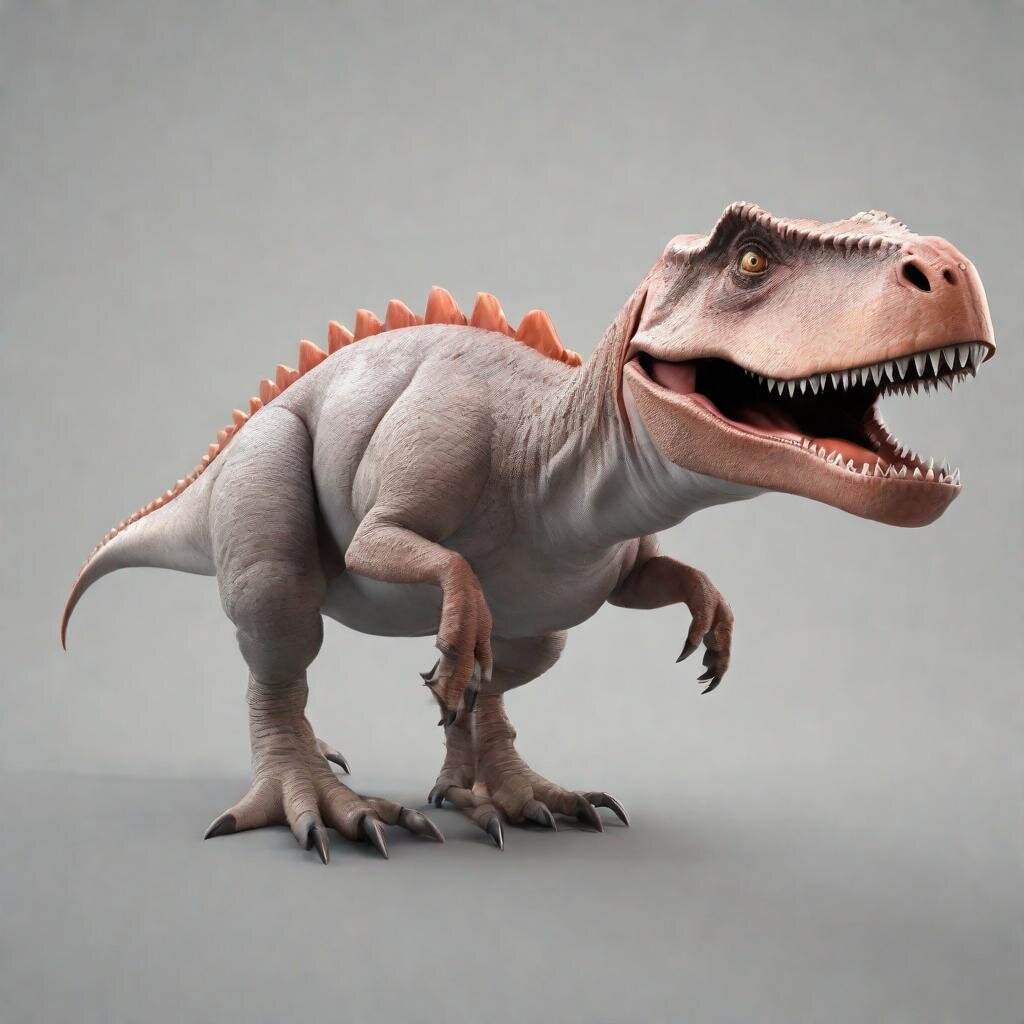}
        \end{minipage} 
        \vspace{-2pt} 
        \\
        \vspace{-1pt} 
        \scriptsize ``A rocket" & \scriptsize ``A dinosaur" &
        \multicolumn{5}{c}{\scriptsize ``...in 3D render, animation studio style."} \\

        \begin{minipage}{0.12\textwidth}
            \includegraphics[width=\textwidth]{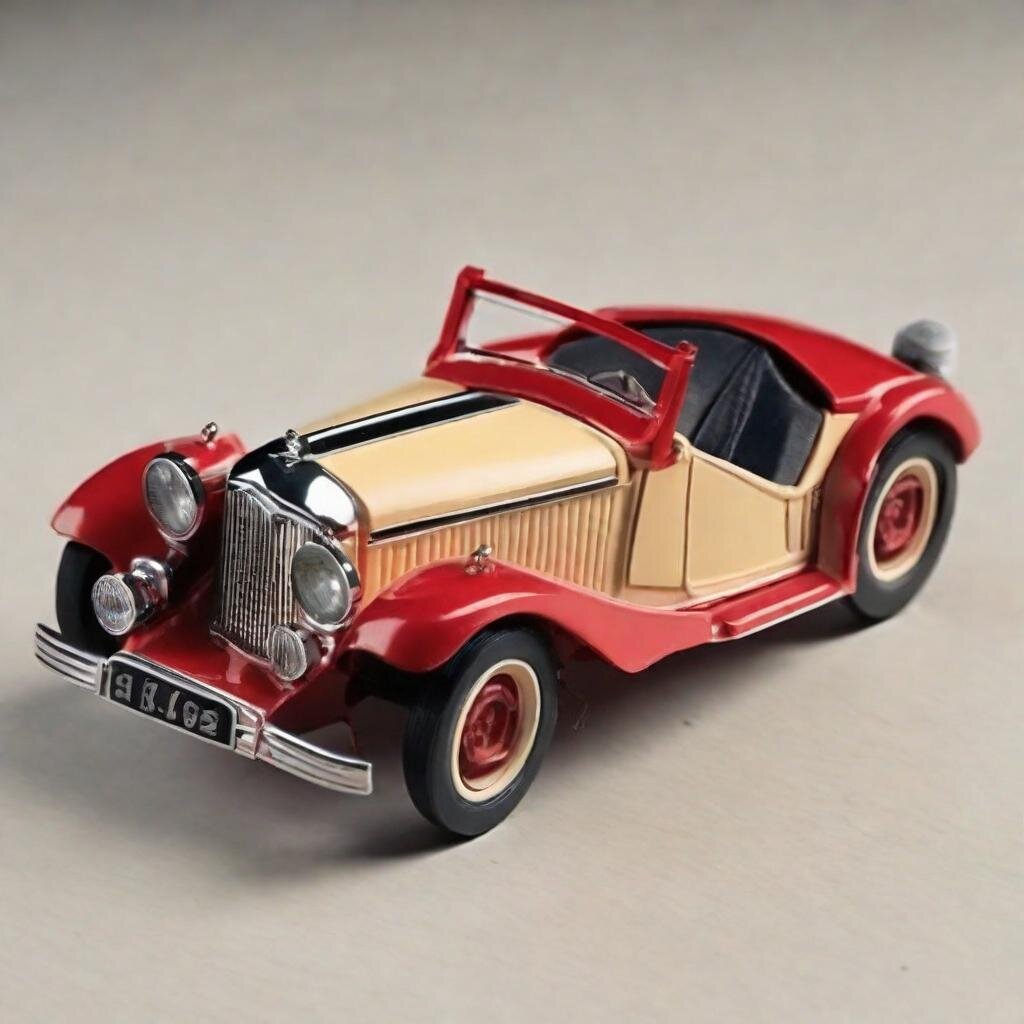}
        \end{minipage} &
        \begin{minipage}{0.12\textwidth}
            \includegraphics[width=\textwidth]{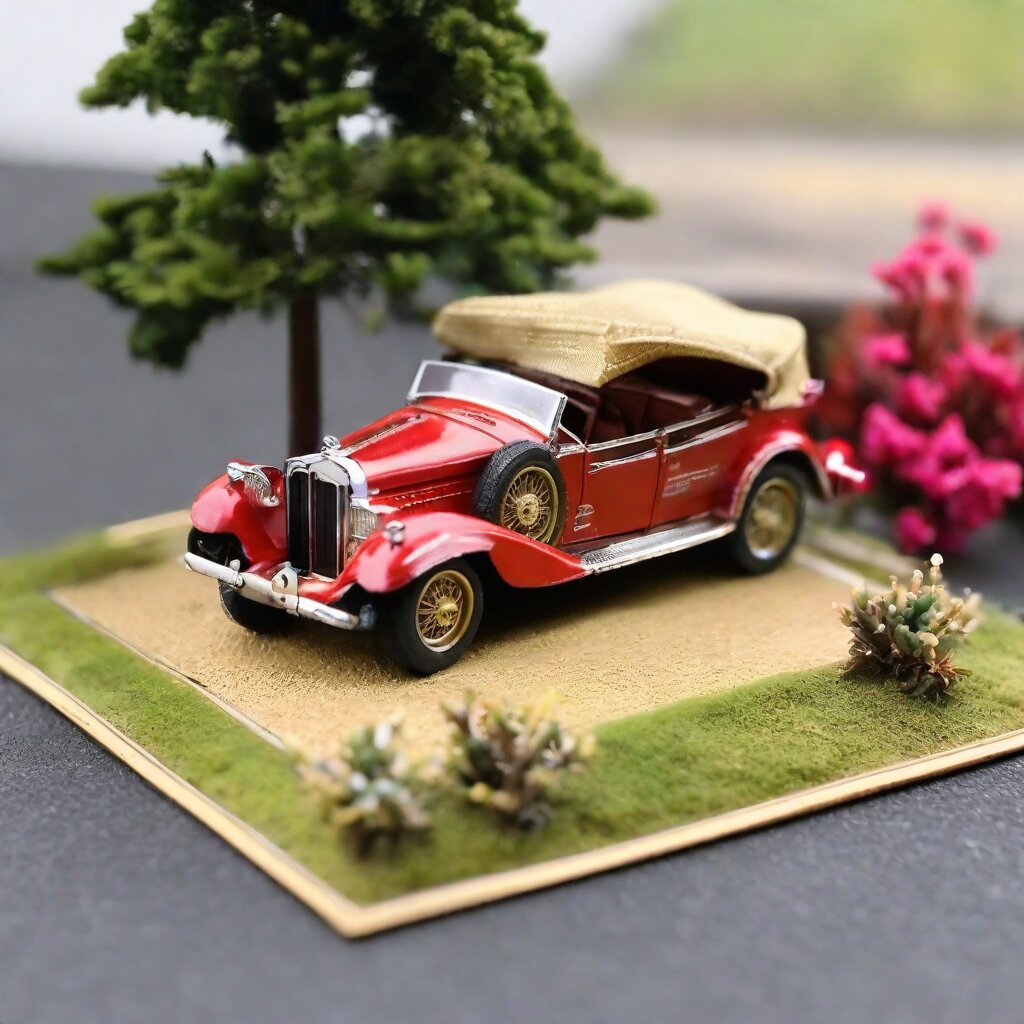}
        \end{minipage} &
        \begin{minipage}{0.12\textwidth}
            \includegraphics[width=\textwidth]{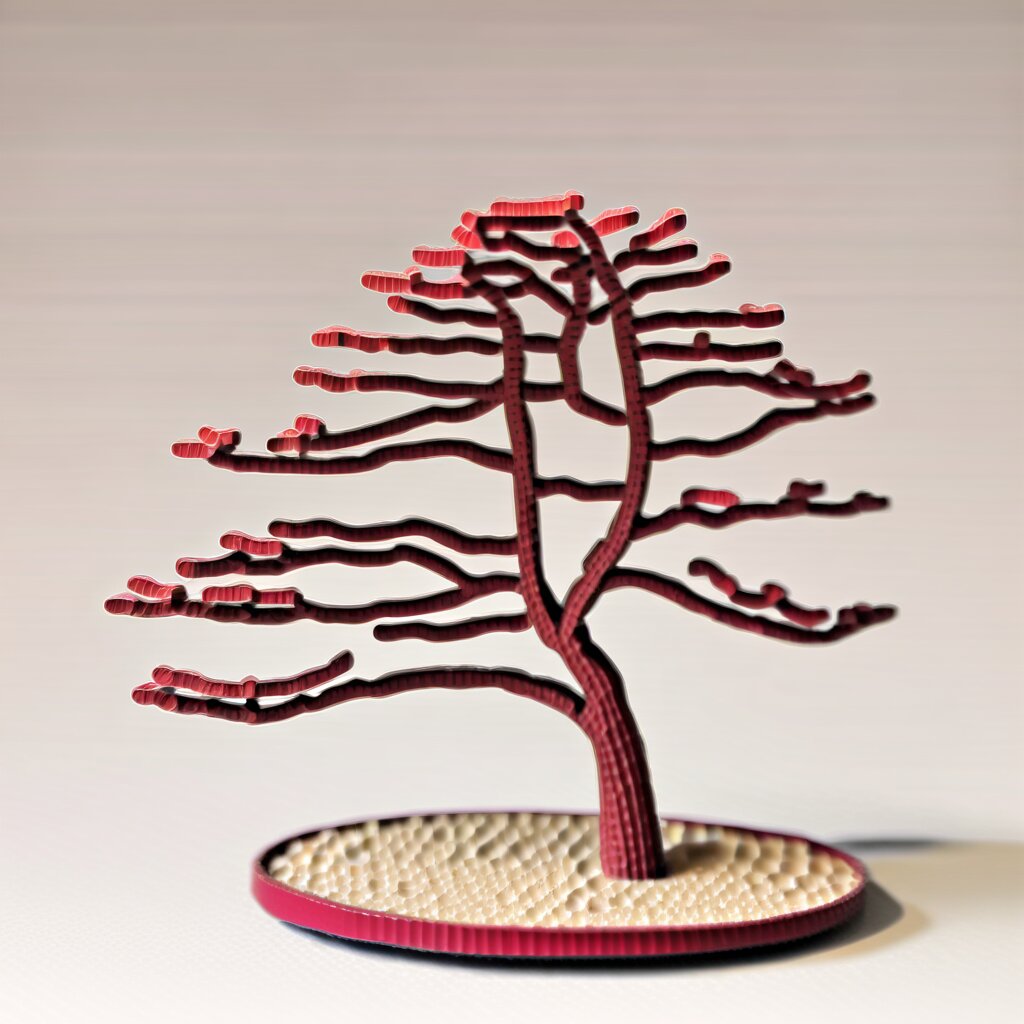}
        \end{minipage} &
        \begin{minipage}{0.12\textwidth}
            \includegraphics[width=\textwidth]{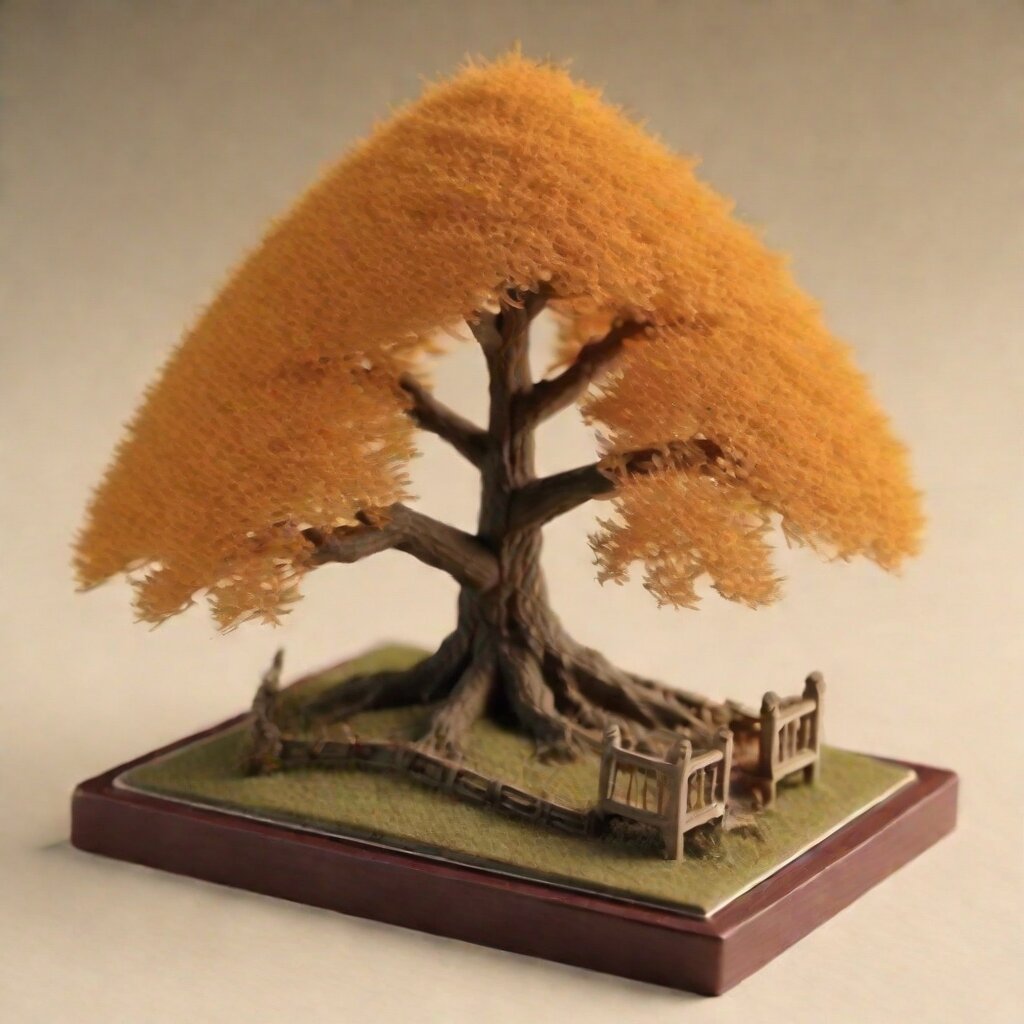}
        \end{minipage} &
        \begin{minipage}{0.12\textwidth}
            \includegraphics[width=\textwidth]{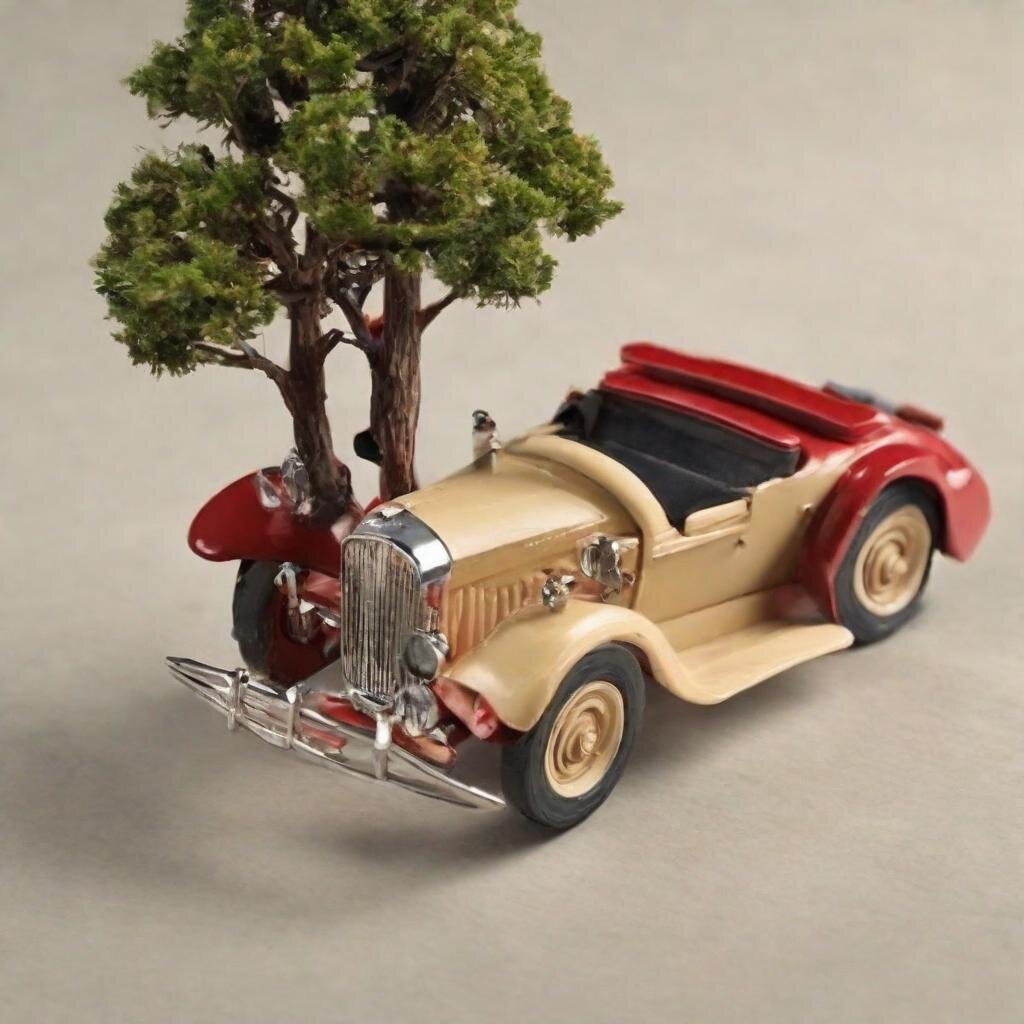}
        \end{minipage} &
        \begin{minipage}{0.12\textwidth}
            \includegraphics[width=\textwidth]{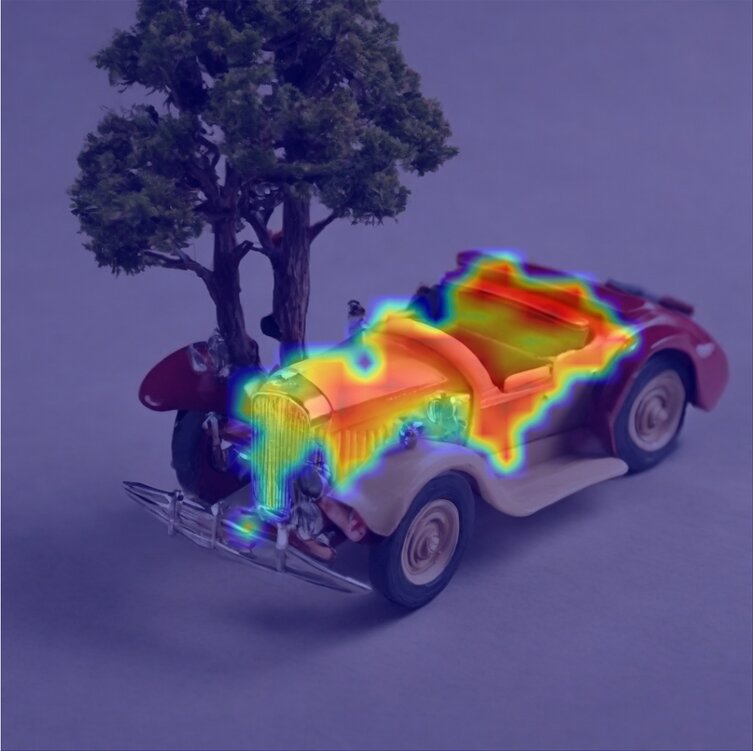}
        \end{minipage} &
        \begin{minipage}{0.12\textwidth}
            \includegraphics[width=\textwidth]{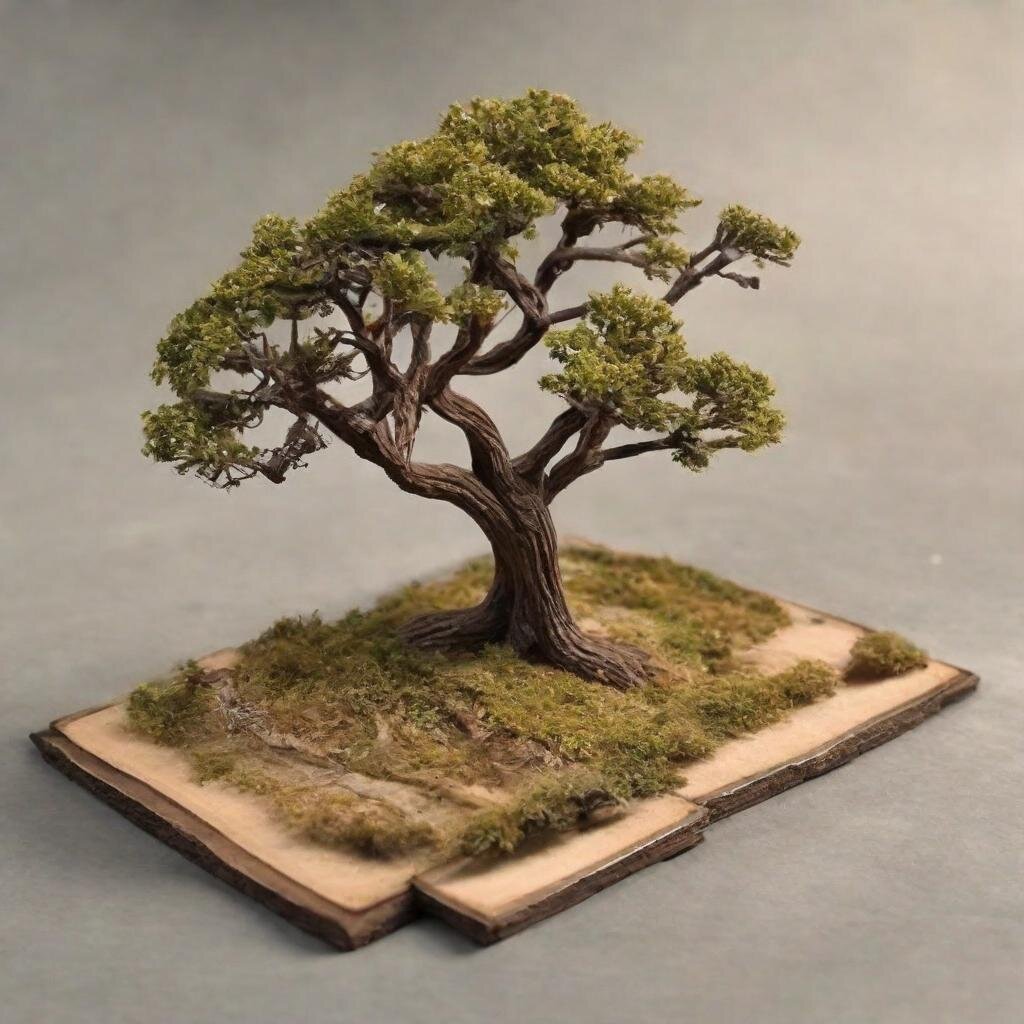}
        \end{minipage}  
        \vspace{-2pt} 
        \\
        \vspace{-1pt} 
        \scriptsize ``A car" & \scriptsize ``A tree" & \multicolumn{5}{c}{\scriptsize ``...in miniature model style."}  \\

        \begin{minipage}{0.12\textwidth}
        \includegraphics[width=\textwidth]{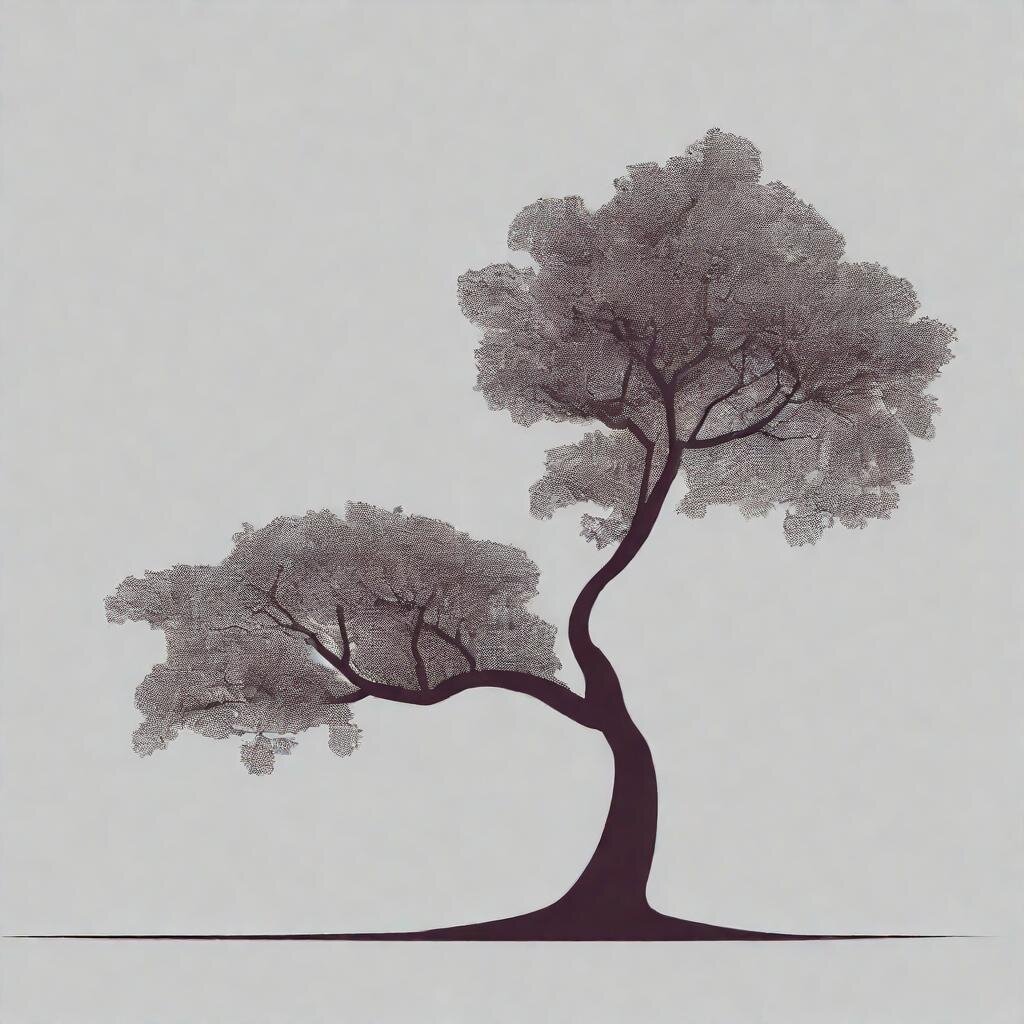}
        \end{minipage} &
        \begin{minipage}{0.12\textwidth}
            \includegraphics[width=\textwidth]{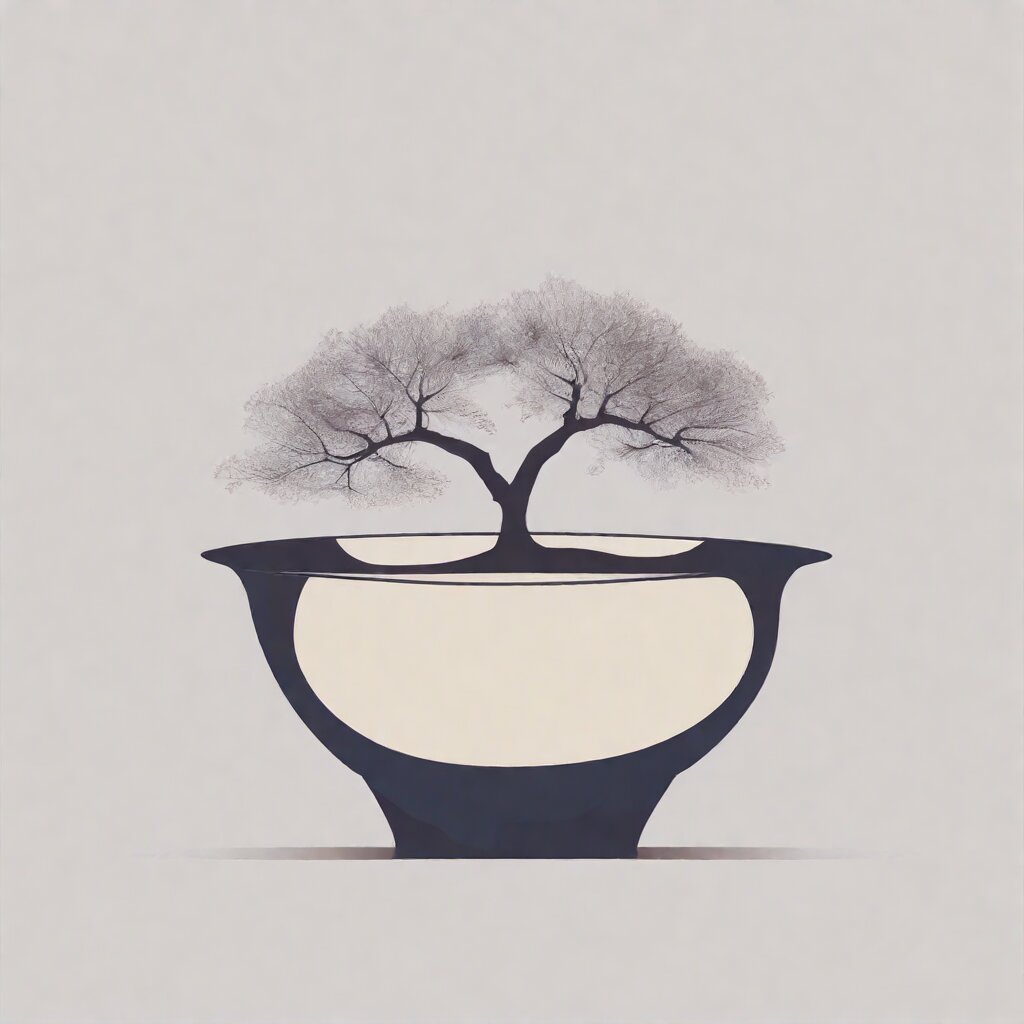}
        \end{minipage} &
        \begin{minipage}{0.12\textwidth}
            \includegraphics[width=\textwidth]{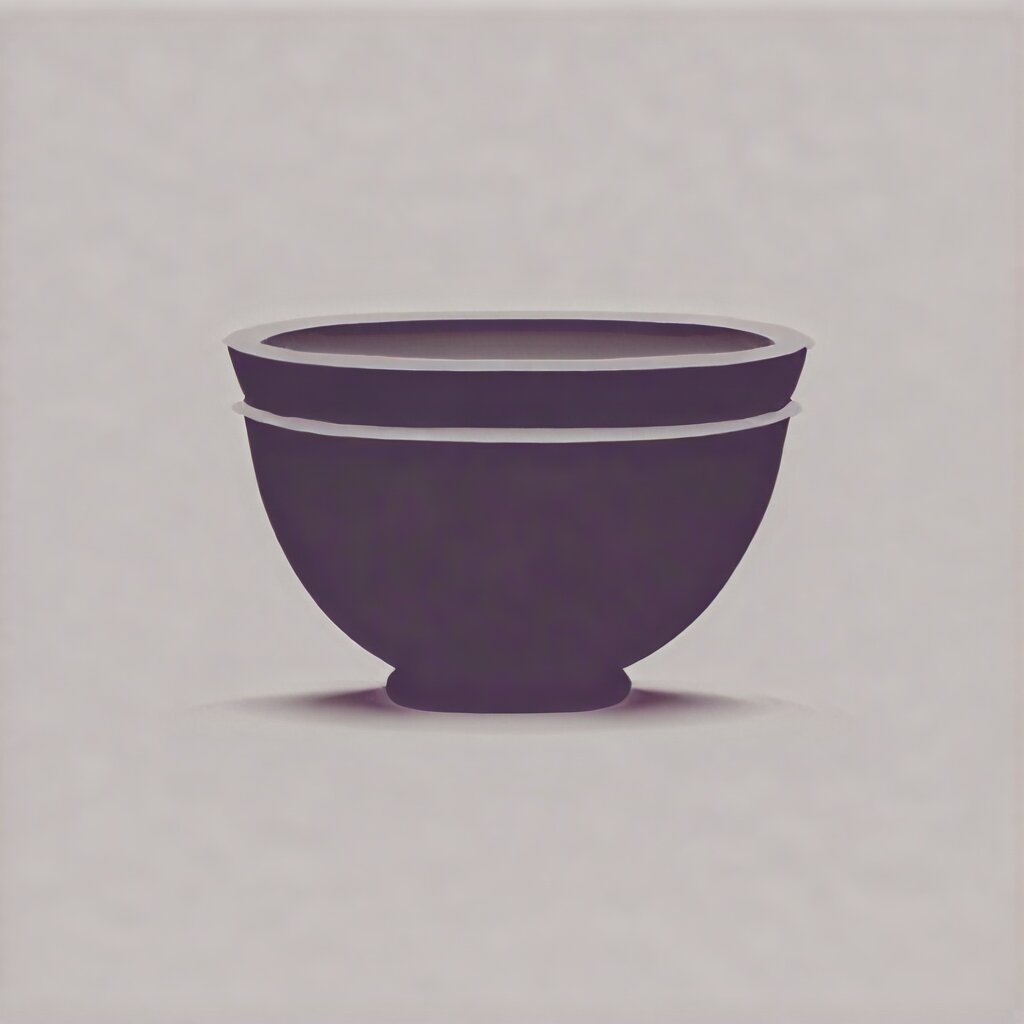}
        \end{minipage} &
        \begin{minipage}{0.12\textwidth}
            \includegraphics[width=\textwidth]{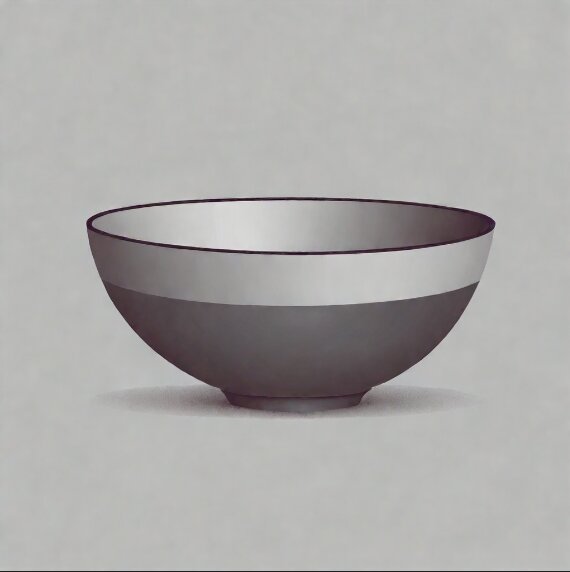}
        \end{minipage} &
        \begin{minipage}{0.12\textwidth}
            \includegraphics[width=\textwidth]{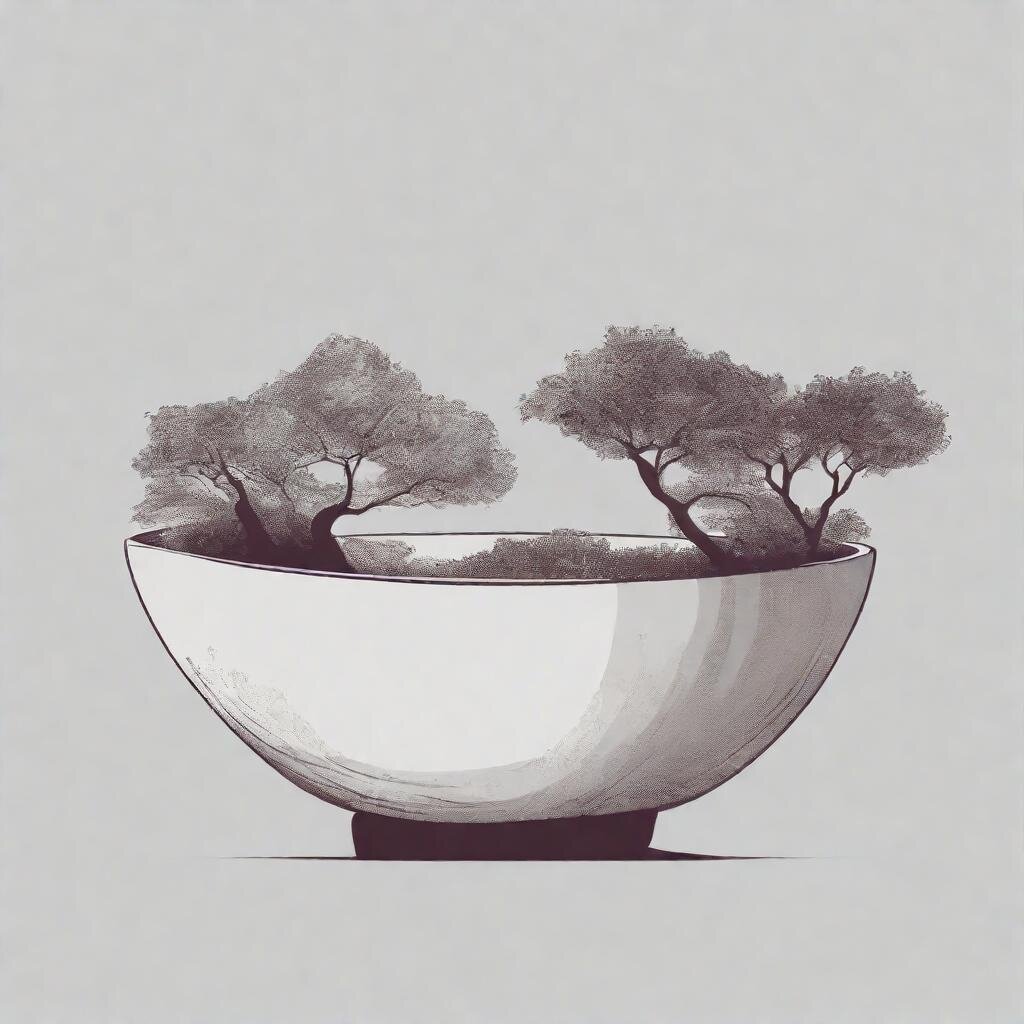}
        \end{minipage} &
        \begin{minipage}{0.12\textwidth}
            \includegraphics[width=\textwidth]{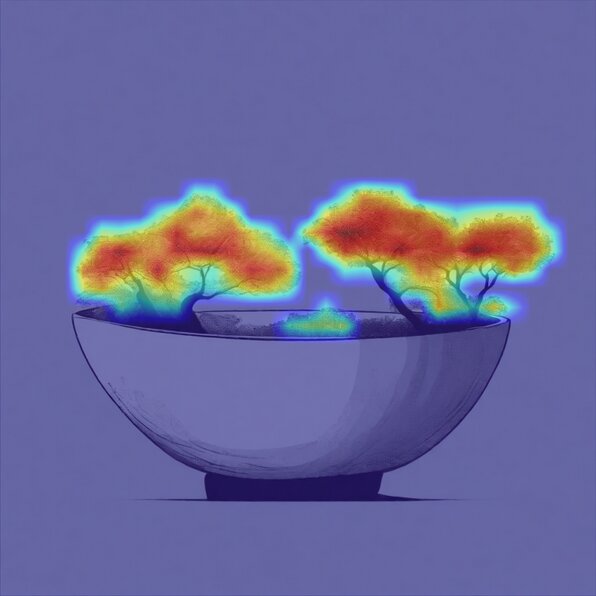}
        \end{minipage} &
        \begin{minipage}{0.12\textwidth}
            \includegraphics[width=\textwidth]{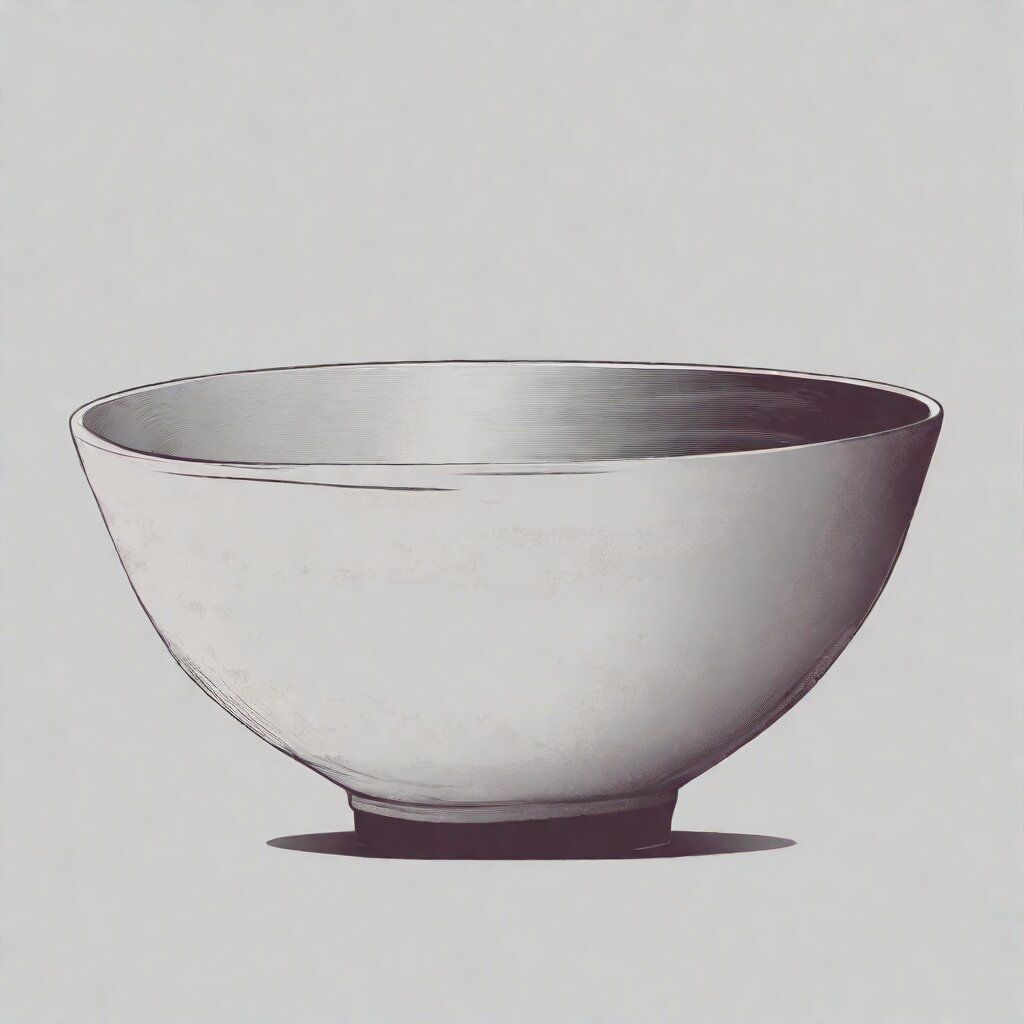}
        \end{minipage}
        \vspace{-2pt} 
        \\
        \vspace{-1pt} 
        \scriptsize ``A tree" & \scriptsize ``A bowl" &
        \multicolumn{5}{c}{\scriptsize ``...in minimal vector art style."} \\
        \begin{minipage}{0.12\textwidth}
            \includegraphics[width=\textwidth]{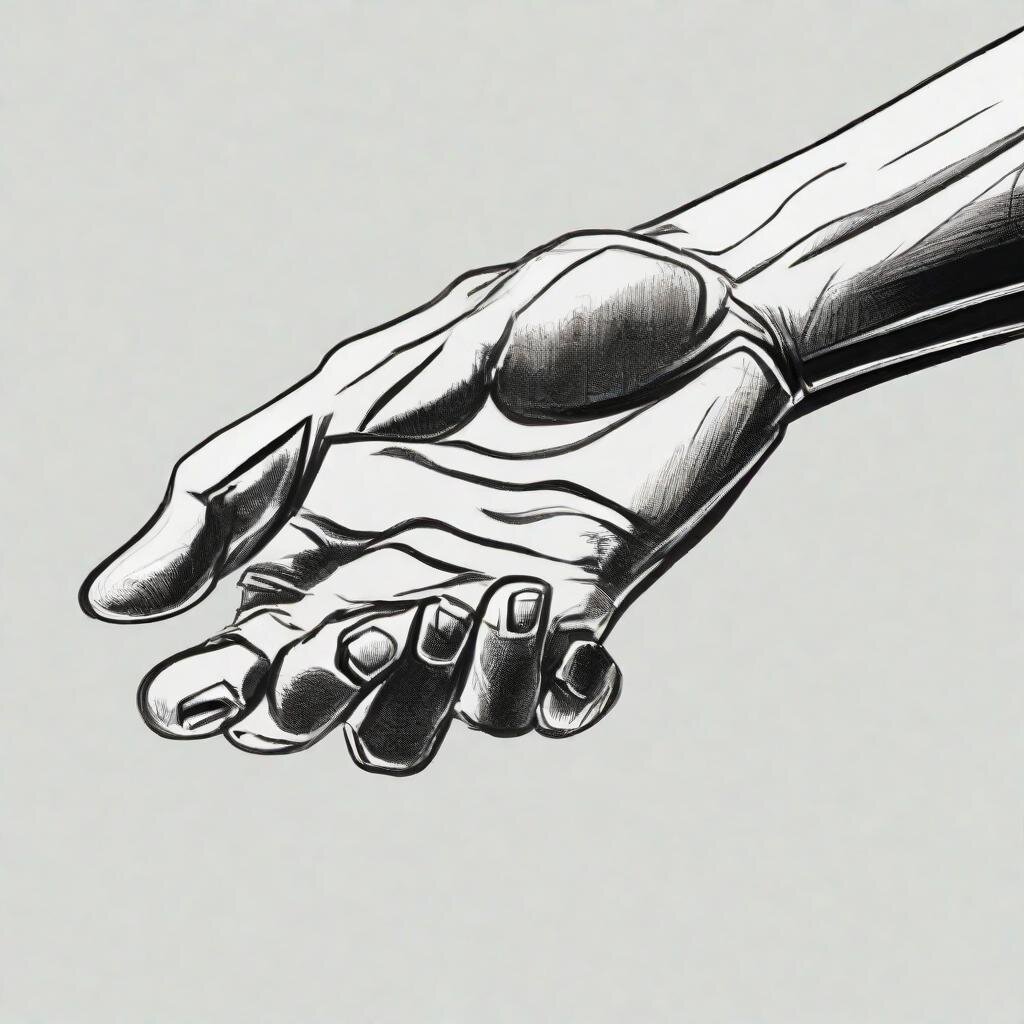}
        \end{minipage} &
        \begin{minipage}{0.12\textwidth}
            \includegraphics[width=\textwidth]{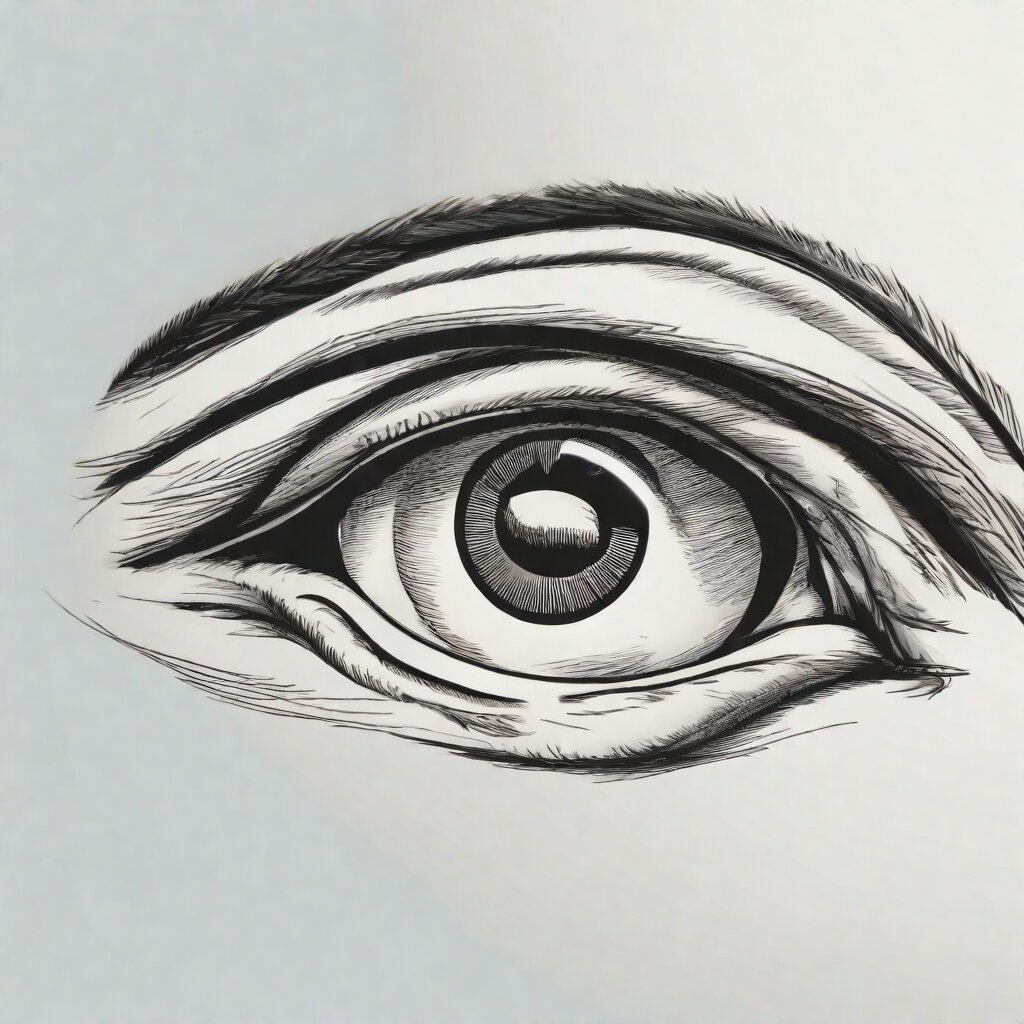}
        \end{minipage} &
        \begin{minipage}{0.12\textwidth}
            \includegraphics[width=\textwidth]{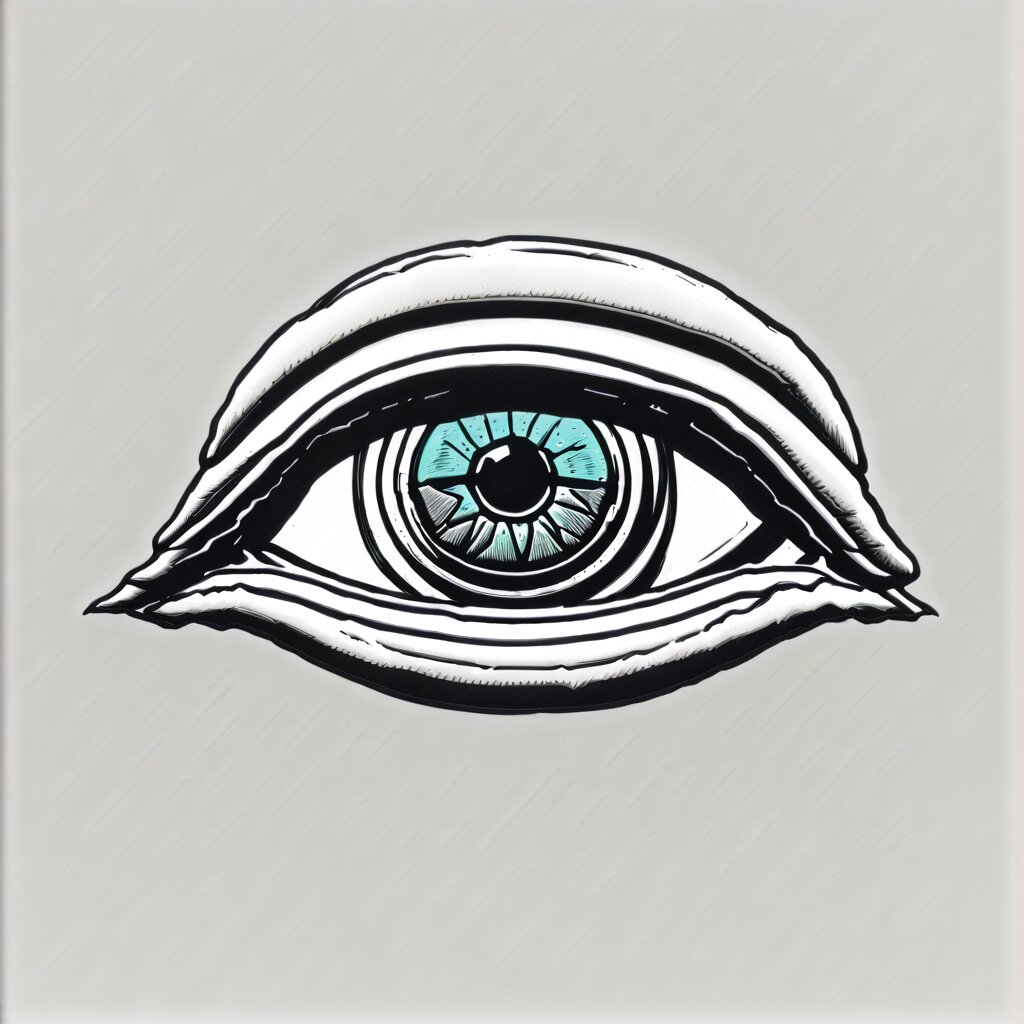}
        \end{minipage} &
        \begin{minipage}{0.12\textwidth}
            \includegraphics[width=\textwidth]{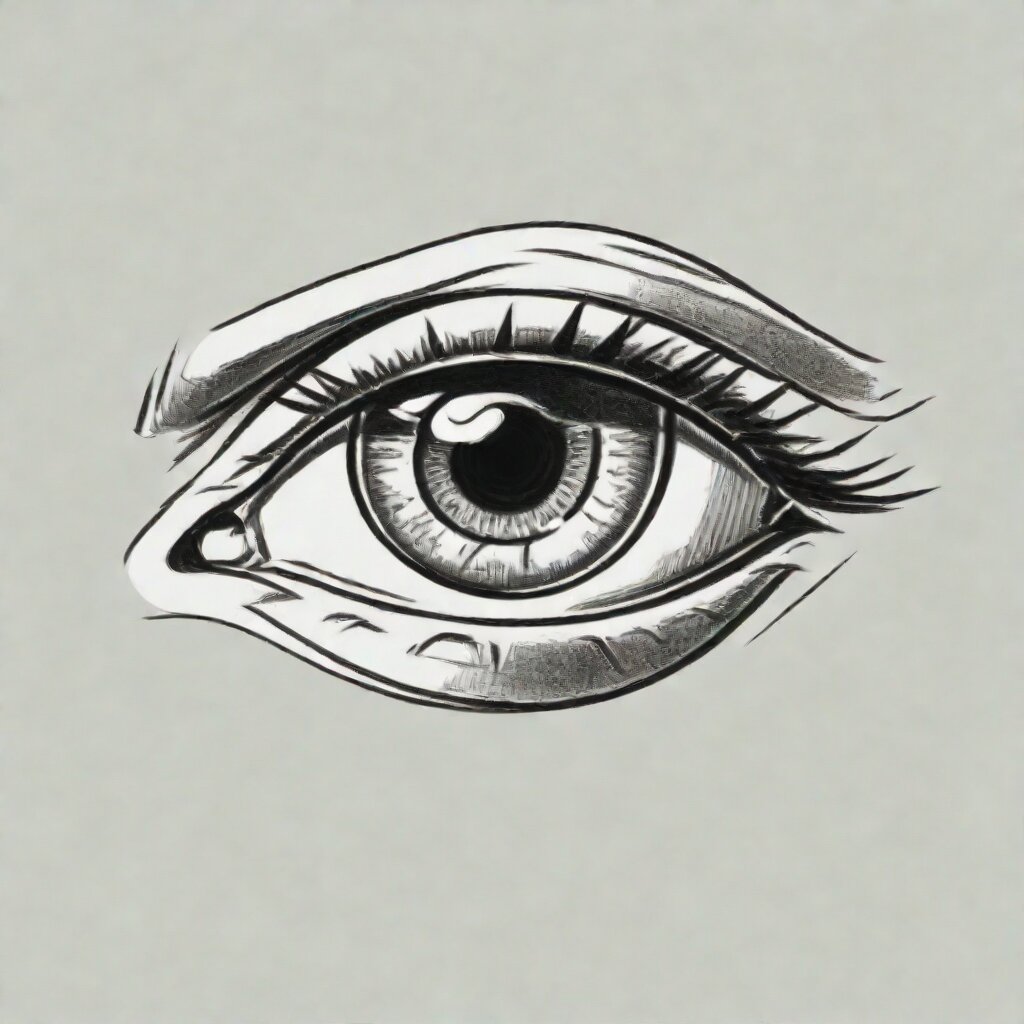}
        \end{minipage} &
        \begin{minipage}{0.12\textwidth}
            \includegraphics[width=\textwidth]{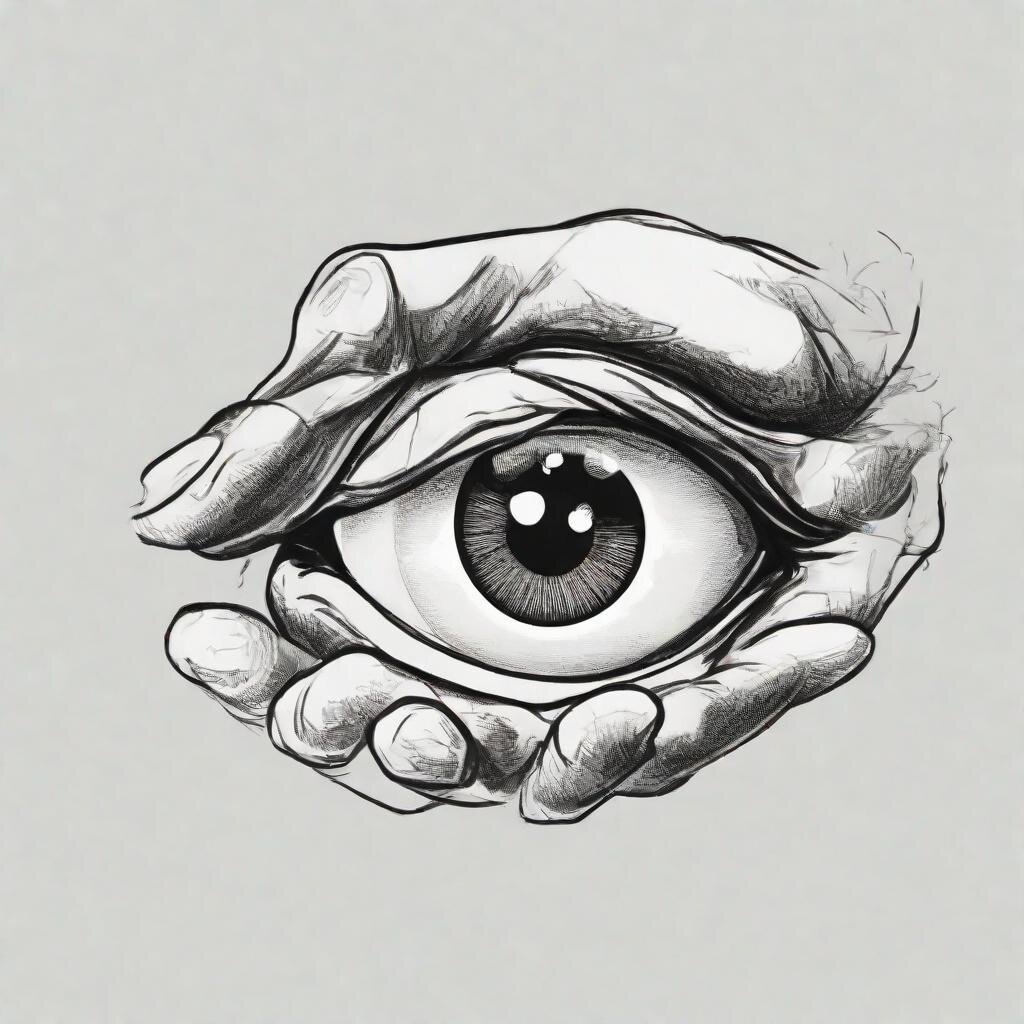}
        \end{minipage} &
        \begin{minipage}{0.12\textwidth}
            \includegraphics[width=\textwidth]{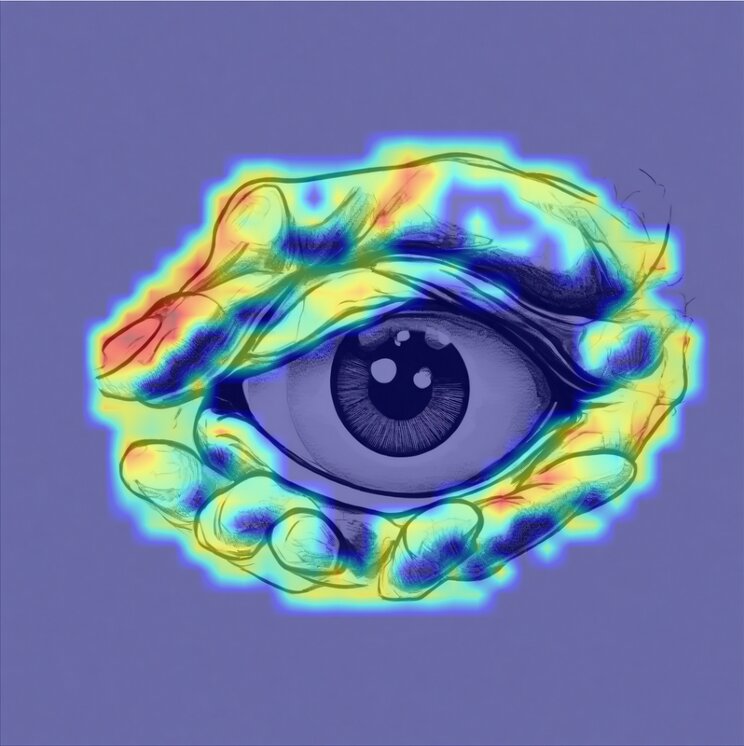}
        \end{minipage} &
        \begin{minipage}{0.12\textwidth}
            \includegraphics[width=\textwidth]{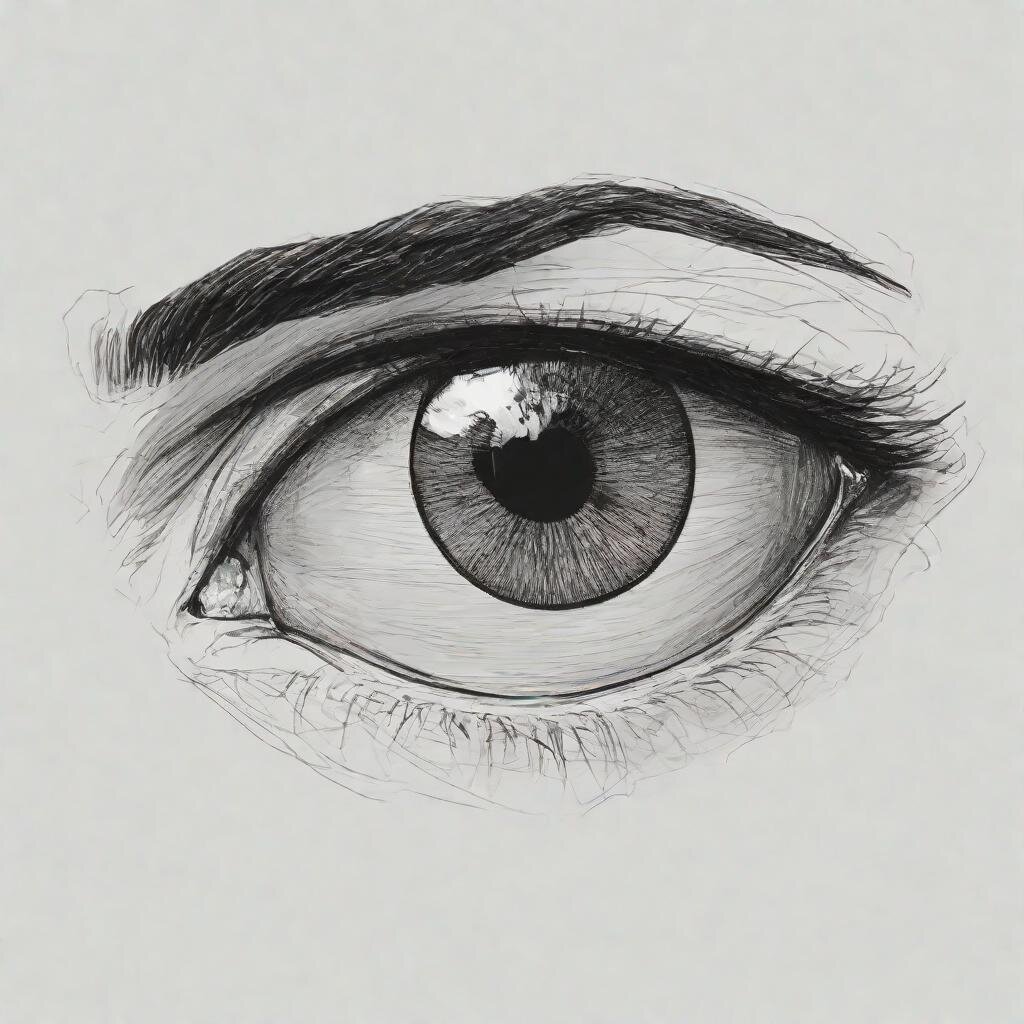}
        \end{minipage}  
        \vspace{-2pt} 
        \\
        \vspace{-1pt} 
        \scriptsize ``A hand" & \scriptsize ``An eye" & \multicolumn{5}{c}{\scriptsize ``... in line drawing style."} \\
        
        \begin{minipage}{0.12\textwidth}
        \includegraphics[width=\textwidth]{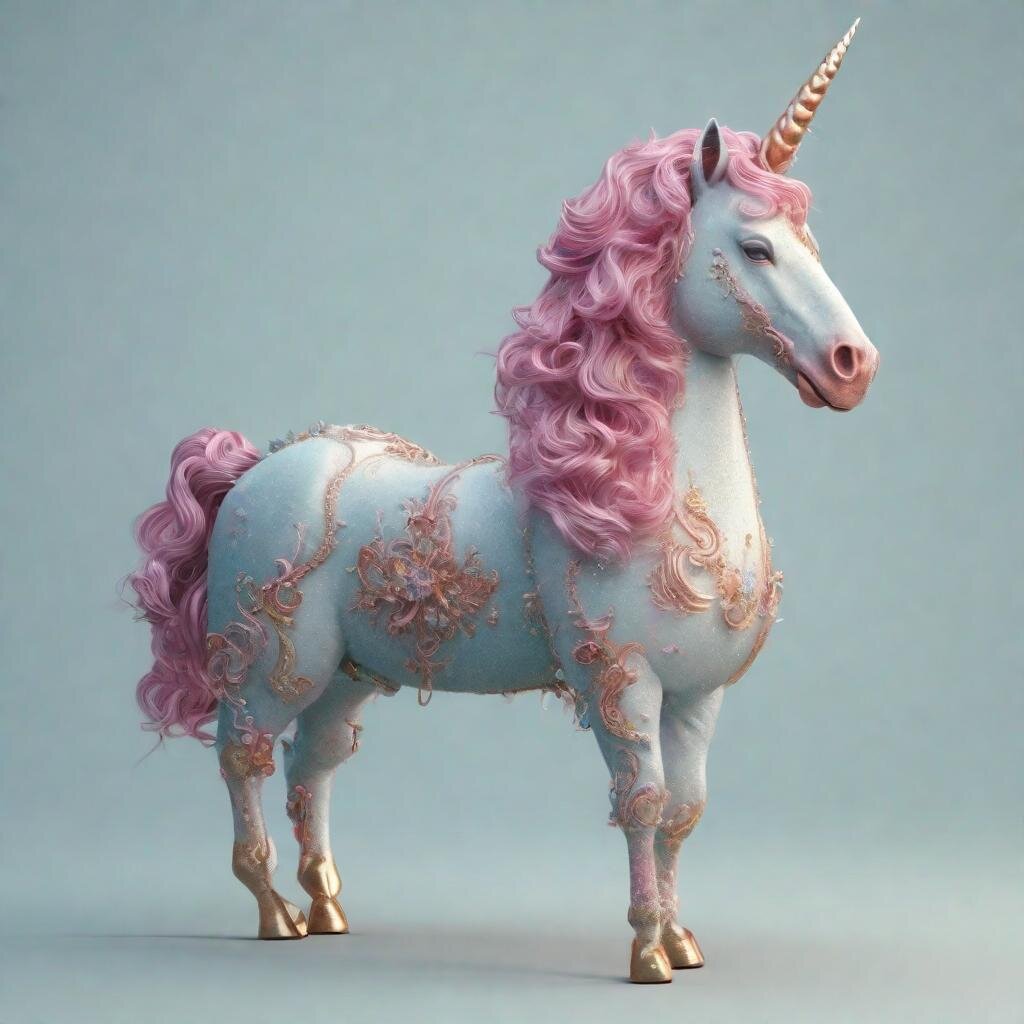}
        \end{minipage} &
        \begin{minipage}{0.12\textwidth}
            \includegraphics[width=\textwidth]{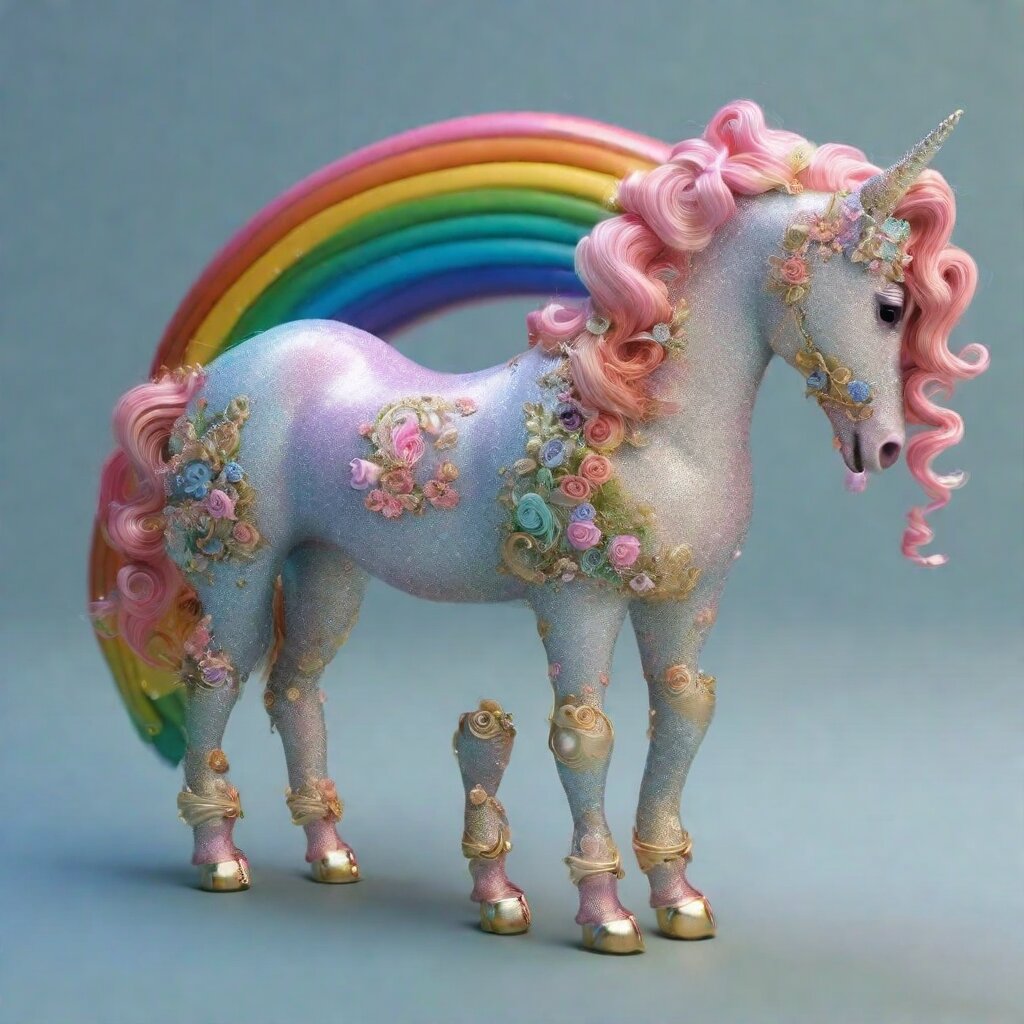}
        \end{minipage} &
        \begin{minipage}{0.12\textwidth}
            \includegraphics[width=\textwidth]{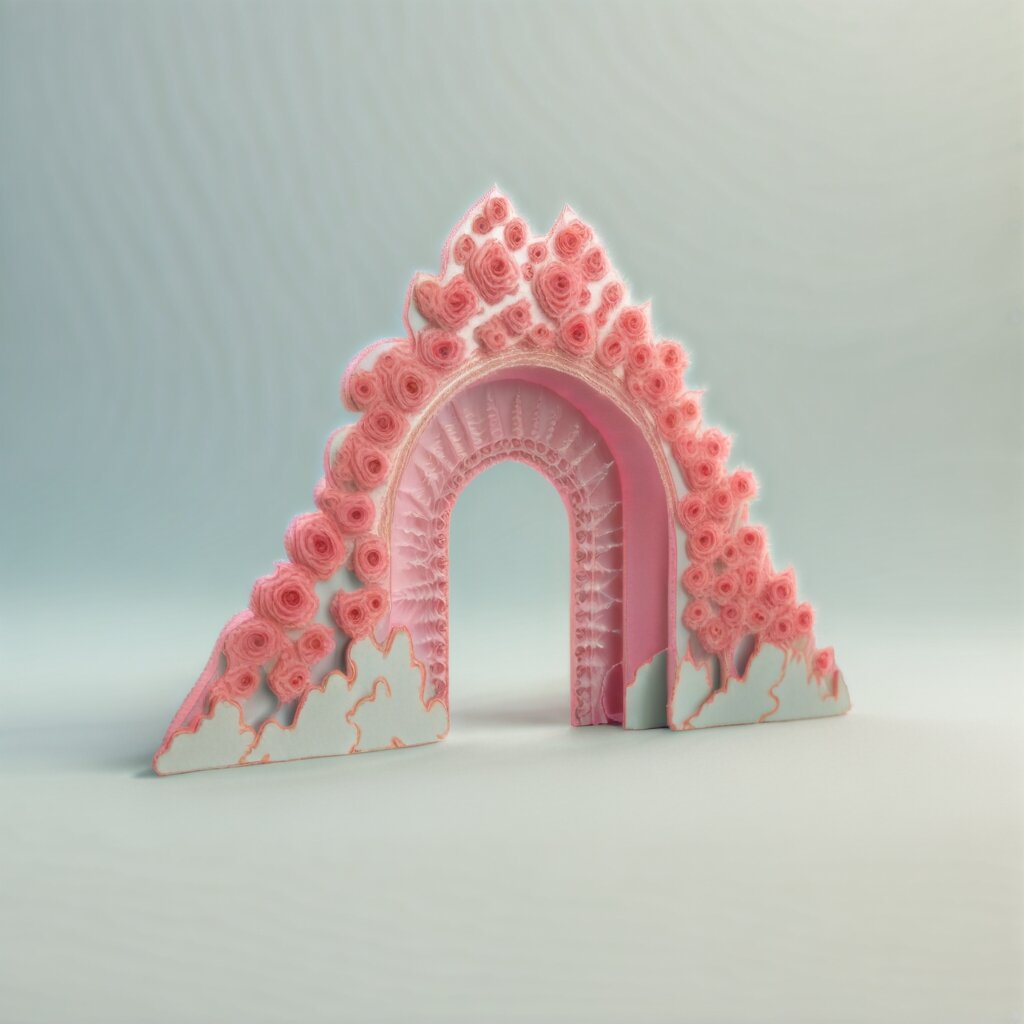}
        \end{minipage} &
        \begin{minipage}{0.12\textwidth}
            \includegraphics[width=\textwidth]{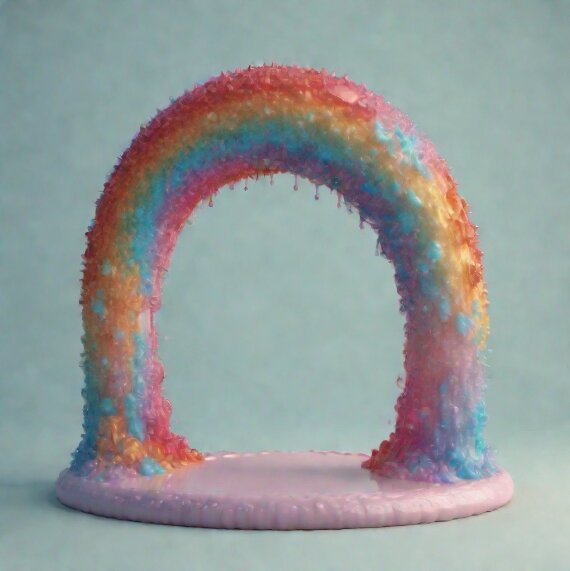}
        \end{minipage} &
        \begin{minipage}{0.12\textwidth}
            \includegraphics[width=\textwidth]{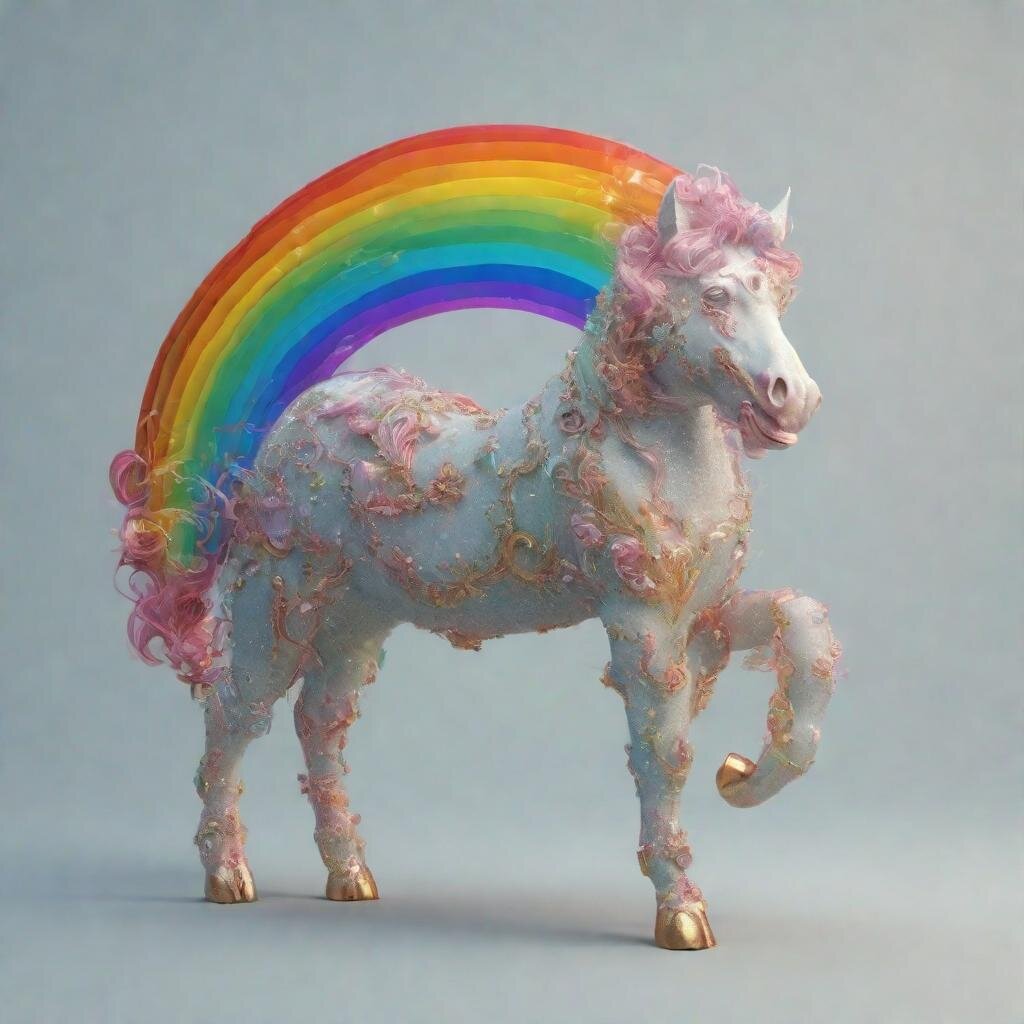}
        \end{minipage} &
        \begin{minipage}{0.12\textwidth}
            \includegraphics[width=\textwidth]{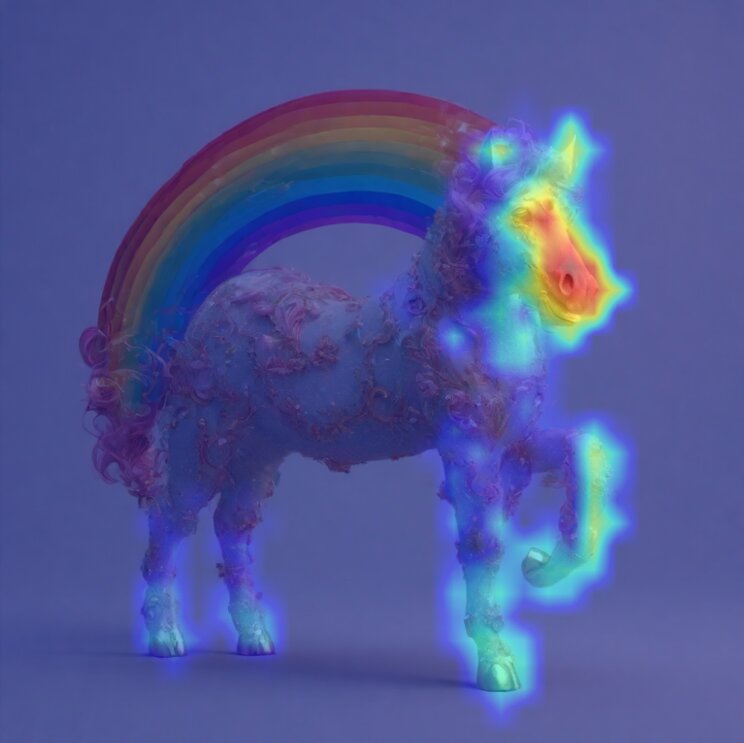}
        \end{minipage} &
        \begin{minipage}{0.12\textwidth}
            \includegraphics[width=\textwidth]{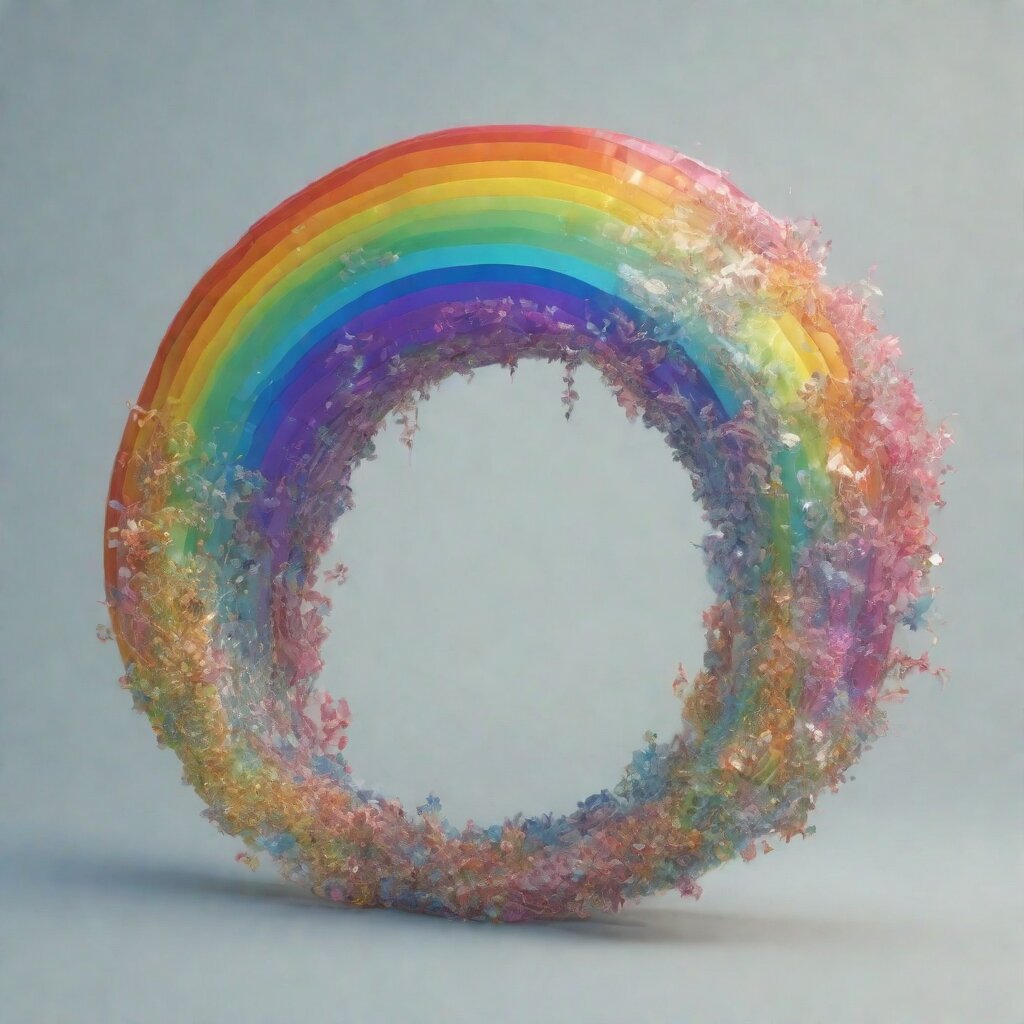}
        \end{minipage}
        \vspace{-2pt} 
        \\
        \vspace{-1pt} 
        \scriptsize ``A unicorn" & \scriptsize ``A rainbow" & \multicolumn{5}{c}{\scriptsize ``...in enchanted 3D rendering style."} \\
        
        \begin{minipage}{0.12\textwidth}
        \includegraphics[width=\textwidth]{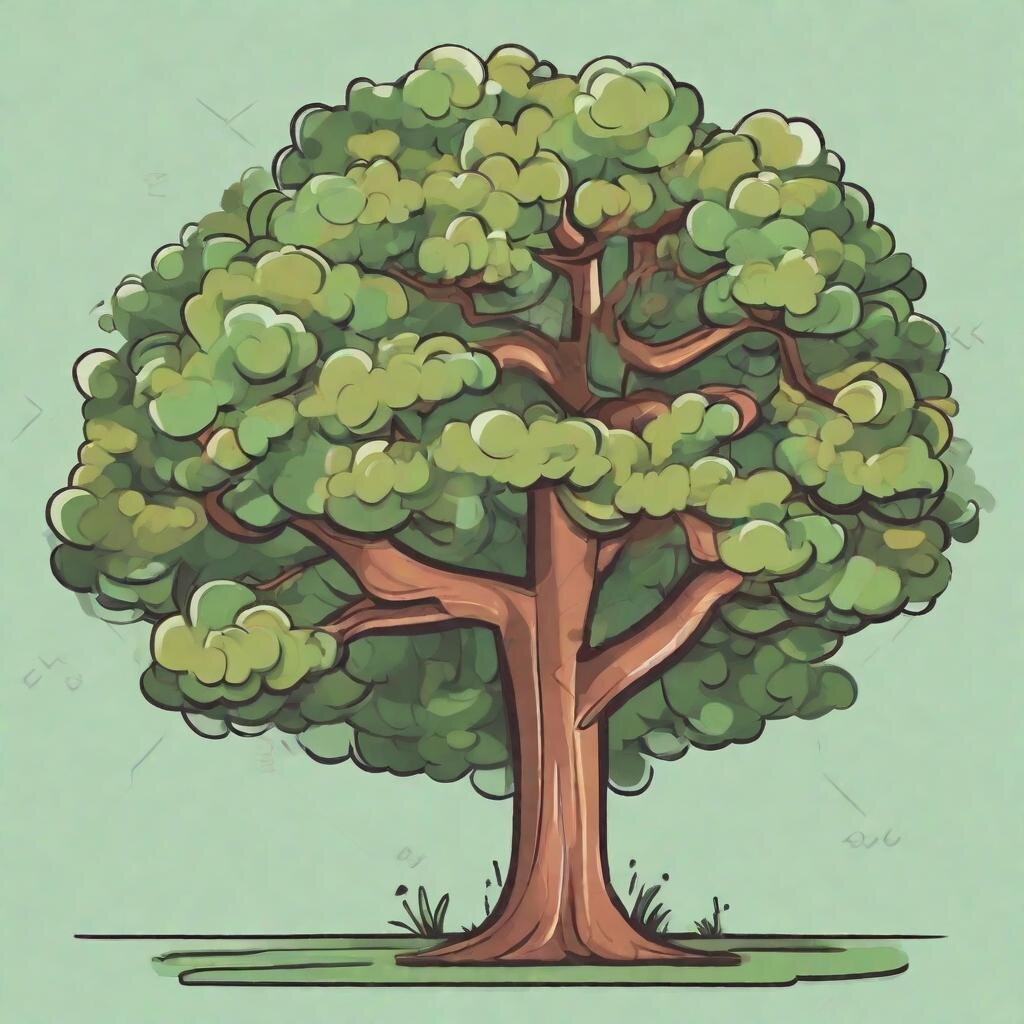}
        \end{minipage} &
        \begin{minipage}{0.12\textwidth}
            \includegraphics[width=\textwidth]{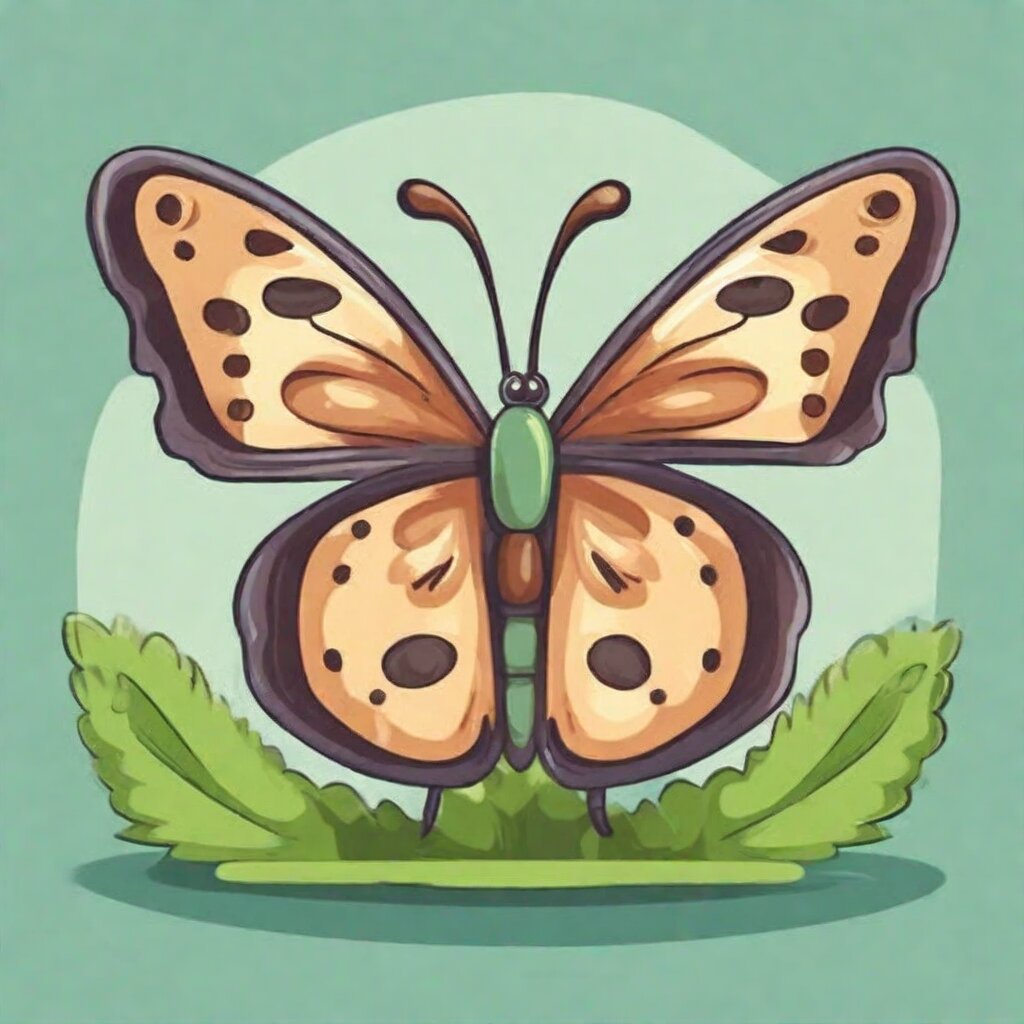}
        \end{minipage} &
        \begin{minipage}{0.12\textwidth}
            \includegraphics[width=\textwidth]{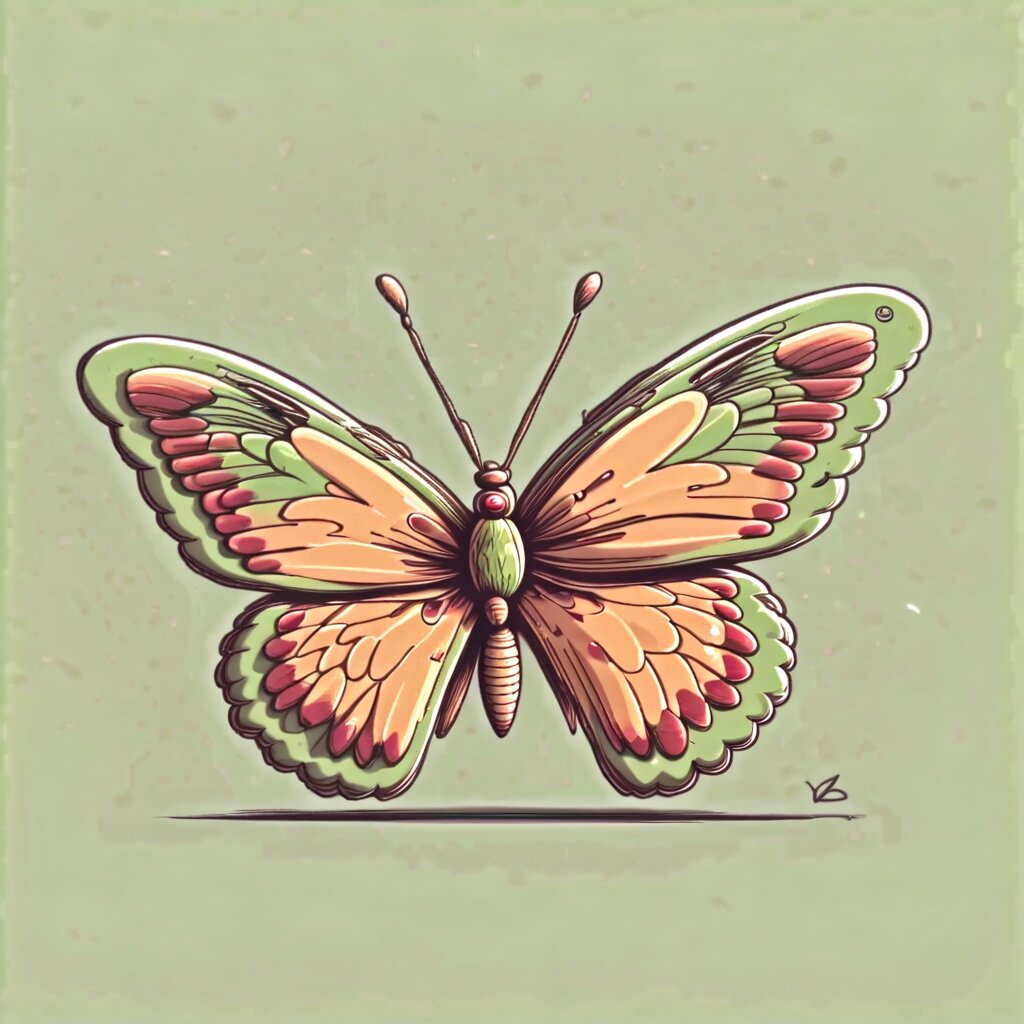}
        \end{minipage} &
        \begin{minipage}{0.12\textwidth}
            \includegraphics[width=\textwidth]{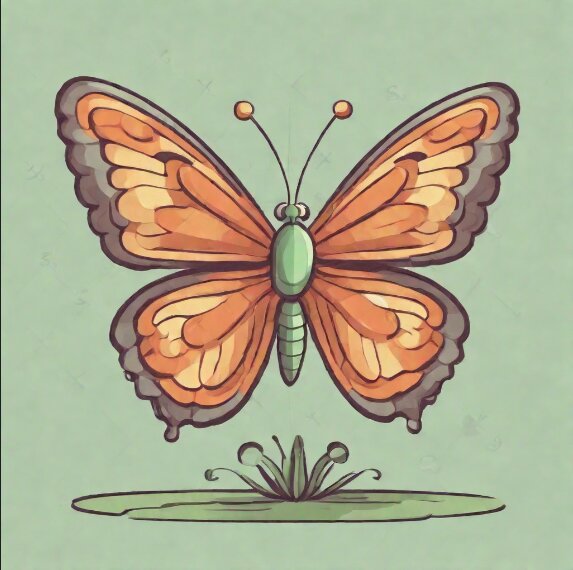}
        \end{minipage} &
        \begin{minipage}{0.12\textwidth}
            \includegraphics[width=\textwidth]{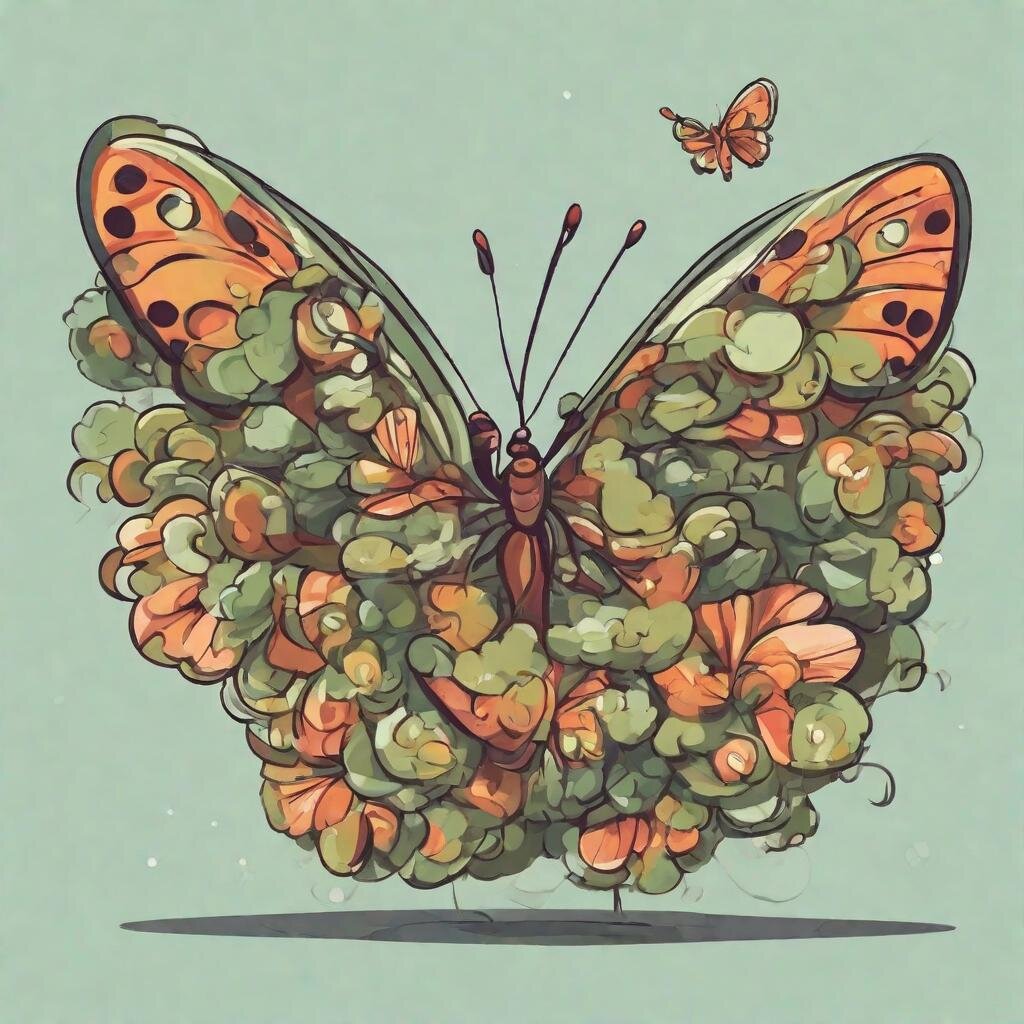}
        \end{minipage} &
        \begin{minipage}{0.12\textwidth}
            \includegraphics[width=\textwidth]{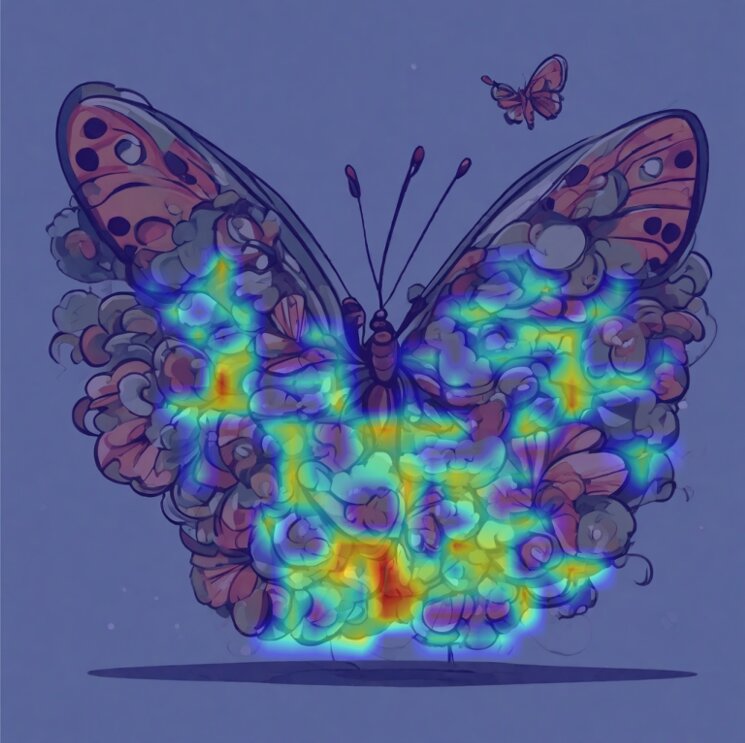}
        \end{minipage} &
        \begin{minipage}{0.12\textwidth}
            \includegraphics[width=\textwidth]{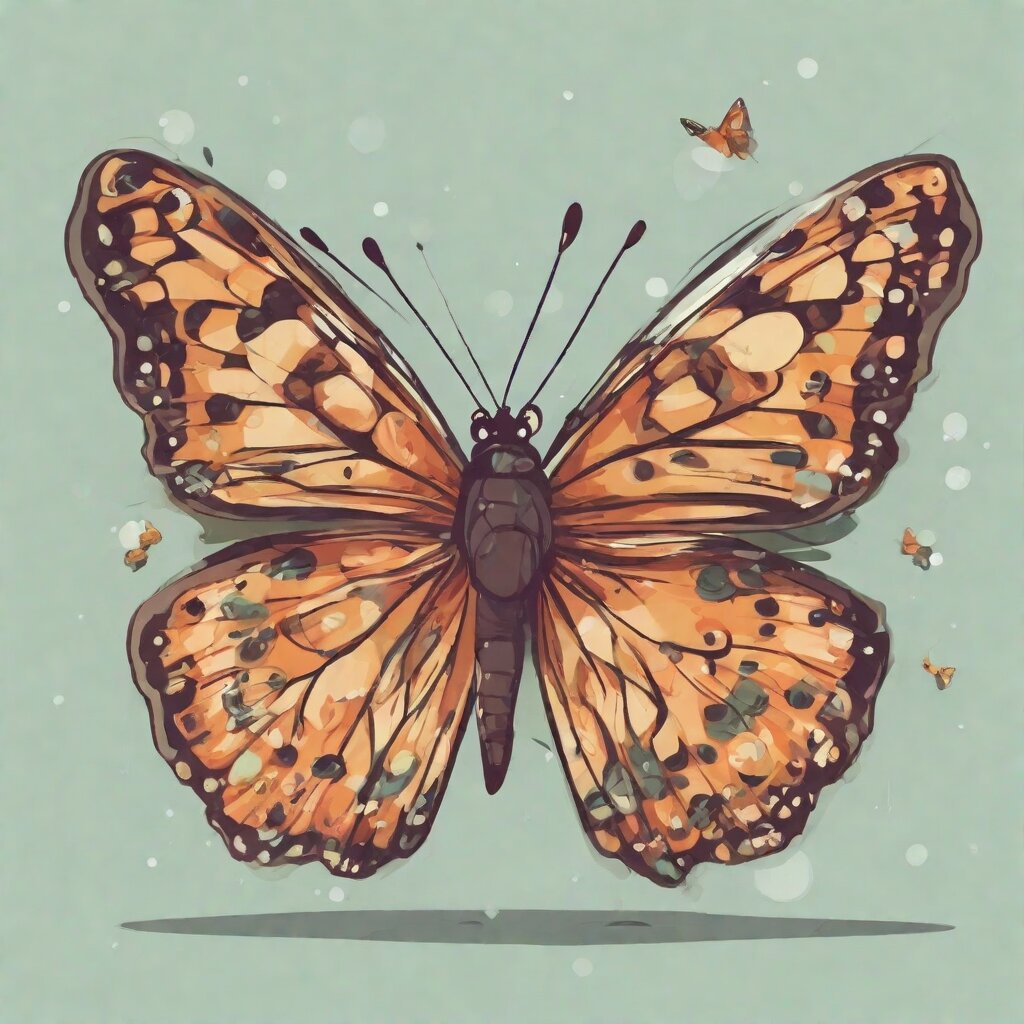}
        \end{minipage}

        \vspace{-2pt} 
        \\
        \vspace{-1pt} 
        \scriptsize ``A tree" & \scriptsize ``A butterfly" & \multicolumn{5}{c}{\scriptsize ``...in flat cartoon illustration style."} \\
        
       \begin{minipage}{0.12\textwidth}
        \includegraphics[width=\textwidth]{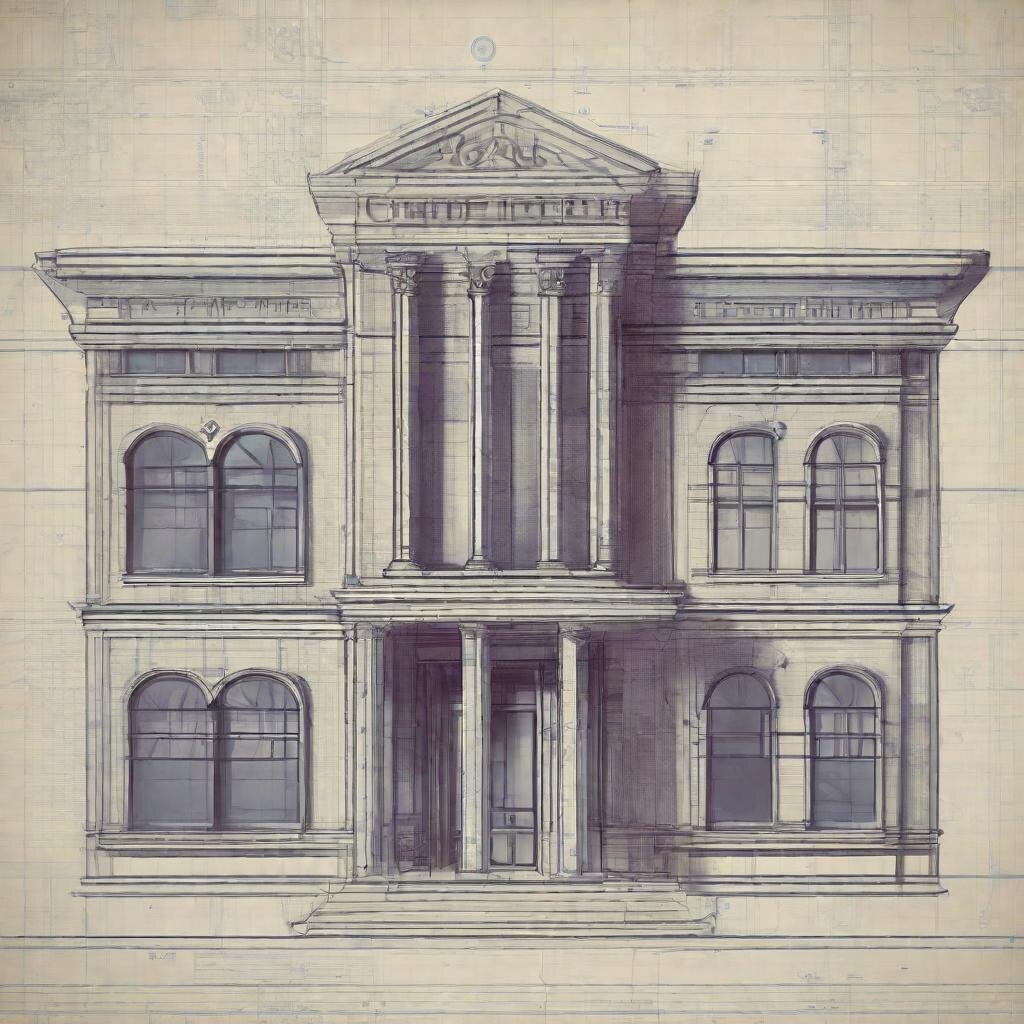}
        \end{minipage} &
        \begin{minipage}{0.12\textwidth}
            \includegraphics[width=\textwidth]{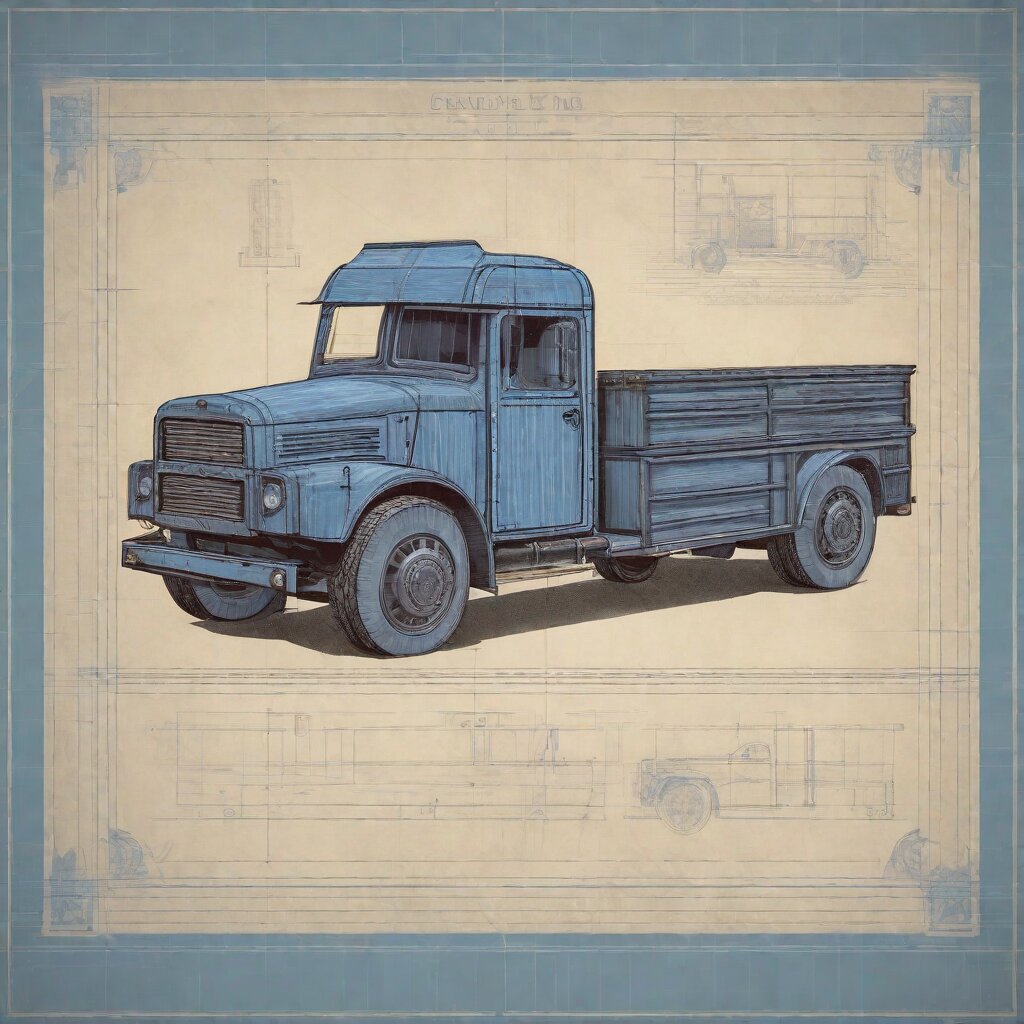}
        \end{minipage} &
        \begin{minipage}{0.12\textwidth}
            \includegraphics[width=\textwidth]{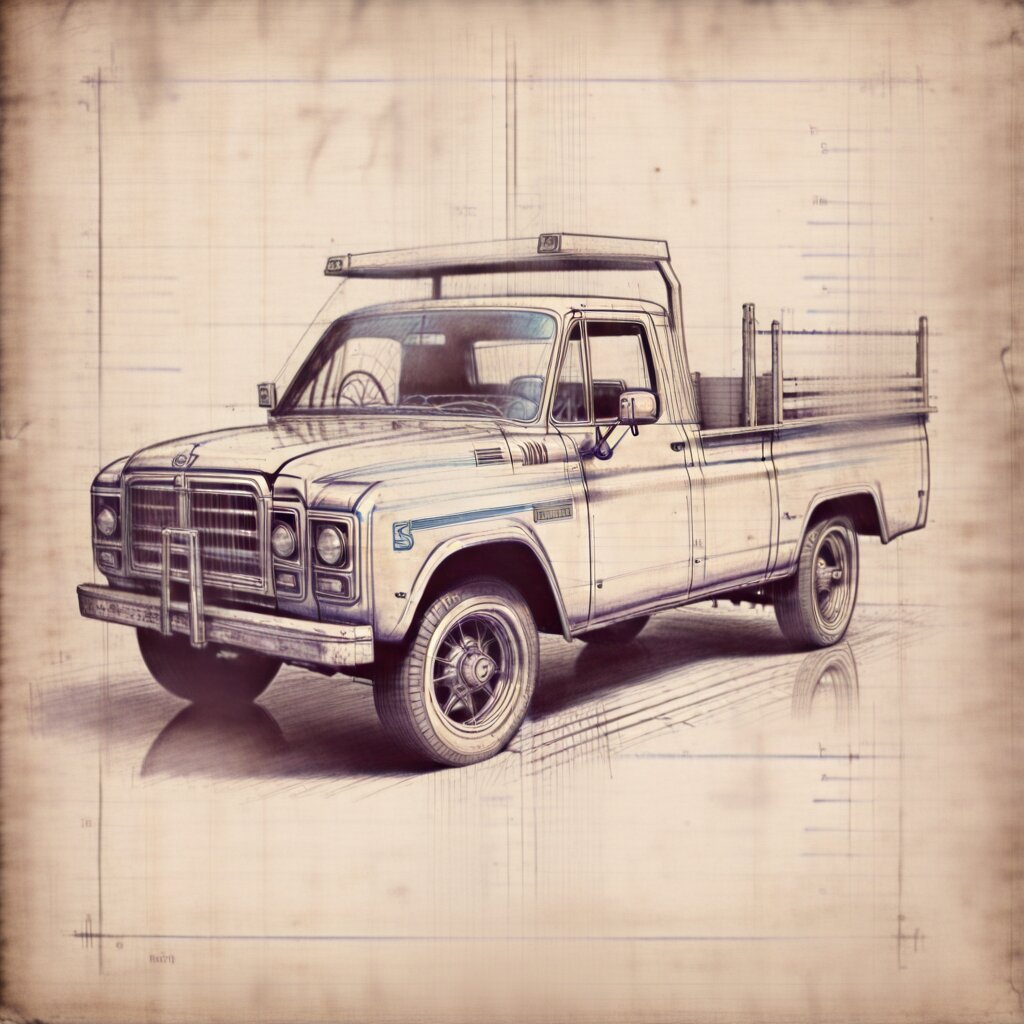}
        \end{minipage} &
        \begin{minipage}{0.12\textwidth}
            \includegraphics[width=\textwidth]{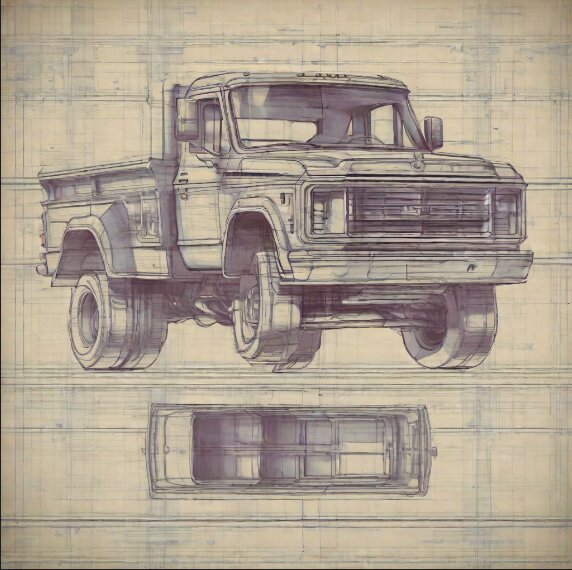}
        \end{minipage} &
        \begin{minipage}{0.12\textwidth}
            \includegraphics[width=\textwidth]{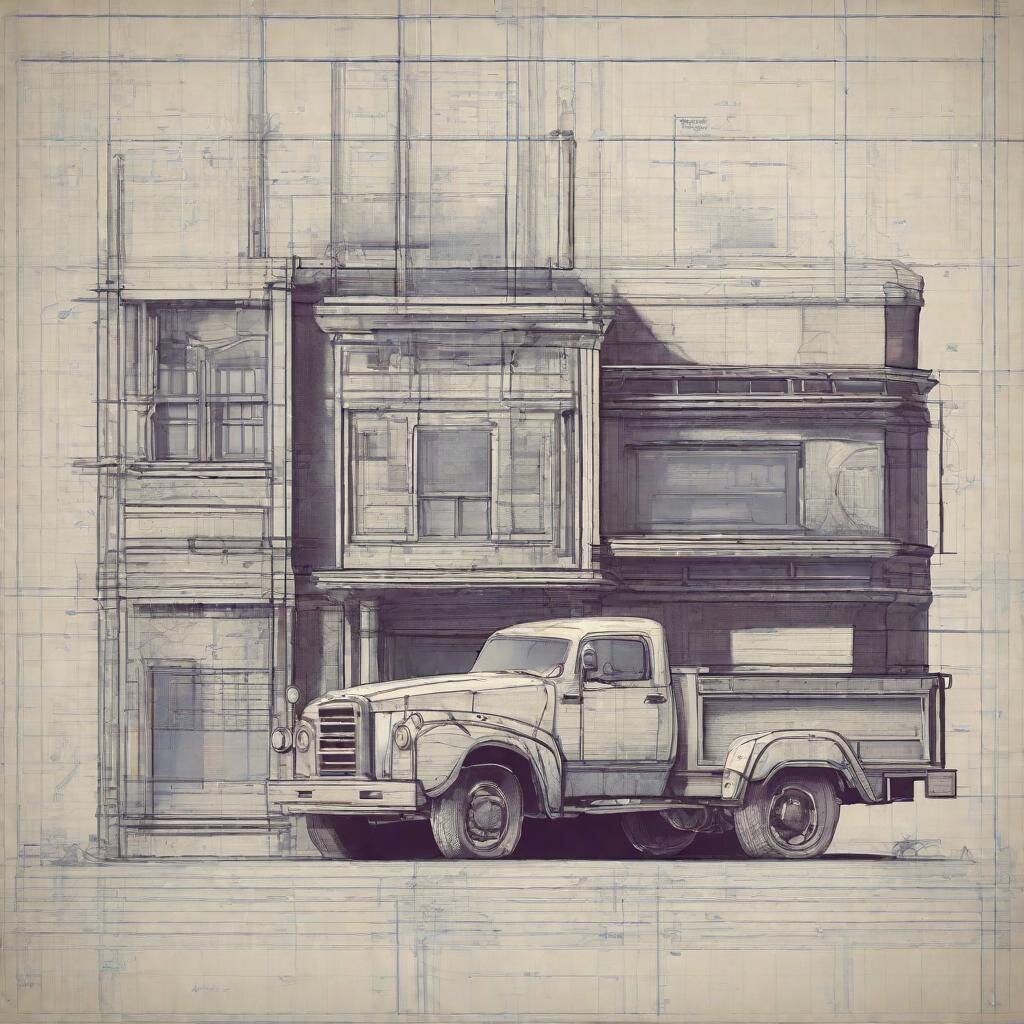}
        \end{minipage} &
        \begin{minipage}{0.12\textwidth}
            \includegraphics[width=\textwidth]{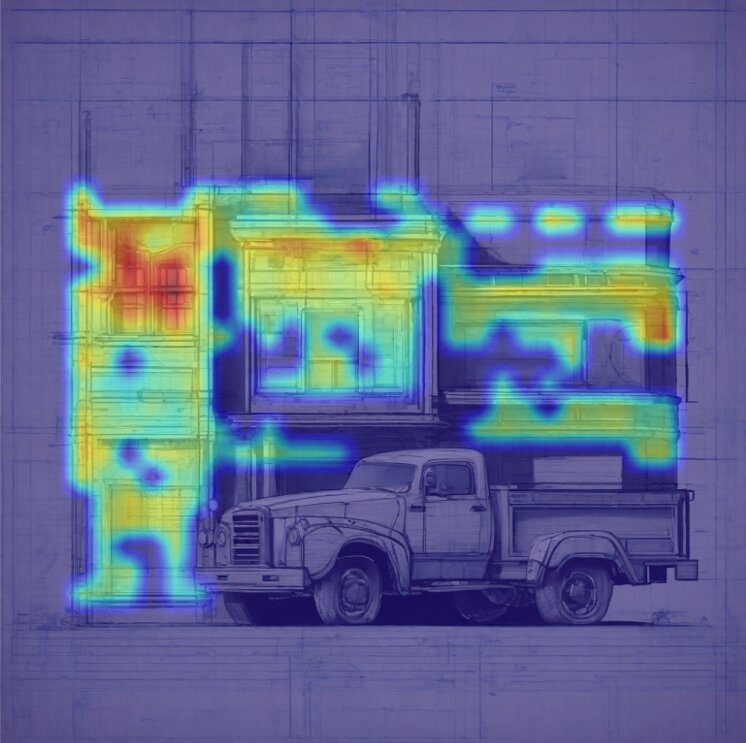}
        \end{minipage} &
        \begin{minipage}{0.12\textwidth}
            \includegraphics[width=\textwidth]{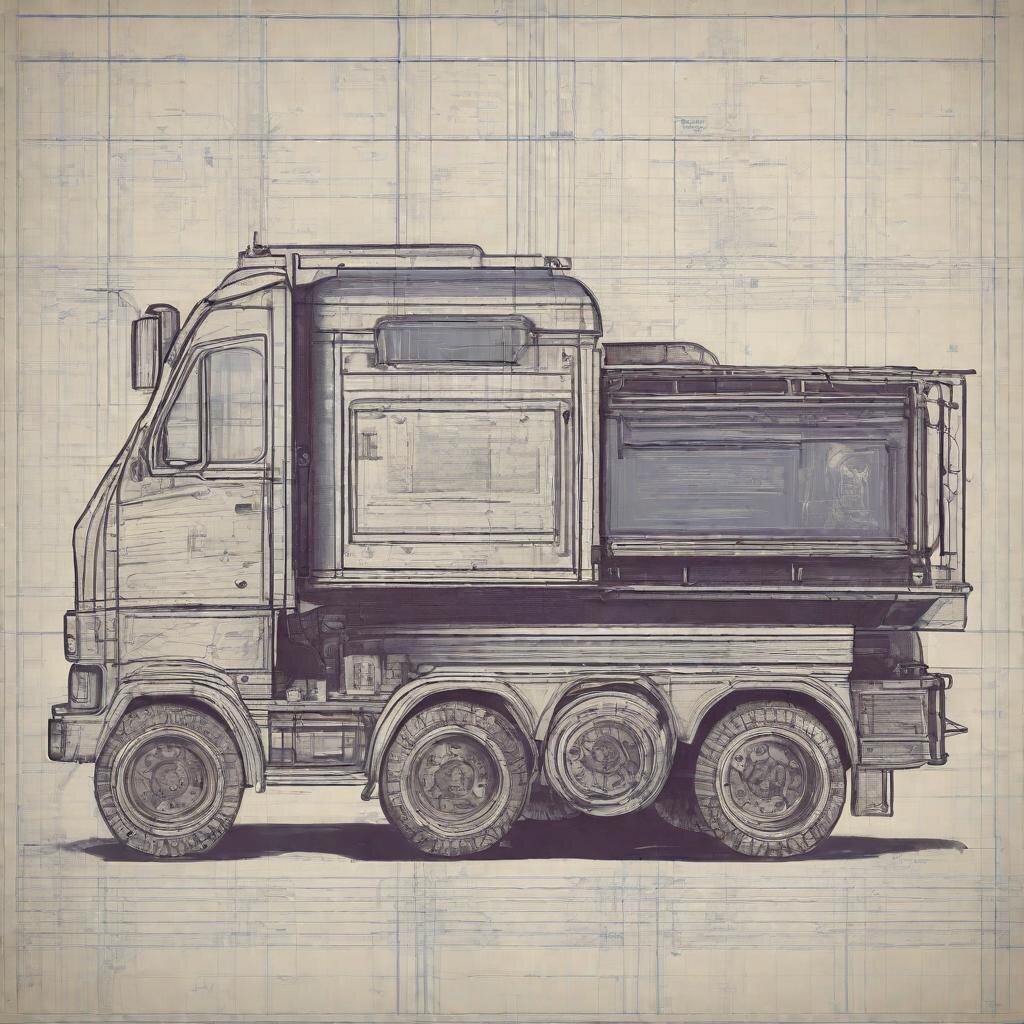}
        \end{minipage}
        \vspace{-2pt} 
        \\
        \vspace{-1pt} 
        \scriptsize ``A building" & \scriptsize ``A truck" & \multicolumn{5}{c}{\scriptsize ``...in blueprint style."}         
    \end{tabular}
    }
    \vspace{-5pt}
    \caption{\textbf{Additional Qualitative results}. We compare Only-Style against StyleAligned~\cite{hertz2024style}, IP-Adapter~\cite{Ye2023IPAdapter}, CSGO~\cite{Xing2024CSGO} and DB-LoRA~\cite{ruiz2023dreamboothfinetuningtexttoimage}. In the next-to-last column we also highlight the content leakage observed in StyleAligned, which is localized and effectively mitigated by our method.
    \vspace{-10pt}
    }
    \label{fig:Qualitative_Suppl}
\end{figure*}

\begin{table}[t]
\footnotesize
\centering
\begin{tabular}{|l|c|c|c|c|}
\hline
Metric         & IP-Adapter & CSGO & DB-LoRA & \emph{Only-Style} \\ \hline
CL (\textdownarrow)          & 0.232        & 0.223      & 0.229      & \textbf{0.215}      \\ \hline
Q1 Success (\textuparrow)               & 0.467        & 0.614      & 0.607      & \textbf{0.683}      \\ \hline
Q2 Success (\textuparrow)               & 0.542        & 0.908      & 0.737      & \textbf{0.830}      \\ \hline
Q3 Success (\textuparrow)               & 0.886        & 0.732      & 0.857      & \textbf{0.957}      \\ \hline
\end{tabular}
\caption{Quantitative comparison between IP-Adapter~\cite{Ye2023IPAdapter}, CSGO~\cite{Xing2024CSGO}, DB-LoRA~\cite{ruiz2023dreamboothfinetuningtexttoimage}, and \emph{Only-Style}, in the metrics discussed in the main manuscript that quantify content leakage.}
\label{tab:additional_comparisons}
\end{table}

\section{Additional Results}

\subsection{Additional Comparisons}

To further highlight the effectiveness of the proposed approach, we additionally compare with the following state-of-the-art methods for style consistent image generation, namely IP-Adapter~\cite{Ye2023IPAdapter}, CSGO~\cite{Xing2024CSGO} and Dreambooth~\cite{ruiz2023dreamboothfinetuningtexttoimage}, using the LoRA~\cite{hu2021loralowrankadaptationlarge} variant (\emph{DB-LoRA}).
The first two are adapter-based methods that introduce additional layers to condition the diffusion model on the CLIP image representation of the stylistic reference, similar to InstantStyle~\cite{Wang2024InstantStyle}. The latter is an optimization-based method, which first fine-tunes the model on the reference image of a specific style by learning a compact set of adaptations (LoRA) that capture the visual characteristics of that style and then when generating new images, these learned LoRA weights are applied to transfer the original style to different subjects. All considered methods use SDXL as their base model as well. We provide both quantitative comparisons, based on the metrics outlined in the main manuscript, in Figures~\ref{tab:additional_comparisons} and \ref{fig:Quantitative_add}, as well as qualitative results in Figure~\ref{fig:Qualitative_Suppl}, evaluated on our test prompt set. Notably, these methods also exhibit significant content leakage across all quantitative metrics assessing leakage, in contrast to \emph{Only-Style}, emphasizing how frequently the problem occurs.

\begin{table}[t]
\footnotesize
\centering
\begin{tabular}{|l|c|}
\hline
\textbf{Method} & \textbf{Set Consistency} (DINO $\uparrow$) \\ \hline
StyleAligned                & $0.372 \pm 0.22$ \\ \hline
Consistory (fixed object)   & $0.326 \pm 0.19$ \\ \hline
Standard T2I (fixed object) & $0.218 \pm 0.2$  \\ \hline
Standard T2I (fixed style)  & $0.225 \pm 0.21$ \\ \hline
\emph{Only-Style}           & $0.345 \pm 0.2$  \\ \hline
\end{tabular}
\caption{\textbf{Detailed Quantitative Results on Stylistic Set Consistency}. We evaluate the generated image sets in terms of set consistency (DINO embedding similarity). $\pm X$ denotes the standard deviation of the score across the evaluation set.}
\label{tab:quantitative_comparison}
\end{table}

\begin{figure}[t]
    \begin{tabular}{c@{\hspace{.1cm}}c@{\hspace{.1cm}}c@{\hspace{.1cm}}c}
        
        \begin{minipage}{0.11\textwidth}
            \centering
            \scriptsize Standard T2I 
        \end{minipage} &
        \begin{minipage}{0.11\textwidth}
            \includegraphics[width=\textwidth]{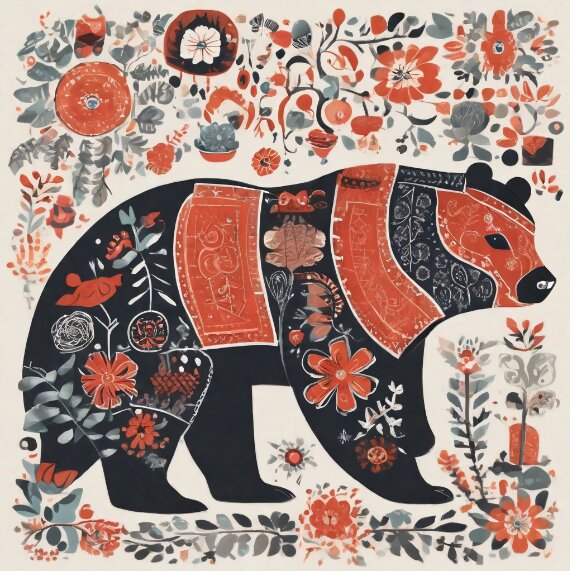}
        \end{minipage} &
        \begin{minipage}{0.11\textwidth}
            \includegraphics[width=\textwidth]{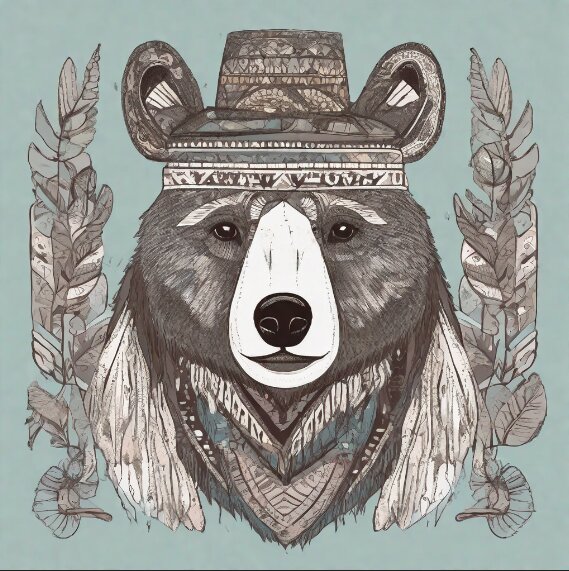}
        \end{minipage} &
        \begin{minipage}{0.11\textwidth}
            \includegraphics[width=\textwidth]{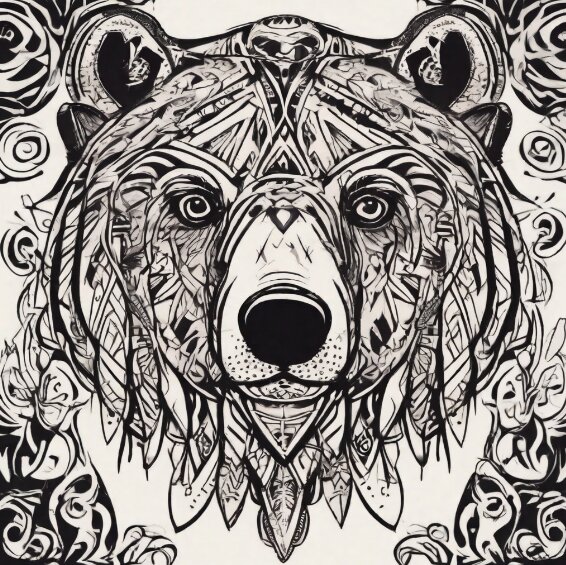}
        \end{minipage} 
    \end{tabular}

    \vspace{.1cm}

    \begin{tabular}{c@{\hspace{.1cm}}c@{\hspace{.1cm}}c@{\hspace{.1cm}}c}  
        \begin{minipage}{0.11\textwidth}
            \centering
            \scriptsize Consistory
        \end{minipage} &
        \begin{minipage}{0.11\textwidth}
            \includegraphics[width=\textwidth]{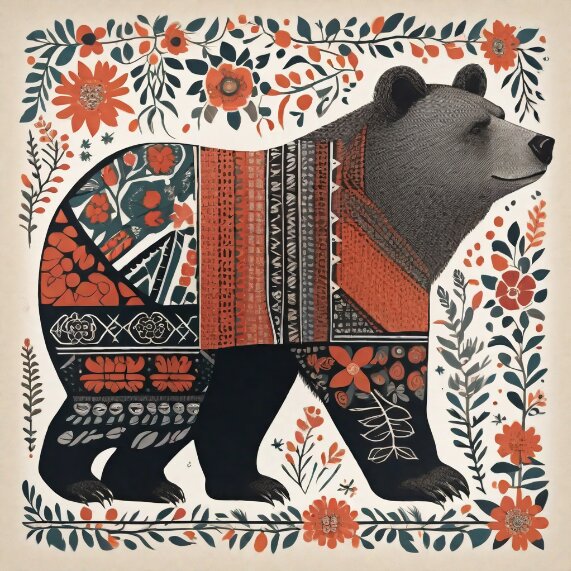}
        \end{minipage} &
        \begin{minipage}{0.11\textwidth}
            \includegraphics[width=\textwidth]{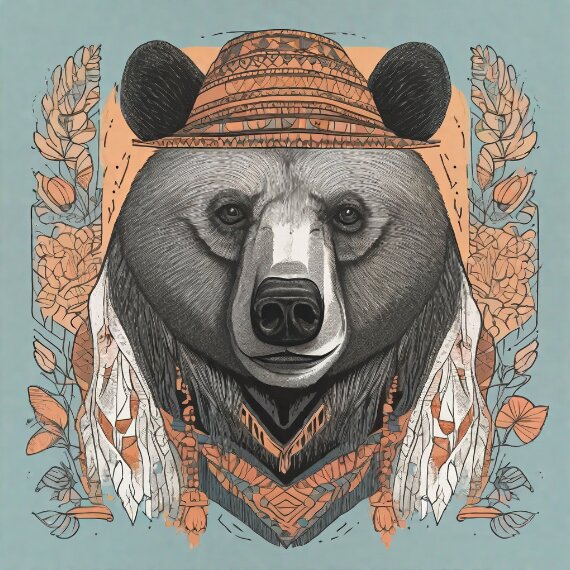}
        \end{minipage} &
        \begin{minipage}{0.11\textwidth}
            \includegraphics[width=\textwidth]{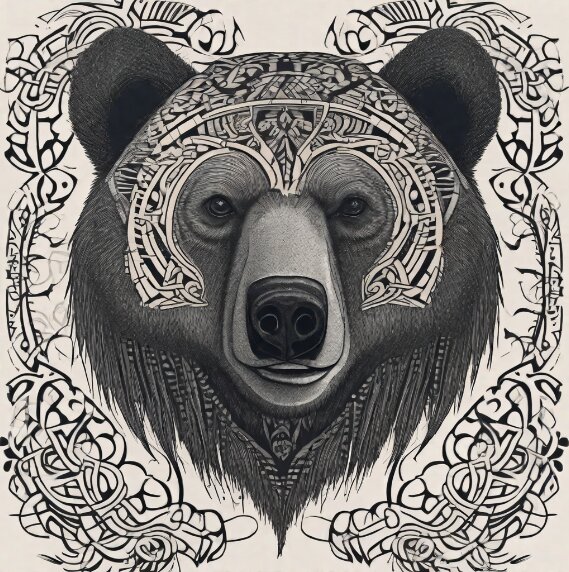}
        \end{minipage} 
    \end{tabular}
    
    \caption{\textbf{Consistent subject in different styles.} We employ Standard T2I~\cite{podell2023sdxlimprovinglatentdiffusion} to generate images of the same subject in different styles (first row). Since the identity of the subject is not preserved within different generations, it does not accurately simulate the effect of content leakage. To achieve this we employ a subject identity preservation method, ConsiStory~\cite{10.1145/3658157}, rendering the same object in different stylistic descriptors (second row). }
    \label{fig:fixed_obj}
    \vspace{-0.25cm}
\end{figure}

\subsection{Discussion on Stylistic Set Consistency} 

As we discussed in the main manuscript, we follow state-of-the-art style alignment methods~\cite{hertz2024style, jeong2024visual} and evaluate \emph{stylistic set consistency} within a style aligned image set, as the pairwise cosine similarity between DINO~\cite{caron2021emerging} embeddings of the generated target images $I_{tgt}$ with their stylistic reference images $I_{ref}$. 
However, although the aforementioned metric promotes the stylistic consistency between images, it also promotes semantic and structural consistency, which is undesired in stylistic alignment. 

We argue that this is because the metric is favored by semantic content leakage of the reference image subject in the target image. 
To quantitatively showcase this phenomenon, we employ two baselines that consist of generated sets of images in diverse styles but consistent depicted subjects. 
We reverse the logic of our evaluation prompt set (different objects in the same style) and generate the same object in different styles. 
For example: \textit{A bear {`in Scandinavian folk art style.', `in bohemian style.', `in tribal tattoo style.'}}

First, we employ the standard text-to-image model and generate images of an object in different styles. 
Note that the object generated in different styles is not the same for different generations (e.g., different kinds of bears are generated as shown in Fig.~\ref{fig:fixed_obj}). 
This does not exactly simulate the problem of content leakage, which refers to the leak of semantic attributes of the specific visual interpretation of the reference object across the target images. 
To address this problem, mimicking the effect of content leakage, we employ a state-of-the-art subject identity preservation method, ConsiStory~\cite{10.1145/3658157}. 
This method generates the same object (e.g., the same bear as illustrated in Fig.~\ref{fig:fixed_obj}) across different styles, effectively consisting of a content leakage baseline w.r.t. the aforementioned evaluation process.  

We observe that semantic consistency, expressed by the baselines we introduced, is favored as much as stylistic consistency within the \emph{stylistic set consistency} metric. 
Specifically, the fixed-subject-in-different-styles variant of standard text-to-image generation achieves a set consistency score comparable to the different-subjects-in-a-fixed-style variant. 
Furthermore, when the identity of the generated subjects is preserved across styles using the ConsiStory approach, the pairwise set consistency achieves a level close to state-of-the-art style alignment methods line \emph{Only-Style} and StyleAligned~\cite{hertz2024style}, even though the stylistic alignment is diminished on purpose. This suggests that reducing unwanted content leakage while ensuring stylistic alignment can be penalized by this metric, which fails to fully reflect the effectiveness of our approach.

\subsection{Additional Ablation Studies}

\begin{figure*}[t]
    \centering
    \begin{minipage}{\textwidth}
        \centering        
        \begin{minipage}{0.17\textwidth}
            \centering
            \includegraphics[width=\linewidth]{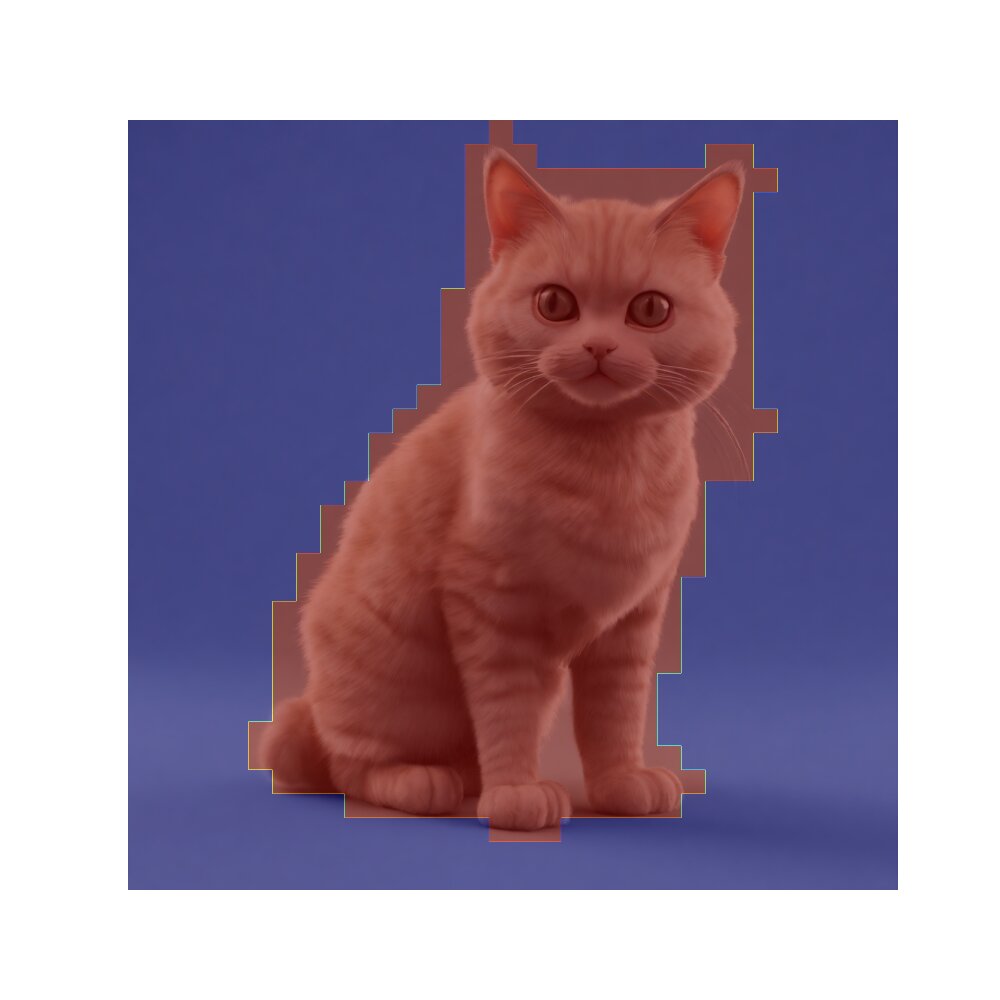}
        \end{minipage}%
        \begin{minipage}{0.17\textwidth}
            \centering
            \includegraphics[width=\linewidth]{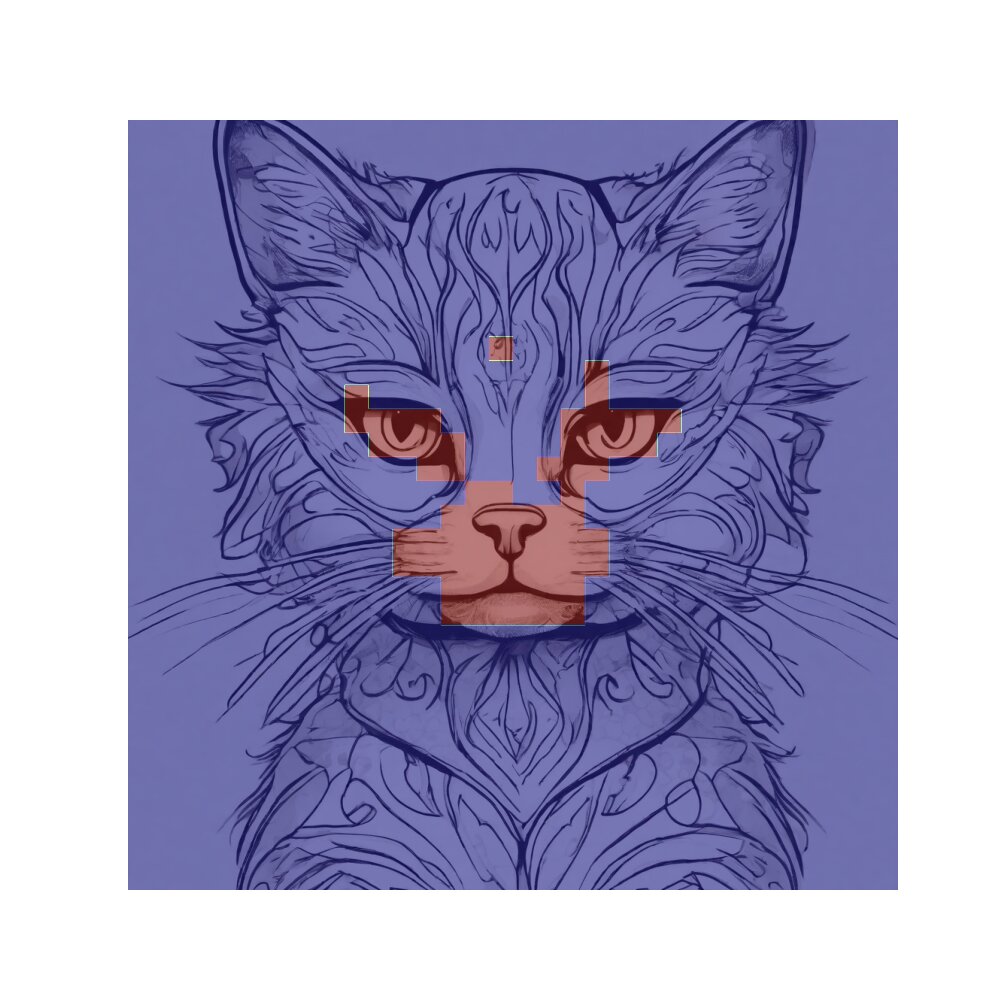}
        \end{minipage}%
        \begin{minipage}{0.17\textwidth}
            \centering
            \includegraphics[width=\linewidth]{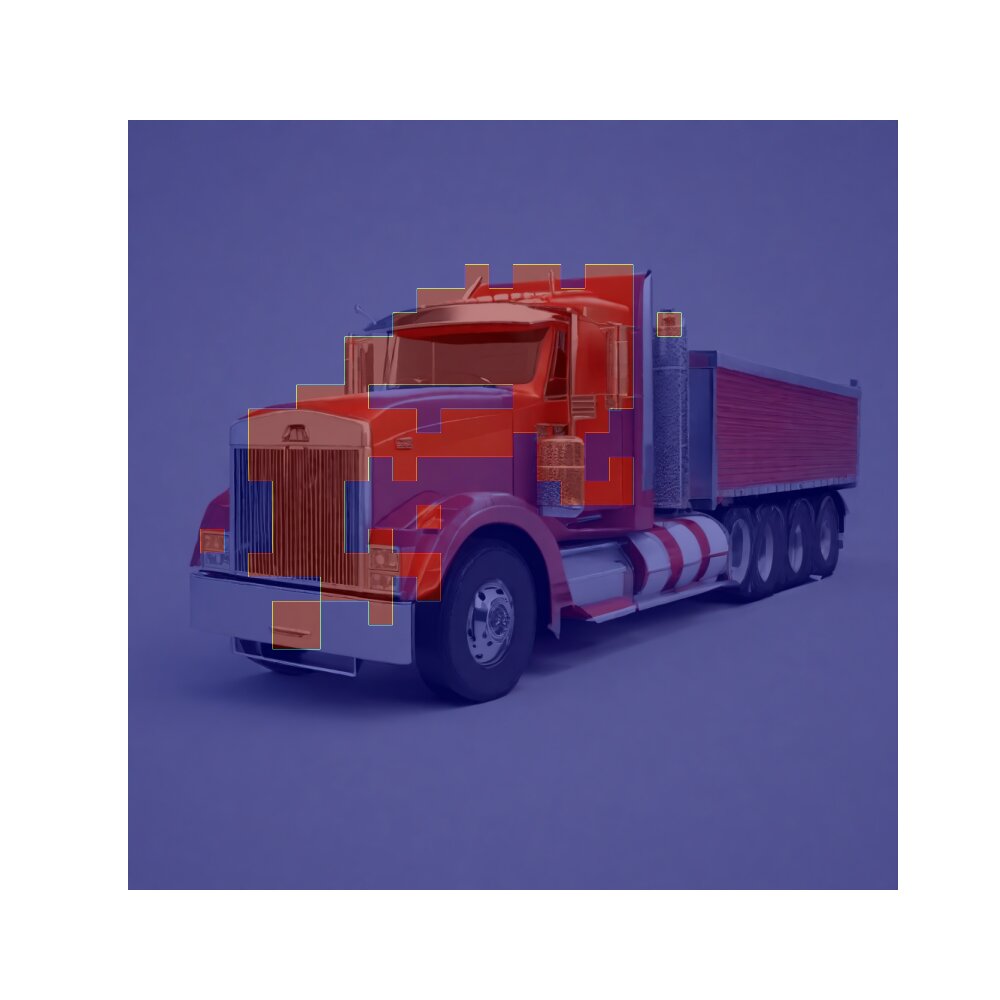}
        \end{minipage}%
        \begin{minipage}{0.17\textwidth}
            \centering
            \includegraphics[width=\linewidth]{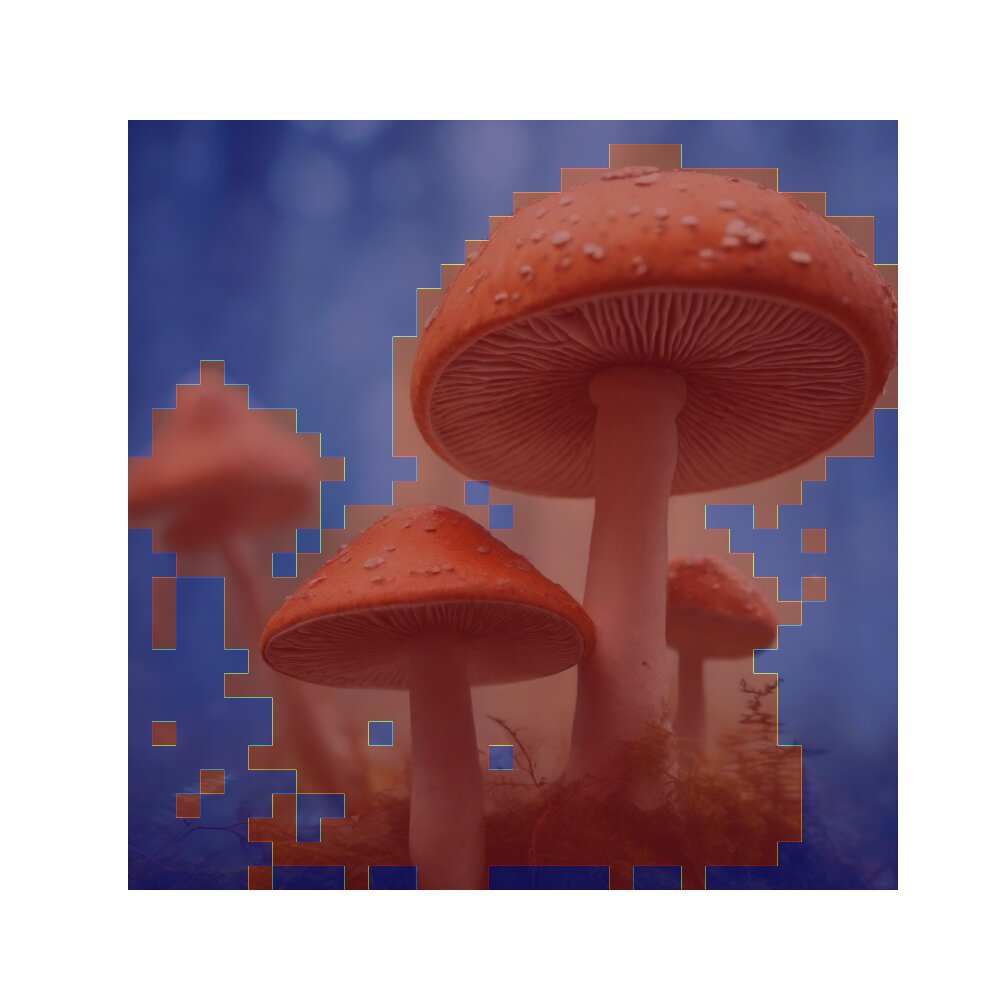}
        \end{minipage}%
        \begin{minipage}{0.17\textwidth}
            \centering
            \includegraphics[width=\linewidth]{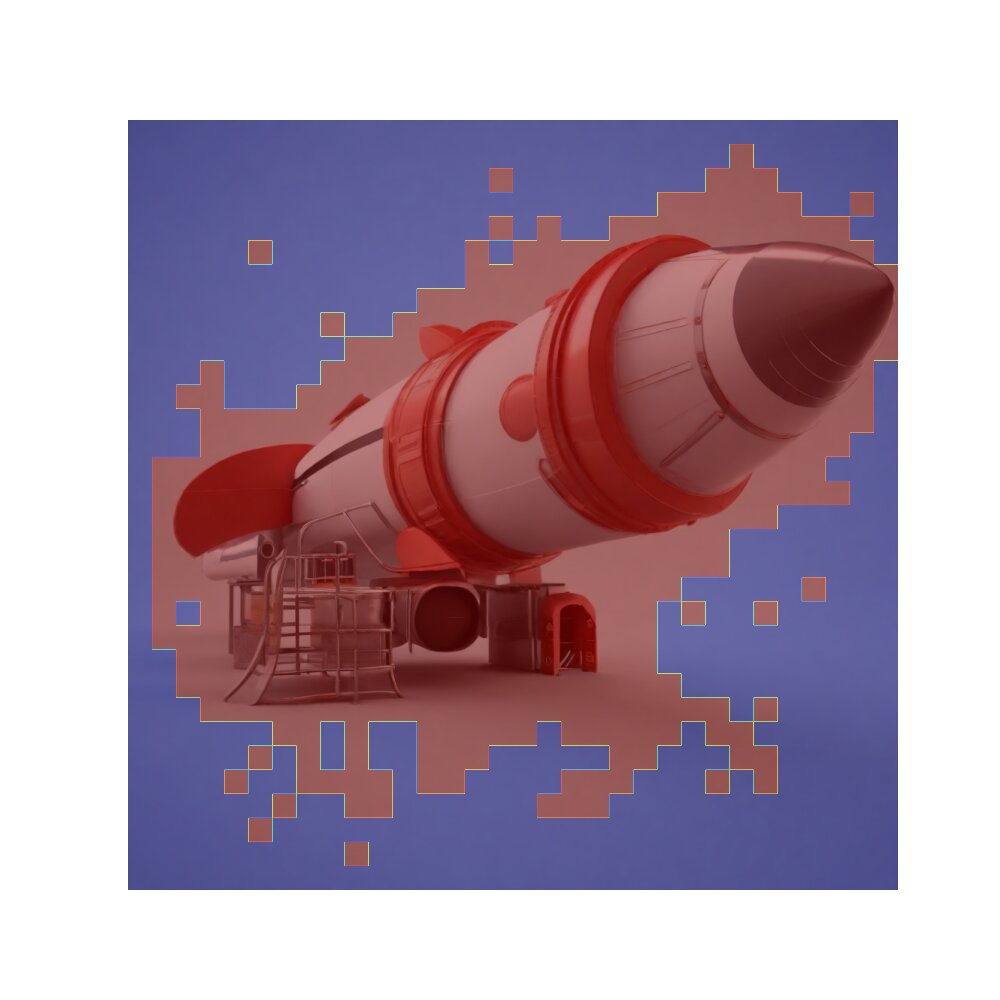}
        \end{minipage}%
        \begin{minipage}{0.17\textwidth}
            \centering
            \includegraphics[width=\linewidth]{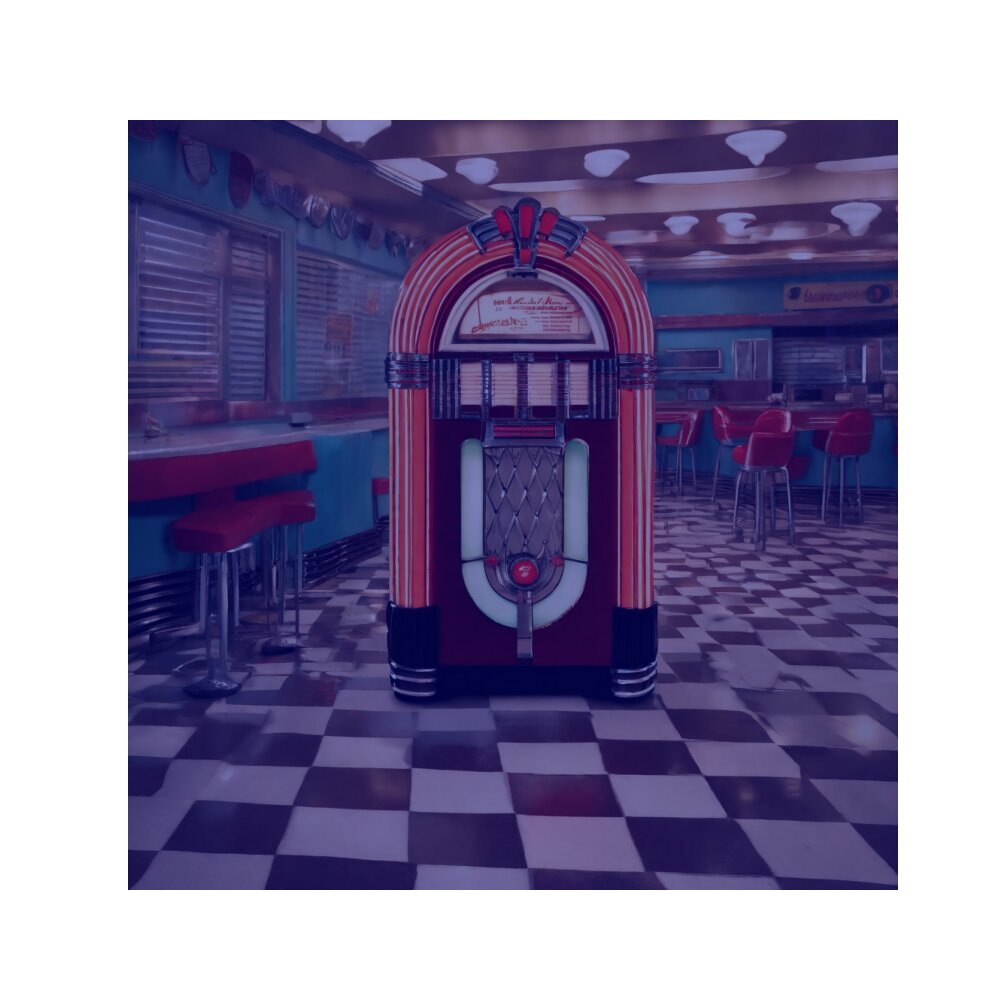}
        \end{minipage} 
        \vspace{0.1cm}
        \begin{minipage}{0.17\textwidth}
            \centering
            \includegraphics[width=\linewidth]{graphics/results/ablation1/cat_cluster.jpg}
            \subcaption{``Cat"}\label{fig:cat_thres}
        \end{minipage}%
        \begin{minipage}{0.17\textwidth}
            \centering
            \includegraphics[width=\linewidth]{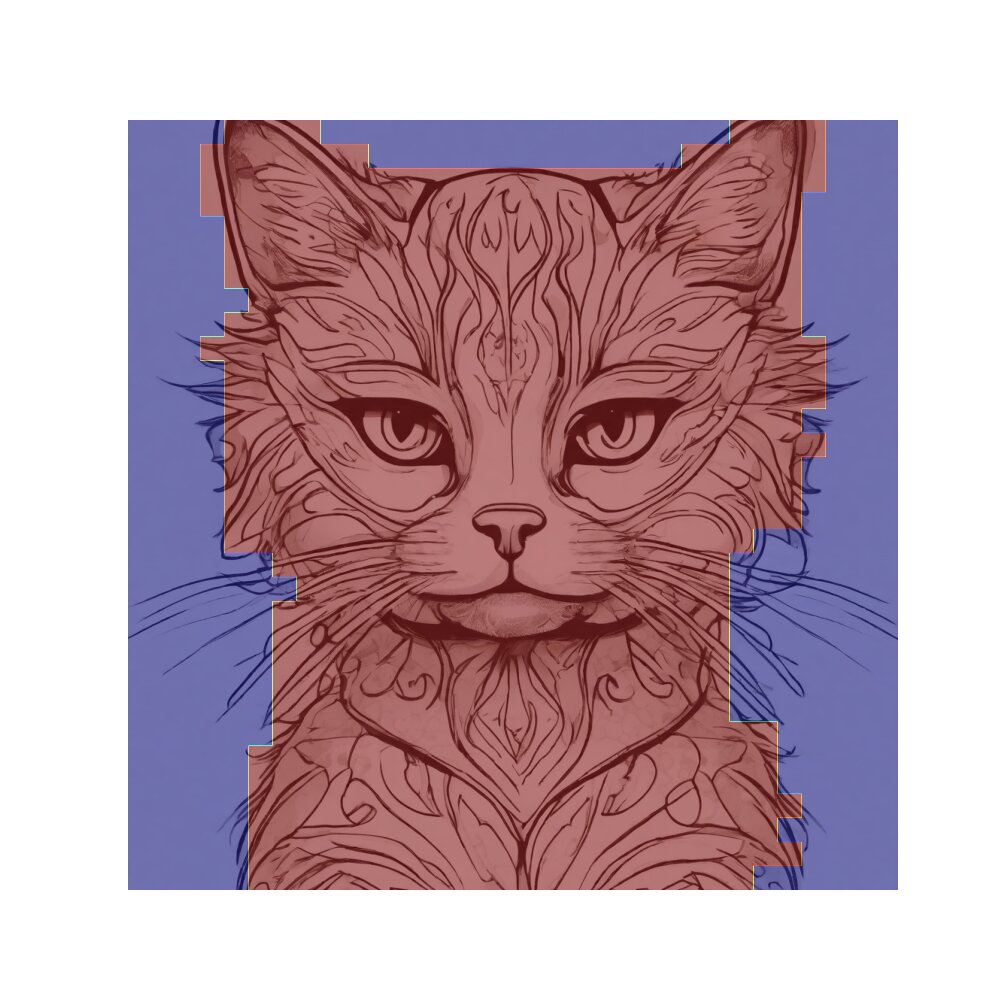}
            \subcaption{``Cat"}\label{fig:cat2_thres}
        \end{minipage}%
        \begin{minipage}{0.17\textwidth}
            \centering
            \includegraphics[width=\linewidth]{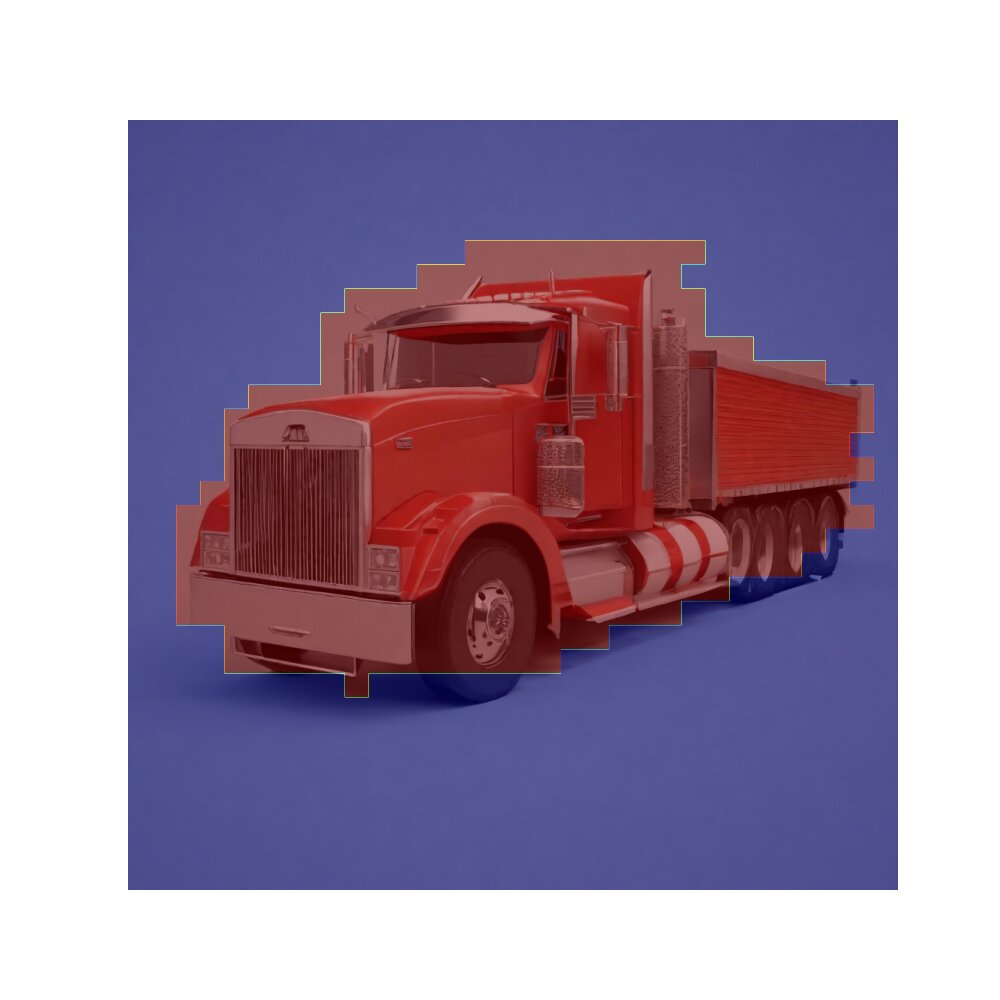}
            \subcaption{``Truck"}\label{fig:truck_thres}
        \end{minipage}%
        \begin{minipage}{0.17\textwidth}
            \centering
            \includegraphics[width=\linewidth]{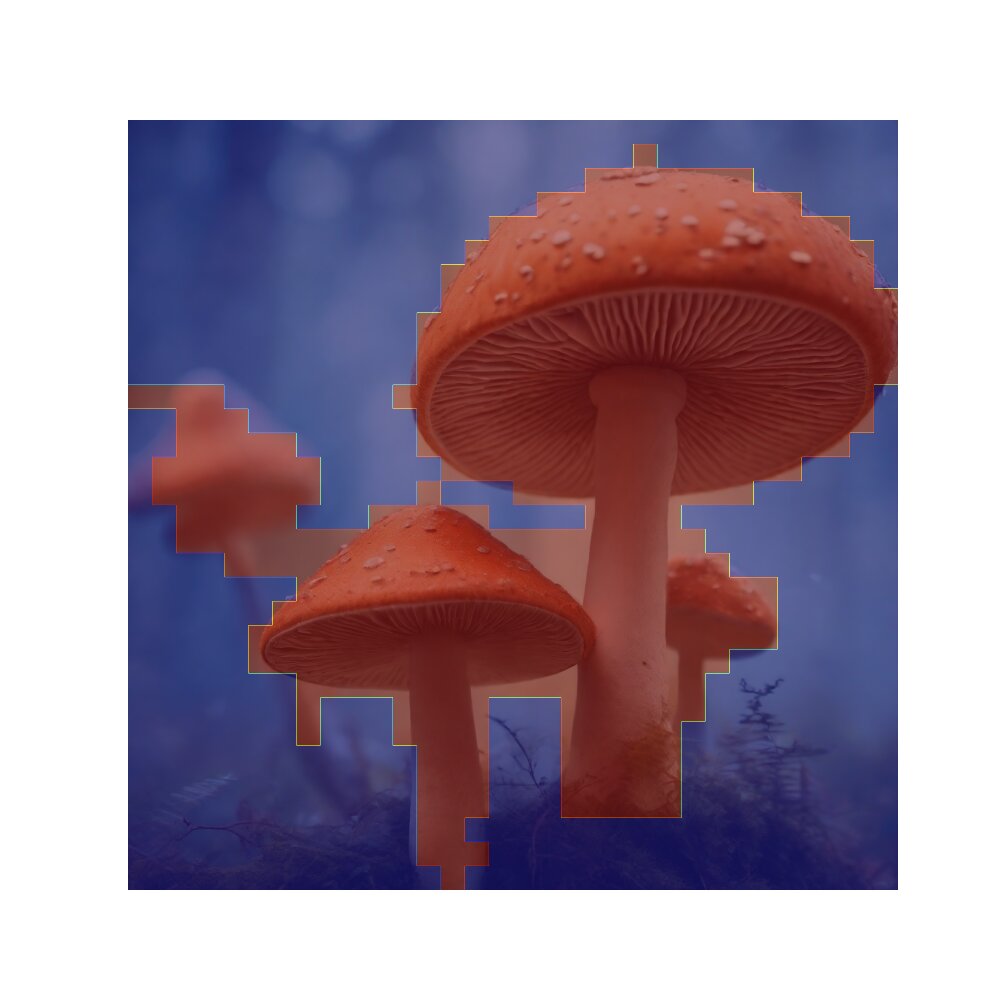}
            \subcaption{``Mushroom"}\label{fig:mushroom_thres}
        \end{minipage}%
        \begin{minipage}{0.17\textwidth}
            \centering
            \includegraphics[width=\linewidth]{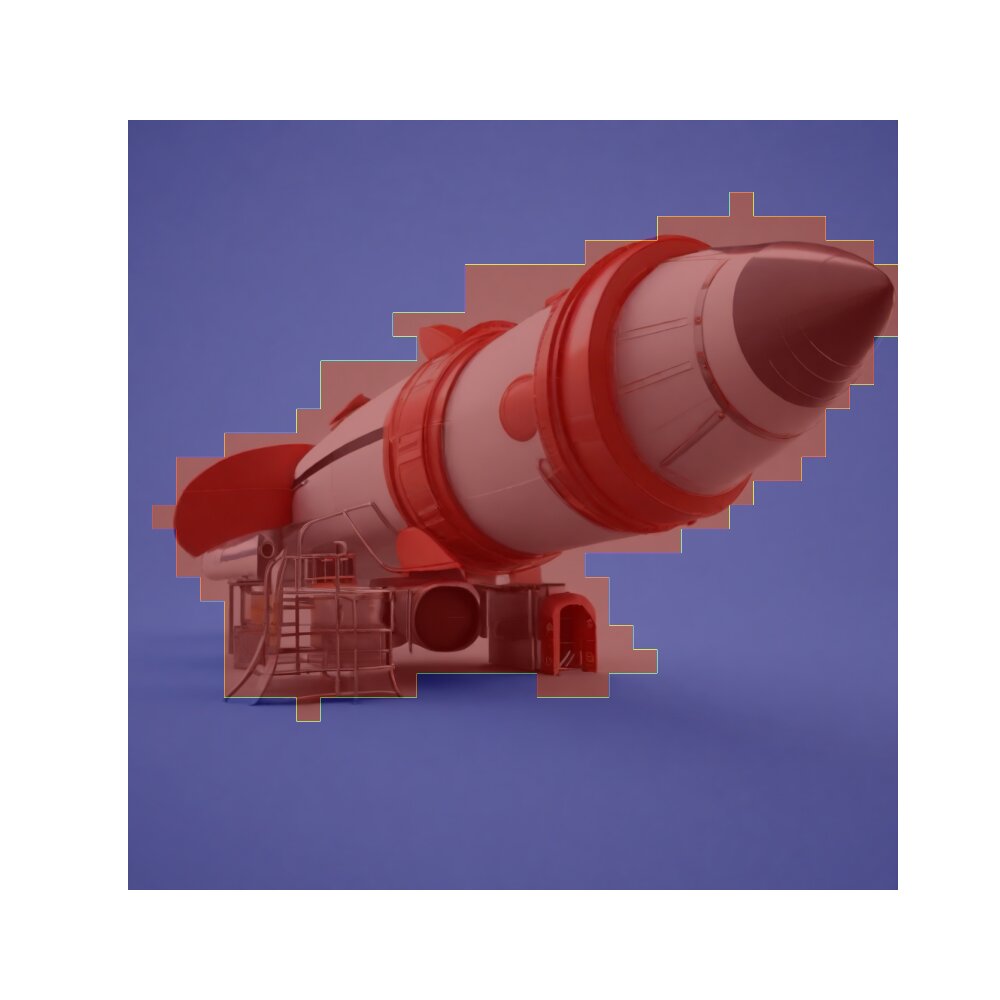}
            \subcaption{``Rocket"}\label{fig:rocket_thres}
        \end{minipage}%
        \begin{minipage}{0.17\textwidth}
            \centering
            \includegraphics[width=\linewidth]{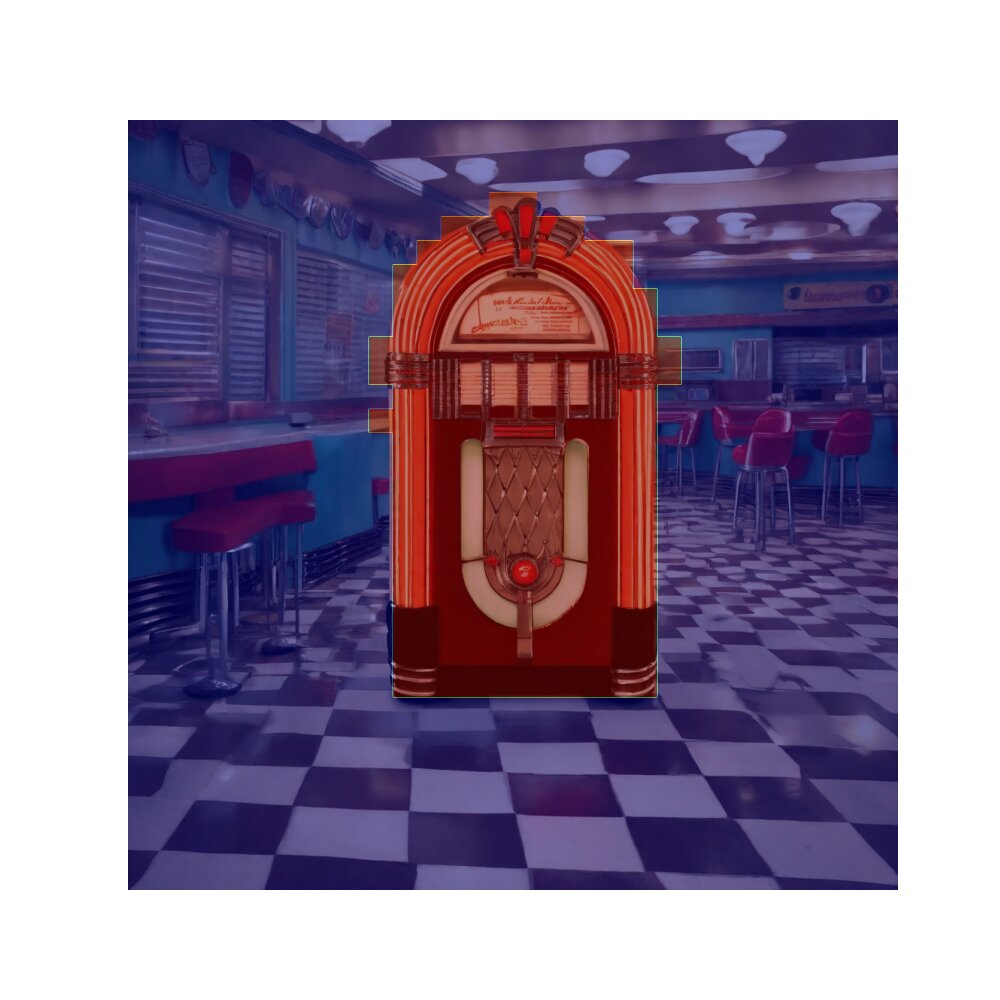}
            \subcaption{``Jukebox"}\label{fig:jukebox_thres}
        \end{minipage}
        
    \end{minipage}
    \caption{\textbf{Insufficiency of Fixed Thresholding:} Binarization of $\hat{\mathbf{A}}_{sub}$ for different images, indicating the ability to correctly localize the subject, either via fixed thresholding (top row) or via the proposed approach (bottom row).}
    \label{fig:Ablation_1}
\end{figure*}

\textbf{Insufficiency of Fixed Thresholding.}

To access and control content leakage we rely on the binary mask $\mathbf{R}$ to scale down only subject-related patches (see Sec. 3.3 of main manuscript and Sec. 1.2 of Supp. Material). 
As we described in Sec. 1.2 of this manuscript, the proposed extraction of $\mathbf{R}$ effectively calculates a different threshold for each $\hat{\mathbf{A}}_{sub}$ of the reference image. The same rationale was followed by~\cite{10.1145/3658157}, as opposed to the fixed threshold assumption of~\cite{patashnik2023localizingobjectlevelshapevariations}.
The fixed threshold alternative can be motivated by the fact that the $\hat{\mathbf{A}}_{sub}$ map corresponds to the aggregated cross-attention probabilities and thus a suitable probability-motivated threshold can work for all cases. 
Nonetheless, such an approach is inadequate in practice, as different text prompts result in varying attention probabilities. 
This stems from the variability of text-tokens within the prompt, which leads to distinct cross-attention distributions that cannot be modeled in advance. 

We visually illustrate the effectiveness of our approach and highlight the insufficiency of fixed thresholding in Fig.~\ref{fig:Ablation_1}. For the fixed thresholding case, we tune the threshold to faithfully capture the image's subject in the first column and fix it across all the other instances. 
As shown, the fixed threshold often fails, either being too high or too low, whereas our method consistently captures the visual elements of the object across all generated instances.

\begin{figure}[t]
    \centering
    \scriptsize
    \begin{tabular}{c@{\hspace{.1cm}}c@{\hspace{.1cm}}c}  
        Reference & With Subject Detection & Without Subject Detection \\

        \vspace{.1cm}

        \begin{minipage}{0.15\textwidth}
            \includegraphics[width=\textwidth]{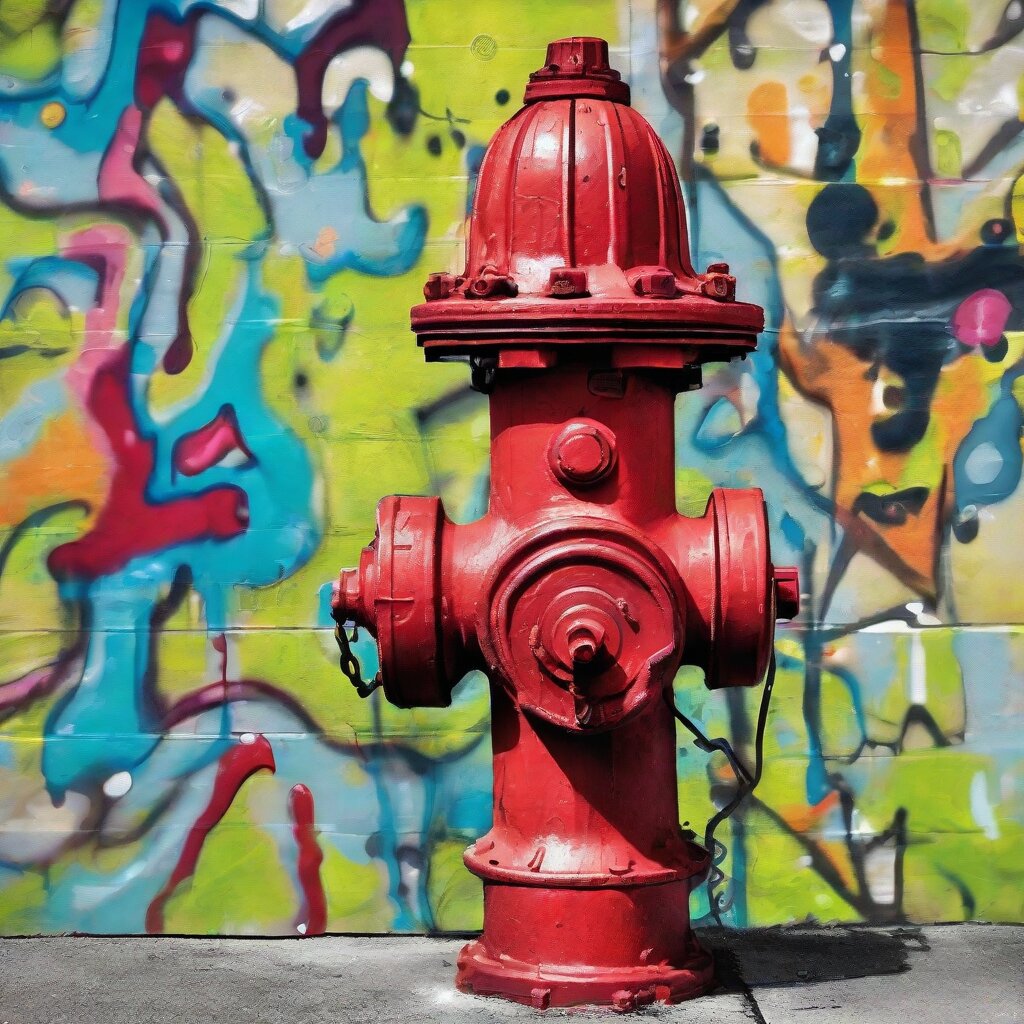}
        \end{minipage} &
        \begin{minipage}{0.15\textwidth}
            \includegraphics[width=\textwidth]{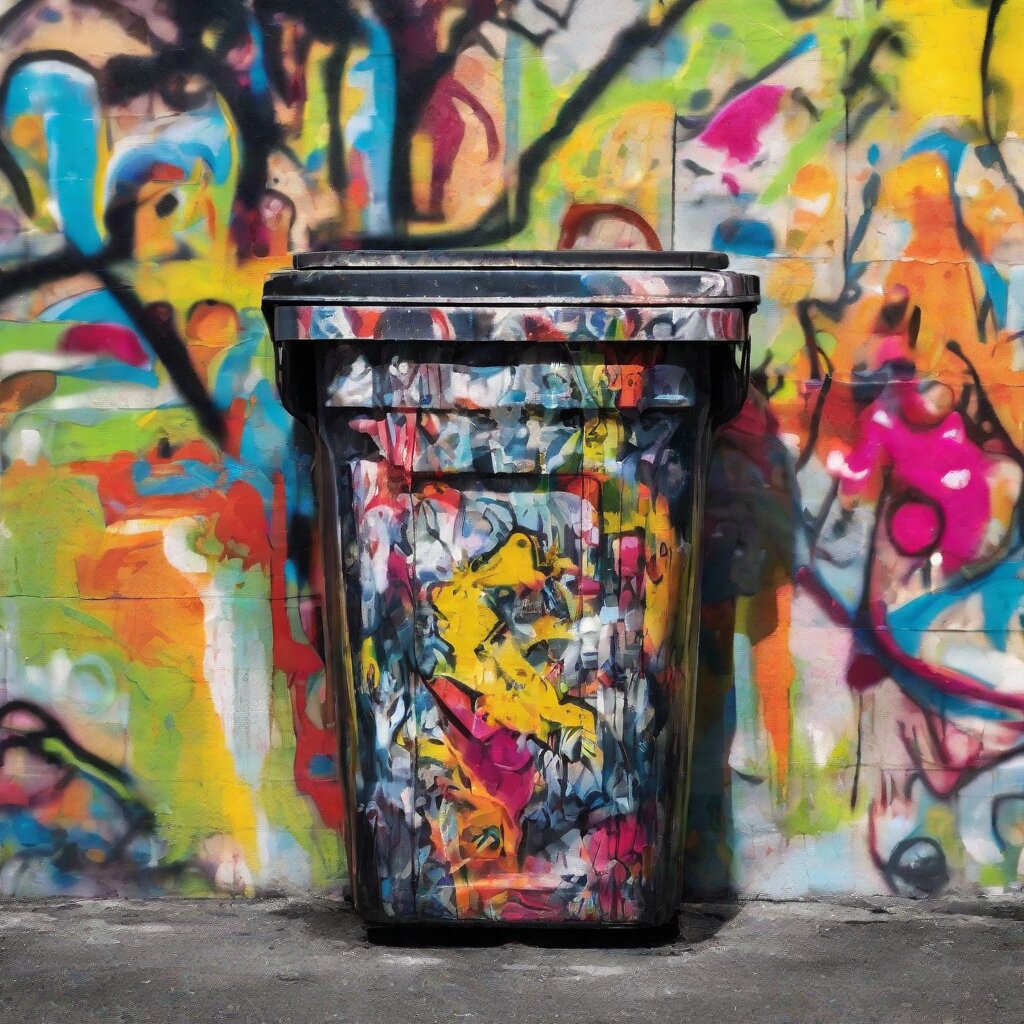}
        \end{minipage} &
        \begin{minipage}{0.15\textwidth}
            \includegraphics[width=\textwidth]{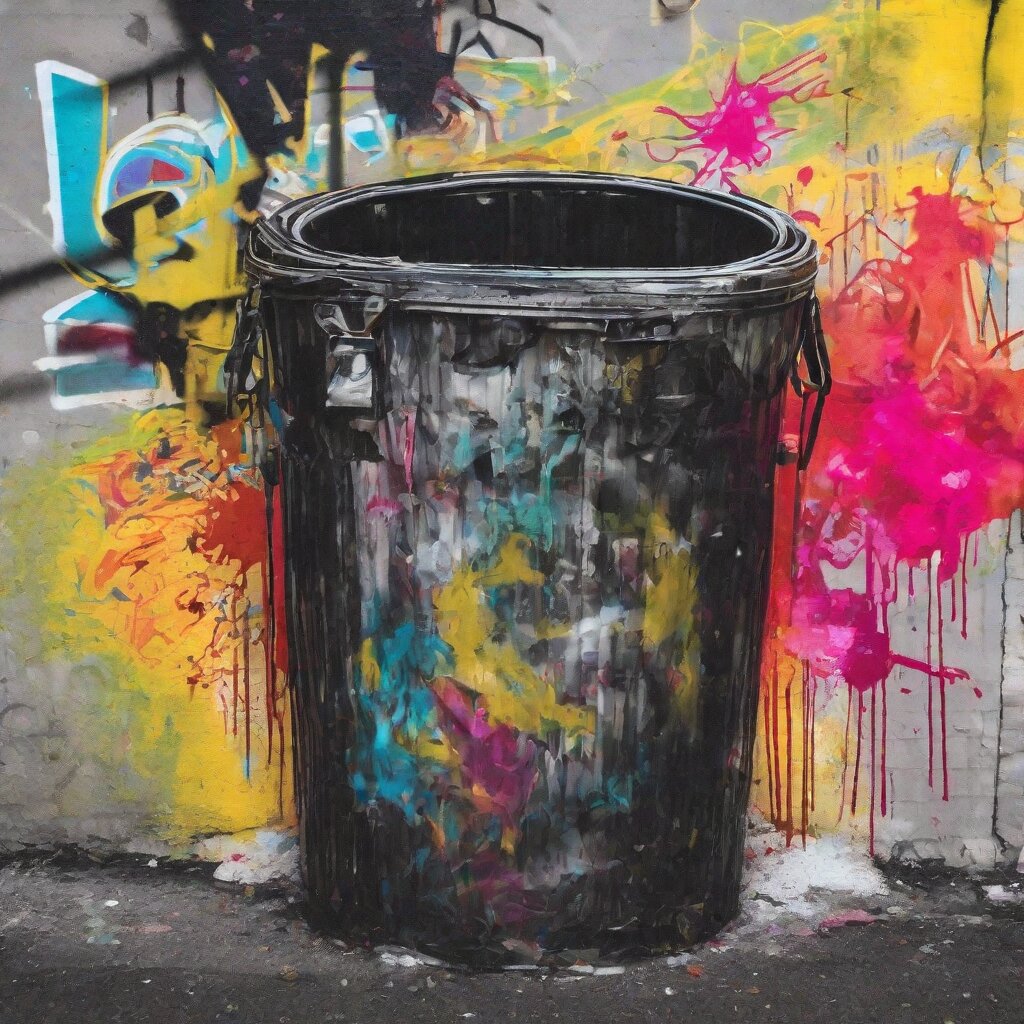}
        \end{minipage} \\

        \vspace{.1cm}
        
        \begin{minipage}{0.15\textwidth}
            \includegraphics[width=\textwidth]{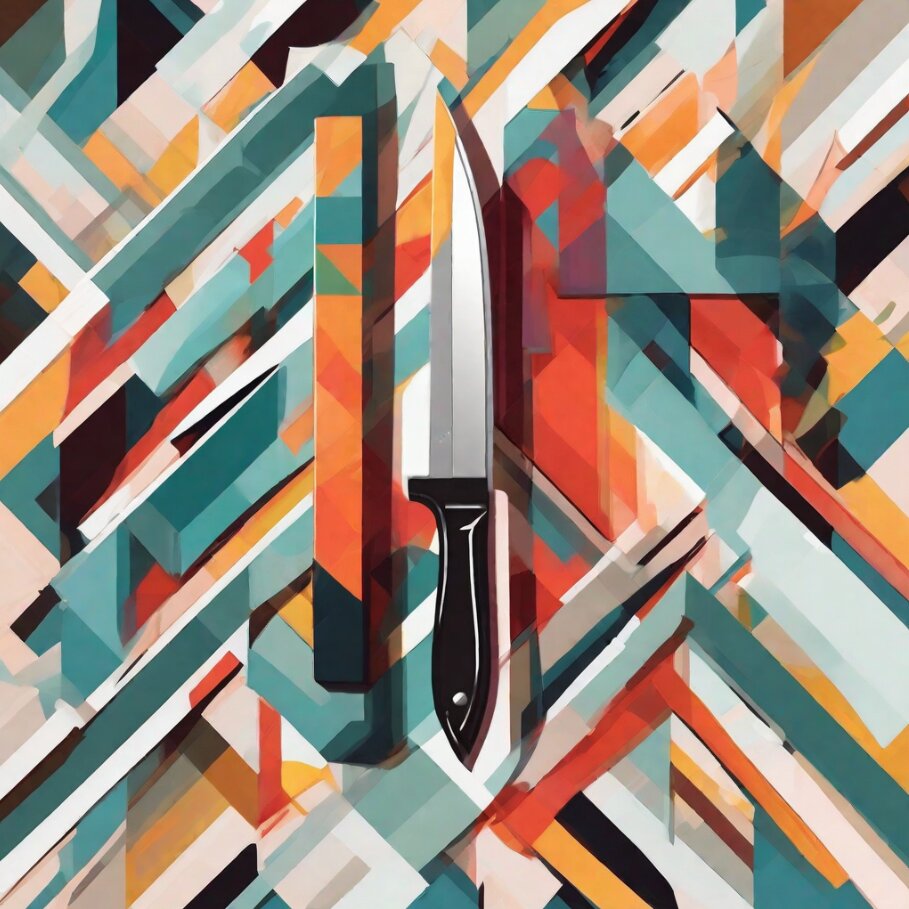}
        \end{minipage} &
        \begin{minipage}{0.15\textwidth}
            \includegraphics[width=\textwidth]{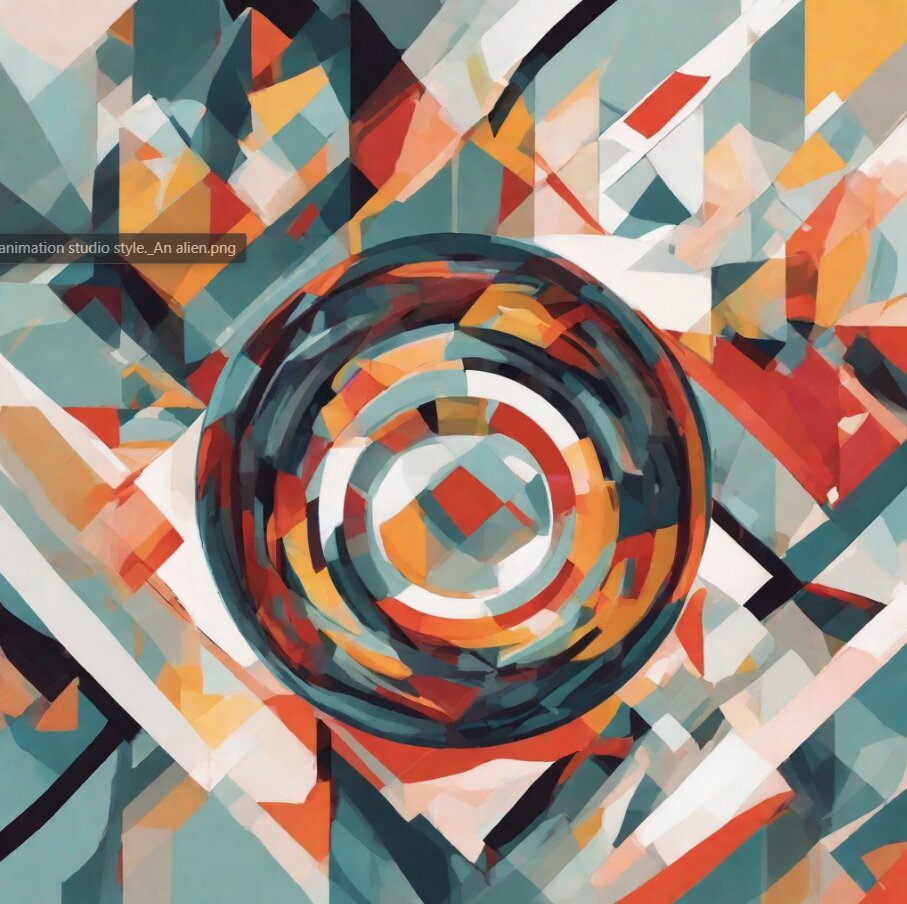}
        \end{minipage} &
        \begin{minipage}{0.15\textwidth}
            \includegraphics[width=\textwidth]{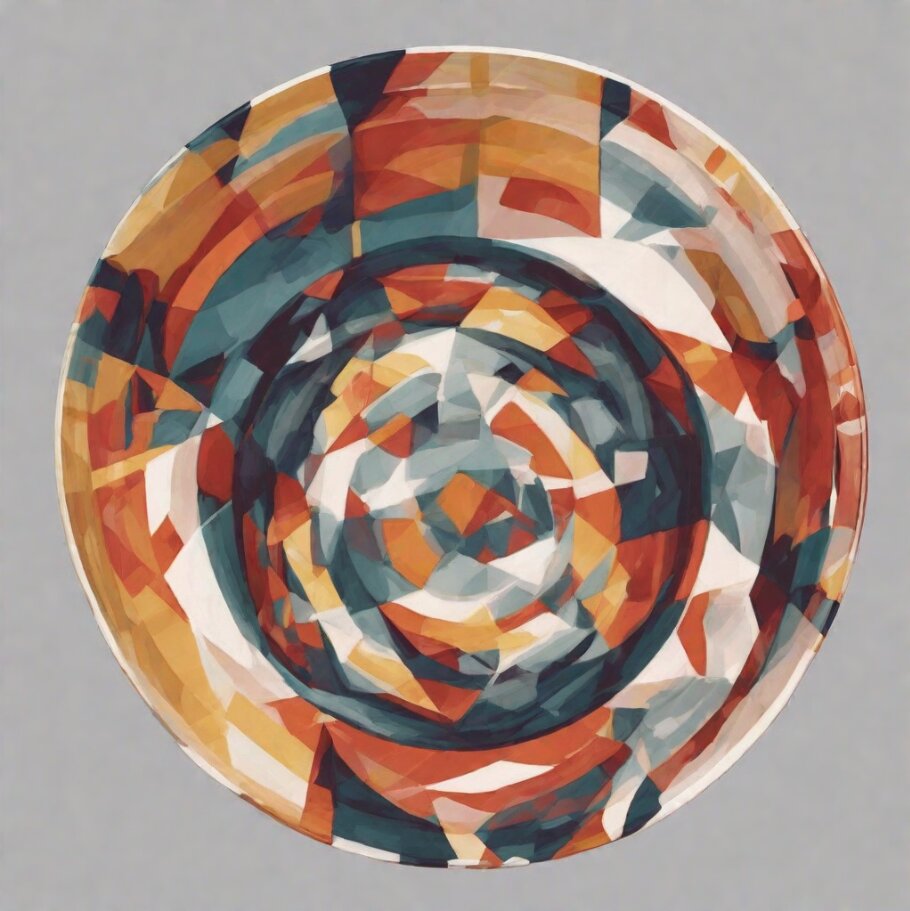}
        \end{minipage} \\

        \vspace{.1cm}

        \begin{minipage}{0.15\textwidth}
            \includegraphics[width=\textwidth]{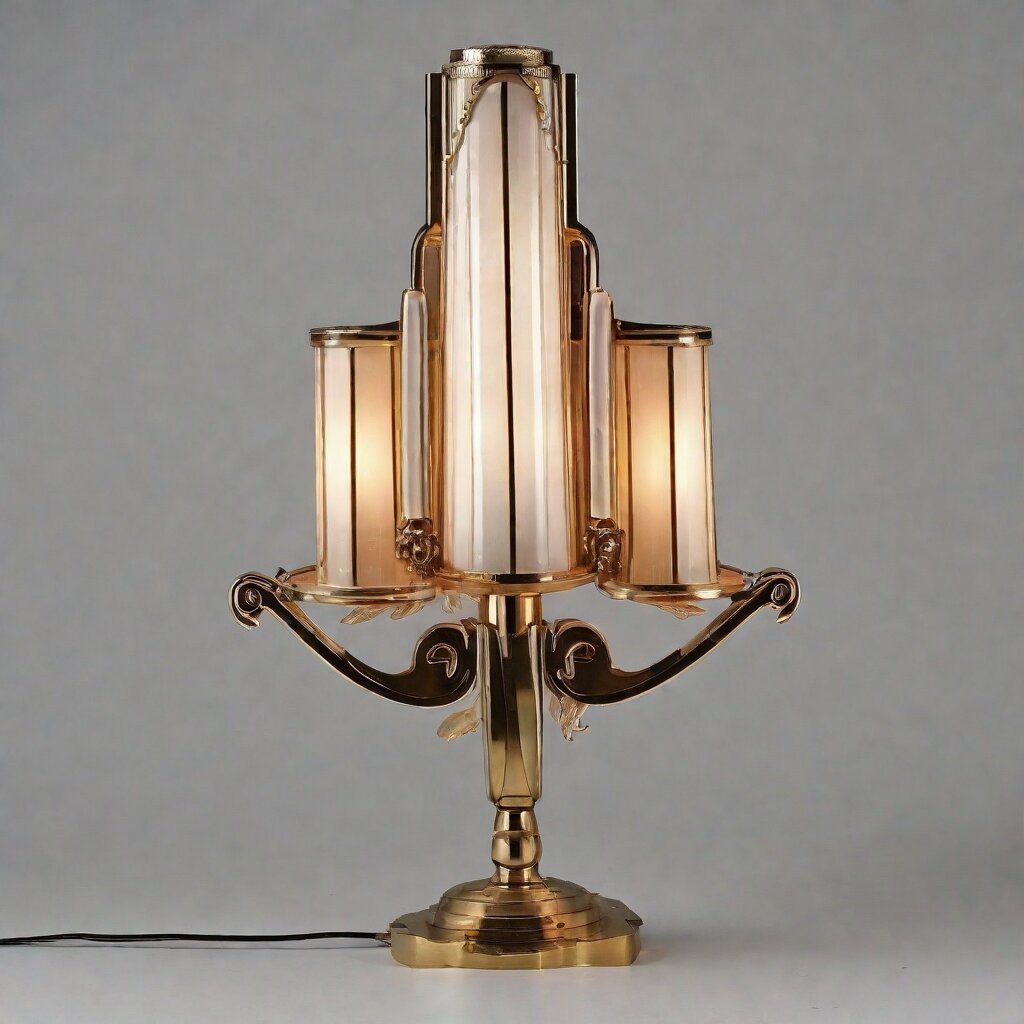}
        \end{minipage} &
        \begin{minipage}{0.15\textwidth}
            \includegraphics[width=\textwidth]{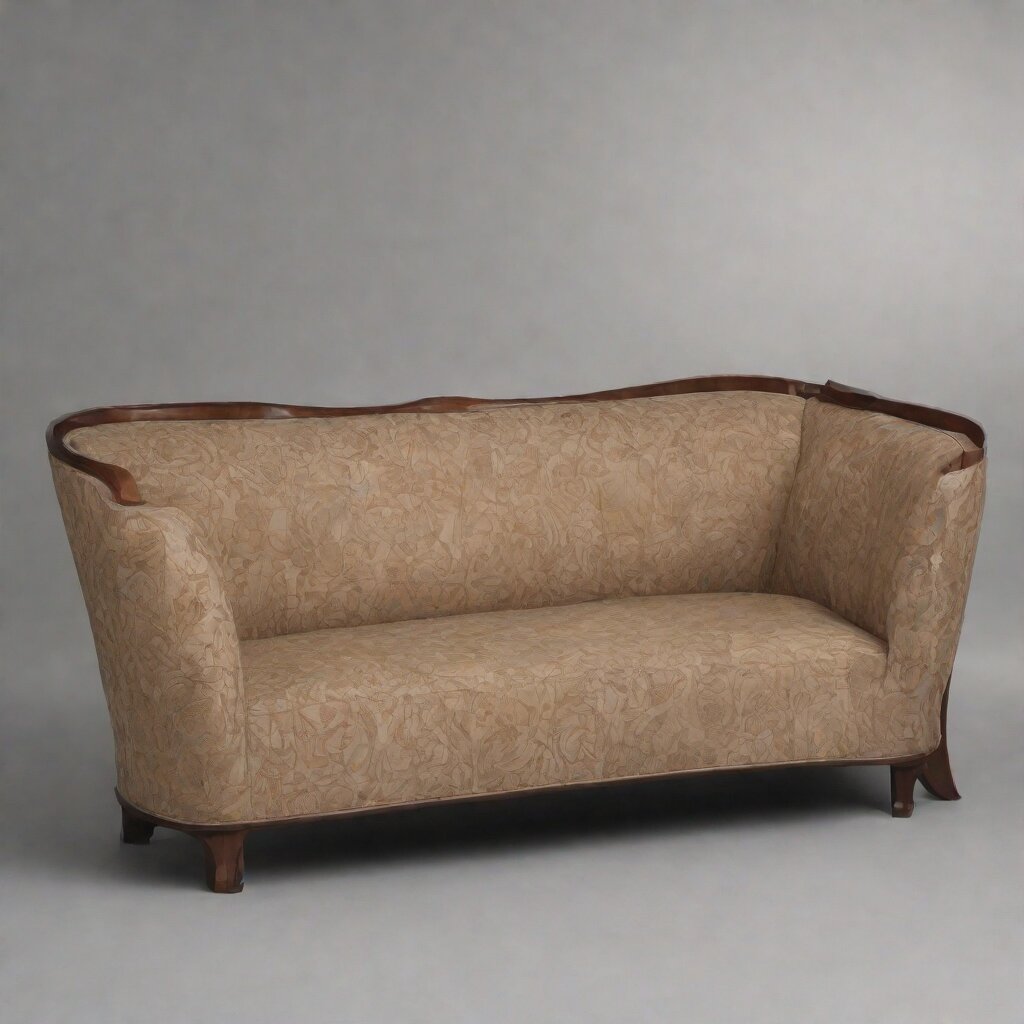}
        \end{minipage} &
        \begin{minipage}{0.15\textwidth}
            \includegraphics[width=\textwidth]{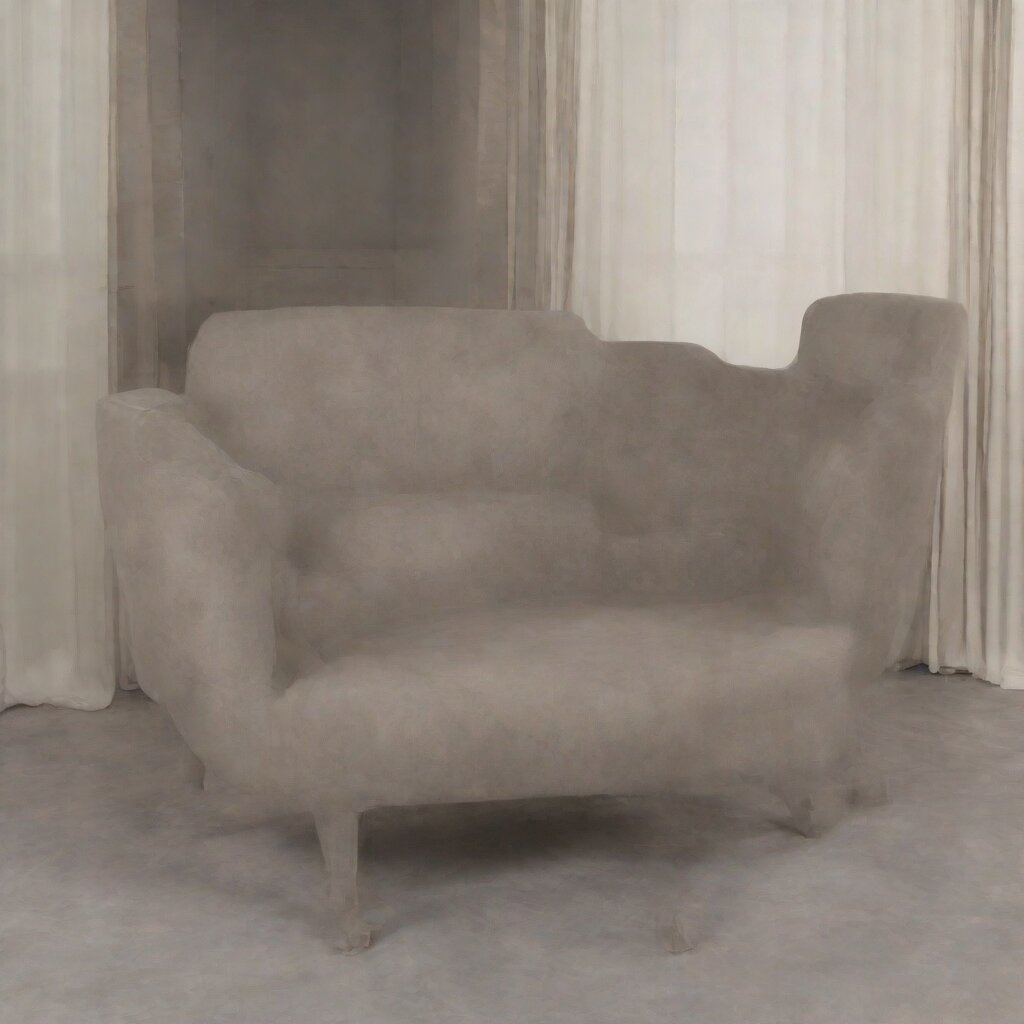}
        \end{minipage} \\

        \vspace{.1cm}
        
        \begin{minipage}{0.15\textwidth}
            \includegraphics[width=\textwidth]{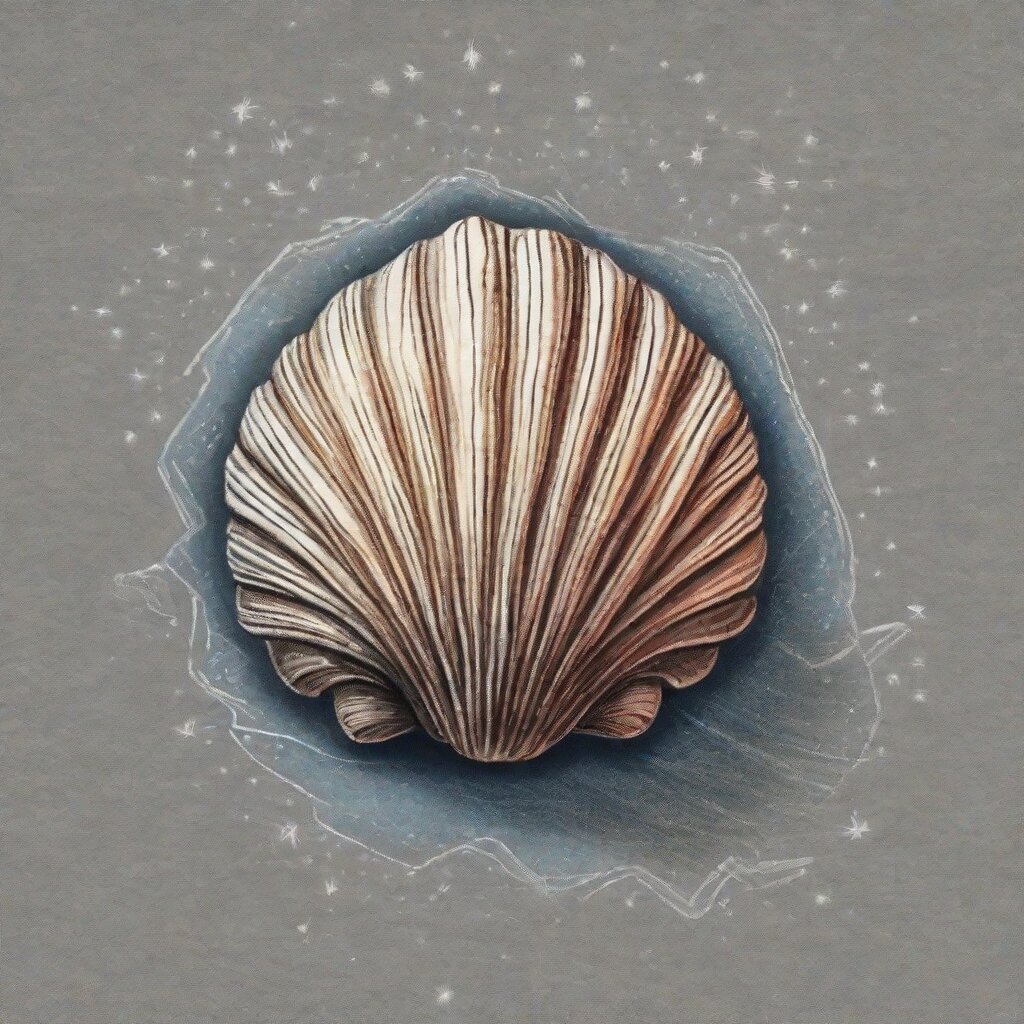}
        \end{minipage} &
        \begin{minipage}{0.15\textwidth}
            \includegraphics[width=\textwidth]{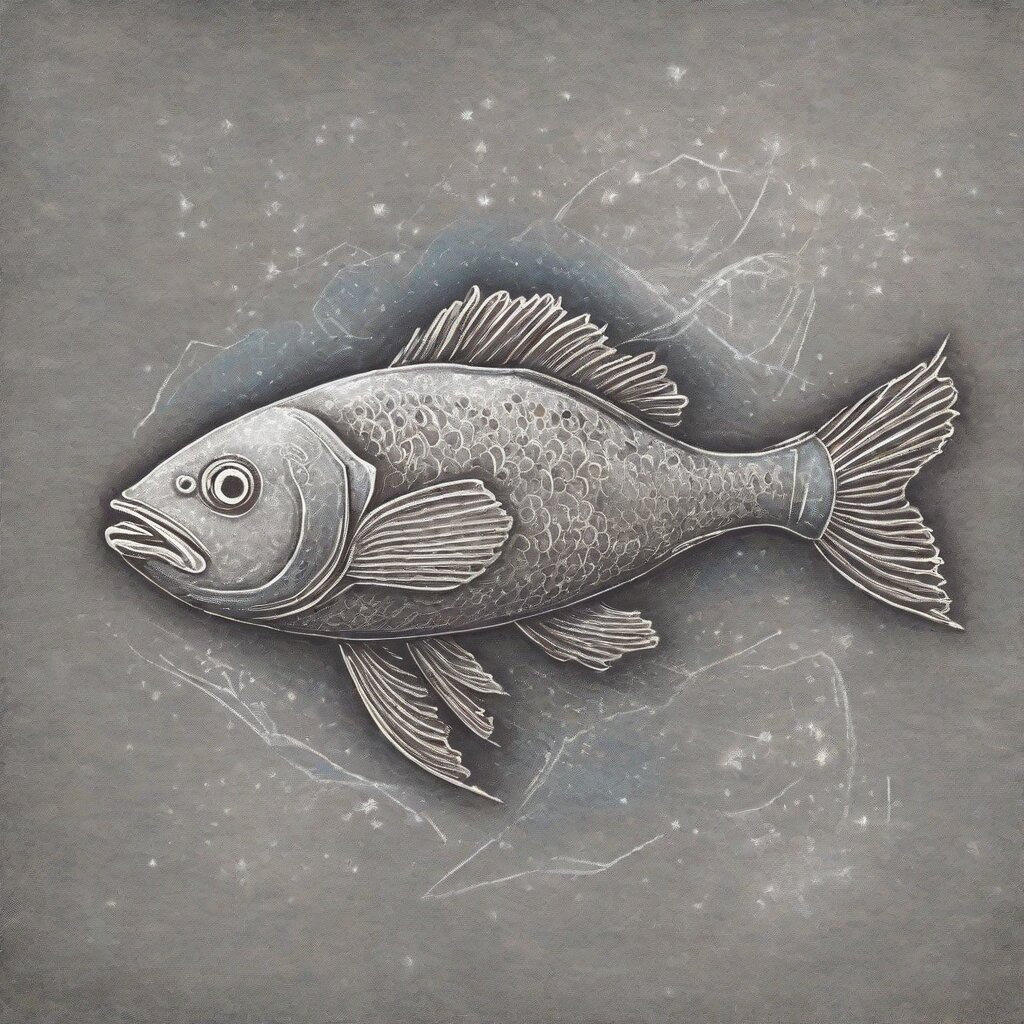}
        \end{minipage} &
        \begin{minipage}{0.15\textwidth}
            \includegraphics[width=\textwidth]{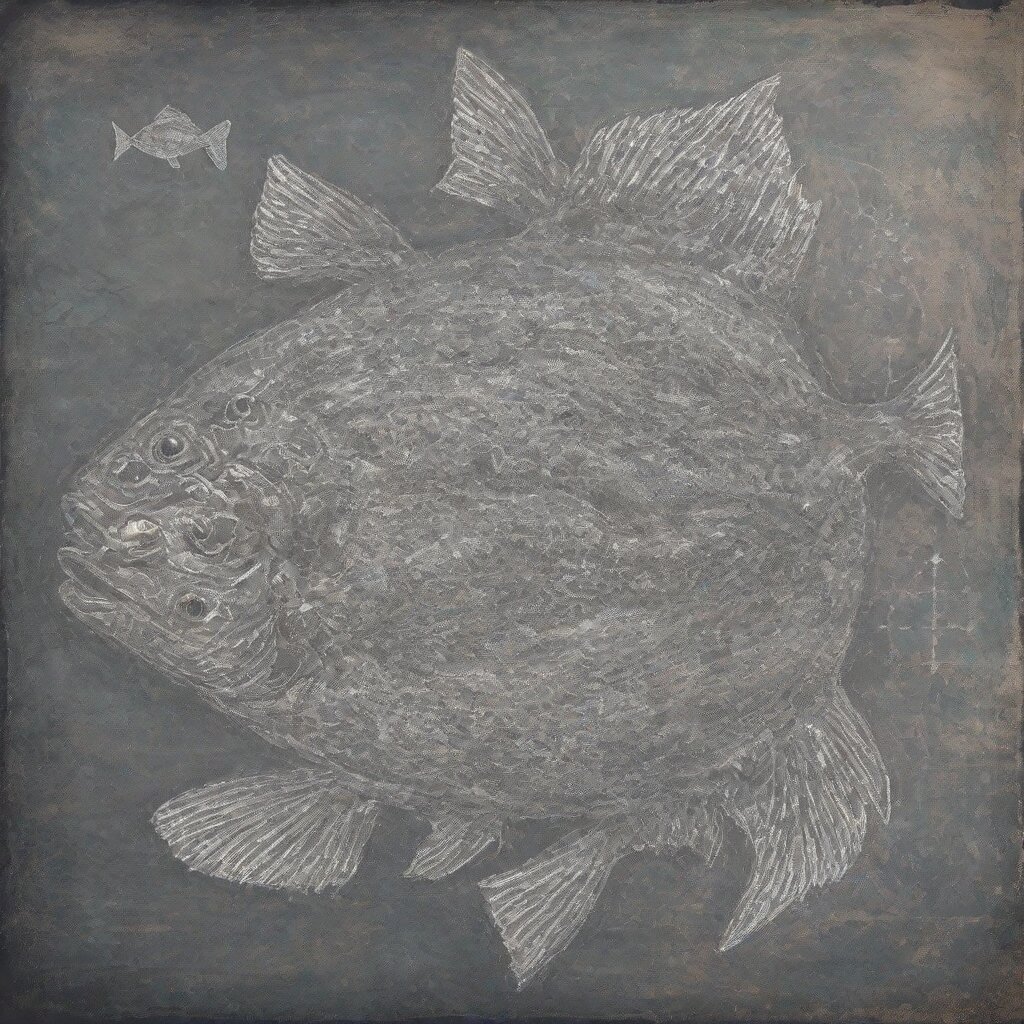}
        \end{minipage} \\
    \end{tabular}
    
    \caption{\textbf{Impact of Subject Detection.} We compare the results of the scaling method presented in Sec. 3.3, with and without the fine-grained choice on the reference image subject patches. Specifically, we fix the scaling parameter across our experiment ($\alpha = .5$), controlling the transfer of, on the one hand, the reference subject image patches (proposed - middle column), and on the other hand, all reference image patches (right column). We observe that scaling all patches ruins the stylistic alignment (top two rows), or exhibits destructive results (bottom two rows).}
    \label{fig:Ablation_3}
\end{figure}

\textbf{Impact of Subject Detection.} 

To motivate the annotation of the reference subject patches, we visually illustrate the effect of scaling all the reference image patches, essentially setting $\mathbf{R} = \mathbf{1}_{H \times W}$.
We compare the results of our fixed scaling pipeline ($\alpha = .5$), presented within the ablation study in Sec. 4.2, with and without this fine-grained choice of patches, and visually display the results in Fig.~\ref{fig:Ablation_3}. 
It is obvious that scaling agnostically the reference image patches ruins the stylistic alignment of the target image with respect to the reference. 
Moreover, in many cases the structure and semantics of the image are ruined as well. 
Note that this approach (i.e., scaling all patches) has been employed in \cite{hertz2024style} in order to mitigate the transfer of extremely popular reference image assets, which can result in disregarding the target prompt.
 
\begin{figure}[t]
    \centering
    \begin{tabular}{c@{\hspace{.1cm}}c@{\hspace{.1cm}}c@{\hspace{.1cm}}c}  
        \scriptsize Reference style & \scriptsize Target image & \scriptsize Ours & \scriptsize Cross attention\\
        
        \begin{minipage}{0.11\textwidth}
            \includegraphics[width=\textwidth]{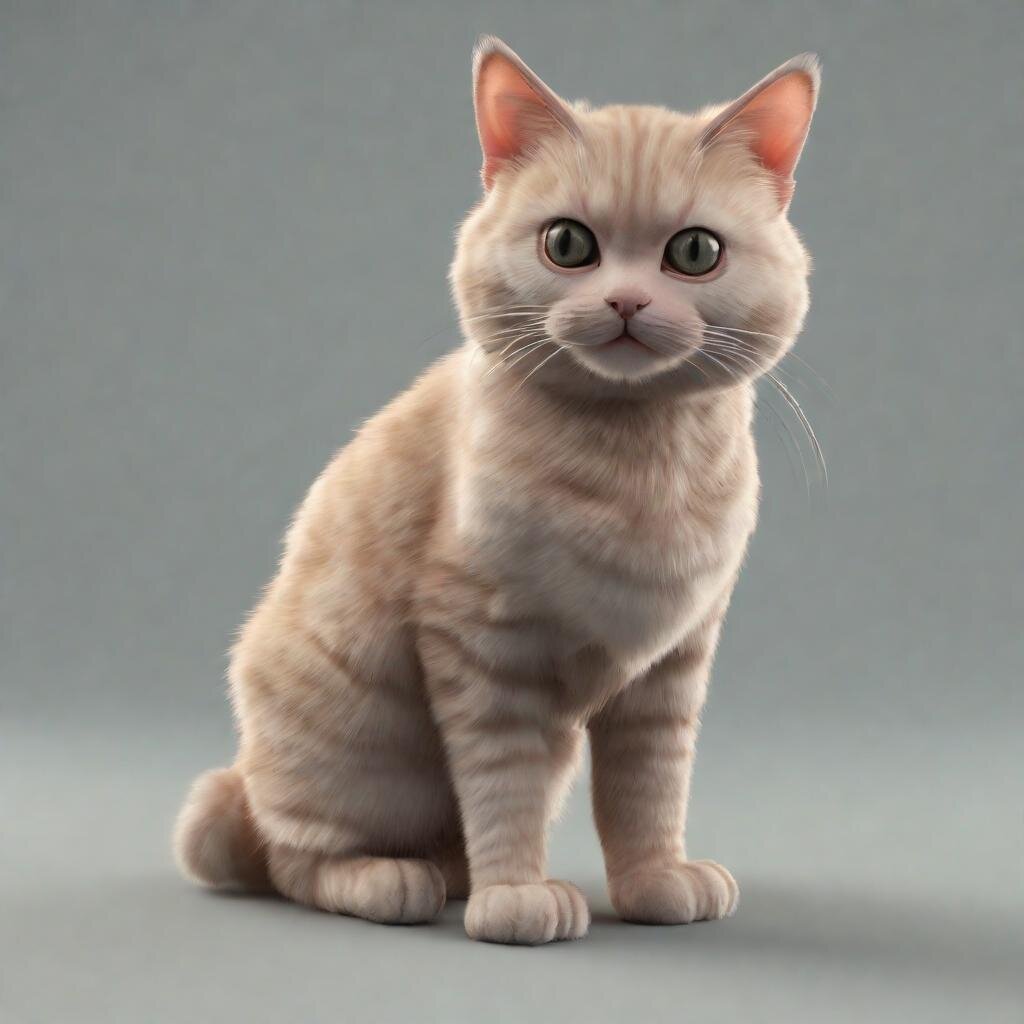}
            \subcaption*{\scriptsize A cat}
        \end{minipage} &
        \begin{minipage}{0.11\textwidth}
            \includegraphics[width=\textwidth]{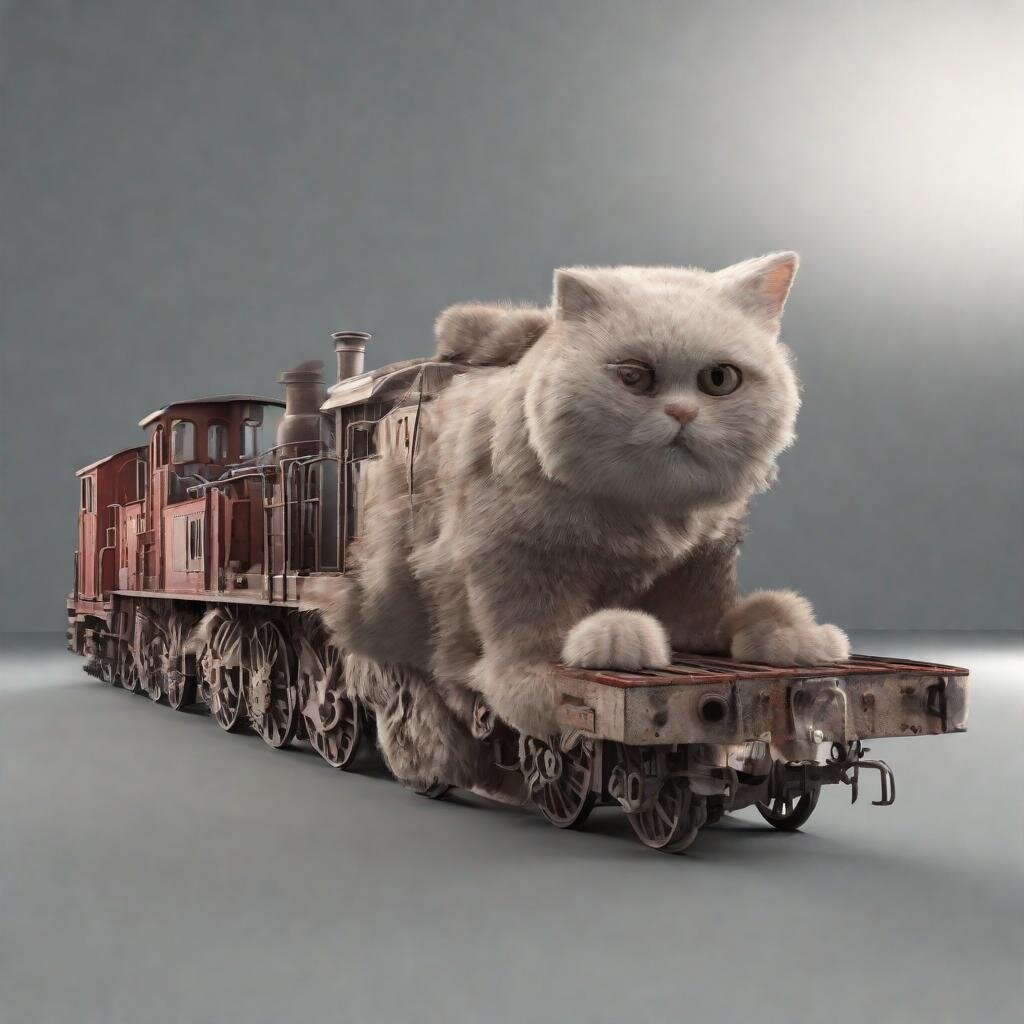}
            \subcaption*{\scriptsize A train}
        \end{minipage} &
        \begin{minipage}{0.11\textwidth}
            \includegraphics[width=\textwidth]{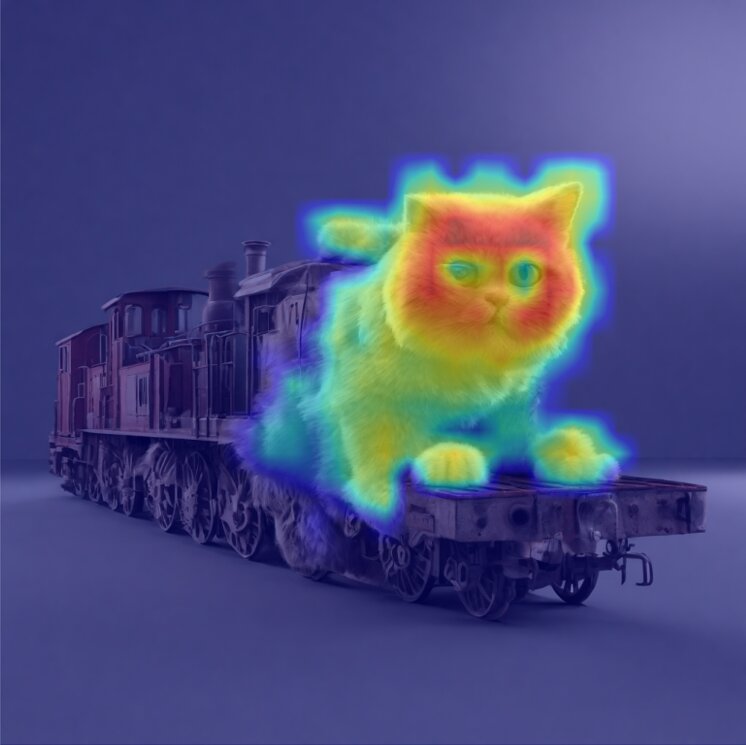}
            \subcaption*{\scriptsize cat - train}
        \end{minipage} &
        \begin{minipage}{0.11\textwidth}
            \includegraphics[width=\textwidth]{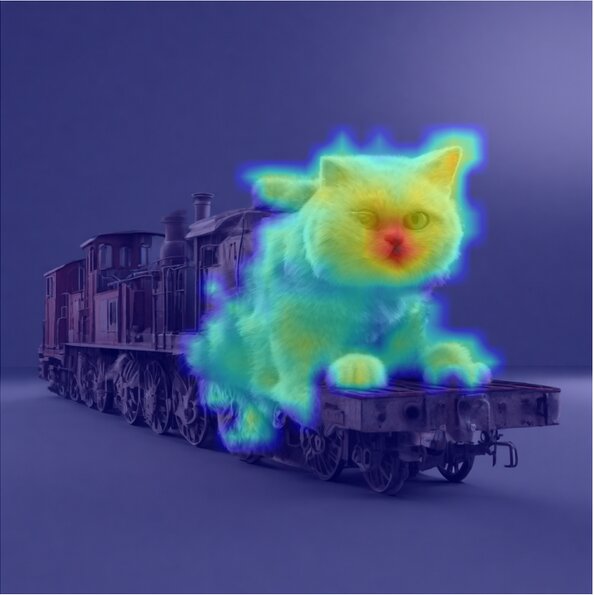}
            \subcaption*{\scriptsize cat - train}
        \end{minipage} \\
        \vspace{4pt}
        \begin{minipage}{0.11\textwidth}
            \includegraphics[width=\textwidth]{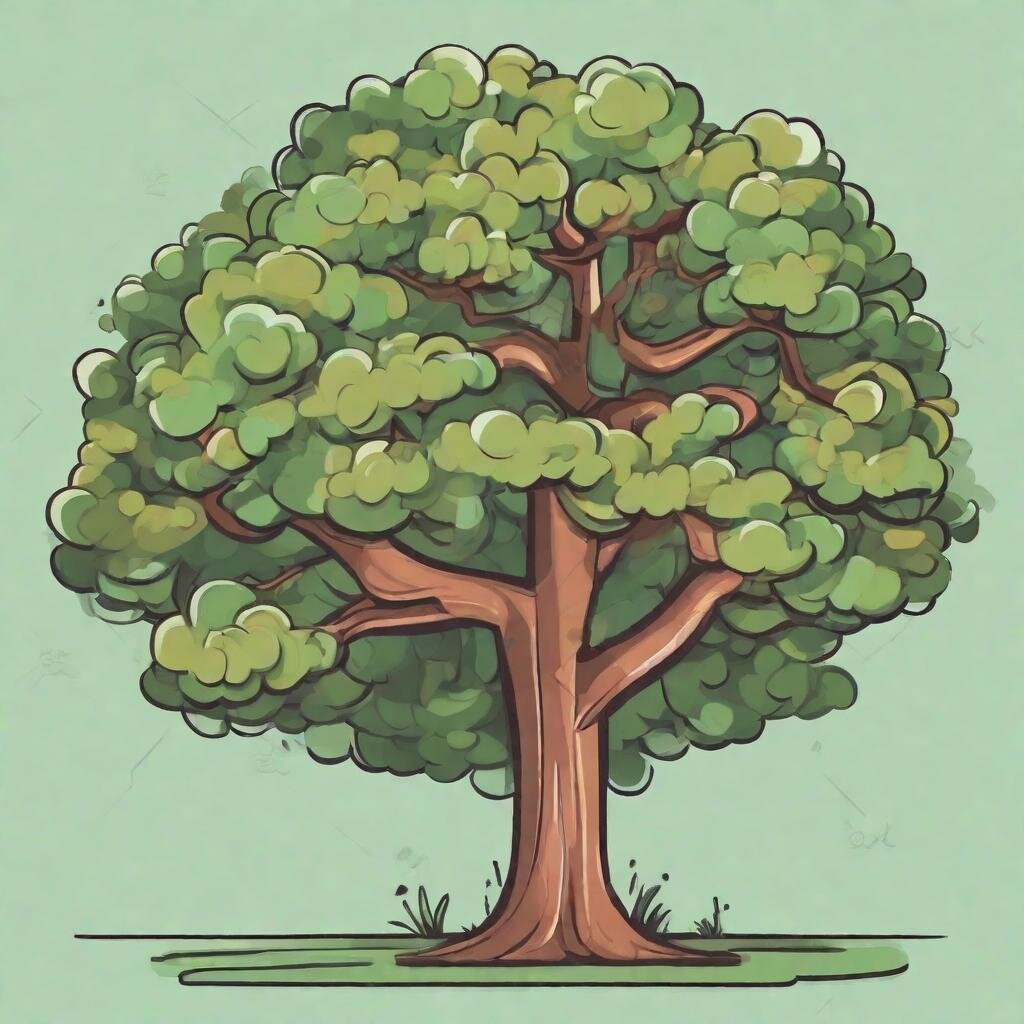}
            \subcaption*{\scriptsize A tree}
        \end{minipage} &
        \begin{minipage}{0.11\textwidth}
            \includegraphics[width=\textwidth]{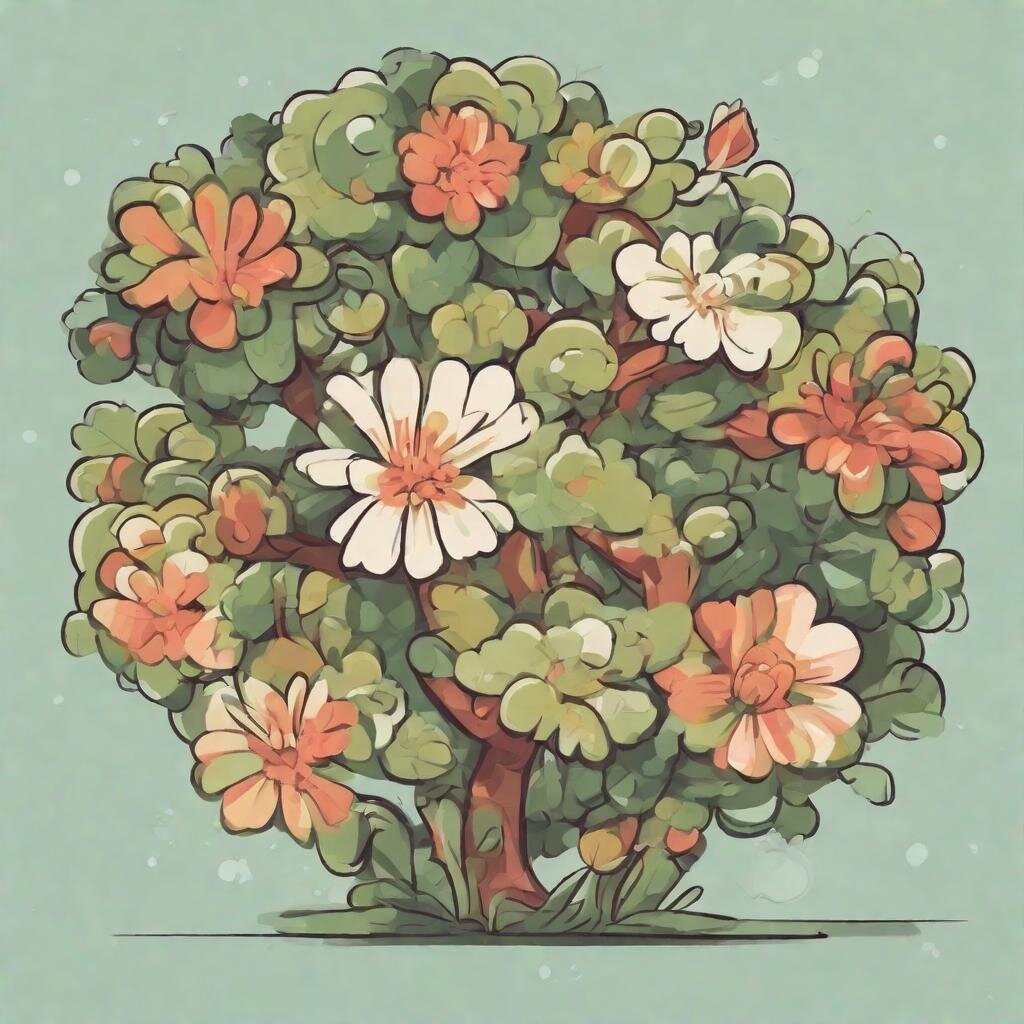}
            \subcaption*{\scriptsize A flower}
        \end{minipage} &
        \begin{minipage}{0.11\textwidth}
            \includegraphics[width=\textwidth]{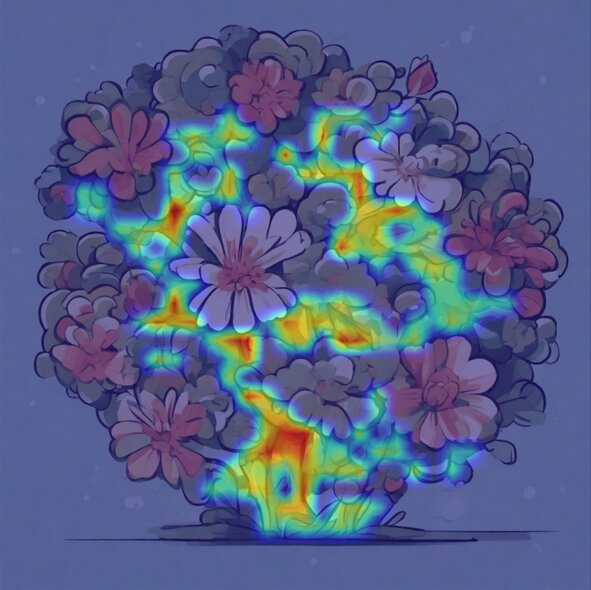}
            \subcaption*{\scriptsize tree - flower}
        \end{minipage} &
        \begin{minipage}{0.11\textwidth}
            \includegraphics[width=\textwidth]{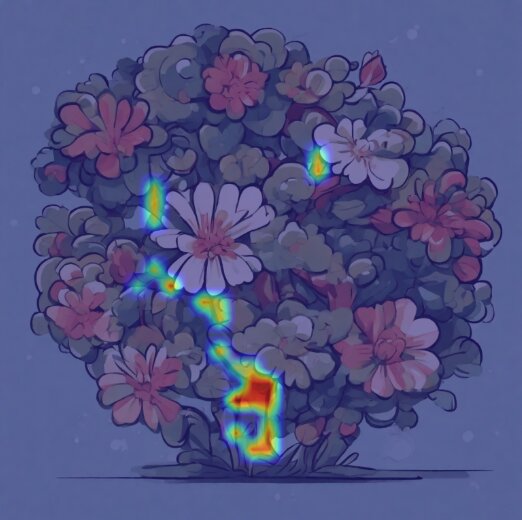}
            \subcaption*{\scriptsize tree - flower}
        \end{minipage} \\
        \vspace{4pt}
        \begin{minipage}{0.11\textwidth}
            \includegraphics[width=\textwidth]{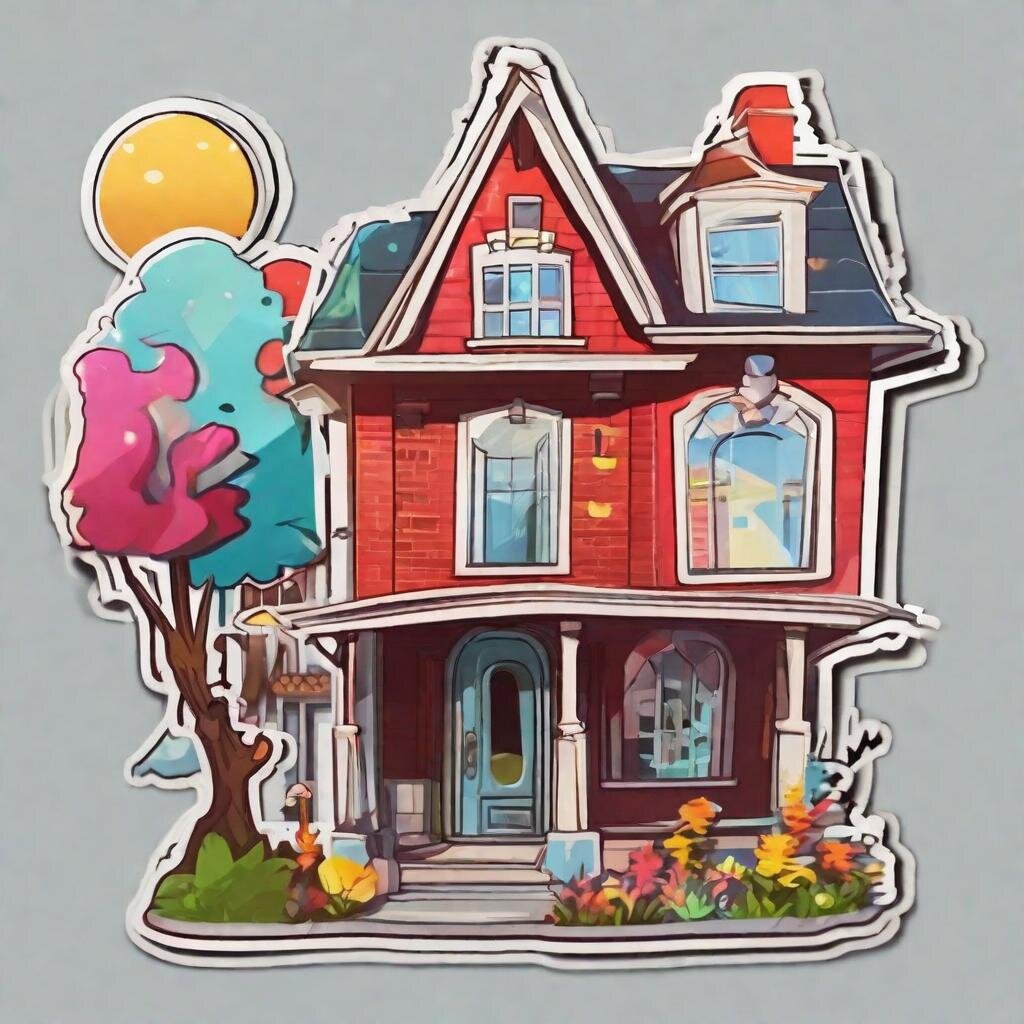}
            \subcaption*{\scriptsize A house}
        \end{minipage} &
        \begin{minipage}{0.11\textwidth}
            \includegraphics[width=\textwidth]{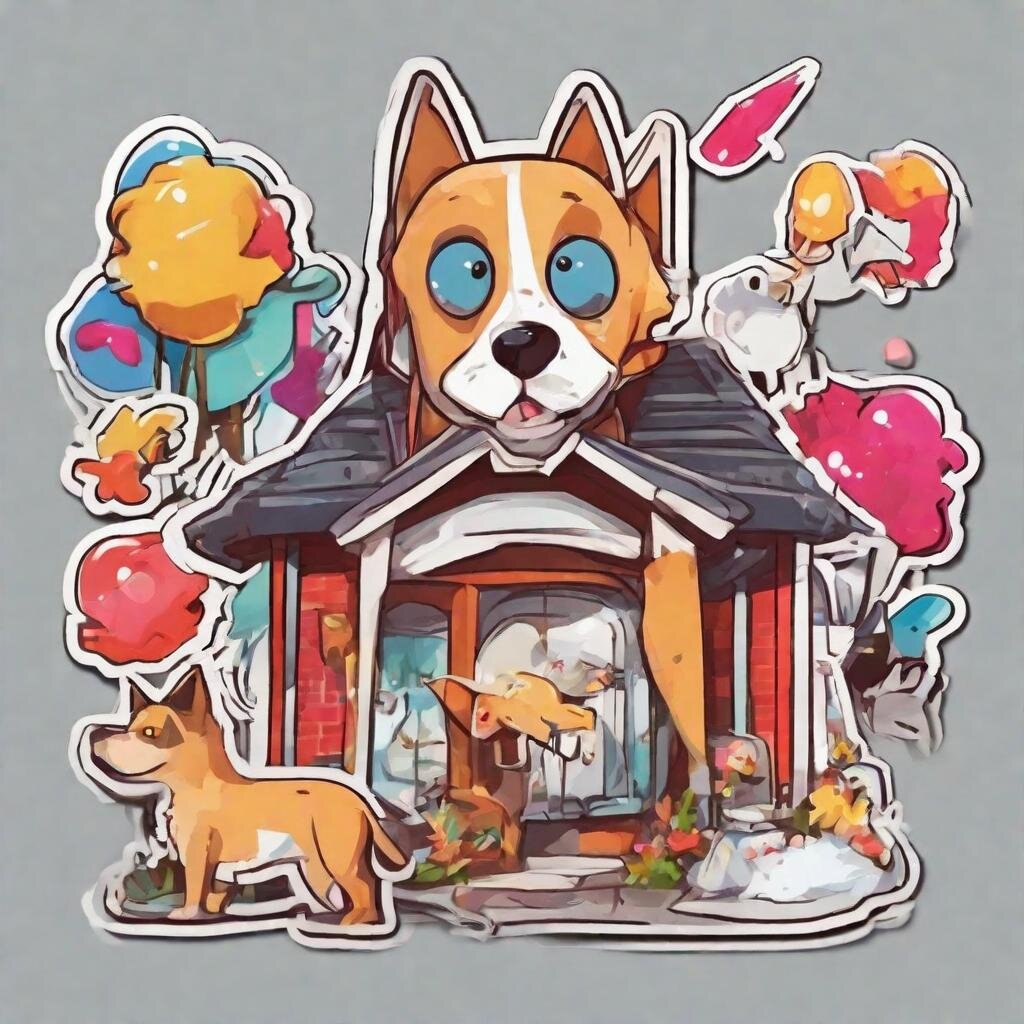}
            \subcaption*{\scriptsize A dog}
        \end{minipage} &
        \begin{minipage}{0.11\textwidth}
            \includegraphics[width=\textwidth]{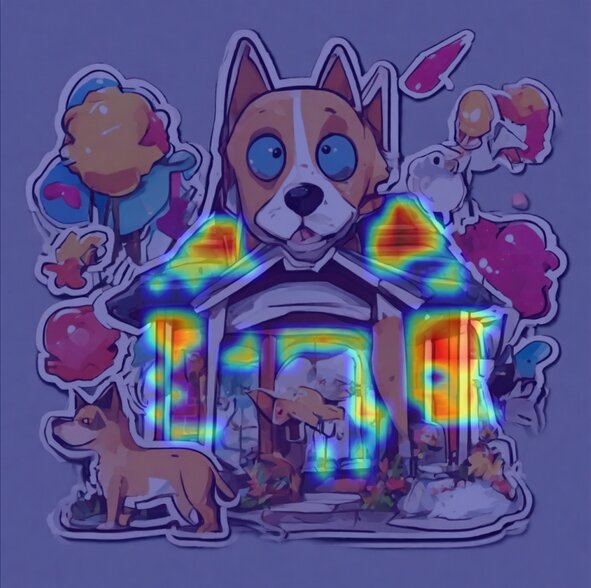}
            \subcaption*{\scriptsize house - dog}
        \end{minipage} &
        \begin{minipage}{0.11\textwidth}
            \includegraphics[width=\textwidth]{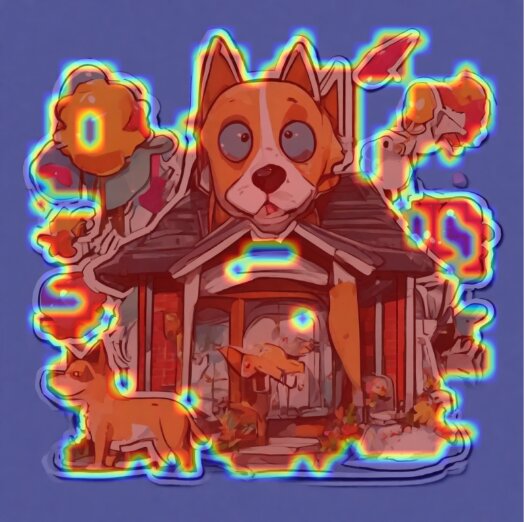}
            \subcaption*{\scriptsize house - dog}
        \end{minipage} \\
        \vspace{4pt}
        \begin{minipage}{0.11\textwidth}
            \includegraphics[width=\textwidth]{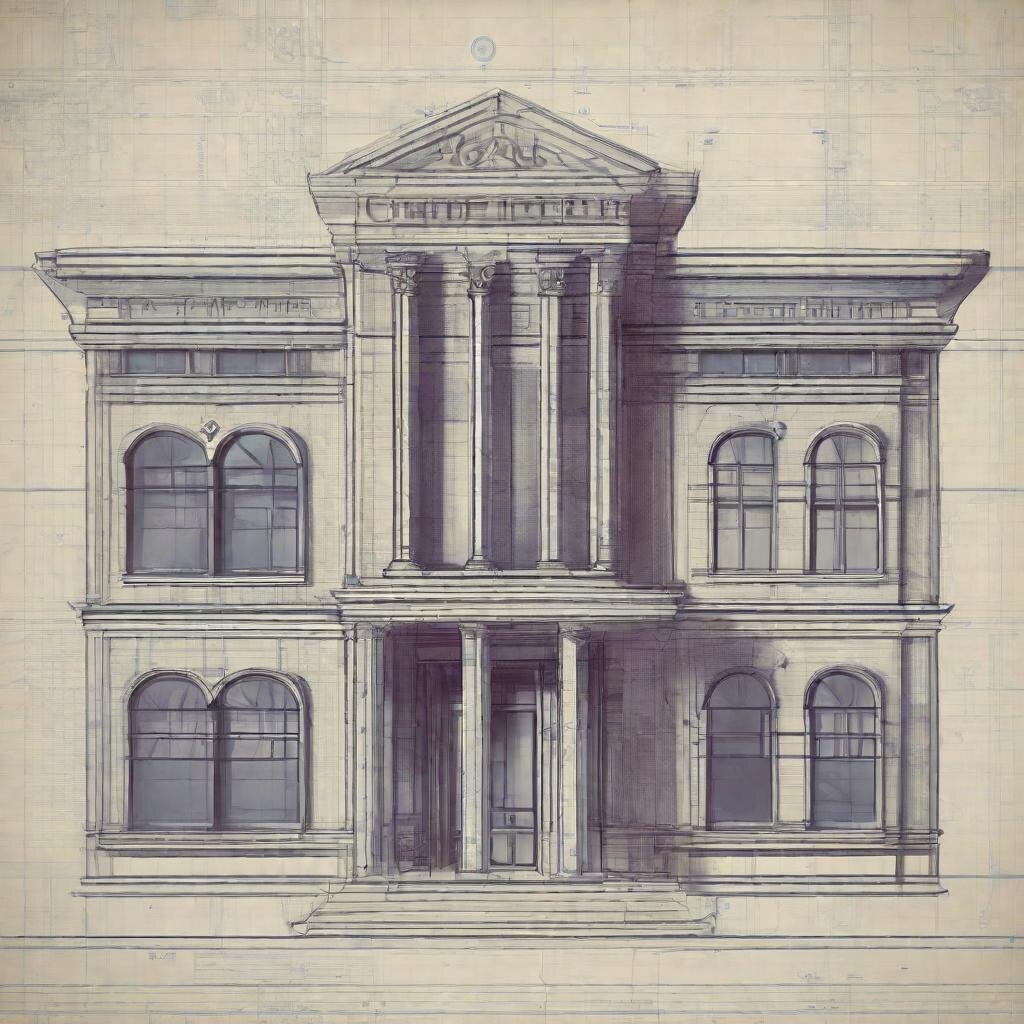}
            \subcaption*{\scriptsize A building}
        \end{minipage} &
        \begin{minipage}{0.11\textwidth}
            \includegraphics[width=\textwidth]{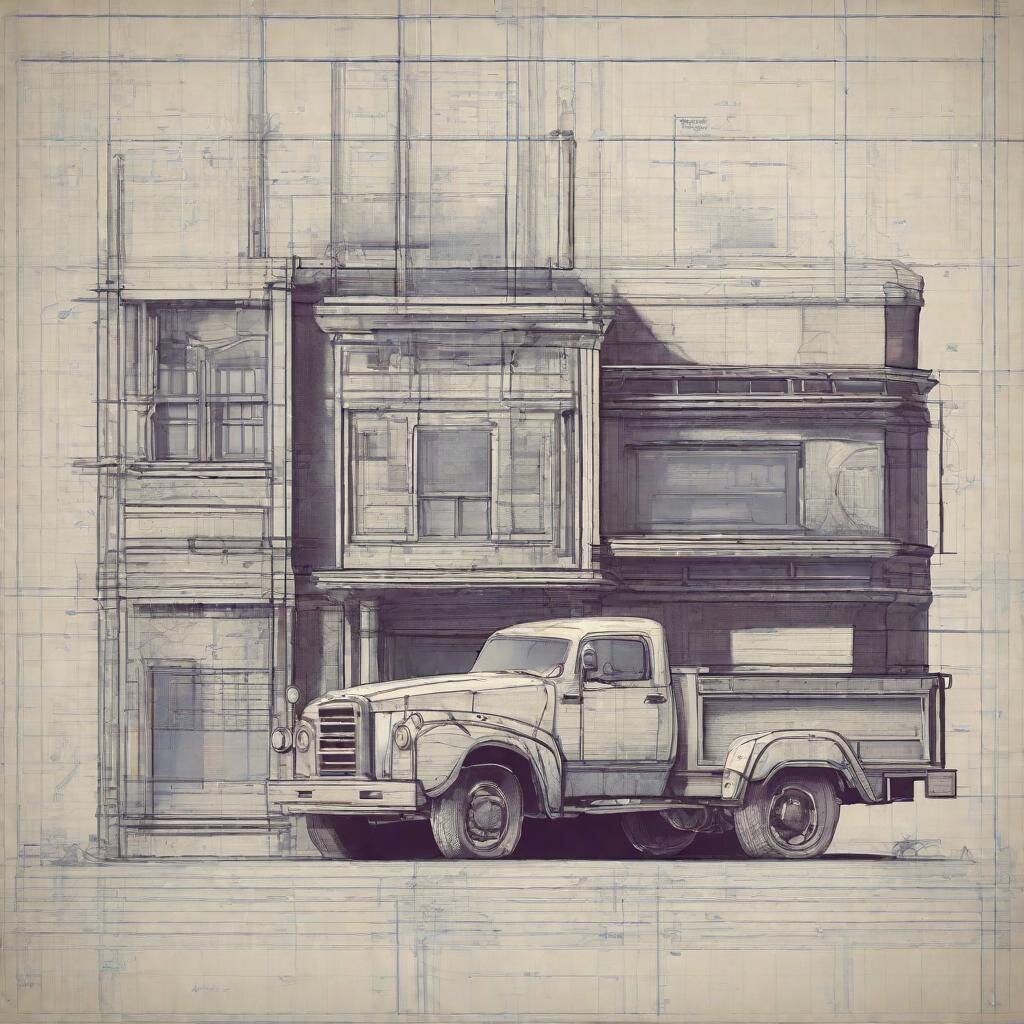}
            \subcaption*{\scriptsize A truck}
        \end{minipage} &
        \begin{minipage}{0.11\textwidth}
            \includegraphics[width=\textwidth]{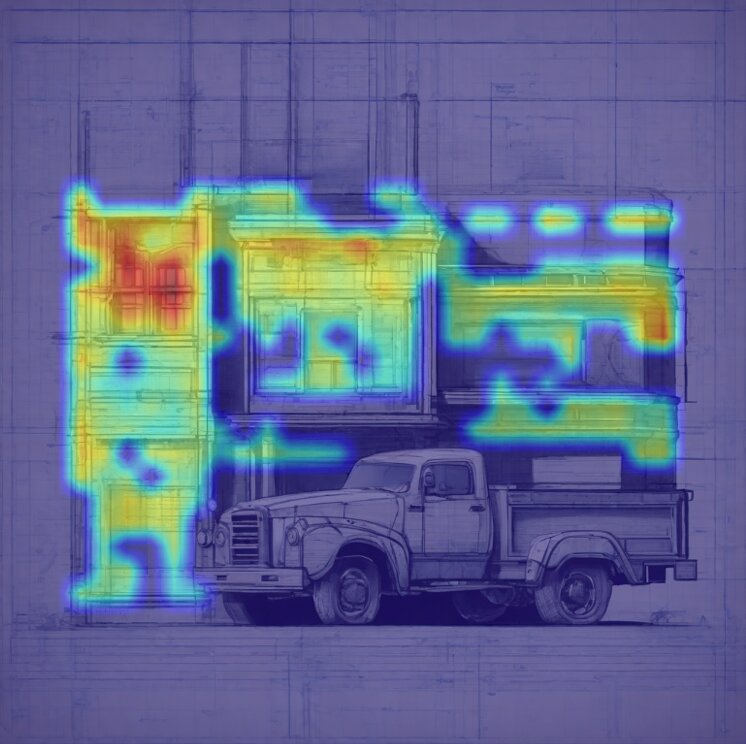}
            \subcaption*{\scriptsize building - truck}
        \end{minipage} &
        \begin{minipage}{0.11\textwidth}
            \includegraphics[width=\textwidth]{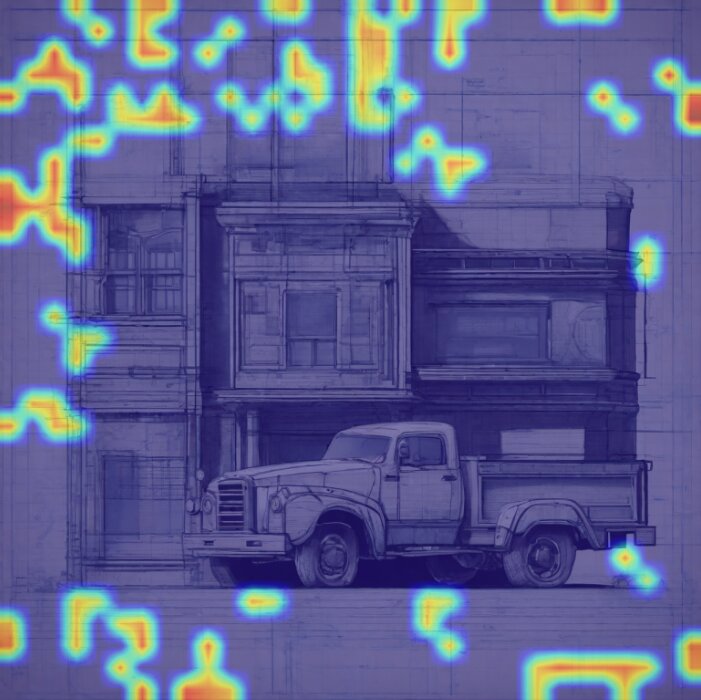}
            \subcaption*{\scriptsize building - truck}
        \end{minipage} \\
    \end{tabular}
    
    \caption{\textbf{CLIP text embeddings vs Subject Representations.} 
    The first two columns are the reference and the target images, while the next two rows visualize the localization difference between the target and the reference, as defined by $\mathbf{L} \odot (\mathbf{C}_{ref} - \mathbf{C}_{tgt})$ (see Sec. 3.4 of the main manuscript). 
    Our content leakage localization method, based on the extraction of subject representations on the image feature space, faithfully localizes the content leakage, if exists. 
    On the contrary, the cross attention scores between image features and textual CLIP features of the subject token, even though semantically explainable, are not a trustworthy metric to perform this localization.}
    \label{fig:Ablation_4}
\end{figure}

\textbf{CLIP text embeddings vs Subject Representations.} 
As discussed in Section 3.4 of the main manuscript, we use a patch-level localization method during inference to annotate reference subject features in the target image, which indicates content leakage. Given the semantic nature of this problem, a natural starting point is to explore the layers responsible for determining the semantics in text-to-image (T2I) generation. 
These semantics are primarily guided by the cross-attention layers. Thus, an intuitive initial experiment involves performing the cross-attention mechanism between the target image features and the reference subject's textual description, identifying dominant cross-attention values in the aggregated subject map. 
Patches in the target image that exhibit content leakage are then defined as those that ``attend" significantly more to the reference subject token than to the target subject token. 
However, the CLIP token embeddings used in the cross-attention mechanism are not always sufficiently expressive to localize subtle visual features of the reference subject, especially when these features overlap with those of the target subject in the generated image.

To showcase this limitation and motivate our subject representation extraction, we perform the localization using the cross-attention values, as we described above, and visually illustrate the results in Figure \ref{fig:Ablation_4}. 
It becomes clear that while this approach can work in cases that the leakage is semantically evident or the CLIP representations of the subjects are expressive enough to distinguish the reference from the target one (e.g., top row of Fig. \ref{fig:Ablation_4}), it fails to systematically localize the subtle content leakage features in the target image. 
This is because the visual representation features of our approach are by definition more descriptive of the per-case generated image and can accurately detect patches that are correlated with either the reference or the target subject.
On the contrary, the textual CLIP features used in the cross-attention mechanism are limited to a more general semantic representation of the subject that can be hurtful in the context of accurate leakage detection.

\subsection{Time requirements of state-of-the-art methods}

In Table~\ref{tab:time_compl}, we present the requirements in terms of time for the state-of-the-art style alignment methods evaluated. The reported time reflects the duration each method requires to generate a stylistically aligned set of two images. For optimization-based methods like B-LoRA~\cite{frenkel2024blora}, StyleDrop (SDRP)~\cite{sohn2023styledrop} DB-LoRA~\cite{ruiz2023dreamboothfinetuningtexttoimage}, we account for both the fine-tuning process on the reference image and the final inference to produce the stylistically aligned target. StyleAligned~\cite{hertz2024style} generates a batch of two images, using the first as the reference and the second as the target.

Adapter-based methods, such as IP-Adapter~\cite{Ye2023IPAdapter}, CSGO~\cite{Xing2024CSGO}, and InstantStyle~\cite{Wang2024InstantStyle}, operate by encoding the reference image and subsequently generating the target while integrating the reference information through cross-attention layers. However, these methods necessitate large-scale training to effectively enable this image conditioning within a diffusion model.

For our method, we first infer only the reference image to detect the reference subject, followed by a binary search to determine the optimal scaling factor \(\alpha\), which results in the final stylistic alignment. 
As discussed in the main manuscript, we set a binary search precision of \(p = 0.03125\), requiring the generation process to be repeated five times. 
All methods are implemented on top of the SDXL~\cite{podell2023sdxlimprovinglatentdiffusion} framework and evaluated in a NVIDIA GeForce RTX 3090.

\begin{table}[t]
\scriptsize
\centering
\begin{tabular}{|l|c|c|c|}
\hline
\textbf{Method} & \textbf{Pretraining} & \textbf{Opt.} & \textbf{Time Requirement} \\ \hline
IP-Adapter       & \checkmark & \ding{55} & 0 min 14 sec \\ \hline
InstantStyle     & \checkmark & \ding{55} & 0 min 16 sec \\ \hline
CSGO           & \checkmark & \ding{55} & 0 min 20 sec \\ \hline
B-LoRA          & \ding{55}  & \checkmark & 11 min 13 sec \\ \hline
DreamBooth-LoRA & \ding{55}  & \checkmark & 8 min 42 sec \\ \hline
SDRP           & \ding{55}  & \checkmark & 13 min 09 sec \\ \hline
StyleAligned   & \ding{55}  & \ding{55}  & 0 min 29 sec \\ \hline
\emph{Only-Style}& \ding{55}  & \ding{55}  & 1 min 46 sec \\ \hline
\end{tabular}
\caption{\textbf{Time requirements of different Style Consistent Generation Methods}. We report for each method the time required to generate a stylistically aligned set of two images on an NVIDIA RTX 3090. All methods are implemented on top of SDXL. ``Pretraining" denotes methods that use large scale training to incorporate image conditioning. ``Opt." denotes methods that require per instance optimization to capture a style.}
\label{tab:time_compl}
\end{table}

\begin{figure}[t]
    \centering
        \begin{tabular}{c@{\hspace{.1cm}}c@{\hspace{.1cm}}c@{\hspace{.1cm}}c}  
            \scriptsize Reference & \scriptsize & \scriptsize & \scriptsize \\
            \begin{minipage}{0.11\textwidth}
                \includegraphics[width=\textwidth]{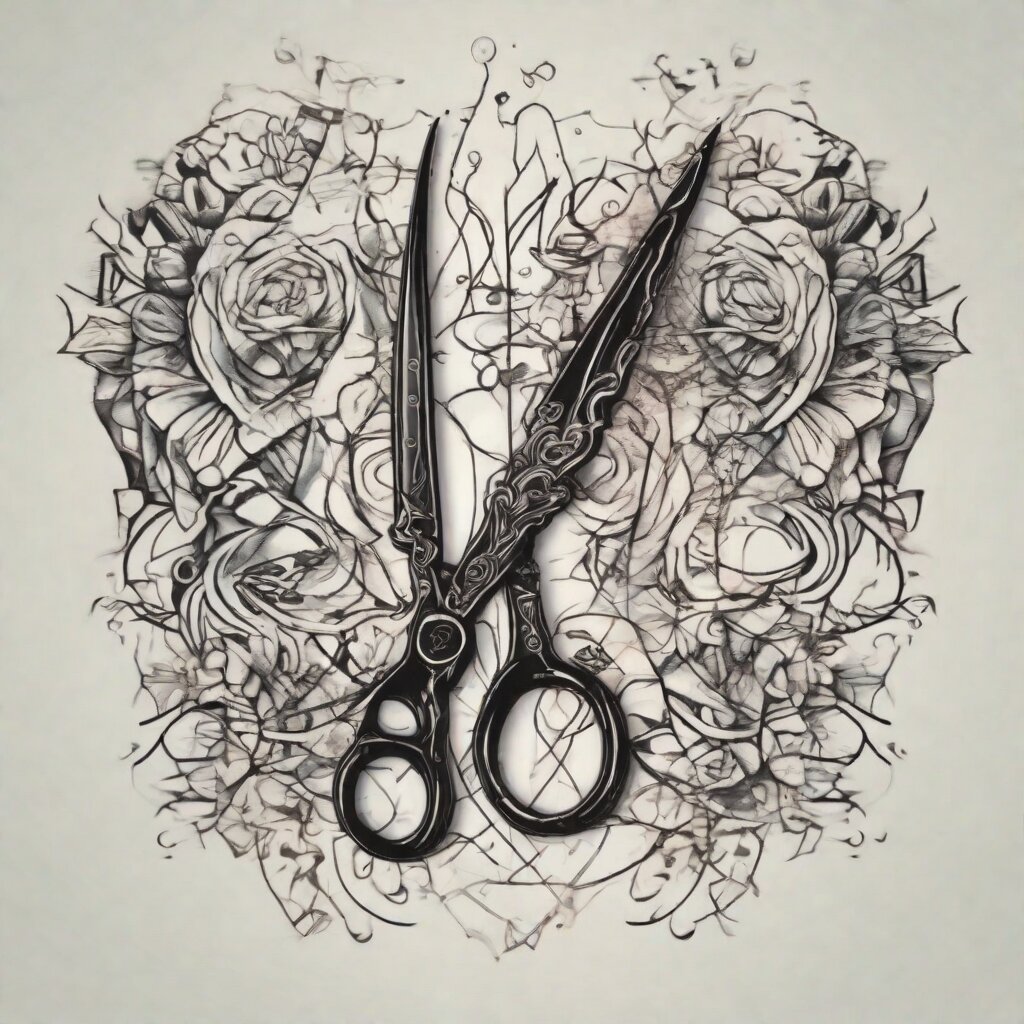} 
            \end{minipage} &
            \begin{minipage}{0.11\textwidth}
                \scriptsize \centering Q1: Are there any \textit{scissors} visual features in this \textit{face} image?
            \end{minipage}  &
            \begin{minipage}{0.11\textwidth}
                \includegraphics[width=\textwidth]{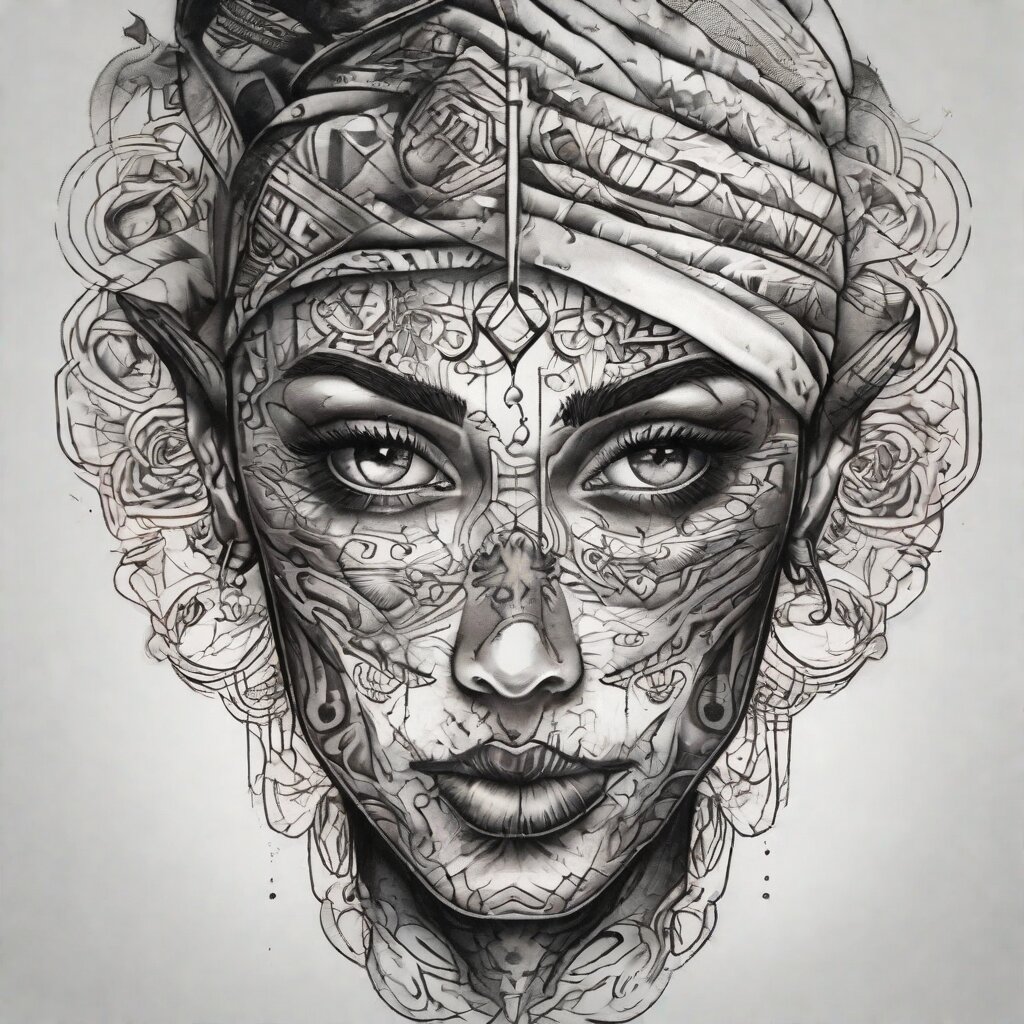} 
            \end{minipage} &
            \begin{minipage}{0.11\textwidth}
                \scriptsize \centering A: Yes, there are scissors visible in the image as part of the face's design.
            \end{minipage} 
        \end{tabular}

        \begin{tabular}{c@{\hspace{.1cm}}c@{\hspace{.1cm}}c@{\hspace{.1cm}}c}  
            \begin{minipage}{0.11\textwidth}
                \includegraphics[width=\textwidth]{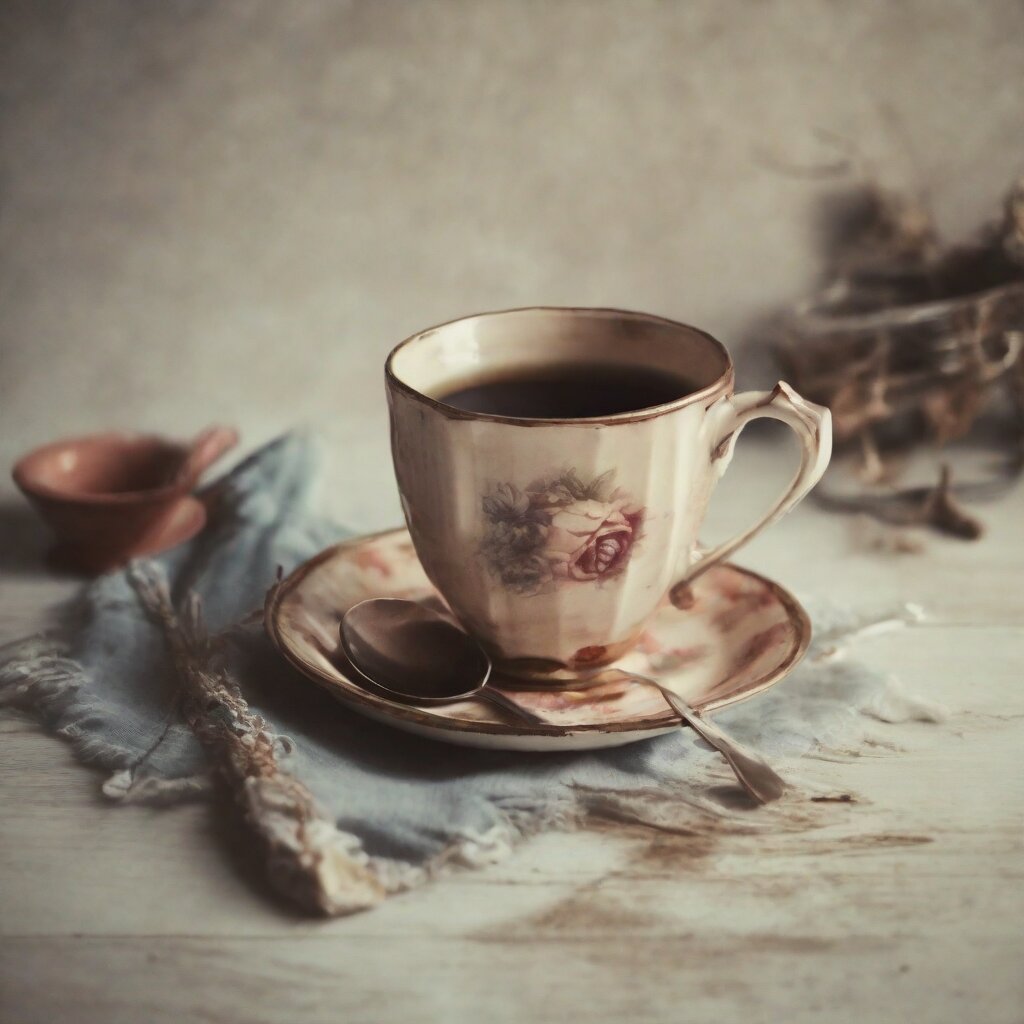} 
            \end{minipage} &
            \begin{minipage}{0.11\textwidth}
                \scriptsize \centering Q1: Are there any \textit{cup} visual features in this \textit{plate} image?
            \end{minipage}  &
            \begin{minipage}{0.11\textwidth}
                \includegraphics[width=\textwidth]{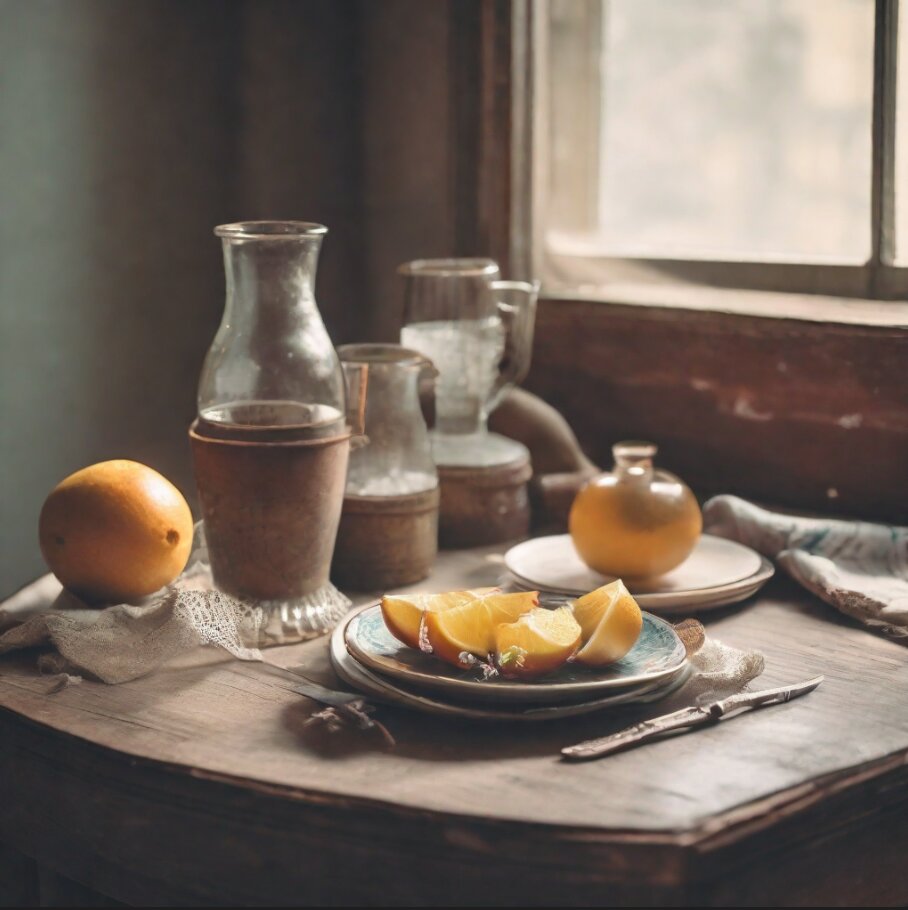} 
            \end{minipage} &
            \begin{minipage}{0.11\textwidth}
                \scriptsize \centering A: Yes, there are visual features of cups in the image.
            \end{minipage} 
        \end{tabular}

        \begin{tabular}{c@{\hspace{.1cm}}c@{\hspace{.1cm}}c@{\hspace{.1cm}}c}     
            \begin{minipage}{0.11\textwidth}
                \includegraphics[width=\textwidth]{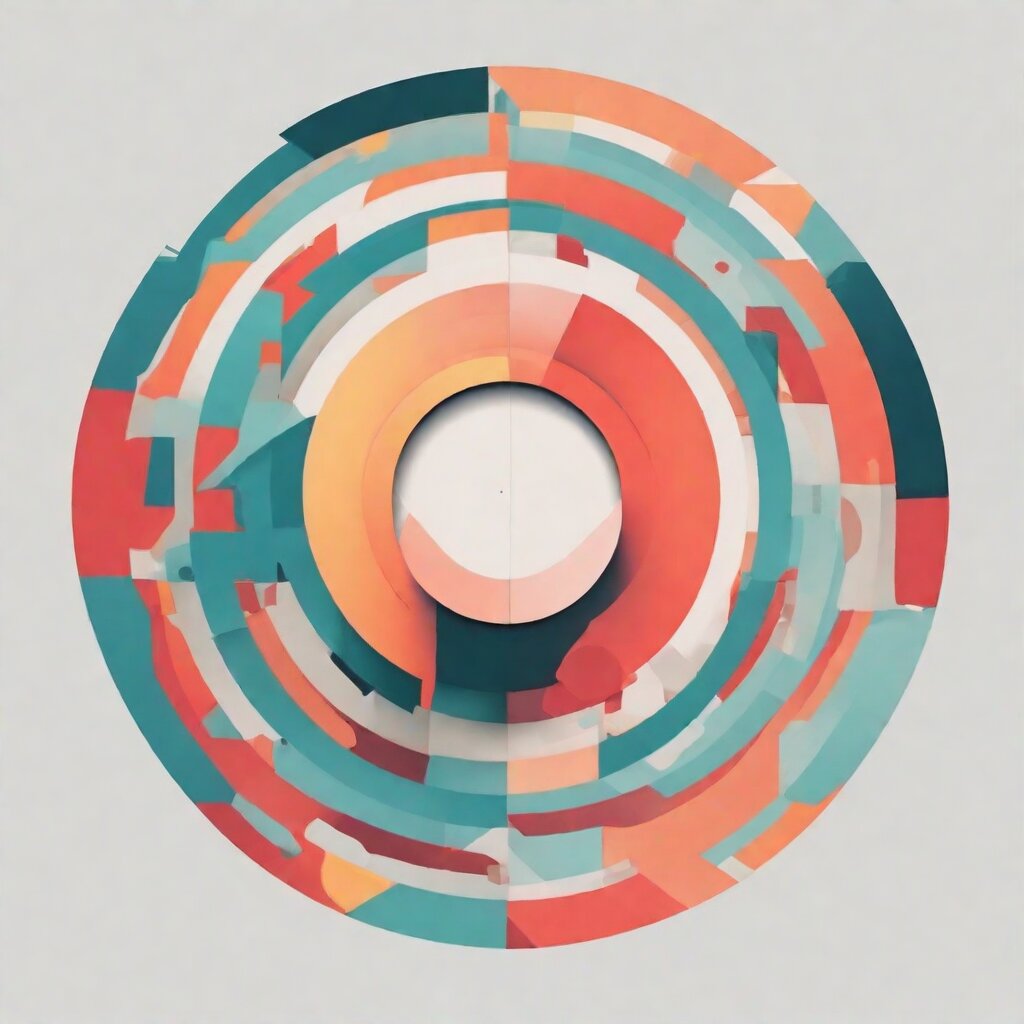} 
            \end{minipage} &
            \begin{minipage}{0.11\textwidth}
                \scriptsize \centering Q3: Is there a \textit{circle} in the image?
            \end{minipage}  &
            \begin{minipage}{0.11\textwidth}
                \includegraphics[width=\textwidth]{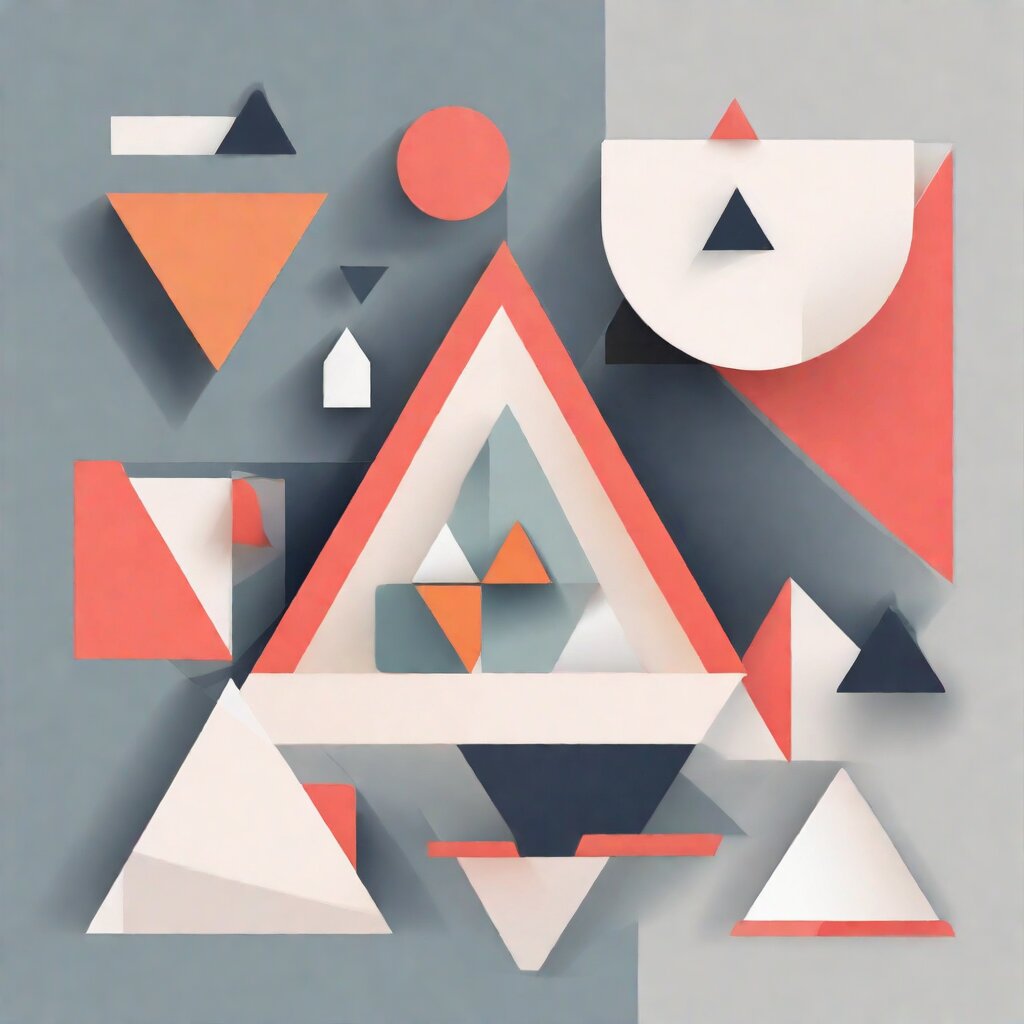} 
            \end{minipage} &
            \begin{minipage}{0.11\textwidth}
                \scriptsize \centering A: Yes, there is a circle in the image.
            \end{minipage} 
        \end{tabular}   
    \caption{\textbf{Failure cases of the LVLM evaluation framework.} The target image in the first row is generated with \emph{Only-Style} while the images in the bottom two rows are standard text-to-image generations. The subject of each image can be inferred from the respective questions.}
    \label{fig:LVLM_fail}
\end{figure}

\subsection{LVLM-based Evaluation Protocol}

\begin{figure*}[t]
    \centering

        \begin{tabular}{c@{\hspace{.1cm}}c@{\hspace{.1cm}}c@{\hspace{.1cm}}c@{\hspace{.6cm}}c@{\hspace{.1cm}}c}  
            \scriptsize  & \scriptsize Reference & \scriptsize StyleAligned~\cite{hertz2024style} & \scriptsize Answer & \scriptsize \emph{Only-Style} & \scriptsize Answer \\

            \vspace{0.1cm}
            
            \begin{minipage}{0.11\textwidth}
                \scriptsize \centering Q1: Are there any \textit{rollercoaster} visual features in this \textit{carousel} image?
            \end{minipage}  &
            \begin{minipage}{0.14\textwidth}
                \includegraphics[width=\textwidth]{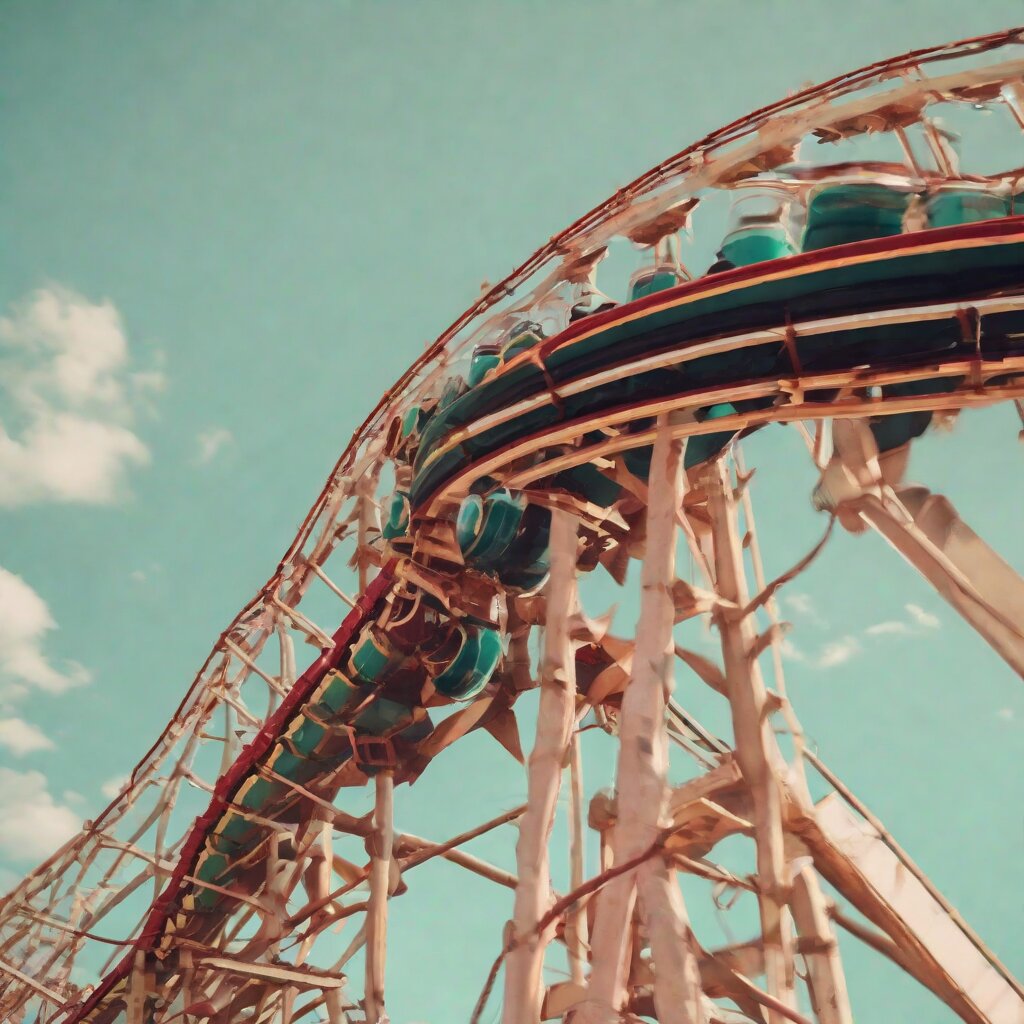} 
            \end{minipage} &
            \begin{minipage}{0.14\textwidth}
                \includegraphics[width=\textwidth]{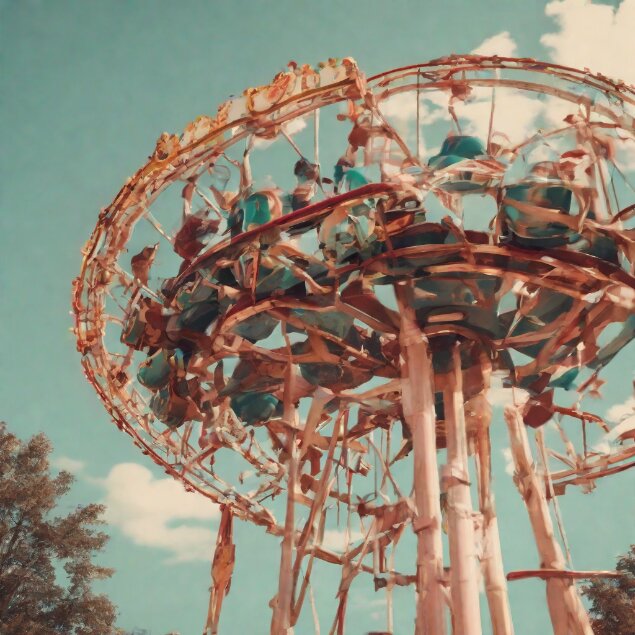}
            \end{minipage} &
            \begin{minipage}{0.2\textwidth}
                \scriptsize \centering Yes, the image shows a carousel with a rollercoaster-like structure, featuring a series of loops and turns that resemble the track of a rollercoaster.
            \end{minipage} &
            \begin{minipage}{0.14\textwidth}
                \includegraphics[width=\textwidth]{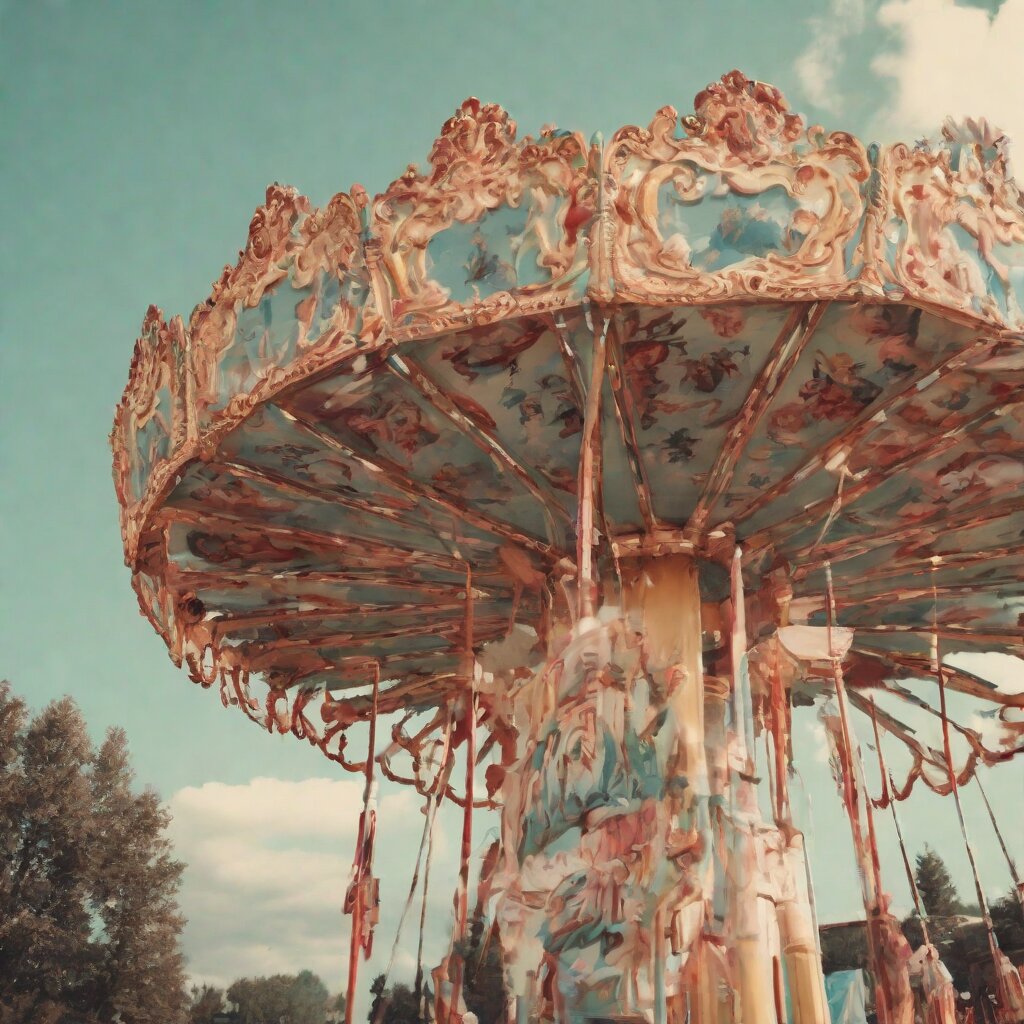}
            \end{minipage} &
            \begin{minipage}{0.2\textwidth}
                \scriptsize \centering No, there is no rollercoaster in the image.
            \end{minipage} \\

            \vspace{0.1cm}
            
            \begin{minipage}{0.11\textwidth}
                \scriptsize \centering Q1: Are there any \textit{rocket} visual features in this \textit{robot} image?
            \end{minipage}  &
            \begin{minipage}{0.14\textwidth}
                \includegraphics[width=\textwidth]{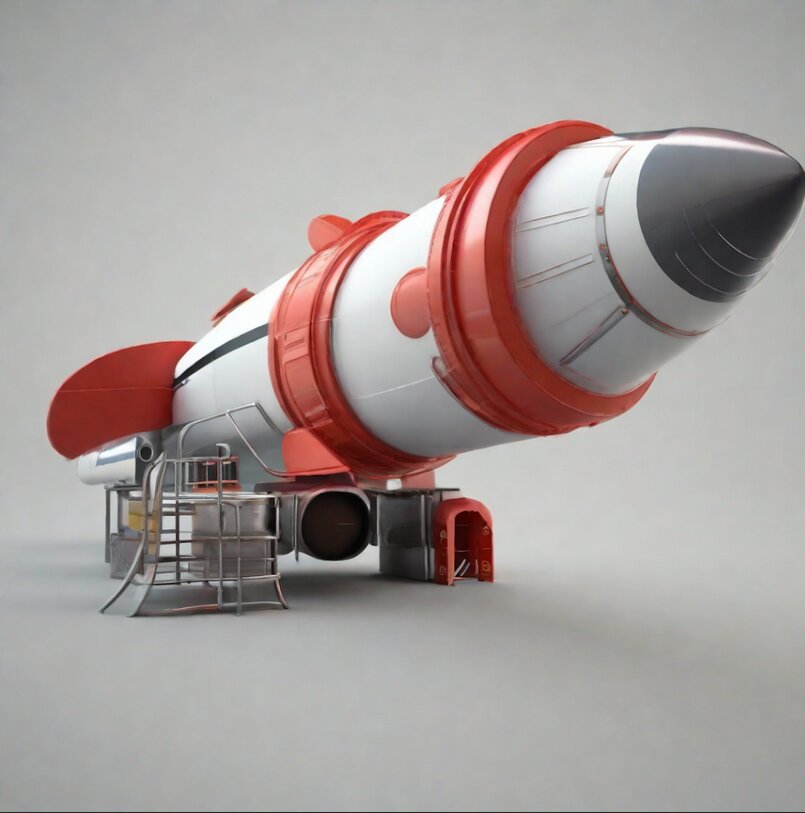} 
            \end{minipage} &
            \begin{minipage}{0.14\textwidth}
                \includegraphics[width=\textwidth]{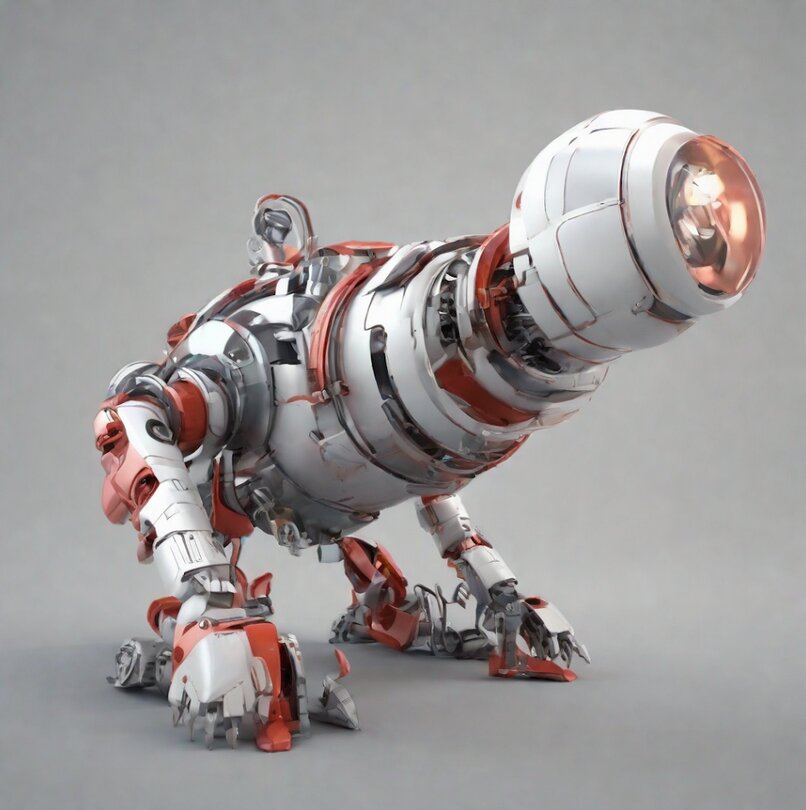}
            \end{minipage} &
            \begin{minipage}{0.2\textwidth}
                \scriptsize \centering Yes, the robot in the image has a prominent cylindrical body with a red and white color scheme that resembles a rocket.
            \end{minipage} &
            \begin{minipage}{0.14\textwidth}
                \includegraphics[width=\textwidth]{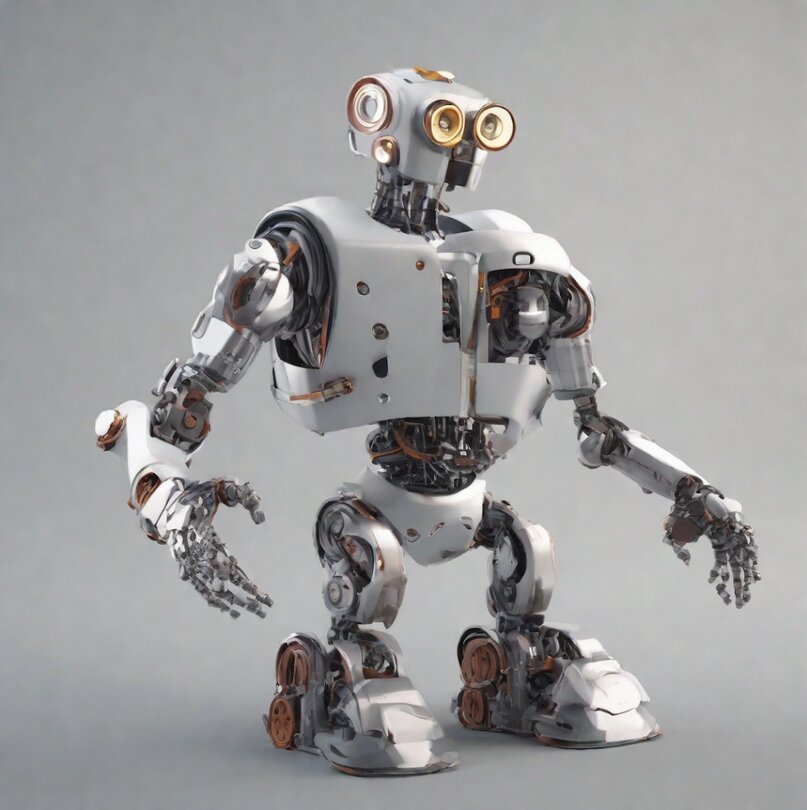}
            \end{minipage} &
            \begin{minipage}{0.2\textwidth}
                \scriptsize \centering No, there are no rocket visual features in this robot image. The robot appears to be a humanoid figure with a white and metallic color scheme ... It does not have any visible rockets or other space-related elements.
            \end{minipage} \\

            \vspace{0.1cm}
            
            \begin{minipage}{0.11\textwidth}
                \scriptsize \centering Q1: Are there any \textit{dragon} visual features in this \textit{skateboard} image?
            \end{minipage}  &
            \begin{minipage}{0.14\textwidth}
                \includegraphics[width=\textwidth]{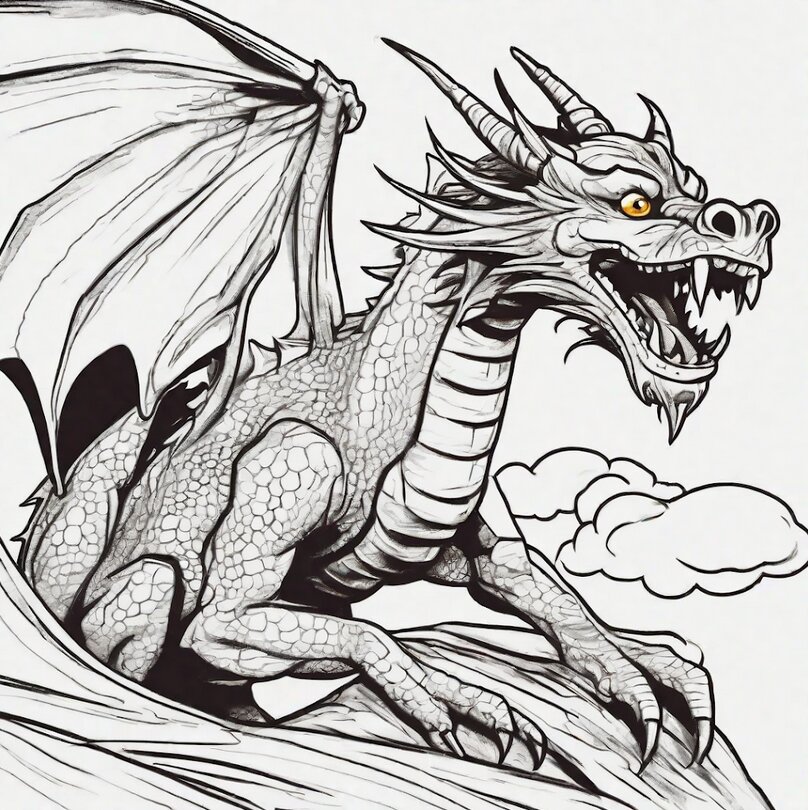} 
            \end{minipage} &
            \begin{minipage}{0.14\textwidth}
                \includegraphics[width=\textwidth]{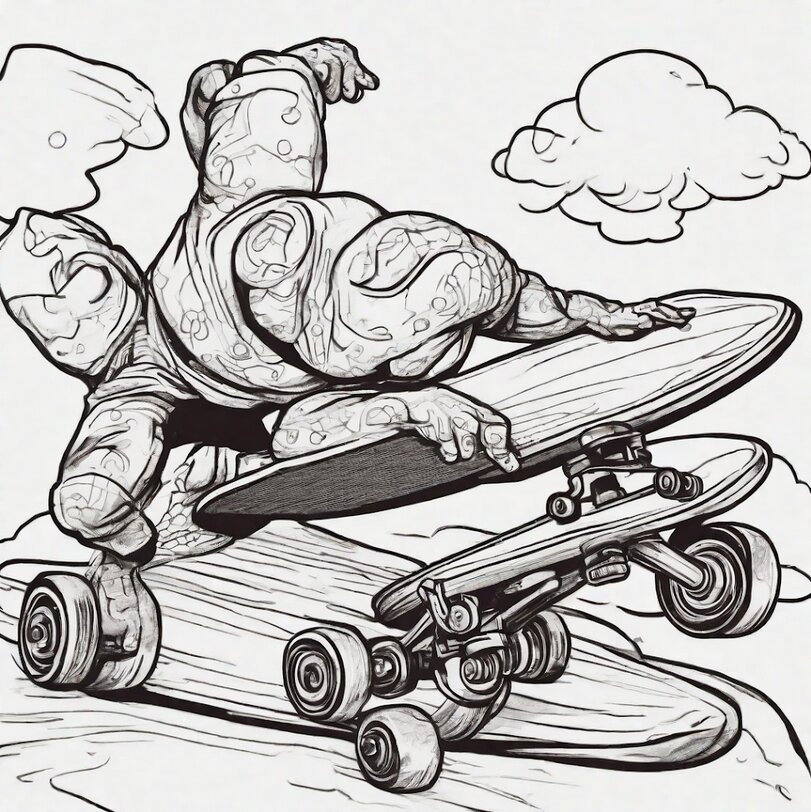}
            \end{minipage} &
            \begin{minipage}{0.2\textwidth}
                \scriptsize \centering Yes, the skateboarder in the image has dragon-like features such as wings and a tail.
            \end{minipage} &
            \begin{minipage}{0.14\textwidth}
                \includegraphics[width=\textwidth]{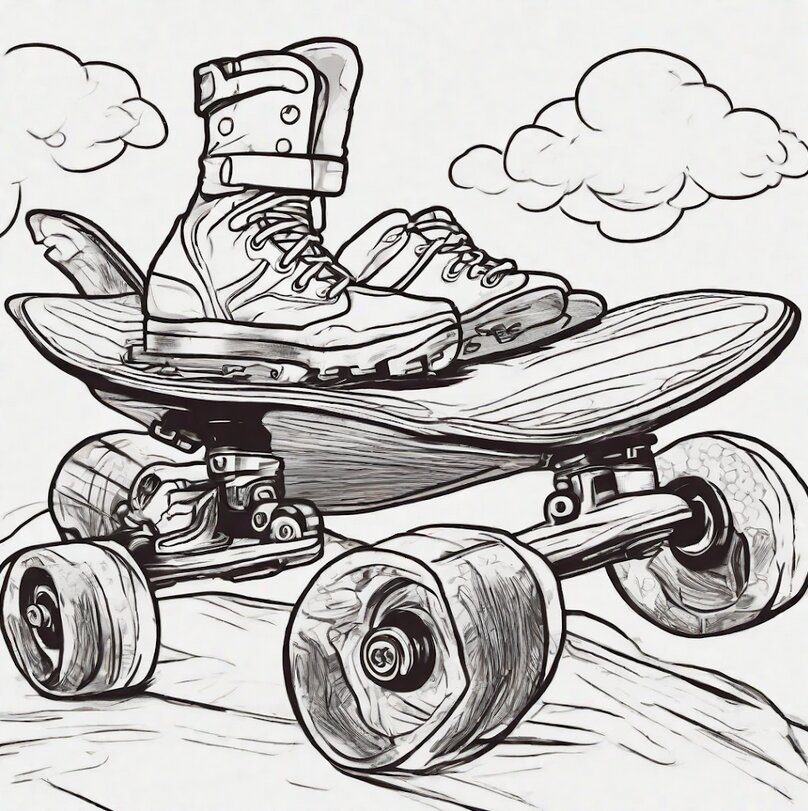}
            \end{minipage} &
            \begin{minipage}{0.2\textwidth}
                \scriptsize \centering No, there are no dragon visual features in this skateboard image. The image is a black and white line drawing of a skateboard with two boots placed on top of it, against a background of clouds and the sky.
            \end{minipage} \\

            \vspace{0.1cm}
            
            \begin{minipage}{0.11\textwidth}
                \scriptsize \centering Q1: Are there any \textit{chair} visual features in this \textit{couch} image?
            \end{minipage}  &
            \begin{minipage}{0.14\textwidth}
                \includegraphics[width=\textwidth]{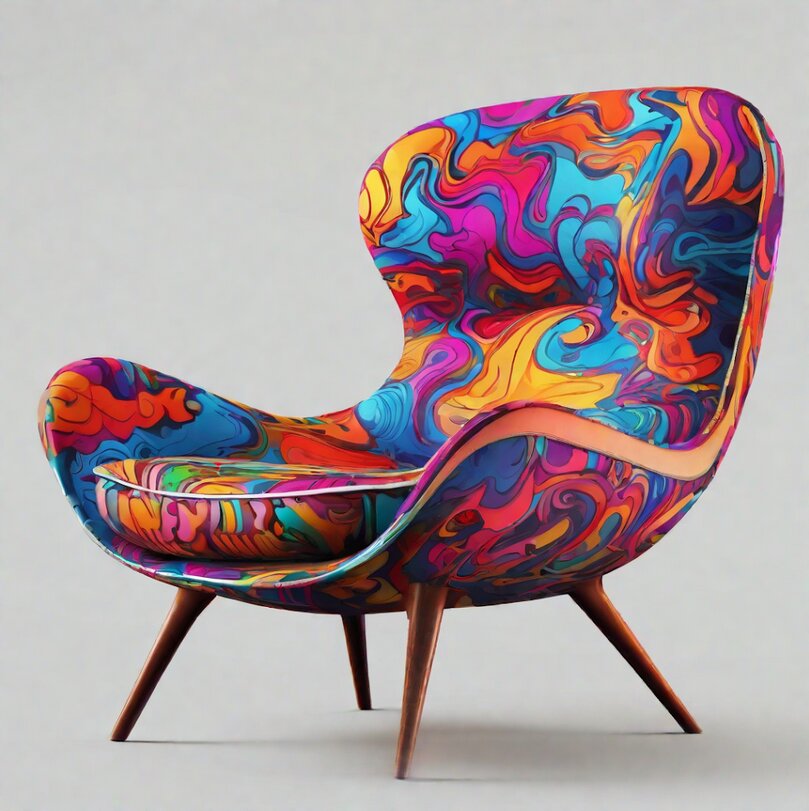} 
            \end{minipage} &
            \begin{minipage}{0.14\textwidth}
                \includegraphics[width=\textwidth]{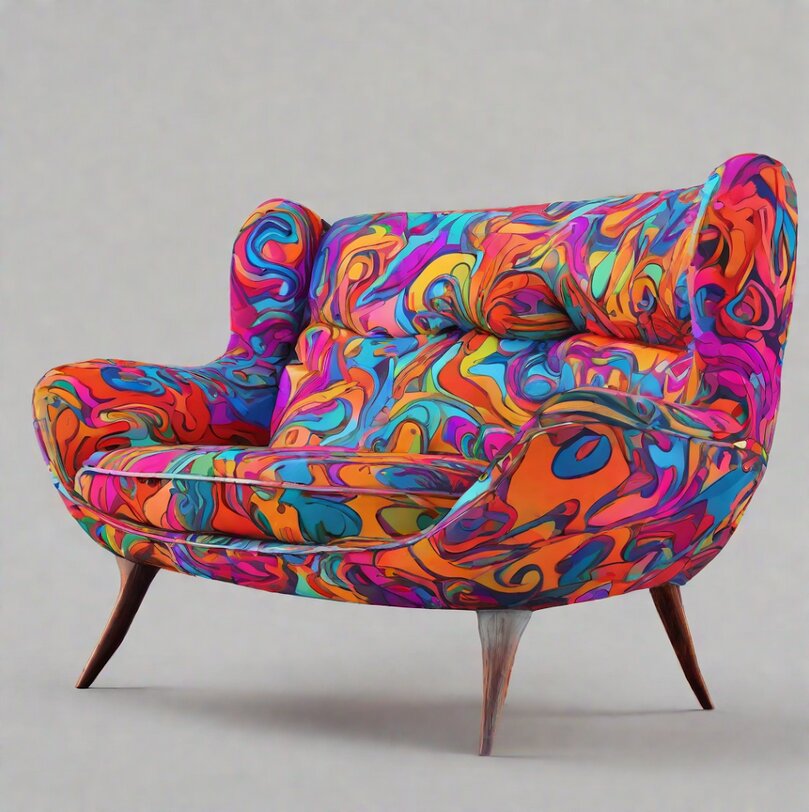}
            \end{minipage} &
            \begin{minipage}{0.2\textwidth}
                \scriptsize \centering Yes, the couch in the image has armrests and legs, which are typical features of a chair.
            \end{minipage} &
            \begin{minipage}{0.14\textwidth}
                \includegraphics[width=\textwidth]{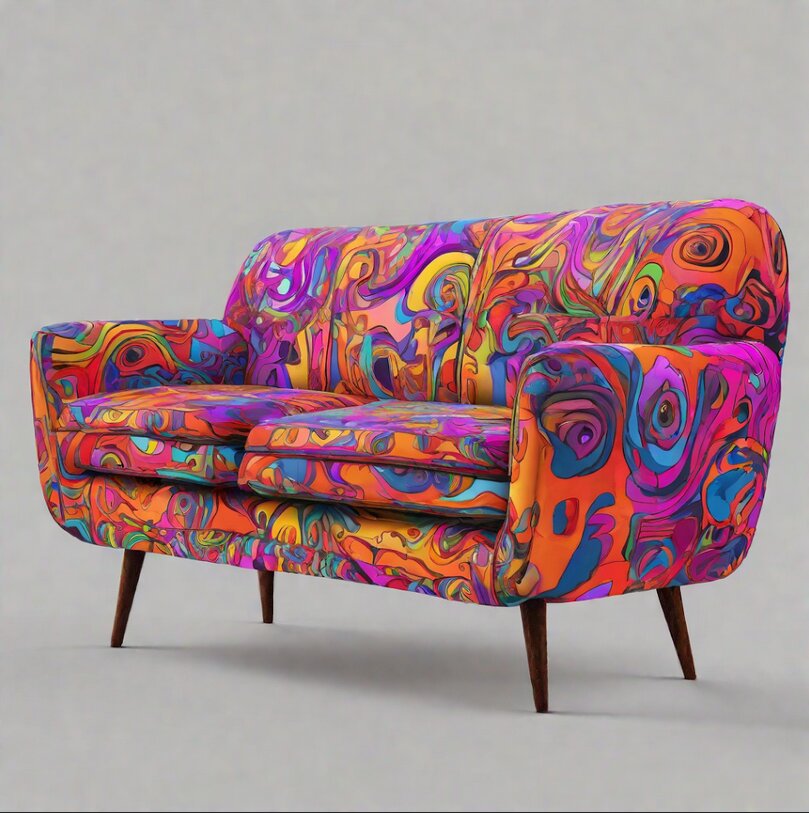}
            \end{minipage} &
            \begin{minipage}{0.2\textwidth}
                \scriptsize \centering No, there are no chair visual features in this couch image. The image shows a sofa with a vibrant, abstract pattern on its upholstery, but it does not depict any chairs.
            \end{minipage} \\

            \vspace{0.1cm}
            
            \begin{minipage}{0.11\textwidth}
                \scriptsize \centering Q2: Is there any \textit{cube} in this image?
            \end{minipage}  &
            \begin{minipage}{0.14\textwidth}
                \includegraphics[width=\textwidth]{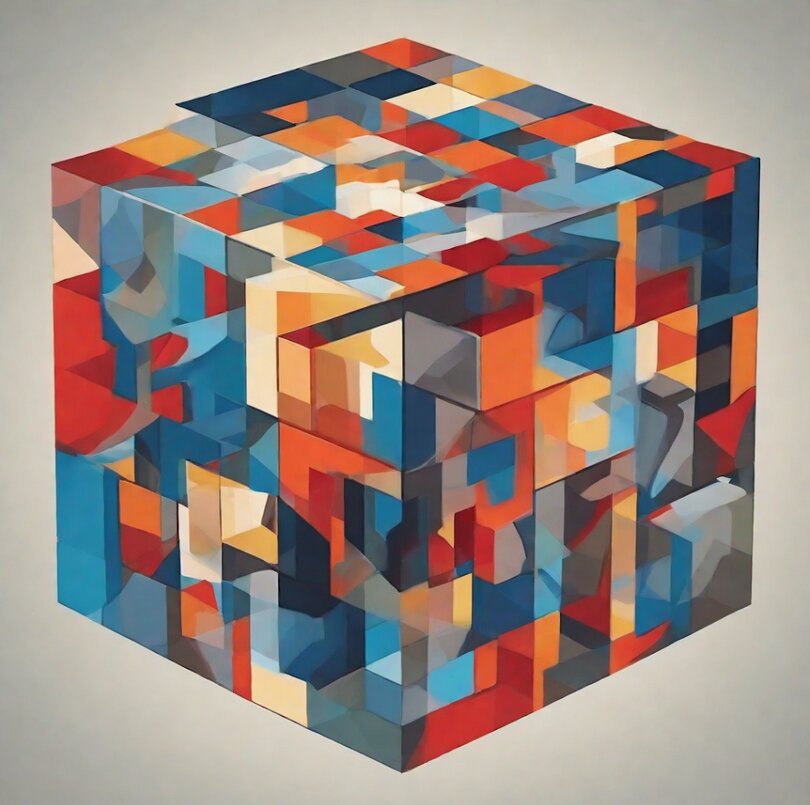} 
            \end{minipage} &
            \begin{minipage}{0.14\textwidth}
                \includegraphics[width=\textwidth]{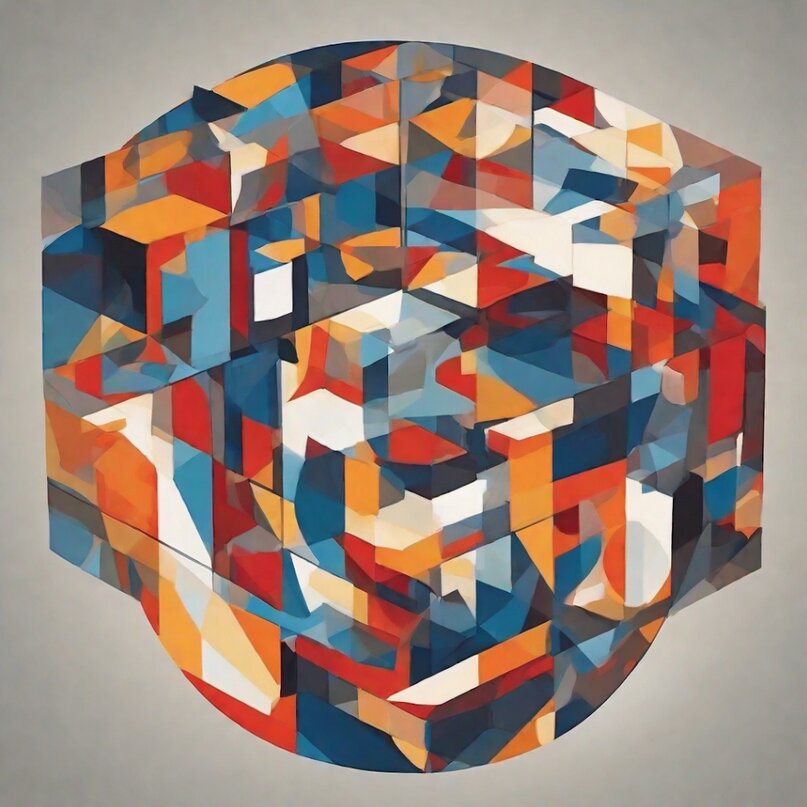}
            \end{minipage} &
            \begin{minipage}{0.2\textwidth}
                \scriptsize \centering Yes, there is a cube in this image. It is a central element of the composition and appears to be composed of various geometric shapes and colors.
            \end{minipage} &
            \begin{minipage}{0.14\textwidth}
                \includegraphics[width=\textwidth]{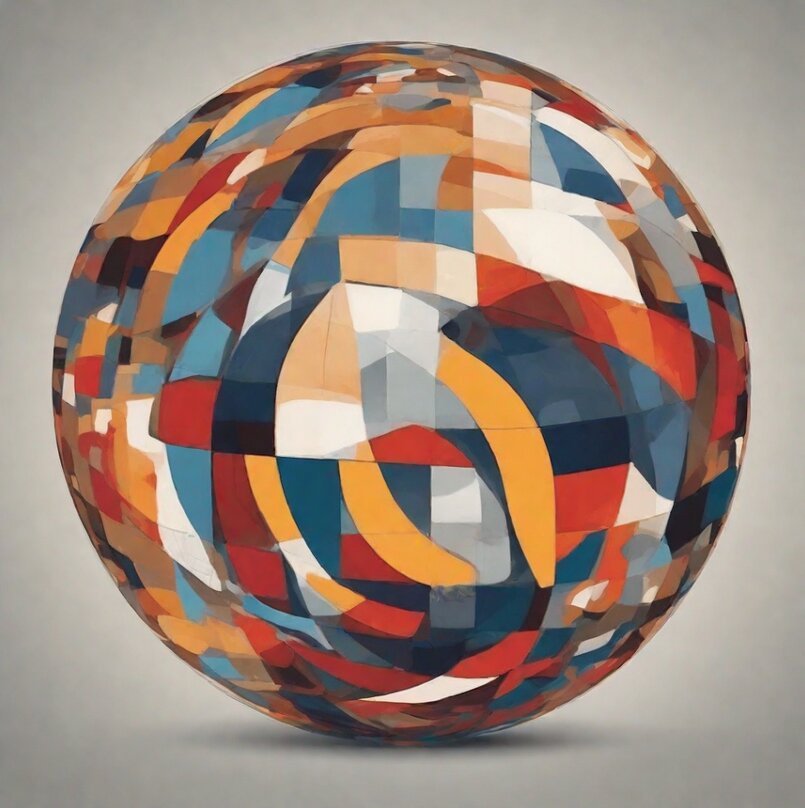}
            \end{minipage} &
            \begin{minipage}{0.2\textwidth}
                \scriptsize \centering No, there is no cube in the image. The image features a colorful, abstract sphere with multiple geometric shapes and patterns.
            \end{minipage} \\

            \vspace{0.1cm}
            
            \begin{minipage}{0.11\textwidth}
                \scriptsize \centering Q2: Is there any \textit{car} in this image?
            \end{minipage}  &
            \begin{minipage}{0.14\textwidth}
                \includegraphics[width=\textwidth]{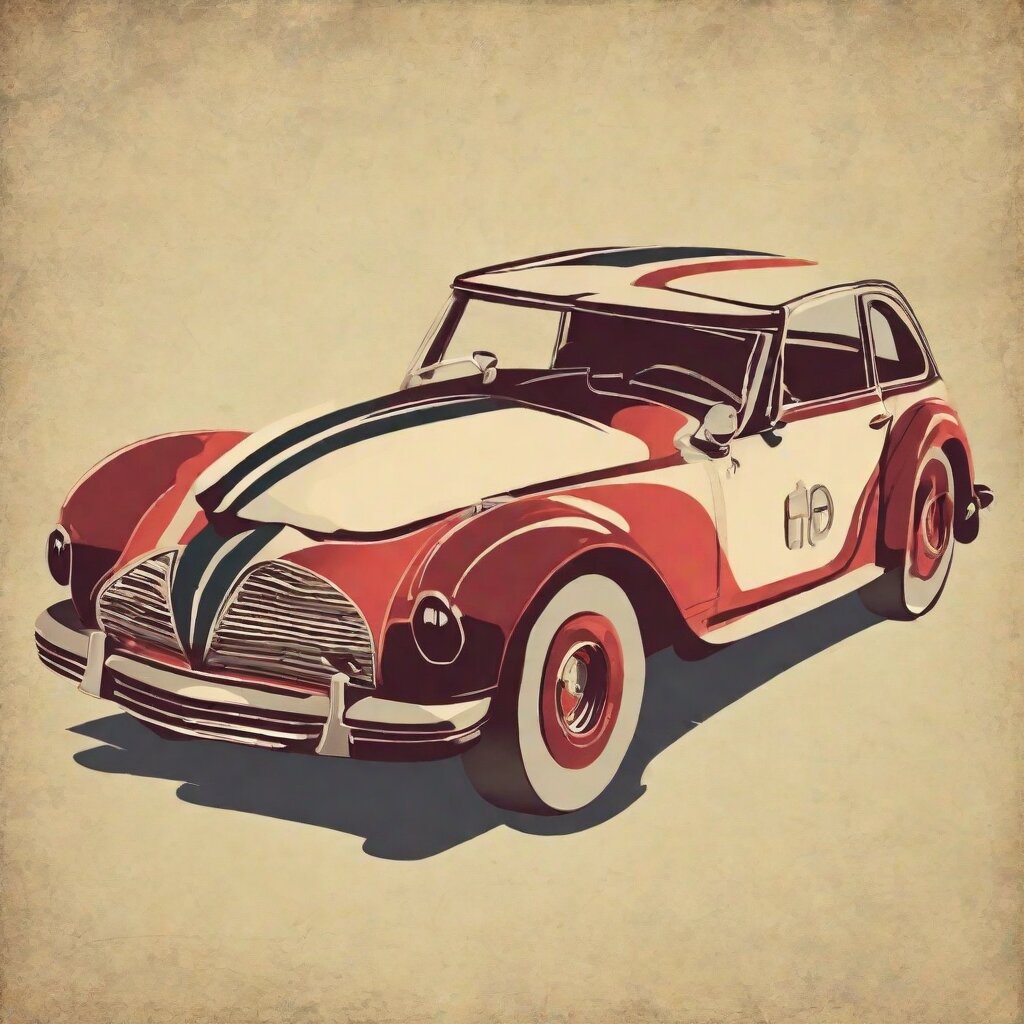} 
            \end{minipage} &
            \begin{minipage}{0.14\textwidth}
                \includegraphics[width=\textwidth]{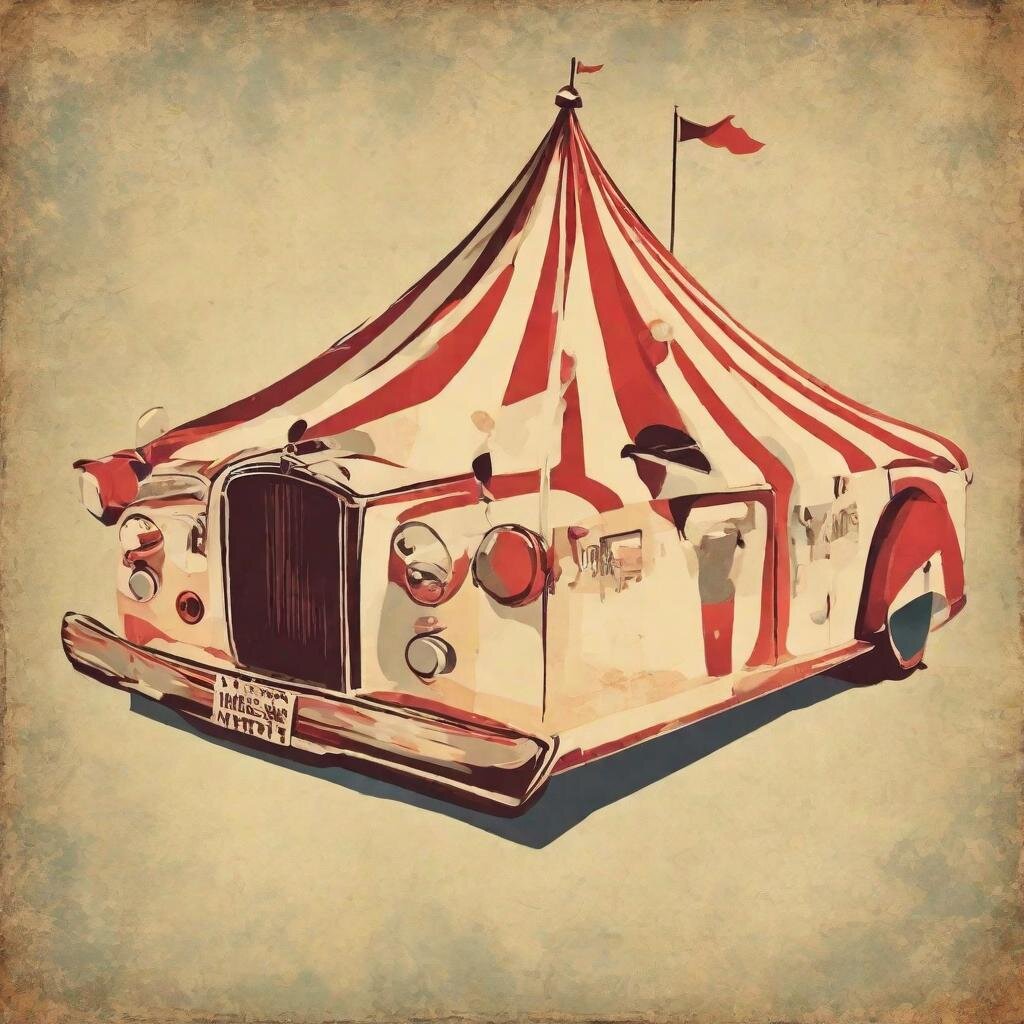}
            \end{minipage} &
            \begin{minipage}{0.2\textwidth}
                \scriptsize \centering Yes, the image features a car that has been creatively transformed into a circus tent...
            \end{minipage} &
            \begin{minipage}{0.14\textwidth}
                \includegraphics[width=\textwidth]{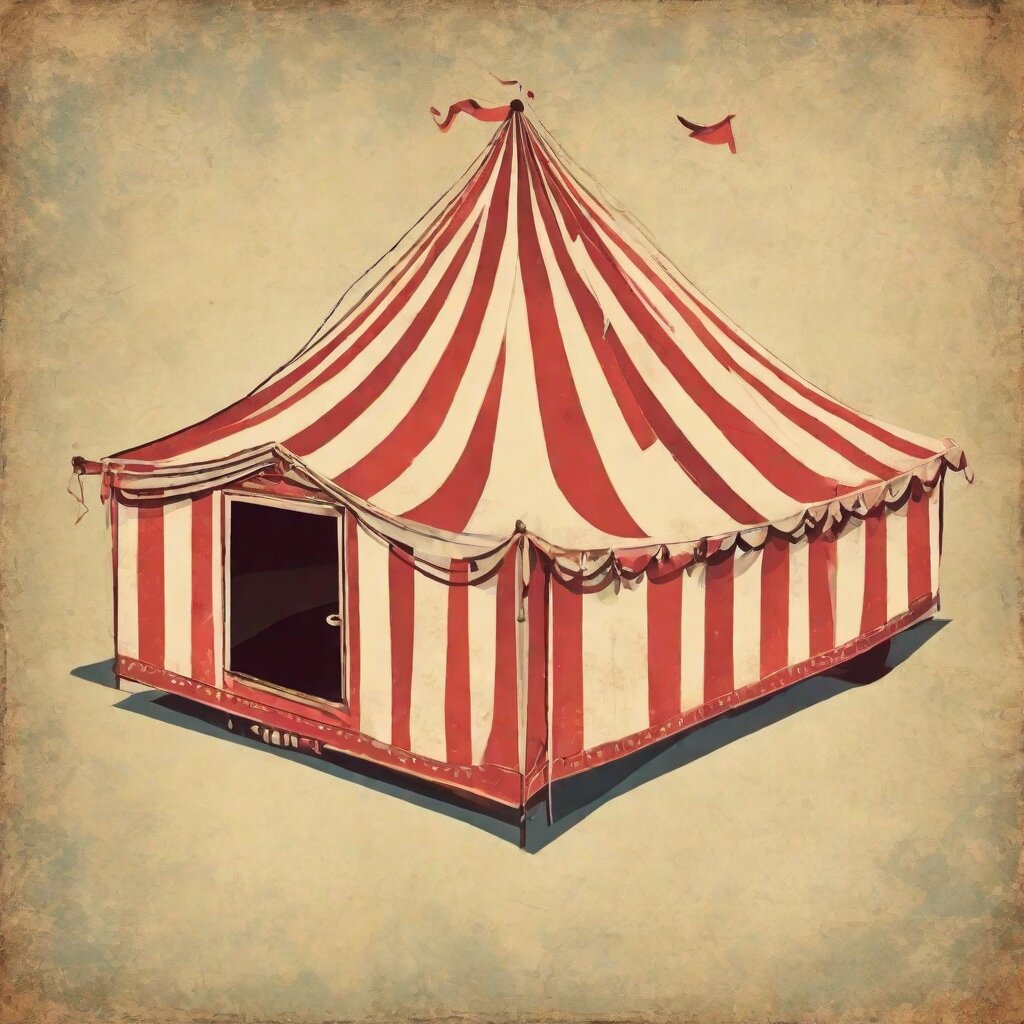}
            \end{minipage} &
            \begin{minipage}{0.2\textwidth}
                \scriptsize \centering No.
            \end{minipage} \\

            \vspace{0.1cm}
            
            \begin{minipage}{0.11\textwidth}
                \scriptsize \centering Q3: Is there any \textit{fork} in this image?
            \end{minipage}  &
            \begin{minipage}{0.14\textwidth}
                \includegraphics[width=\textwidth]{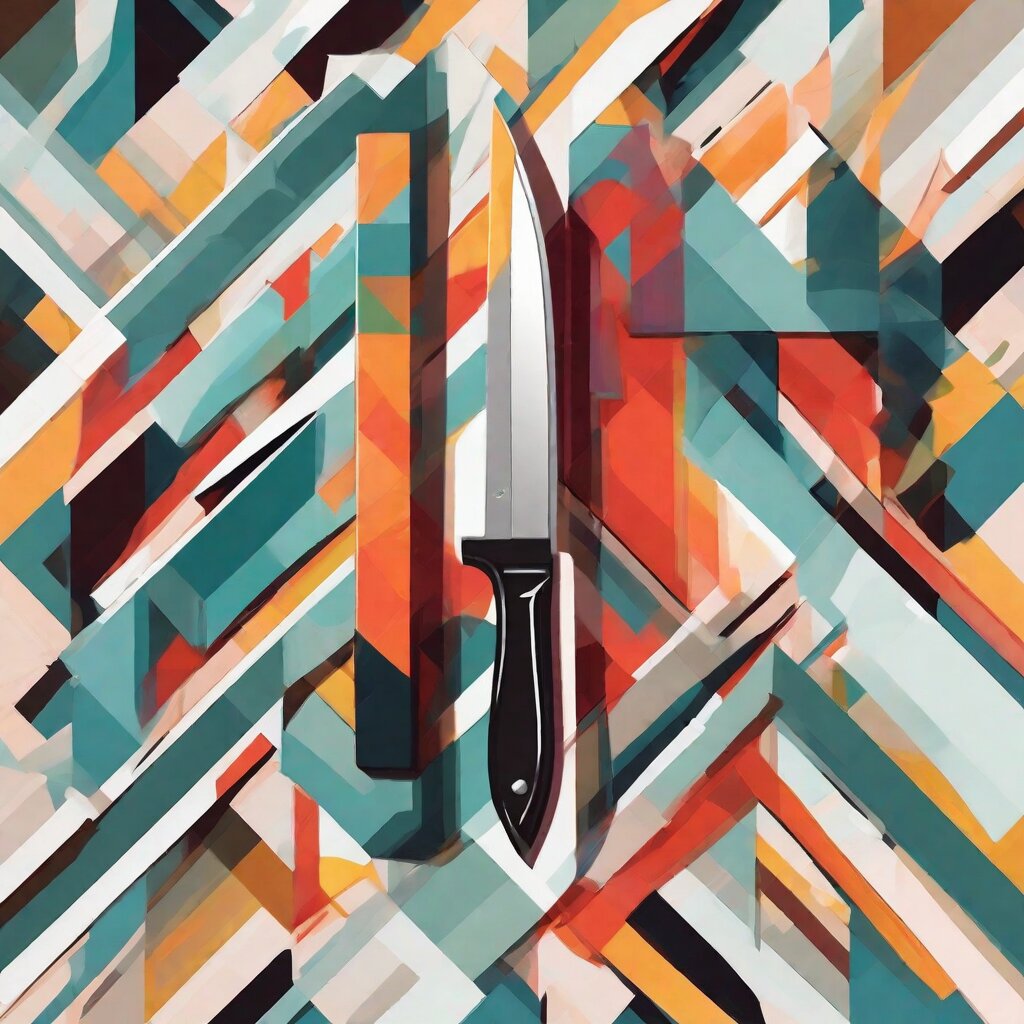} 
            \end{minipage} &
            \begin{minipage}{0.14\textwidth}
                \includegraphics[width=\textwidth]{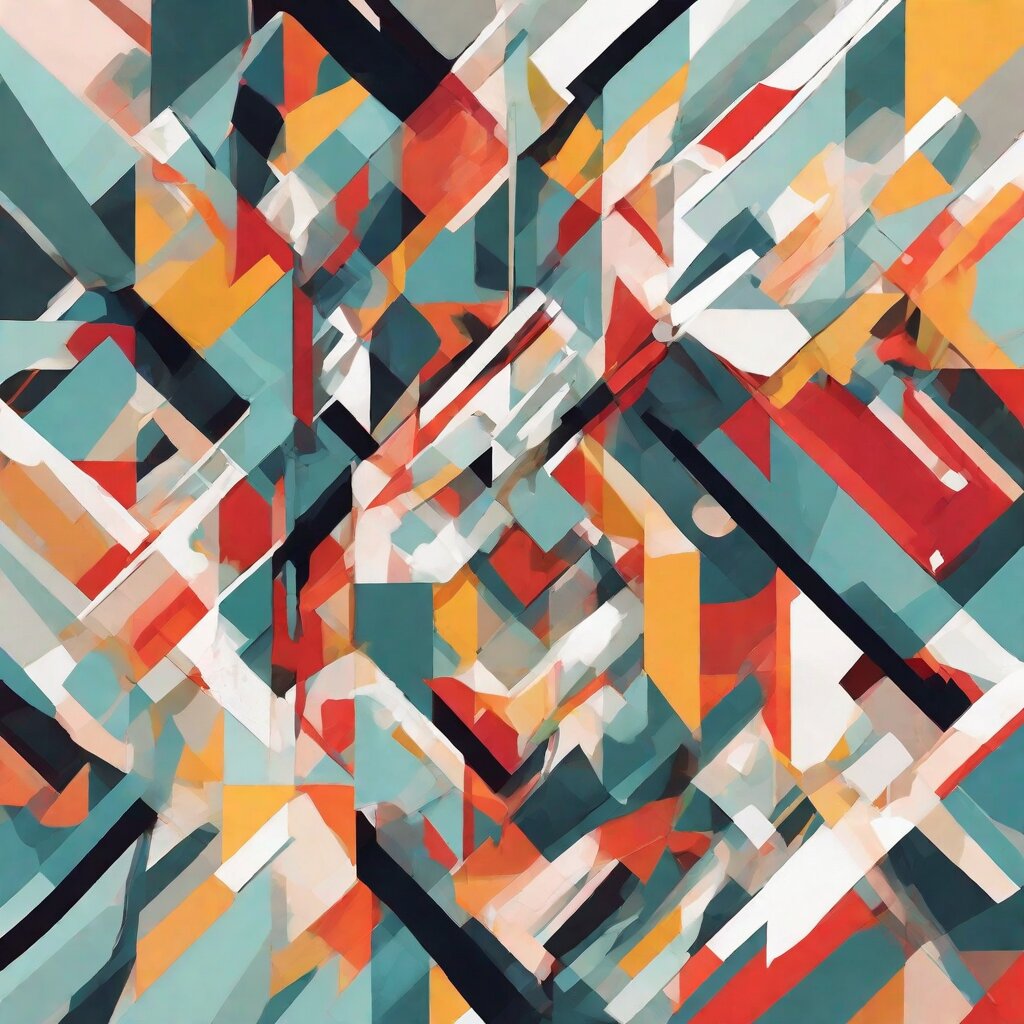}
            \end{minipage} &
            \begin{minipage}{0.2\textwidth}
                \scriptsize \centering No, there is no fork in this image. The image appears to be a digital artwork or abstract design with geometric shapes and vibrant colors. It does not depict any objects that resemble a fork.
            \end{minipage} &
            \begin{minipage}{0.14\textwidth}
                \includegraphics[width=\textwidth]{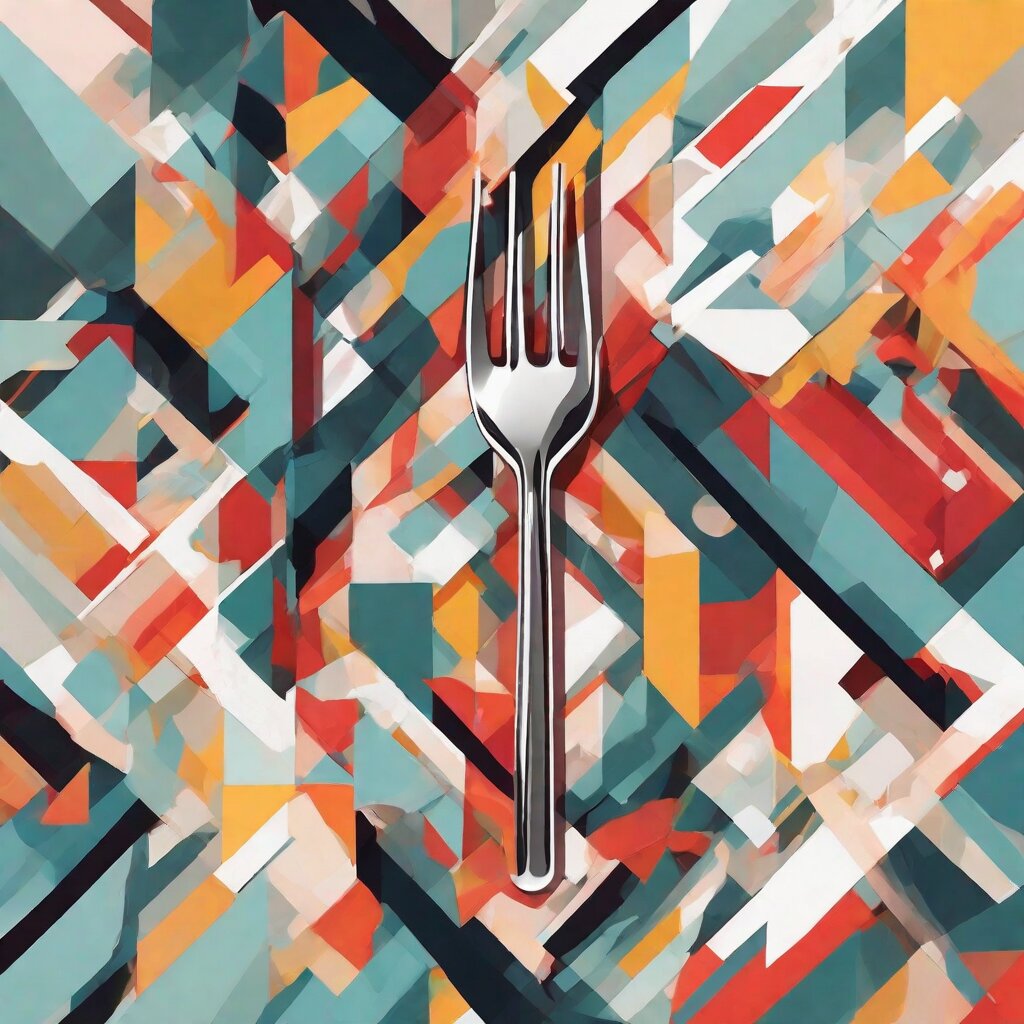}
            \end{minipage} &
            \begin{minipage}{0.2\textwidth}
                \scriptsize \centering Yes, there is a fork in the center of the image.
            \end{minipage} \\

            \vspace{0.1cm}
            
            \begin{minipage}{0.11\textwidth}
                \scriptsize \centering Q3: Is there any \textit{train} in this image?
            \end{minipage} &
            \begin{minipage}{0.14\textwidth}
                \includegraphics[width=\textwidth]{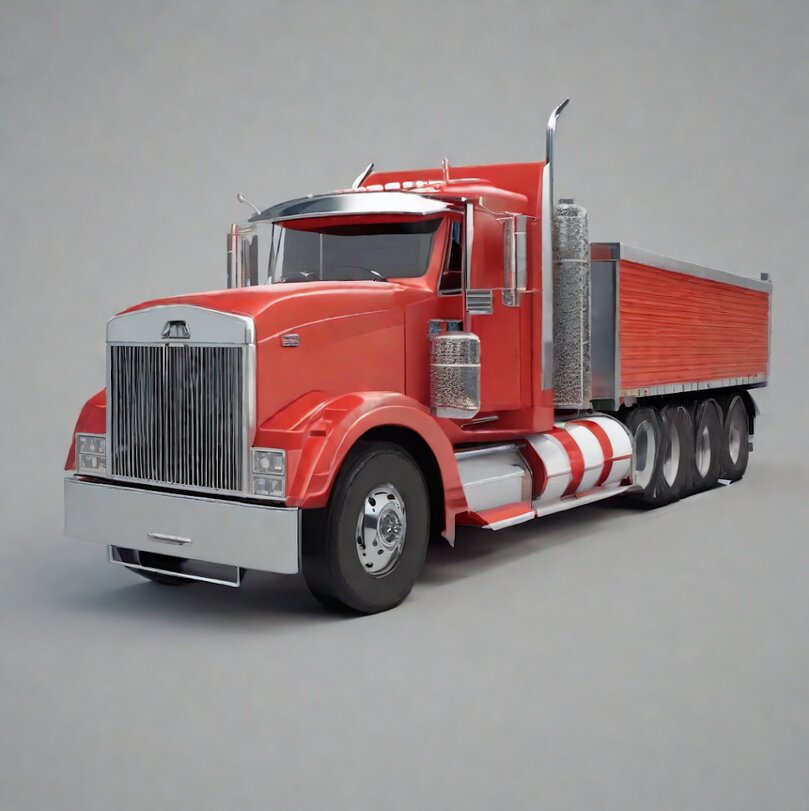} 
            \end{minipage} &
            \begin{minipage}{0.14\textwidth}
                \includegraphics[width=\textwidth]{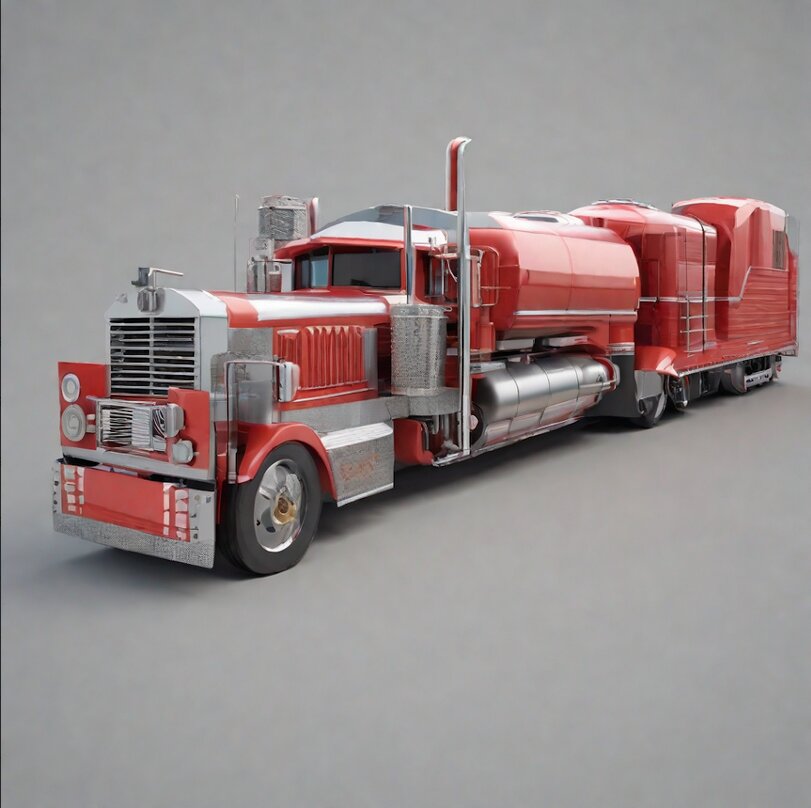}
            \end{minipage} &
            \begin{minipage}{0.2\textwidth}
                \scriptsize \centering No, there is no train in this image. The image features a red and silver 3D model of a semi-truck with a tanker trailer attached to it.
            \end{minipage} &
            \begin{minipage}{0.14\textwidth}
                \includegraphics[width=\textwidth]{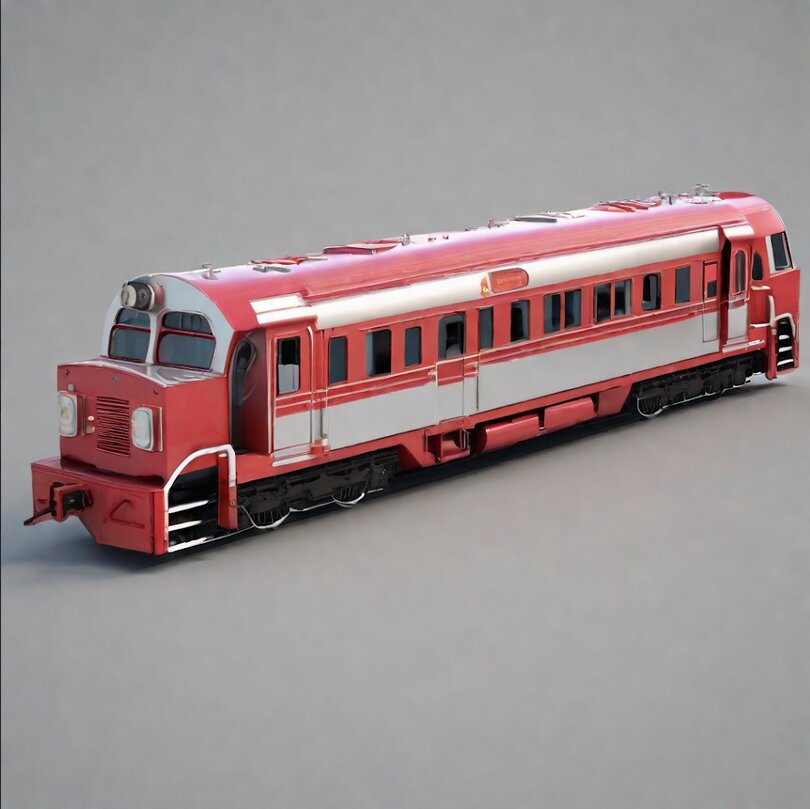}
            \end{minipage} &
            \begin{minipage}{0.2\textwidth}
                \scriptsize \centering Yes, there is a red and white train in the image.
            \end{minipage}
         
        \end{tabular}
   
    \caption{\textbf{Qualitative examples of our LVLM-based evaluation protocol}. We present results from StyleAligned~\cite{hertz2024style}, a method prone to content leakage, and \emph{Only-Style} that mitigates this undesired effect.}
    \label{fig:LVLM_Qual}
\end{figure*}

We also display multiple qualitative results of our evaluation framework in Figure \ref{fig:LVLM_Qual}, to better showcase the purpose of the questions we pose to evaluate content leakage (see Sec. 4.3 of the main manuscript).
Given only the target image and the respective question, we observe that the large multimodal model can understand even subtle content leakage features. 
Moreover, it can unveil cases where the prompt specified target subject is not rendered at all, due to severe content leakage or dominance of background stylistic features (bottom two rows).

Notably, the LVLM systems cannot always provide correct answers for such an intricate task as the content leakage detection of fine-grained visual features.
We showcase such failure cases in Fig. \ref{fig:LVLM_fail}.
First, the LVLM frameworks are prone to hallucinations~\cite{liu2024survey}, sometimes forcing the response to fit the question.  For example, we come across a few object hallucinations, especially when we prompt the model to identify ``visual features" which are subtle by definition (top row of Fig. \ref{fig:LVLM_fail}). 
Moreover, content leakage refers to the presence of the reference image subject in the target image, where the generated target image is not semantically consistent to the target prompt anymore.
Nonetheless, the generated target image may include visual features related to the reference subject that are in line with the requested style and do not affect the correct generation of the target subject, without displaying any content leakage.  
In such cases, the LVLM can correctly detect visual features of the reference subject that, however, do not correspond to a leakage case.
This is particularly evident when it is semantically natural for the reference and the target subject to co-occur in a stylistic alignment scenario (bottom rows of figure \ref{fig:LVLM_fail}). 

However, these limitations do not consistently favor one method over another, so the mean success rates reported on our evaluation dataset serve as a reliable indicator of content leakage for each method. 

\subsection{Details on the User Study}

\begin{figure*}[t]
    \centering
    \includegraphics[width=.7\textwidth]{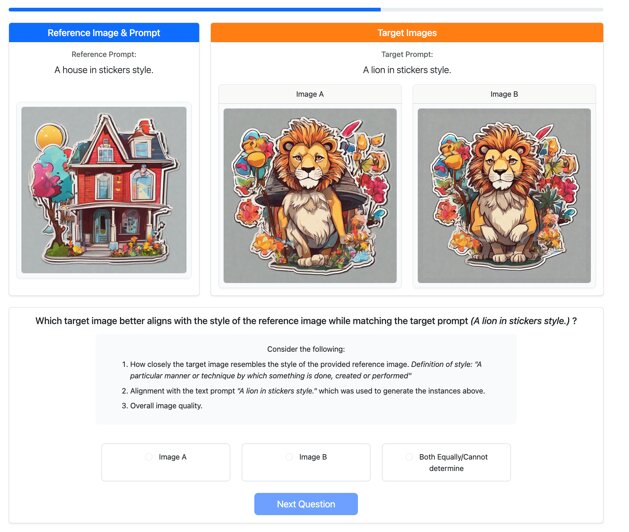} 
    \caption{\textbf{An example screenshot of a question from the conducted perceptual User Study.}}
    \label{fig:User_Study}
\end{figure*}

In the study, users were shown a stylistic reference image alongside two target images, one generated by \emph{Only-Style} and one by a competitor method. The images were accompanied by their generating text prompts. Participants were asked to select their preferred target image based on the following criteria, stylistic alignment to the reference, alignment with the target image prompt and overall image quality, an option cannot decide was also provided, as illustrated in the example of Fig.~\ref{fig:User_Study}. The question aimed to provide an overall evaluation of the factors contributing to successful stylistic alignment. Detailed results of our perceptual User Study with human participants are presented in table~\ref{tab:User_study_absolute_numbers}. As can be observed by the number of undetermined responses, participants often faced challenges in selecting a preferred method due to the conflicting evaluation criteria (style alignment versus text alignment) they were asked to consider simultaneously. Nonetheless, \emph{Only-Style} was significantly preferred over all other baselines. It is worth noting that the significant number of undetermined responses against StyleAligned is due to instances where StyleAligned does not exhibit leakage, resulting in our method producing an identical target image. 

\begin{table}[t]
    \centering
    \resizebox{\linewidth}{!}{%
        \begin{tabular}{lccc}
        \hline
        \textbf{Competitor} & \textbf{Our Method} & \textbf{Competitor Method} & \textbf{Tie/Undetermined} \\ \hline
        StyleAligned~\cite{hertz2024style}         & 357                & 137                        & 306                       \\
        IS~\cite{Wang2024InstantStyle}            & 319                & 210                        & 271                       \\
        B-LoRA~\cite{frenkel2024blora}           & 419                & 155                        & 226                       \\
        SDRP~\cite{sohn2023styledrop}             & 321                & 202                        & 277                       \\ \hline
        \end{tabular}
    }
    \caption{\textbf{Absolute Numbers of our Perceptual User Study}. A total of 800 pairwise comparisons were performed against each competitor method.}
    \label{tab:User_study_absolute_numbers}
\end{table}

\section{Limitations}

Although \emph{Only-Style} consistently localizes the semantic content of the reference image within the target and removes it while preserving stylistic alignment, it exhibits the following limitations.

\begin{itemize}
\item
\textbf{Localization Accuracy:}

Since our goal is to reveal the semantic visual features of the reference subject that ``leaked" in the target image, we want the subject representations $\mathbf{v}_{sub}$ (see Sec. 3.4 of the main manuscript) to focus solely on the semantic features of the image subject. However, in some cases, the retrieved representations also capture stylistic features alongside the semantic ones. 
This results in the unintended identification of stylistic features from the reference subject within the target image as content leakage.

\item
\textbf{Monotonicity Assumption:}
The proposed binary search for determining the optimal scaling operates under the assumption that lower values of the scaling parameter $\alpha$ correspond to reduced content leakage, while higher values increase it.
While this monotonicity assumption relies on a straight-forward intuition (``if we reduce the contribution of the reference subject patches, we will will reduce leakage phenomena") and has been experimentally validated, it lacks a formal theoretical guarantee, especially given the complexity of the diffusion backbone.
Moreover, potential non-accurate localization of the leakage (due to the way that we measure leakage - see 1st limitation) can also disrupt the monotonicity assumption, even though we have not encountered such problem in practice.

\item
\textbf{Computational Complexity:}
Finally, one already discussed issue is the increased overhead of the proposed method compared to the vanilla StyleAligned approach. 
This overhead mainly stems from the iterations of the binary search.
Thus, further reducing the complexity is one of the major directions for future research. 

\end{itemize}

\section{Future work}

As a potential future enhancement we believe that it is worth exploring the ability to adaptively change the scaling parameter $\alpha$ during a single style alignment generation - adopting scheduling tactics or more sophisticated mechanisms.

Moreover, in a different direction, it is imperative to further establish and validate well-suited metrics, such as the proposed LVLM evaluation protocol. The main goal of such an effort is to minimize metric-induced biases (to avoid issues we met while using the set consistency metric for example). Towards this end, we can extend the concept of LVLM acting as ``critics" beyond the content leakage detection.

\onecolumn
\section*{Evaluation prompt set:}
\vspace{-12cm}
{\small
{A house, A dog, A lion, A hippo} in stickers style. \\ 
{A kite, A skateboard, A canoe, A hammock} in watercolor painting style. \\ 
{A hand, A leaf, An eye, A feather} in line drawing style. \\ 
{A dragon, A teapot, A skateboard, A cactus} in cartoon line drawing style. \\ 
{A truck, A boat, A train, A car} in 3D rendering style. \\ 
{A mushroom, A dragon, A dwarf, A fairy} in glowing style. \\ 
{A bottle, A wine glass, A teapot, A cup} in glowing 3D rendering style. \\ 
{A bear, A frisbee, A ball, A torch} in kid crayon drawing style. \\ 
{A couch, A table, A bird, A fish} in wooden sculpture style. \\ 
{An elephant, A zebra, A rhino, A giraffe} in oil painting style. \\ 
{A tree, A flower, A mushroom, A butterfly} in flat cartoon illustration style. \\ 
{A clock, A chameleon, A candle, A cupcake} in abstract rainbow colored flowing smoke wave design. \\ 
{A fork, A spoon, A knife, A glass} in melting golden 3D rendering style. \\ 
{A train, A van, An airplane, A bicycle} in minimalist round BW logo style. \\ 
{A stop sign, A traffic light, A cone, A lighthouse} in neon graffiti style. \\ 
{A car, A bear, A circus tent, A clown} in vintage poster style. \\ 
{A wine glass, A cup, A bowl, A pitcher} in woodblock print style. \\ 
{A surfboard, A wave, A dolphin, A palm} in retro surf art style. \\ 
{A swan, An umbrella, A boat, An airplane} in minimal origami style. \\ 
{A robot, A spaceship, A drone, Godzilla} in cyberpunk art style. \\ 
{A scissors, A bug, A face, A rose} in tattoo art style. \\ 
{A lamp, A chair, A sofa, A mirror} in art deco style. \\ 
{A plant, A bed, A wave, A sunbed} in vintage travel poster style. \\ 
{A rollercoaster, A wheel, A carousel, Balloons} in retro amusement park style. \\ 
{A rocket, A dinosaur, A robot, An alien} in 3D render, animation studio style. \\ 
{A jukebox, A milkshake, A bench, A record player} in 1950s diner art style. \\ 
{A bird, A fox, A cactus, A deer} in Scandinavian folk art style. \\ 
{A dragon, A potion, A sword, A shield} in fantasy poison book style. \\ 
{A giraffe, An elephant, A flamingo, A parrot} in Hawaiian sunset paintings style. \\ 
{A guitar, A balloon, A drum, A microphone} in paper cut art style. \\ 
{A car, A vase, A camera, A watch} in retro hipster style. \\ 
{A suitcase, A ship, A train, A map} in vintage postcard style. \\ 
{A mask, A feather, A tent, A sword} in tribal tattoo style. \\ 
{A wave, A mountain, A cherry, A crane} in Japanese ukiyo-e style. \\ 
{A castle, A knight, A dragon, A wizard} in fantasy book cover style. \\ 
{A fireplace, A blanket, A cup, A book} in hygge style. \\ 
{A stone, A rake, A leaf, A lantern} in Zen garden style. \\ 
{A star, A planet, A comet, The moon} in celestial artwork style. \\ 
{A zebra, A giraffe, A horse, A lion} in medieval fantasy illustration style. \\ 
{A unicorn, A fairy, A castle, A rainbow} in enchanted 3D rendering style. \\ 
{A suitcase, A globe, A plane, A map} in travel agency logo style. \\ 
{A cup, Beans, A croissant, A teapot} in cafe logo style. \\ 
{A book, An owl, A globe, A lantern} in educational institution logo style. \\ 
{A screwdriver, A wrench, A hammer, A toolbox} in mechanical repair shop logo style. \\ 
{A stethoscope, A pill, A syringe, A thermometer} in healthcare and medical clinic logo style. \\ 
{A cloud, A heart, A balloon, A blossom} in doodle art style. \\ 
{A knife, A spoon, A fork, A bowl} in abstract geometric style. \\ 
{A kangaroo, A skyscraper, A lighthouse, A bridge} in mosaic art style. \\ 
{A butterfly, A flamingo, A flower, The sun} in paper collage style. \\ 
{A sunflower, A saxophone, A compass, A guitar} in origami style. \\ 
{A fire hydrant, A trash can, A mailbox, A streetlamp} in abstract graffiti style. \\ 
{A bench, A wolf, A can, A dragon} in street art style. \\ 
{A leaf, A clock, A cloud, A star} in mixed media art style. \\ 
{A snowboard, Skis, A helmet, A ski pole} in abstract expressionism style. \\ 
{A mouse, A keyboard, A laptop, A monitor} in digital glitch art style. \\ 
{A chair, A couch, A mirror, A lamp} in psychedelic art style. \\ 
{A clock, A vase, A painting, A torch} in street art graffiti style. \\ 
{A shoe, A phone, A bottle, A rose} in pop art style. \\ 
{A key, A bird, A door, A lock} in minimalist surrealism style. \\ 
{A cube, A sphere, A pyramid, A circle} in abstract cubism style. \\ 
{A woman, A bicycle, A camera, A bat} in abstract impressionism style. \\ 
{A chair, A table, A lamp, A bookshelf} in post-modern art style. \\ 
{A cat, A car, An android, A drone} in neo-futurism style. \\ 
{A lollipop, A ladder, A star, A rocket} in abstract constructivism style. \\ 
{Lava, Smoke, Water, Fire} in fluid art style. \\ 
{A butterfly, A bug, A blade, A moth} in macro photography style. \\ 
{A burger, A pizza, A salad, A soda} in professional food photography style for a menu. \\ 
{A cup, A wine glass, A plate, A bottle} in vintage still life photography style. \\ 
{A car, A cat, A tree, A bus} in miniature model style. \\ 
{A tent, A campfire, A backpack, A sleeping bag} in outdoor lifestyle photography style. \\ 
{A cat, A train, A serpent, A fish} in realistic 3D render. \\ 
{A record, A cassette, A microphone, A guitar} in retro music and vinyl photography style. \\ 
{A bed, A chair, A fireplace, A table} in cozy winter lifestyle photography style. \\ 
{A candle, A blossom, A light, A vase} in bokeh photography style. \\ 
{A circle, A triangle, A square, A hexagon} in minimal flat design style. \\ 
{A tree, A bird, A bowl, A corn} in minimal vector art style. \\ 
{A cloud, Waves, A blade, A sun} in minimal pastel colors style. \\ 
{A kitten, A tree, A house, A fence} in minimal digital art style. \\ 
{A fish, A bat, A star, A seashell} in minimal abstract illustration style. \\ 
{A mountain, A river, A cloud, A bush} in minimal monochromatic style. \\ 
{A wolf, A skull, A horse, A raven} in woodcut print style. \\ 
{A seashell, A fish, A hand, A starfish} in chalk art style. \\ 
{A heart, A moon, A satellite, Cotton} in pixel art style. \\ 
{A superhero, A villain, A city, A spaceship} in comic book style. \\ 
{A rocket, A planet, A spaceship, A dragon} in vector illustration style. \\ 
{A house, A car, A tree, A cat} in isometric illustration style. \\ 
{A computer, A phone, A camera, A tablet} in wireframe 3D style. \\ 
{A leaf, A cloud, A fish, A wave} in paper cutout style. \\ 
{A building, A bridge, A truck, A leopard} in blueprint style. \\ 
{A hero, A monster, A spaceship, A robot} in retro comic book style. \\ 
{A flowchart, An advertisement, A map, A graph} in infographic style. \\ 
{A microscope, A crystal, A flag, A telescope} in geometric shapes style. \\ 
{A cat, A dog, A bird, A rabbit} in cartoon line drawing style. \\ 
{A flower, A tree, A river, A mountain} in watercolor and ink wash style. \\ 
{A mushroom, A clock, A fish, A key} in dreamy surreal style. \\ 
{A car, A clock, A pipe, A gear} in steampunk mechanical style. \\ 
{Clock, Globe, Map, A compass} in 3D realism style. \\ 
{A bus, A scooter, A car, A bicycle} in retro poster style. \\ 
{A flower, A feather, A bat, A cactus} in bohemian hand-drawn style. \\ 
{Panda, Rhino, Telescope, Hippo} in vintage stamp style. \\ 

\twocolumn

\end{document}